\documentclass{phdunitartu}
\usepackage{phdstyle}
\ThesisType{COLLECTION} 
\author{Fatemeh Rastgar}
\title{Towards reliable real-time trajectory optimization}
\institute{Institute of Technology} 
\faculty{Faculty of Science and Technology} 
\affiliation{University of Tartu} 

\ISSNPRINT{2228-0855}
\ISBNPRINT{978-9916-27-550-4} 
\ISSNPDF{2806-2620}
\ISBNPDF{978-9916-27-551-1} 

\dissertationseriesname{DISSERTATIONES TECHNOLOGIAE UNIVERSITATIS TARTUENSIS}  
\dissertationseriesnumber{83}  
\thesisfield{Physical Engineering} 
\commencementdate{May 28th, 2024,} 
\defensedate{.........} 
\defensetime{...} 
\defenseloc{Nooruse 1-121} 

\addSup{Supervisors:}{&Arun Kumar Singh, PhD\\
   & Associate Professor of Collaborative Robotics\\
   & Institute of Technology, University of Tartu\\
    &Tartu, Estonia\\
\\
     &Alvo Aabloo, PhD\\
   &Professor of Polymeric Materials, Materials Science\\
   &Institute of Technology, University of Tartu\\
   & Tartu, Estonia

}

\addREW{Reviewer:}{
    &Mozhgan Pourmoradnasseri, PhD\\
    &Lecturer in Mobility Modelling \\
    & Institute of Computer Science, University of Tartu \\
    & Tartu, Estonia
}
\addOpp{Opponent:}{
    &Andreas Müller, PhD\\
    & Prof. Dr.-Ing. Habil, Institute of Robotics \\
    & Johannes Kepler University, Linz, Austria\\
}

\addCOM{Commencement:}{
    &Auditorium 121, Nooruse 1, Tartu, Estonia, at 10.15\\
    & on June 21st, 2024\\
}

\bibliography{main}

\DeclareUnicodeCharacter{2212}{-}
\begin{document}
    \maketitle
    \thispagestyle{empty}
    \sloppy
    \newpage\null\vskip35mm
\begin{flushright}
\itshape To my dearest mother, Masoumeh,\\
my beloved father, Torabali,\\
and my loving husband, Iman.
\end{flushright}
    \setcounter{page}{6}
    \newpage\pagestyle{plain}

\chapter*{Abstract}

Motion planning is a key aspect of robotics, allowing robots to move through complex and changing environments. A common approach to address motion planning problems is trajectory optimization. Trajectory optimization can represent the high-level behaviors of robots through mathematical formulations. However, current trajectory optimization approaches have two main challenges. Firstly, their solution heavily depends on the initial guess, and they are prone to get stuck in local minima. Secondly, they face scalability limitations by increasing the number of constraints.

This thesis endeavors to tackle these challenges by introducing four innovative trajectory optimization algorithms to improve reliability, scalability, and computational efficiency.

There are two novel aspects of the proposed algorithms. The first key innovation is remodeling the kinematic constraints and collision avoidance constraints. Another key innovation lies in the design of algorithms that effectively utilize parallel computation on GPU accelerators. By using reformulated constraints and leveraging the computational power of GPUs, the proposed algorithms of this thesis demonstrate significant improvements in efficiency and scalability compared to the existing methods. Parallelization enables faster computation times, allowing for real-time decision-making in dynamic environments. Moreover, the algorithms are designed to adapt to changes in the environment, ensuring robust performance even in unknown and cluttered conditions.

Extensive benchmarking for each proposed optimizer validates their efficacy. Through comprehensive evaluation, the proposed algorithms consistently outperform state-of-the-art methods across various metrics, such as smoothness costs and computation time. These results highlight the potential of the proposed trajectory optimization algorithms to significantly advance the state-of-the-art in motion planning for robotics applications.

Overall, this thesis makes a significant contribution to the field of trajectory optimization algorithms. It introduces innovative solutions that specifically address the challenges faced by existing methods. The proposed algorithms pave the way for more efficient and robust motion planning solutions in robotics by leveraging parallel computation and specific mathematical structures.

    
    %
    \tableofcontents
    \listoffigures
    \listoftables
    \chapter*{List of abbreviations}

\renewcommand{\glossarysection}[2][]{\section*{#1}}
\printglossary[type=\acronymtype,nonumberlist]
\printglossary[title=Nomenclature]

    \IfThesisTypeIsCollection{\chapter*{List of original publications}
\addcontentsline{toc}{chapter}{List of original publications}
\section{Publications Included In The Thesis}

\begin{enumerate}

\item \textbf{F. Rastgar}, A. K. Singh, H. Masnavi, K. Kruusamae, A. Aabloo, "A novel trajectory optimization for affine systems: Beyond convex-concave procedure," 2020 IEEE/RSJ International Conference on Intelligent Robots and Systems (IROS), Las Vegas, NV, USA, 2020, pp. 1308-1315, doi: 10.1109/IROS45743.2020.9341566. 

\item  \textbf{F. Rastgar}, H. Masnavi, K. Kruusamäe, A. Aabloo and A. K. Singh, "GPU Accelerated Batch Trajectory Optimization for Autonomous Navigation," 2023 American Control Conference (ACC), San Diego, CA, USA, 2023, pp. 718-725 doi: 10.23919/ACC55779.2023.10156088

\item \textbf{F. Rastgar}, H. Masnavi, B. Sharma, A. Aabloo, J. Swevers, A. K. Singh, "PRIEST: Projection Guided Sampling-Based Optimization For Autonomous Navigation," in IEEE Robotics and Automation Letters, January 2024, doi: 10.1109/LRA.2024.3357311


\item  \textbf{F. Rastgar}, H. Masnavi, J. Shrestha, K. Kruusamäe, A. Aabloo and A. K. Singh, "GPU Accelerated Convex Approximations for Fast Multi-Agent Trajectory Optimization," in IEEE Robotics and Automation Letters, vol. 6, no. 2, pp. 3303-3310, April 2021, doi: 10.1109/LRA.2021.3061398.

\end{enumerate}

\section{Author's Contributions}

In Publication I \cite{rastgar2020novel}, the author proposed a novel algorithm for trajectory optimization for affine systems Beyond the convex-concave procedure and compared the proposed optimizer with the \gls{sota} methods. The author was also responsible for writing different sections of the paper. 

\noindent In Publication II \cite{rastgar2023gpu}, the author extended the previous work and proposed a novel batch trajectory optimization algorithm for autonomous navigation problems. The author conducted the simulations and implementations. She also compared with \gls{sota} methods and drafted various sections of the paper.

\noindent In Publication III \cite{rastgar2024priest}, the author expanded upon previous research on batch trajectory optimization algorithms and designed a projection guided sampling-based optimization algorithm for autonomous navigation. The author conducted simulations and implementations on real-world applications, comparing them with \gls{sota} methods. Additionally, she took responsibility for drafting various sections of the paper.


\noindent In Publication IV \cite{RASTGAR2021}, the author developed a novel algorithm for multi-agent trajectory optimization and compared the proposed optimizer with the \gls{sota} method. In addition, the author undertook the task of writing different sections of the paper.

\section{Other Publications}

\begin{enumerate}

\item D. Guhathakurta, \textbf{F. Rastgar}, M. A. Sharma, K M. Krishna, A. K. Singh, "Fast Joint Multi-Robot Trajectory Optimization by GPU Accelerated Batch Solution of Distributed Sub-Problems," in Frontiers in robotics and AI, 9, 890385, doi: https://doi.org/10.3389/frobt.2022.890385, \cite{guhathakurtafast}.

\item V. K. Adajania, H. Masnavi, \textbf{F. Rastgar}, K. Kruusamäe and A. K. Singh, "Embedded Hardware Appropriate Fast 3D Trajectory Optimization for Fixed Wing Aerial Vehicles by Leveraging Hidden Convex Structures," 2021 IEEE/RSJ International Conference on Intelligent Robots and Systems (IROS), Prague, Czech Republic, 2021, pp. 571-578, doi: 10.1109/IROS51168.2021.9636337, \cite{adajania2021embedded}.

\item \textbf{F. Rastgar}, M. Rahmani, "Distributed robust filtering with hybrid consensus strategy for sensor networks," in IET Wireless Sensor Systems, vol. 10, no. 1, pp. 37-46, 2020/2, doi: https://doi.org/10.1049/iet-wss.2019.0093 \cite{rastgar2019distributed}.

\item \textbf{F. Rastgar}, "Exploiting Hidden Convexities for Real-time and Reliable Optimization Algorithms for Challenging Motion Planning and Control Applications," Proceedings of the 20th International Conference on Autonomous Agents and MultiAgent Systems, May 3-7, 2021, Online, \cite{fat}.

\end{enumerate}}
    
    %
    \include{sections/preface}
\chapter{Motion Planning Challenges and Objectives}\label{intro}

\section{Introduction}

\begin{figure}[h]
    \centering
\includegraphics[scale=0.85]{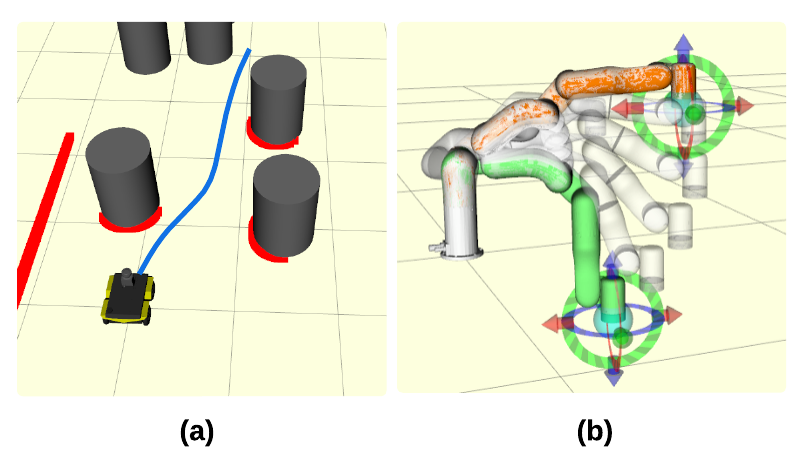}
    \vspace{-0.2cm}
    \caption[Motion planning example]{\small{Motion planning examples in a (a) mobile robot (the sequence of positions are shown as a blue trajectory). (b): manipulator ( the sequence of joint motions is shown in transparent color)}}
    \label{motion_example}
\end{figure}

Motion planning is a critical component of any robotic application. In simple terms, it involves computing how different independent parts of a robot will move over a specific time horizon to perform tasks such as object manipulation or navigation \cite{la2011motion, schwartz1988survey, kavraki2016motion}. For instance, in the case of a manipulator, motion planning can be thought of as computing the sequence of joint motions to grasp particular objects \cite{sandakalum2022motion}. Likewise, for a mobile robot, motion planning is expressed in terms of computing the sequence of spatial positions for the robot to navigate through a cluttered environment \cite{sun2021motion}(see Figure \ref{motion_example}).

There are three broad classes of approaches for motion planning, namely graph-search, sampling-based, and trajectory optimization methods. Graph-search methods like $A^*$ \cite{hart1968formal,erke2020improved} and Dijkstra \cite{dijkstra2022note,zhang2014multiple} represent the environment as a graph, where nodes represent potential robot states, and edges show possible transitions between states. Through traversing this graph, graph-based algorithms determine the shortest path from the starting point to the destination. However, these methods are computationally heavy, especially when the robot has many degrees of freedom or when we have to plan over a long horizon in cluttered and dynamic environments \cite{huang2017motion,zang2024graphmp}.

Sampling-based methods, such as Rapidly-exploring Random Trees (RRTs) \cite{kuffner2000rrt, rodriguez2006obstacle, wang2022rapidly} and Probabilistic Roadmaps (PRMs) \cite{kavraki1996probabilistic, geraerts2004comparative,kannan2016robot}, offer a robust approach to exploring the configuration space of a robot.
Nonetheless, a common challenge associated with these methods is the tendency to generate non-smooth trajectories, leading to suboptimal paths and potential inefficiencies in meeting tight constraints \cite{huang2017motion}.


In recent years, trajectory optimization methods have become the default standard for motion planning since they allow for encoding a robot's behavior through carefully designed cost functions and a set of constraints \cite{gopalakrishnan2014time, zhao2018efficient,howell2022trajectory}. For example, cost functions may be designed to facilitate smooth motions \cite{zucker2013chomp} or track a specific path, while constraints encompass boundary conditions on position, velocity, acceleration \cite{adajania2021embedded}, and collision avoidance criteria \cite{zucker2013chomp}. By modeling mathematical functions to define the robot's behavior, optimization-based approaches generate smooth trajectories capable of performing complex navigation and manipulation tasks. Thus, this thesis concentrates on improving trajectory optimization approaches.

\vspace{0.45cm}
\noindent \textbf{Core Challenge:} Trajectory optimization problems are straightforward when the underlying cost and constraint functions have a property called convexity. We discuss the exact mathematical description of convexity later in section \ref{convex_opt}. But intuitively, convexity ensures that the trajectory optimization problem has only one solution (global minimum), and we are guaranteed to find it. Moreover, there is a large collection of optimization algorithms (or optimizers) with efficient open-source implementations that can be applied to convex problems \cite{andersen2020cvxopt}.

Unfortunately, the majority of trajectory optimization problems encountered in robot motion planning are non-convex \cite{singh2020bi}. More precisely, the cost and constraint functions modeling the motion planning problem do not have the convexity property. For example, one major source of non-convexity stems from the collision avoidance constraints that are quintessential in any navigation or manipulation problems \cite{singh2020bi}. Intuitively, non-convexity results in having multiple solutions (local-minima), and it is often difficult to predict which one of the potential solutions will be returned by a particular optimizer. Nevertheless, many optimizers are proposed in the existing literature that can efficiently compute local minima in a wide class of problems \cite{metz2003rockit, schulman2013finding, augugliaro2012generation}. However, the following fundamental challenges still remain, especially if the aim is to use trajectory optimization for real-time motion planning in dynamic environments.




    

\begin{itemize}
    \item \textbf{C1 Scalability:} Motion planning in highly cluttered environments and over long horizons requires considering a large number of collision avoidance and kinematic constraints. However, 
existing \gls{sota} optimizers like \gls{sqp}\cite{stoer1985principles, betts2000very}, Interior-point \cite{roos2005interior}, etc, implemented in software libraries like ROCKIT \cite{metz2003rockit}, FATROP \cite{vanroye2023fatrop}, ACADO \cite{houska2011acado,nilsson2018trajectory}, IPOPT \cite{lipp2016variations} do not scale well with the increase in the number of non-convex constraints. More precisely, their computation time increases sharply with either the planning horizon or the number of obstacles.
    
    \item \textbf{C2 Initialization:} Existing optimizers for non-convex problems heavily rely on the user providing a good guess of potential optimal solutions. Poor initial guesses can result in the optimizer running for a long time without even converging to a feasible solution or converging to a bad local minimum.
\end{itemize}

This thesis aims to provide a solution to the two challenges described above and develop novel non-convex optimizers that push the boundary of robot motion planning. 

\section{Objective and Contributions of the Thesis}

The overall objective of this thesis is to tackle the challenges associated with trajectory optimization and improve their reliability, scalability, and computational performance. We focus primarily on problems encountered for robot navigation, although results can find potential utility in manipulation as well. To achieve these goals, we are focusing on two core ideas:

\begin{itemize}
    \item \textbf{I1: Exploiting specific structures in the trajectory optimization problem:}
    Our key idea is to reformulate constraints, such as collision avoidance, into a suitable form that exposes hidden convex structures and allows us to leverage these structures for efficient computation.  
  
    \item \textbf{I2: Leveraging parallel computing abilities of modern computing hardware like \gls{gpu}:} Another important idea behind the work presented in the thesis is to find ways to exploit the parallel computation ability of \gls{gpu}s. For example, we can reformulate the collision-avoidance or kinematic constraints in a way that allows for breaking trajectory optimization into smaller parallelizable sub-problems.
    
\end{itemize}

In the following, we provide a brief summary of the publications associated with each core idea presented above and how they solve the scalability (C1) and initialization (C2) bottlenecks of existing approaches.
Also, a diagram that shows the interrelation among papers is provided in Figure \ref{diagram}.
To enhance clarity in this research, the work is divided into two distinct parts: single robots and multi-agent robots. Moving forward, I will always begin by discussing works related to single robots and then transition to multi-agent robots.

\begin{itemize}

 \item Paper I (Addressing Challenge C1 based on Idea I1):
 We present a new approach to formulating collision avoidance constraints. We demonstrate that this novel representation has some multi-convex structures that can be exploited through techniques such as \gls{am} and \gls{admm} \cite{boyd2004convex,byrne2011alternating}. We demonstrate that our resulting optimizer has a better scaling with the number of obstacles. We validate our optimizer by comparing it with the \gls{sota} method, \gls{ccp} \cite{lipp2016variations}, which utilizes affine approximations of collision avoidance constraints in terms of both optimal cost and computation time.

 \item Paper II (Addressing Challenge C2 based on Idea I2):  This work is based on a simple idea that one way to by-pass the local-minima issue in non-convex optimization is to run the optimizer from multiple initialization. We can then choose the best solution among the different local minima obtained. In this work,
 we introduce a novel \gls{gpu}-accelerated optimizer that allows us to implement this multiple-initialization idea for real-time navigation.
 We show that the optimization problem can be reduced to just computing large matrix-vector products that can be trivially parallelized across \gls{gpu}s.
 We also demonstrate that our batch optimizer has linear scalability with the number of parallel problem instances (or initialization) and collision avoidance constraints. Additionally, we benchmark our optimizer against the \gls{sota} method, Cross-Entropy Method \gls{cem} \cite{pinneri2021sample}, in terms of success rate and tracking cost.

 \item Paper III (Addressing Challenge C2 based on Idea I2):
 This work provides us insight on how to solve several optimizations in parallel. This paper uses this foundation to combine sampling-based (gradient-free) and convex optimization. In particular, we introduce a projection optimizer with sampling-based optimizer routines to guide the samples towards feasible regions. 
we compare our proposed optimizer against both SOTA gradient-based methods (FATROP and RACKIT ), and Gradient-free approaches, (\gls{ros} navigation stack and Cross-Entropy Method (\gls{cem})), and show improvements in terms of success rate, time-to-reach the goal and computation time.

 



\begin{figure}[t]
    \centering
\includegraphics[scale=0.30]{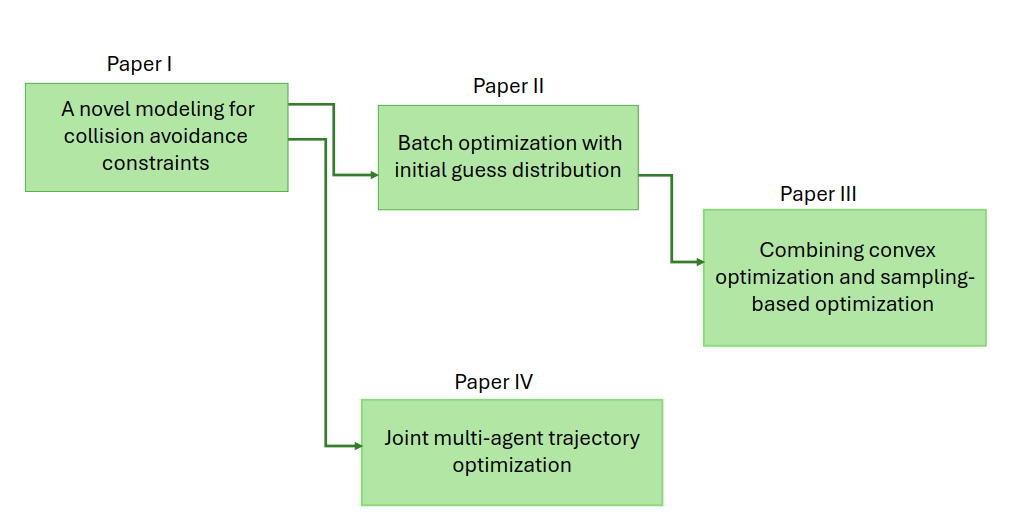}
    \caption[Interrelation among publications]{\small{Publication interrelation: single robot and multi-agent robots }}
    \label{diagram}
\end{figure}

 \item  Paper IV (Addressing Challenges C1 and C2 based on Idea I2): Joint trajectory optimization for multiple agents are generally considered intractable but provides good quality solution due to access to a large feasible space. In paper IV, we make joint optimization more tractable, by reformulating the inter-agent collision avoidance into a certain form allows us to decompose the underlying computations into an offline and online part. The offline part involves expensive matrix factorization and needs to be done only once for a given class of problems. The online part involves computing just matrix-vector products that can be trivially parallelized across \gls{gpu}s. In this work, we introduced a fast joint multi-agent trajectory optimizer and compared it with \gls{sota} methods, Sequential Convex Programming (SCP), in terms of optimal costs and computation time.

\end{itemize}

 This thesis is organized as follows. In Chapter \ref{background}, an overview of the mathematical concepts utilized in this thesis is provided. Chapter \ref{literature} defines the basic trajectory optimization problem and provides a comprehensive literature review. Subsequently, Chapters \ref{papper_1}- \ref{paper_4} elaborate on each publication in detail. Finally, Chapter \ref{conclusion} offers conclusions and outlines future directions.

    
    \chapter{Mathematical Preliminaries } \label{background}


 This chapter offers a concise overview of the foundational mathematics used in this dissertation.  



\section{Convex Set}

\begin{figure}[h]
    \centering
    \includegraphics[scale=0.30]{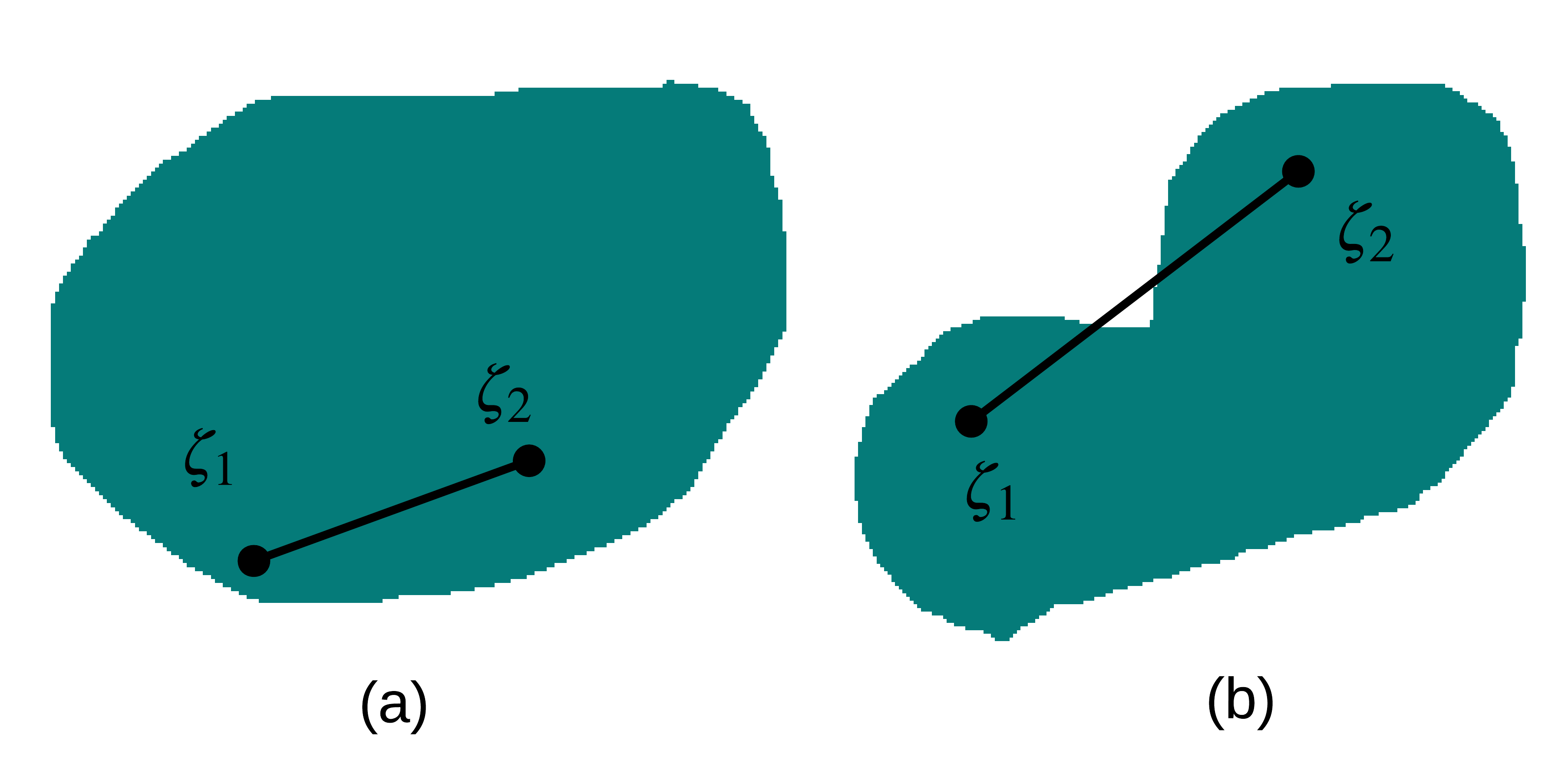}
    \caption[Illustration of a convex and non-convex set.]{\small{Illustration of a convex and non-convex set. (a) The line segment between points $\xi_{1}$ and $\xi_{2}$ is entirely contained within the green set. (b) In contrast, a portion of the line segment connecting $\xi_{1}$ and $\xi_{2}$ extends outside of the green~set.}}
    \label{convex_set}
\end{figure}

\begin{definition}
 A set $\mathcal{C}$ is convex if for all $\xi_{1}$ and $\xi_{2}$ in $\mathcal{C}$, $ \theta \xi_{1}+(1-\theta)\xi_{2} \in \mathcal{C}$~\cite{boyd2004convex}. 
\end{definition}
\noindent In simpler terms, this definition implies that all points on the line segment connecting any two arbitrary points $\xi_{1}$ and $\xi_{2}$ must also belong to the $\mathcal{C}$ (see Figure \ref{convex_set}).

\section{Convex Function}

\begin{figure}[h]
    \centering
    \includegraphics[scale=0.50]{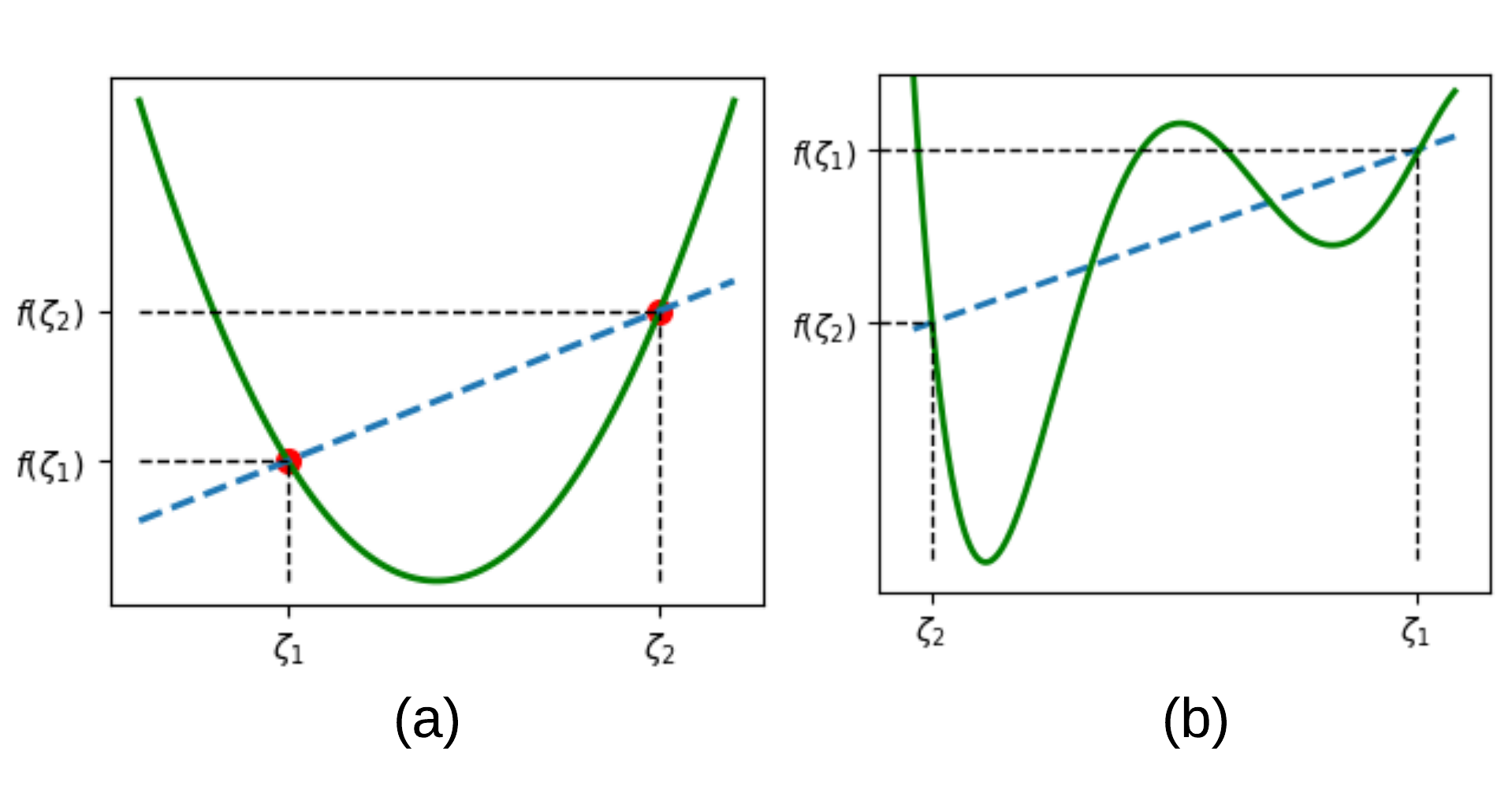}
    \vspace{-0.2cm}
    \caption[Illustration of a convex and non-convex function]{\small{Illustration of a Convex and Non-Convex function. (a) The function remains below or on the line segment, connecting two points. (b) A a portion of $f(\xi)$ is situated above the line segment connecting $\xi_{1}$ and $\xi_{2}$.}}
    \label{convex-fun}
\end{figure}

\begin{definition}
A function $f:\mathbb{R}^{n} \rightarrow\mathbb{R}$ is convex if its domain is a convex set and for all $\xi_{1}$, $\xi_{2}$ in its domain, and all $\theta \in
\left[0,1\right]$, $f(\theta \xi_{1} + (1-\theta)\xi_{2}) \leq \theta f(\xi_{1})+(1-\theta)f(\xi_{2})$ \cite{boyd2004convex}.
\end{definition}
\noindent Geometrically, a function is convex if and only if the line segment connecting any two points on the graph of the function lies above or on the graph between these two points. (see Figure \ref{convex-fun}). 

\section{Convex Optimization Problem} \label{convex_opt}
A standard form of convex optimization problem can be written as:

\begin{align}
    &\min_{\xi} f(\xi) \\
    &\text{s.t.:}~~ g_{j}(\xi) \leq 0,~ j =1,2,..., m \\
    &~~~~~~ h_{i}(\xi) = 0,~ i =1,2,..., p
\end{align}


\noindent where $\xi$ is the optimization variable; $f$ is the objective function and it is convex. The inequality constraint function, which is convex, is shown with $g_{j}$, and equality constraint function, which has affine form, is shown as $h_{i}$.
A non-convex optimization problem is a problem in which at least one part of the optimization problem involves a function that does not satisfy the properties of convexity. 

\section{The Significance of Convexity}

Convexity plays a pivotal role in our pursuit of optimizing functions. It holds a special place in this thesis due to its ability to reveal essential information about the minima, which are the solutions to our optimization problems.

\begin{figure}[!h]
    \centering
    \includegraphics[scale=0.85]{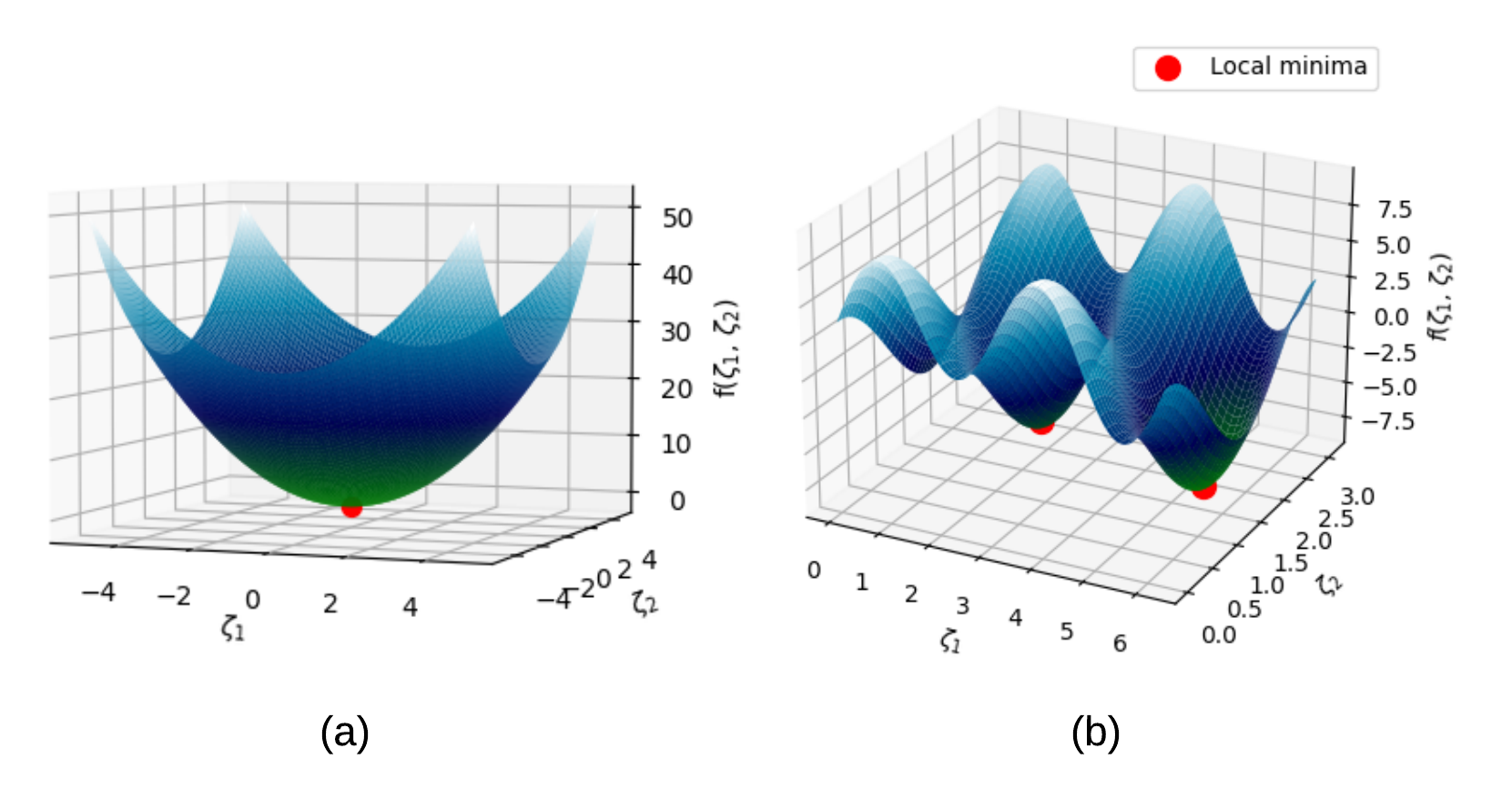}
    \vspace{-0.5cm}
    \caption[Local minima in a convex and non-convex function]{\small{Local minima in a convex (a) and non-convex function(b)}}
    \label{local-minima}
\end{figure}

One of the defining characteristics of convex functions is that any local minimum also serves as the global minimum, given the existence of a unique global minimum (refer to \cite{boyd2004convex} for proof). This is in stark contrast to non-convex functions, which are characterized by the presence of several local minima. It is important to note that in non-convex functions, these local minima may or may not correspond to the global minimum (see Figure \ref{local-minima}). 

\begin{figure}[h]
    \centering
    \includegraphics[scale=0.60]{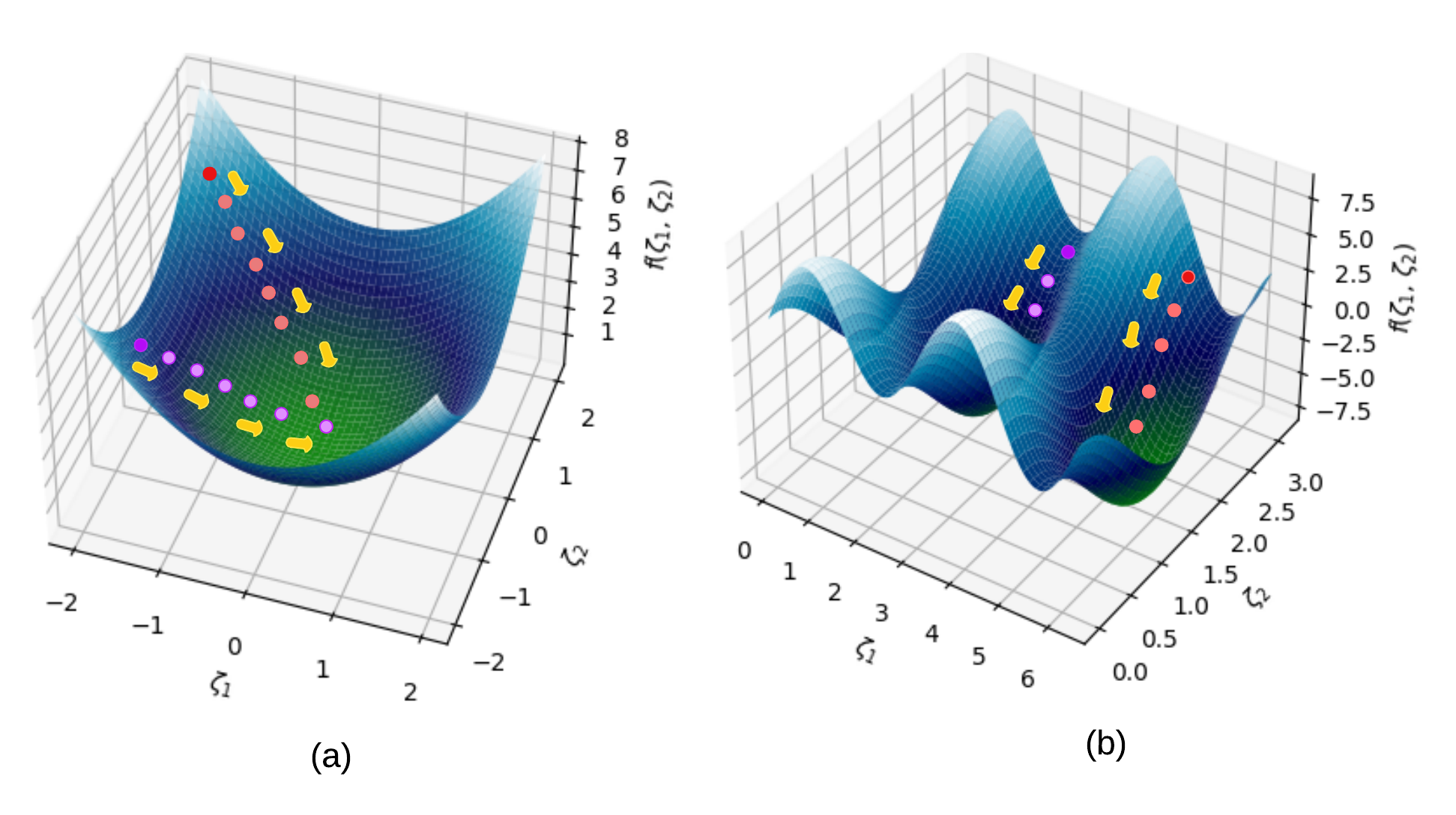}
    \vspace{-0.3cm}
    \caption[Effect of initial guess on an optimization problem]{\small{Demonstration of how initial guesses (red and purple points) can affect the solution of a convex and non-convex function. (a) In a convex function, both the purple and red points converge to the same global minimum. (b) In a non-convex function, distinct local minima are reached for each of these initial guesses.}}
    \label{initial_guess}
\end{figure}

Why does this property matter? Solving an optimization problem requires an initial guess. Convex problems ensure that no matter where we start, we always end up in the same best spot because there is only one minimum. This makes optimization efficient and reliable, leading to globally optimal solutions. However, non-convex problems depend on the initial guess, determining which local minimum is reached in the end (see Figure \ref{initial_guess}). It should be mentioned that whether these minima are acceptable depends on the specific task at hand. 

\section{Multi-Convex Function}\label{def_multi-convex}

\begin{figure}[!h]
    \centering
    \includegraphics[scale=0.39]{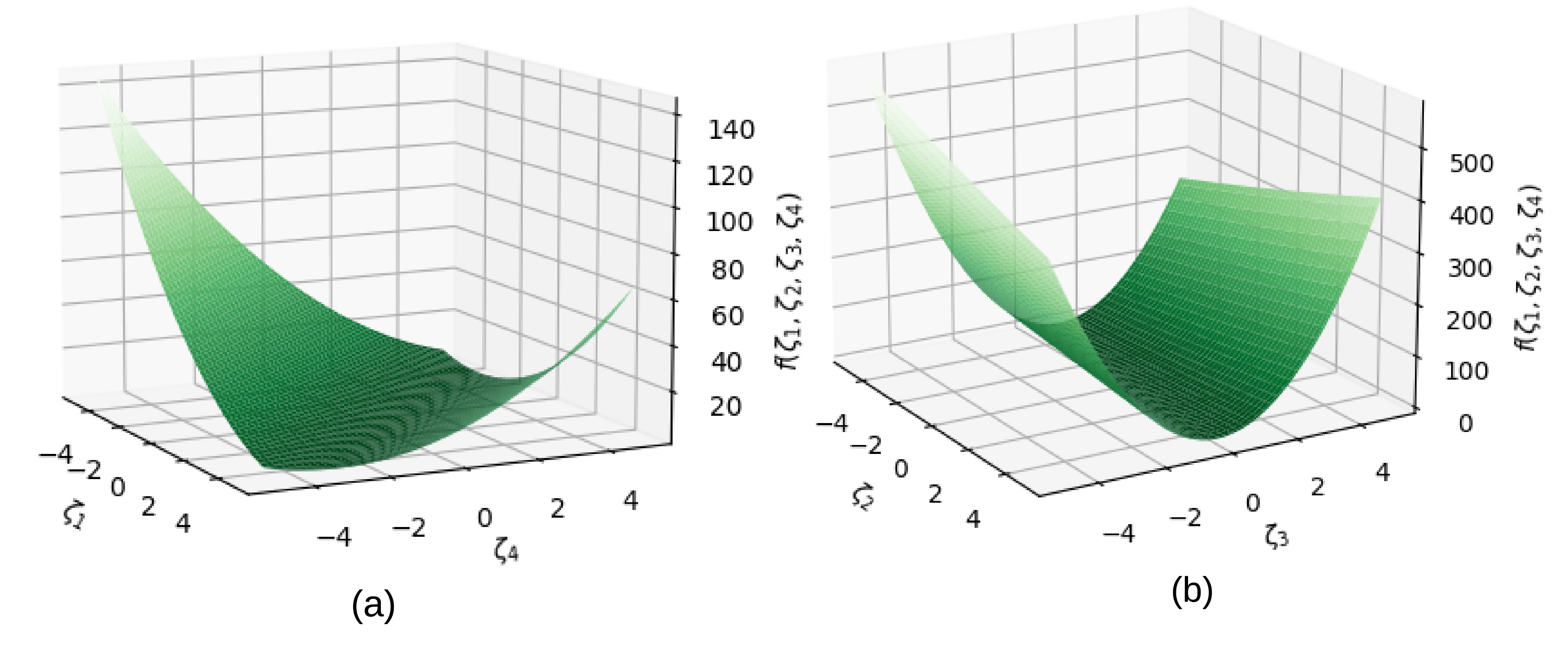}
    \vspace{-0.6cm}
    \caption[Multi-convex function]{\hspace{-0.13cm}\small{An example of a multi-convex function. The function $f\hspace{-0.12cm}= \hspace{-0.11cm}(\hspace{-0.05cm}\xi_{\hspace{-0.02cm}1}\xi_{\hspace{-0.01cm}2}\hspace{-0.07cm}+\hspace{-0.07cm}\xi_{\hspace{-0.02cm}3}\xi_{\hspace{-0.02cm}4}\hspace{-0.1cm}-\hspace{-0.1cm}2\hspace{-0.03cm})^{\hspace{-0.02cm}2}$ is a multi-convex function. We can observe its convexity by fixing certain variables as follows: (a) when $\xi_{2} = 1$ and $\xi_{3}=1$, $f$ remains convex. This can be visualized as a convex function in the $\xi_{1}$ and $\xi_{4}$ variables while holding $\xi_{2}$ and $\xi_{3}$. Similarly, when $\xi_{1}\hspace{-0.1cm}=\hspace{-0.1cm}0.4 $ and  $\xi_{4}=4$, $f$  remains convex. This can be visualized as a convex function in the $\xi_{2}$ and $\xi_{3}$ variables while holding $\xi_{1} $ and  $\xi_{4}$~fixed.}}
    \label{multi-convex}
\end{figure}

\begin{definition}
Multi-convexity refers to a property of optimization problems where the variables can be partitioned into different sets, and within each set, the problem is convex when the other variables are held fixed \cite{shen2017disciplined, jain2017non}. For example, the function $f\hspace{-0.12cm}= \hspace{-0.11cm}(\hspace{-0.05cm}\xi_{\hspace{-0.02cm}1}\xi_{\hspace{-0.01cm}2}\hspace{-0.07cm}+\hspace{-0.07cm}\xi_{\hspace{-0.02cm}3}\xi_{\hspace{-0.02cm}4}\hspace{-0.1cm}-\hspace{-0.1cm}2\hspace{-0.03cm})^{\hspace{-0.02cm}2}$ is a multi-convex function. We can partition the variables into two sets including $\xi_{2},\xi_{3}$ and $\xi_{1},\xi_{4}$. When we consider $\xi_{2},\xi_{3}$ fixed, we can observe that the function is convex in terms of $\xi_{1}$ and $\xi_{4}$. Similarly, by fixing $\xi_{1},\xi_{4}$, the function is convex in terms of $\xi_{2}$ and $\xi_{3}$   (see Figure \ref{multi-convex}).
\end{definition}

Multi-convexity allows us to break complex problems into simpler subproblems. Each of these subproblems are convex and can be solved independently through optimization methods.

\section{Quadratic Programming (QP)} \label{qp_definition}
\begin{definition}
A \gls{qp} optimization problem is a mathematical problem that involves minimizing a quadratic cost function subject to linear inequality and equality constraints \cite{boyd2004convex}. The \gls{qp} problem can be defined as

\small
 \begin{align}
  &\min_{\boldsymbol{\xi}}\frac{1}{2}\boldsymbol{\xi}^{T}\textbf{Q}\boldsymbol{\xi} + \textbf{q}^{T}\boldsymbol{\xi} 
  \nonumber \\
 & \text{s.t.}: 
  \textbf{A}\boldsymbol{\xi} = \textbf{b}
  \label{qp_cost} 
 \end{align}
 \normalsize



\noindent where $\boldsymbol{\xi} \in \mathbb{R}^{n_{v}}$ is the optimization variable and $n_{v}$ shows the number of variables. The symmetric matrix $\textbf{Q}\in \mathbb{R}^{n_{v}\times n_{v}}$ defines the quadratic term. The matrix
$\textbf{A}\in \mathbb{R}^{n_{p}\times n_{v}}$ presents the coefficients of the equality constraints. The vectors $\textbf{b}\in \mathbb{R}^{n_{p}}$ specify the value of the equality constraint. 
\end{definition}

If the \gls{qp} problem has only equality constraints, then the problem \eqref{qp_cost} can be converted to a set of linear equations as 

\small
\begin{align}
    \begin{bmatrix}
        \textbf{Q} & \textbf{A}^{T} \\
        \textbf{A} & \textbf{0}
    \end{bmatrix}\begin{bmatrix}
        \boldsymbol{\xi} \\
        \boldsymbol{\nu} 
    \end{bmatrix} = \begin{bmatrix}
        -\textbf{q} \\
        \textbf{b}
    \end{bmatrix} \label{qp_linear}
\end{align}
\normalsize

\noindent where $\boldsymbol{\nu}\in \mathbb{R}^{n_{p}}$ is the dual optimization variable. Finally, \eqref{qp_linear} can be solved as 

\small
\begin{align}
   \begin{bmatrix}
        \boldsymbol{\xi} \\
        \boldsymbol{\nu} 
    \end{bmatrix} =  \begin{bmatrix}
        \textbf{Q} & \textbf{A}^{T} \\
        \textbf{A} & \textbf{0}
    \end{bmatrix}^{-1} \begin{bmatrix}
        -\textbf{q} \\
        \textbf{b}
    \end{bmatrix}. \label{qp_sol}
\end{align}
\normalsize

\noindent As can be seen, the problem \eqref{qp_cost} simplifies to only a matrix -vector production. 

If I assume that the matrix $\textbf{Q}$ is semi-definite positive, then \eqref{qp_sol} has a unique solution.  
This property of \eqref{qp_sol} is instrumental in my research. In subsequent chapters,  I demonstrate how to convert the non-convex optimization problems into \gls{qp} problems with convex costs. Following this, I further transform them into a system of linear equations. Then, I show that this transformation is beneficial because linear equations can be solved more efficiently and are more easily parallelizable over a GPU. 

     \chapter{Basic Problem Formulation and Review of Existing Approaches } \label{literature}

This chapter introduces a basic trajectory optimization problem and offers an overview of existing solution approaches. Before presenting the problem formulation, the symbols, and notations utilized throughout this thesis are established. Additionally, the concept of differential flatness, a property used in the optimization problem, is elucidated.

\vspace{0.45cm}
\noindent \textbf{Symbols and Notations:} I adopt a notation convention where lowercase normal font letters denote scalars, bold font letters represent vectors, and bold uppercase letters signify matrices. The variables $t$ and $T$ correspond to time stamps and transpose of vectors/matrices, respectively. The left superscript $k$ represents the optimizer's iteration. Table \ref{table_notation} provides a concise summary of some notations utilized in this research. Additional notations will be introduced at their first instance of use.
 
It should be mentioned that for the sake of consistency throughout this thesis, a uniform notation is employed across all chapters. It is acknowledged that each original paper introduces its distinctive set of notations and variable names, which may deviate from those specified in this thesis. Nonetheless, to ensure clarity and precision, symbols and notations in each original paper are explicitly defined within the corresponding paper.

\begin{table}[h]
\caption[Notations]{\small {Notations used throughout the thesis}}
\centering
\label{table_notation}
\begin{tabular}{|
>{\columncolor[HTML]{FFFFC7}}c |
>{\columncolor[HTML]{FFFFC7}}c |ccc}
\cline{1-2}
\cellcolor[HTML]{036400}{\color[HTML]{FFFFFF} Notation} & \cellcolor[HTML]{036400}{\color[HTML]{FFFFFF} Definition}       &  &  &  \\ \cline{1-2}
$(x(t),y(t),z(t))$                                        & Robot position                                            &  &  &  \\ \cline{1-2}
$(x_{o, j}(t),y_{o, j}(t),z_{o, j}(t)) $         & $j^{th}$ obstacle position                                &  &  &  \\ \cline{1-2}
$(x_{des}(t),y_{des}(t),z_{des}(t)) $         &  desired position                                &  &  &  \\ \cline{1-2}
$v_{min},v_{max} $  &  Minimum and maximum velocity                                &  &  &  \\ \cline{1-2}
$a_{min},a_{max} $  &  Minimum and maximum acceleration ~~~~~~~~~~                               &  &  &  \\ \cline{1-2}
$n_{p}$       &   Number of planning steps \\ \cline{1-2}
$n_{o}$ &
Number of obstacles & & &\\ \cline{1-2}
$n_{v}$ &
Number of decision variables & & &\\ \cline{1-2} 
$n_{c}$ &  Number of multi-circles& & & \\ \cline{1-2} 
$N$ &  Number of iterations& & &  \\ \cline{1-2}
$N_{b}$ &  Number of batches& & &  \\ \cline{1-2}
\cline{1-2}
$N_{a}$ &  Number of agents& & &  \\ \cline{1-2}
\cline{1-2}
\end{tabular}
\end{table}

\vspace{0.45cm}
\noindent \textbf{Differentially Flat Robot Motion Model:} 
Throughout this thesis, I assume that the robot motion model has a property called differential flatness. This allows ensures the control inputs $\mathbf{u}=\boldsymbol{\Phi}(x^{(q)}(t),y^{(q)}(t),z^{(q)}(t))$ can be obtained through some analytical mapping $\boldsymbol{\Phi}$ of $q^{th}$ level derivatives of the position-level trajectory. For example, for a simple 2D double integrator robots, the control inputs are simply $\mathbf{u} = (\ddot{x}(t), \ddot{y}(t))$. Similarly, for a car-like robot, we can express the forward acceleration and steering inputs as a function of position derivatives \cite{han2023efficient,ryu2011differential}(More details are provided in Chapter \ref{paper_3}).


\section{Basic Trajectory Optimization Problem for 3D Navigation}

I am interested in addressing the fundamental problem of trajectory optimization, which is crucial for navigating a holonomic robot (e.g., a quadrotor) in 3D space. A key element of this problem is collision avoidance. The robot’s task is twofold: it is required to meet its navigation objectives, such as smoothness or following a desired trajectory, and it also needs to ensure avoiding collisions with obstacles in its environment. To aid in this, I model the obstacles as axis-aligned ellipsoids with $(a, a, b)$ dimensions. Subsequently, collision avoidance can be defined as a series of constraints that keep the robot’s trajectory free from potential collisions.  In Chapters \ref{paper_2}-\ref{paper_4}, I consider a more sophisticated version of this problem. Nevertheless, the basic formulation would allow us to identify the gaps in the existing literature as well as highlight our contribution in the later chapters. The mathematical structure of the optimization problem is defined as follows.

\begin{subequations}
\begin{align}
&\min_{x(t), y(t), z(t)}\sum_{t} c_{x}(x^{(q)}(t))+c_{y}(y^{(q)}(t))+ c_{z}(z^{(q)}(t))  
\label{g_1}\\
   & \text{s.t.: }
    -\frac{\hspace{-0.07cm}(x(t)\hspace{-0.09cm}-
 \hspace{-0.06cm}x_{o, j}(t)\hspace{-0.01cm})^{2}}{a^2}\hspace{-0.09cm} - \hspace{-0.09cm}
 \frac{\hspace{-0.07cm}(y(t) \hspace{-0.075cm}-\hspace{-0.065cm}y_{o, j}(t)\hspace{-0.02cm})^{2}}{a^2} 
  \hspace{-0.09cm}-\hspace{-0.09cm}\frac{\hspace{-0.07cm}(z(t)\hspace{-0.09cm}- 
 \hspace{-0.08cm}z_{o, j}(t)\hspace{-0.01cm})^{2}}{b^2}\hspace{-0.06cm} \hspace{-0.06cm}
  + \hspace{-0.05cm}1\leq 0,1\leq j\leq n_{o}\label{g_3}
\end{align}
\end{subequations}

\noindent where $(x(t),y(t),z(t))$ and $(x_{o, j}(t),y_{o, j}(t),z_{o, j}(t))$ respectively denote the robot and the $j^{th}$ obstacle position at time $t$. The function $c_{x}(.), c_{y}(.)$ and $c_{z}(.)$ are quadratic convex functions encompassing smoothness, trajectory tracking error, or distance to a desired goal position. Even bounds of position and their derivatives can be expressed as quadratic costs and can be included in the cost function. The $(.)^q$ represents the $q^{th}$ derivative of position variables. Constraints \eqref{g_3} also enforce collision avoidance. 

\subsection{Trajectory Parametrization}
Optimization \eqref{g_1}-\eqref{g_3} is defined in terms of trajectory functions. To express it as a standard optimization problem in terms of finite-dimensional variables, I parameterize the $x(t), y(t),$ and $z(t)$ as smooth polynomials.


\begin{align}
    \begin{bmatrix}
    x(t_{1}) \\ \vdots \\ x(t_{n_{p}})
    \end{bmatrix}^{T} \hspace{-0.3cm} = \mathbf{P} \hspace{0.1cm}\boldsymbol{\xi}_{x} ,~~
     \begin{bmatrix}
    y(t_{1})\\ \vdots\\ y(t_{n_{p}})
    \end{bmatrix}^{T} \hspace{-0.2cm}= \mathbf{P} \boldsymbol{\xi}_{y}, ~~
    \begin{bmatrix}
    z(t_{1}) \\ \vdots \\ z(t_{n_{p}})
    \end{bmatrix}^{T} \hspace{-0.3cm} = \mathbf{P} \hspace{0.1cm}\boldsymbol{\xi}_{z} \label{parameter}
\end{align}

\noindent where $\mathbf{P}$ is a matrix created using time-dependent polynomial basis functions that map coefficients $\boldsymbol{\xi}_{x}, \boldsymbol{\xi}_{y}, \boldsymbol{\xi}_{z}$ to the trajectory variables $x(t), y(t), z(t)$. Similar expressions can be applied for derivatives $\dot{x}(t), \dot{y}(t), \dot{z}(t), \ddot{x}(t),\ddot{y}(t)$ and $\ddot{z}(t)$ in terms of trajectory coefficients and derivatives of the basis function matrix $\dot{\mathbf{P}}, \ddot{\mathbf{P}}$.

\vspace{0.45cm}
\begin{remark}
    The choice of matrix $\bold{P}$ includes Bernstein polynomial \cite{sassi2015bernstein}, cubic spline \cite{kolter2009task}, etc. When considering the matrix $\bold{P}$ is identity, the parametrization essentially represents the trajectories as a sequence of waypoints.
\end{remark}

\subsection{Reformulating Trajectory Optimization \beqref{g_1}-\beqref{g_3} Using Trajectory Parametrization}

Using trajectory parametrization, the optimization problem \eqref{g_1}-\eqref{g_3} can be rewritten as

\begin{align}
 &\min_{\boldsymbol{\xi}} \frac{1}{2}\boldsymbol{\xi}^{T}\mathbf{Q}\boldsymbol{\xi}+\mathbf{q}^T\boldsymbol{\xi}  \label{p_1}\\
&~~ \text{s.t.: } \mathbf{g}(\boldsymbol{\xi})\leq \mathbf{0}  \label{eq_1} 
\end{align}

\noindent where $\boldsymbol{\xi} =\begin{bmatrix} \boldsymbol{\xi}_{x}&\boldsymbol{\xi}_{y}&\boldsymbol{\xi}_{z}\end{bmatrix}^{T}$. Due to the presence of collision avoidance constraints, \eqref{eq_1}, our optimization problem becomes non-convex, posing a significant challenge to solve. In the following, I elaborate on how various methods address this issue and discuss their limitations.

\section{Literature Review}
In this section, available methods for solving the \eqref{p_1}- \eqref{eq_1} are reviewed. In addition, it is explained how different methods can handle non-convex inequality constraints and what limitations they have. 

\subsection{Gradient Descent (\gls{gd})} \label{gd_methods}
 Gradient Descent (\gls{gd}) is a common technique for solving unconstrained trajectory optimization problems \cite{bengio2000gradient,boyd2004convex,snyman2005new}. It begins by taking an initial guess of the trajectory parameters and calculating the gradient of the cost function concerning these variables \cite{boyd2004convex}. Subsequently, it updates the trajectory parameters by taking a small step in the direction opposite to the gradient. This process iterates until the decrease in the cost function saturates or the maximum iteration limit is reached.
 
\vspace{0.35cm}
\noindent\textbf{How does \gls{gd} method solve the trajectory optimization problem \eqref{p_1}-\eqref{eq_1}?} 
\gls{gd}-based method is primarily designed for unconstrained problems. Thus, to apply this method to the optimization problem \eqref{p_1}-\eqref{eq_1}, inequality constraints are relaxed as penalties in the cost function. The reformulated optimization problem can be written as:

\begin{align}
 \min_{\boldsymbol{\xi}}\overbrace{\Big( \frac{w_{1}}{2}\boldsymbol{\xi}^{T}\mathbf{Q}\boldsymbol{\xi}+\mathbf{q}^T\boldsymbol{\xi} + w_{2} f_{pen}(\boldsymbol{\xi})) \Big)} ^{f_{gd}}\label{gd_1} 
\end{align}

\noindent where $f_{pen}$ is the penalty function. There are various choices for $f_{pen}$, and \cite{boyd2004convex} provides a good overview of them. Also, $w_{1}$ and $w_{2}$ are weights to make a trade-off between different terms of the cost function. 
The steps to solve \eqref{gd_1} using \gls{gd} are :

\begin{enumerate}
    \item \textbf{Initialization:} Choose an initial guess for the decision variables ${^k}\boldsymbol{\xi}$ at iteration $k=0$
    \item \textbf{Compute Gradient:} Calculate the gradient of the objective function with respect to $\boldsymbol{\xi}$ and evaluate at  ${^k}\boldsymbol{\xi}$ . The gradient is given by: 

\begin{align}
\nabla \mathbf{f}_{gd}(^{k}\boldsymbol{\xi})=\mathbf{Q}\hspace{0.1cm}^{k}\boldsymbol{\xi}+\mathbf{q} + \nabla \mathbf{f}_{pen}(\boldsymbol{\xi}), \label{gd_2}
\end{align}

\noindent and $\mathbf{Q}\boldsymbol{\xi}+\mathbf{q}$ represents the gradient of the quadratic term.
\item \textbf{Update Decision Variables:} Update the decision variables using the gradient descent update rule:

\begin{align}
   ^{k+1} \boldsymbol{\xi} = \hspace{0.1cm}^{k}\boldsymbol{\xi} - \eta \nabla\mathbf{f}_{gd}(^{k}\boldsymbol{\xi}) \label{gd_3}
\end{align}

\noindent where $\eta$ is the step size (or learning rate) and $k$  denotes the iteration number.

\item \textbf{Termination Criterion:} Repeat steps 2-4 until a termination criterion is met, such as reaching a maximum number of iterations, achieving a desired objective function value, or observing small changes in the decision variables.
\end{enumerate}

\vspace{0.2cm}
\noindent \textbf{\gls{gd} method limitations:} The limitations of \gls{gd}-based methods are:

\begin{itemize}
    \item Choosing the appropriate $\eta$ is crucial \cite{saad1998optimal}. Typically, a small value for $ \eta$ is chosen. This slows down the convergence of \gls{gd}. Conversely, selecting a higher $\eta$ can lead to divergence. 

    \item In its original form, \gls{gd} is not designed for constrained problems. In practice, careful choice of the constraint weights $w_i$s are required to make \gls{gd} work. However, the choice of $w_{i}$ is problem-specific and difficult to know apriori. 
\end{itemize}

\vspace{0.35cm}
\noindent\textbf{Existing works:}
A notable example of \gls{gd}-based algorithms in trajectory optimization is the Covariant Hamiltonian Optimization (CHOMP) method, which uses covariant gradient techniques to enhance the quality of sampled trajectories \cite{zucker2013chomp,ratliff2009chomp}. However, as with any \gls{gd}-based method, CHOMP exhibits sensitivity to the selection of parameters, such as the learning rate. Furthermore, it is prone to getting stuck in local minima.

Similarly, the authors in \cite{campana2016gradient} introduced another \gls{gd}-based method that begins with an initial collision-free trajectory. The method employs a basic gradient to shorten the trajectory, thereby optimizing it. However, this method is not immune to the typical issues associated with \gls{gd}-based techniques, such as learning rate selection and getting stuck in local minima.


\subsection{Interior Points}
Interior point methods are a class of optimization algorithms widely used for solving constrained optimization problems \cite{roos2005interior,wright2005interior,renegar2001mathematical}. These methods require the problem to be in a specific form where all constraints are expressed as equality constraints. To ensure this, inequality constraints, such as collision avoidance, are converted into an equality form. The core mechanism of interior point methods involves iteratively updating primal and dual variables to make progress toward the optimal solution while satisfying both constraints and optimality conditions. This update process continues until convergence is achieved, typically when the solution satisfies specified convergence criteria.

\begin{figure}[!h]
    \centering
\includegraphics[scale=0.13]{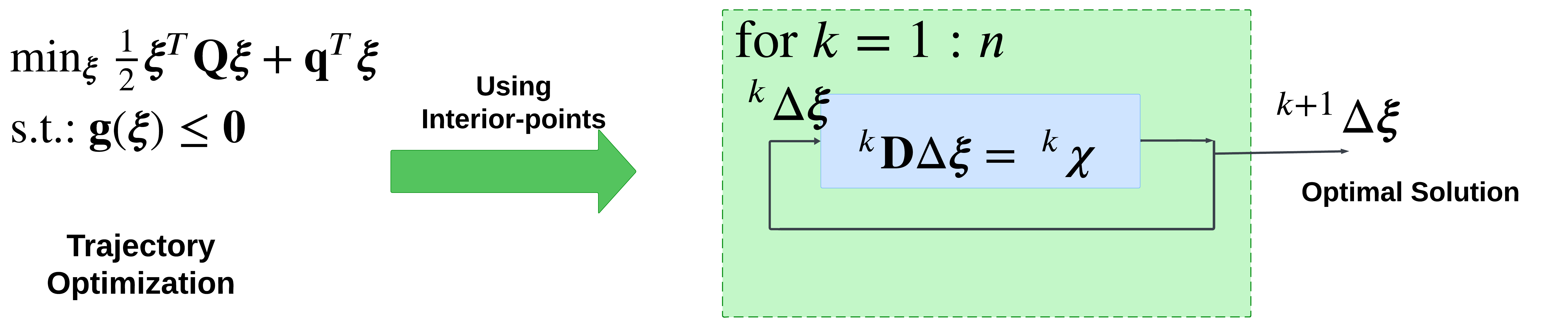}
    \caption[Interior-Points Schematic]{\small{The trajectory optimization can be reduced to solving a set of linear equations iteratively where $^{k}\mathbf{D}$ and $^{k}\boldsymbol{\chi}$ represent a changing matrix and vector for each iteration, respectively. Also, $\Delta\boldsymbol{\xi}$ shows the update  }}
    \label{interior}
\end{figure}

\noindent\textbf{How does Interior Points method solve the trajectory optimization problem \eqref{p_1}-\eqref{eq_1}?} It can be shown that the optimization problem \eqref{eq_1} can be reduced to solving a set of linear equations iteratively (see Figure \ref{interior}). The slack variable $\mathbf{s}$ is introduced to achieve such a form, and the optimization problem is rewritten in the following manner.

\begin{align}
&\min_{\boldsymbol{\xi}, \mathbf{s}} \frac{1}{2}\boldsymbol{\xi}^{T}\mathbf{Q}\boldsymbol{\xi}+\mathbf{q}^T\boldsymbol{\xi} \label{ip-11} \\
 &~~\text{s.t.: } \mathbf{g}(\boldsymbol{\xi})- \mathbf{s}=\mathbf{0} \label{ip_12} \\
 & ~~~~~~~~\mathbf{s}\geq \mathbf{0}\label{ip_20}
\end{align}

\noindent Then, the slack variable is transferred into the cost function using the log-barrier method \cite{guler1996barrier}. Thus, the optimization problem can be written as  

\begin{align}
&\min_{\boldsymbol{\xi}} \frac{1}{2}\boldsymbol{\xi}^{T}\mathbf{Q}\boldsymbol{\xi}+\mathbf{q}^T\boldsymbol{\xi} - \mu\sum_{j}\log s_{j} \label{ip_1}\\
 &\text{s.t.: } \mathbf{g}(\boldsymbol{\xi})- \mathbf{s}=\mathbf{0} \label{ip_2}
\end{align}

\noindent The initial step in solving the barrier problem involves expressing the \gls{kkt} conditions \cite{qi1997semismooth}. By defining a new variable $z=\frac{\mu}{s_{j}}$, the \gls{kkt} conditions can be written as:

\begin{align}
&\nabla_{\xi}L\hspace{-0.05cm}= \hspace{-0.05cm}\mathbf{0}, ~L\hspace{-0.05cm}=\hspace{-0.05cm}\frac{1}{2}\boldsymbol{\xi}^{T}\mathbf{Q}\boldsymbol{\xi}+\mathbf{q}^T\boldsymbol{\xi}\hspace{-0.05cm}+\hspace{-0.05cm}\boldsymbol{\lambda}^{T}_{in}(\mathbf{g}(\boldsymbol{\xi})-\mathbf{s})-\hspace{-0.06cm}\mathbf{z}^{T}\mathbf{s}
\label{ip_4} \\
&\nabla_{s} L = \mathbf{0}
\label{ip_5}\\
&\mathbf{g}(\boldsymbol{\xi})-\mathbf{s}= \mathbf{0} \label{ip_7}\\
&\mathbf{Z}\mathbf{S}-\mu\mathbf{e} =\mathbf{0}, ~~ \mathbf{Z} = \mathbf{diag}\mathbf{z}, \mathbf{S} = \mathbf{diag}s_{j},\mathbf{e}=\begin{bmatrix}
    1\\\vdots \\1 
\end{bmatrix}\label{ip_8}
\end{align}

\noindent The above \eqref{ip_4}-\eqref{ip_8}
are nonlinear and pose a significant challenge when attempting a direct solution. Therefore, in the context of the interior point method, authors in \cite{roos2005interior,wright2005interior,renegar2001mathematical} opt to linearize and approximate the solution. This is achieved by considering the update directions as $\begin{bmatrix}
    \Delta\boldsymbol{\xi}&\Delta\mathbf{s}& \Delta\boldsymbol{\lambda}_{in} & \Delta\mathbf{z}
\end{bmatrix}^{T}$.  Furthermore, the left side of \eqref{ip_4}-\eqref{ip_8} is considered as the residuals of the current states. Then, using the first-order approximation for the $k^{th}$ iteration, the following linear system is obtained. 

\begin{align}
    \begin{bmatrix}
    ^{k}\nabla_{\xi\xi}^{2}L & \mathbf{0}  & ^{k}\nabla_{\mathbf{g}}^{T}& \mathbf{0} \\
        \mathbf{0} & \mathbf{0} &  \mathbf{-I} & \mathbf{-I} \\
    ^{k}\nabla_{\mathbf{g}} & \mathbf{-I} & \mathbf{0} & \mathbf{0} \\
\mathbf{0}&^{k}\mathbf{Z}&\mathbf{0}&^{k}\mathbf{S}
    \end{bmatrix}\begin{bmatrix}
        \Delta\boldsymbol{\xi} \\\Delta\mathbf{s}\\\Delta\boldsymbol{\lambda}_{in}\\\Delta\mathbf{z}
    \end{bmatrix}=\begin{bmatrix}
        ^{k}\nabla_{\xi}L \\^{k}\nabla_{s}L\\\mathbf{g}(^{k}\boldsymbol{\xi})-^{k}\mathbf{s}\\^{k}\mathbf{Z}^{k}\mathbf{S}-\mu\mathbf{e}
    \end{bmatrix} \label{ip_9}
\end{align}

\noindent The update directions can be computed at this point. The current iterate can then be updated iteratively to derive the solution. It is also worth noting that \eqref{ip_9} follows a linear format similar to Figure \ref{interior}, where the left side represents the matrix $^{k}\mathbf{D}$, and the right side represents the vector $^{k}\boldsymbol{\chi}$.

\vspace{0.45cm}
\noindent \textbf{Interior Point method limitations:} As it can be seen, the computation cost of this linear equation depends on computing the right and left side of \eqref{ip_9}, $^{k}\mathbf{D}$, and $^{k}\boldsymbol{\chi}$ respectively, at each iteration. Furthermore, for each iteration, it is necessary to calculate the inverse of matrix $^{k}\mathbf{D}$, which is computationally demanding. It should be mentioned that as the number of constraints increases, the size of this matrix expands, further amplifying the computational complexity. For example, a highly cluttered environment will lead to a large number of collision avoidance constraints and may render the interior-point-based approach too slow for real-time applications \cite{roos2005interior}. Nevertheless, these classes of optimizers are extremely popular and have been packaged in the form of some easy-to-use libraries. 

\vspace{0.45cm}
\noindent\textbf{Existing works:} Some of the studies that are built on top of the interior-point-based approach are as follows.

\subsubsection{Interior Point OPTimizer(IPOPT)} IPOPT is an open-source software package designed to address large-scale nonlinear optimization problems. It employs a variant of the interior point method, customized for nonlinear optimization tasks \cite{wachter2009short, wachter2006implementation, biegler2009large, ma2017trajectory, andrei2017interior}. Like the interior point method, the effectiveness of IPOPT's solution is influenced by the initial guess provided by the user. Furthermore, it may encounter computational challenges when applied to large-scale problems with numerous constraints and variables \cite{liu2022real}.

\subsubsection{ROCKIT}
Rockit is a software framework for optimal control. It utilizes various Non-Linear Programming solvers, including IPOPT, to solve optimization problems through the implementation of a primal-dual interior point method. When a problem is defined in ROCKIT using CasADi's \cite{andersson2019casadi} symbolic representation, IPOPT is utilized as the solver to determine the optimal solution. This process involves iteratively refining an estimate of the solution while considering the problem's constraints and the objective function \cite{metz2003rockit, bos2022multi,mrak2023model,dirckx2022smooth}. 

\subsubsection{FATROP} FATROP is a constrained nonlinear optimal control problem solver that solves the optimization problem through the dual-primal Interior-Point method \cite{vanroye2023fatrop}.

\subsection{Convex-Concave Procedure (CCP)}

\begin{figure}[!h]
    \centering
\includegraphics[scale=0.5]{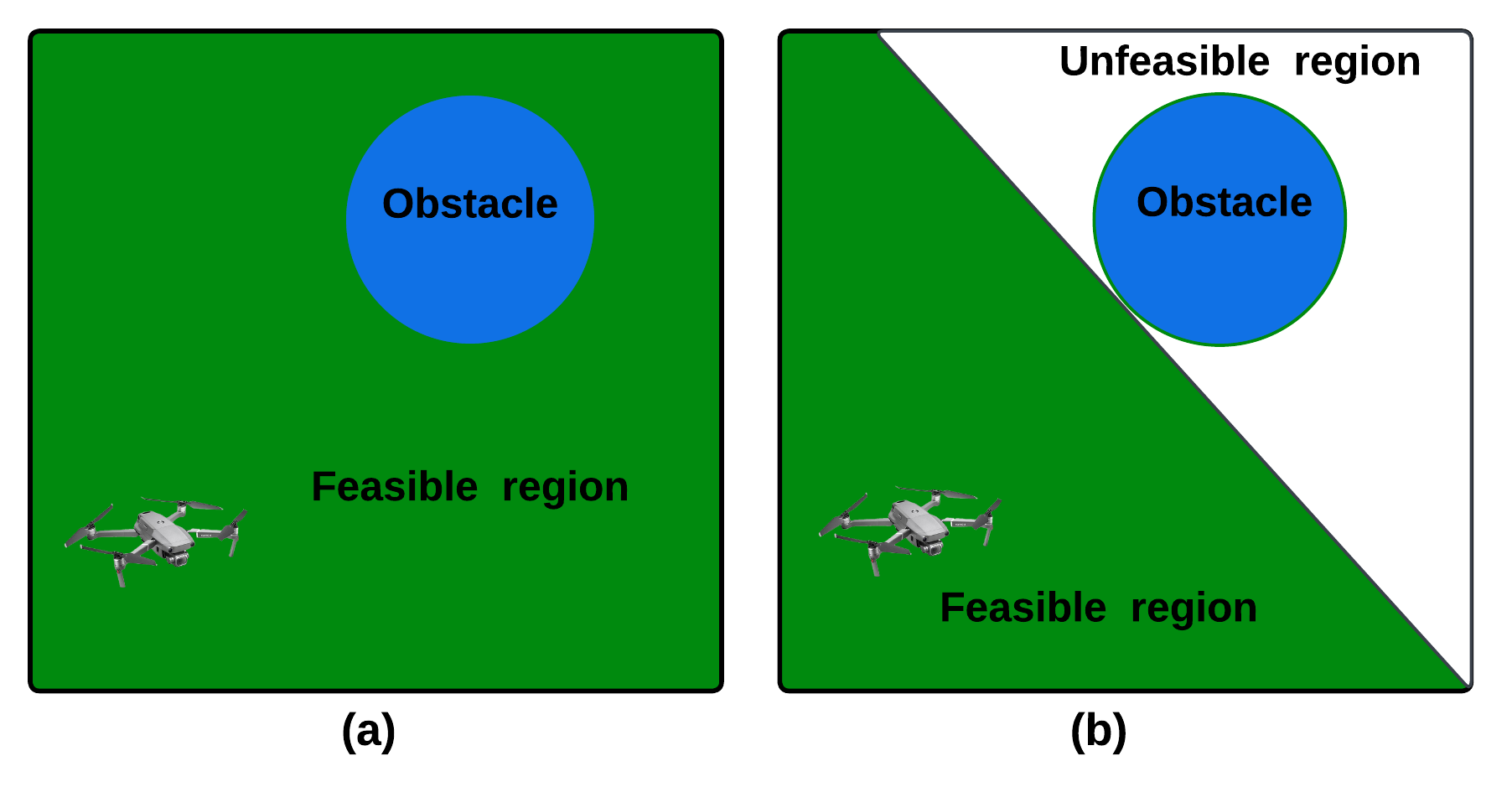}
    \vspace{-0.3cm}
    \caption[Affine approximation of collision-avoidance constraints]{\small{(a)Feasible region (in green) in general, (b) feasible region (in green) using affine approximation of collision-avoidance constraints  }}
    \label{affine_approximation}
\end{figure}

The collision avoidance constraints, denoted as \eqref{g_3}, possess a unique characteristic: they are purely concave. This means that their affine approximation can serve as a global conservative upper bound for the original quadratic constraints. This structure has led to the development of a set of methods known as the Convex-Concave Procedure, or \gls{ccp}, which is used to solve optimization problems \cite{chen2022convex, lipp2016variations,soria2021distributed,shen2016disciplined, lu2021convex, gao2017quadrotor,rey2018fully}. \gls{ccp} simplifies these problems by using the affine approximations of the concave parts (as shown in Figure \ref{affine_approximation}).  By iteratively alternating between solving simplified convex problems and refining the approximation of concave parts, \gls{ccp} gradually converges to the locally optimal solution \cite{gao2017quadrotor,rey2018fully}. 

\noindent\textbf{How does \gls{ccp} method solve the trajectory optimization problem \eqref{p_1}-\eqref{eq_1}?} \gls{ccp} is an iterative process, where at each iteration, we solve a convex approximation of the original problem \cite{lipp2016variations}. To solve the optimization problem using \eqref{p_1}-\eqref{eq_1}, at iteration $k$, the inequality constraints $\mathbf{g}(\boldsymbol{\xi})$ 
 is linearized using the first-order Taylor expansion as:

\begin{align}
   \mathbf{g}(\boldsymbol{\xi}) \approx \nabla\mathbf{g}(^{k}\boldsymbol{\xi})^{T}(\boldsymbol{\xi}-^{k}\boldsymbol{\xi})+ \mathbf{g}(^{k}\boldsymbol{\xi})  = \hspace{0.1cm}^{k}\mathbf{A} \boldsymbol{\xi}- \hspace{0.1cm}^{k}\mathbf{b}\label{ccp_1}
\end{align}

\noindent Typically, the affine approximation of \eqref{ccp_1} are more conservative than the original constraints (see Figure \ref{affine_approximation}). In other words, optimization with \eqref{ccp_1} can be infeasible, even though the original problem might have a solution. To counter such cases, it is common practice to introduce slack variables $\mathbf{s}$ \cite{lipp2016variations}. The final convex approximation that \gls{ccp} solves at iteration $k$ can be defined in the following manner.

\begin{align}
 &\min_{\boldsymbol{\xi}, \mathbf{s}} \frac{1}{2}\boldsymbol{\xi}^{T}\mathbf{Q}\boldsymbol{\xi}+\mathbf{q}^T\boldsymbol{\xi} + \boldsymbol{\mu}^{T}\mathbf{s} \label{sqp_1}  \\
&~~ \text{s.t.: } ^{k}\mathbf{A}\boldsymbol{\xi}-^{k}\mathbf{b} - \mathbf{s} \leq \mathbf{0} \label{sqp_2}\\
 &~~~~~~~~~\mathbf{s}  \geq \mathbf{0} \label{sqp_3}
\end{align}

\begin{figure}[!h]
    \centering
\includegraphics[scale=0.28]{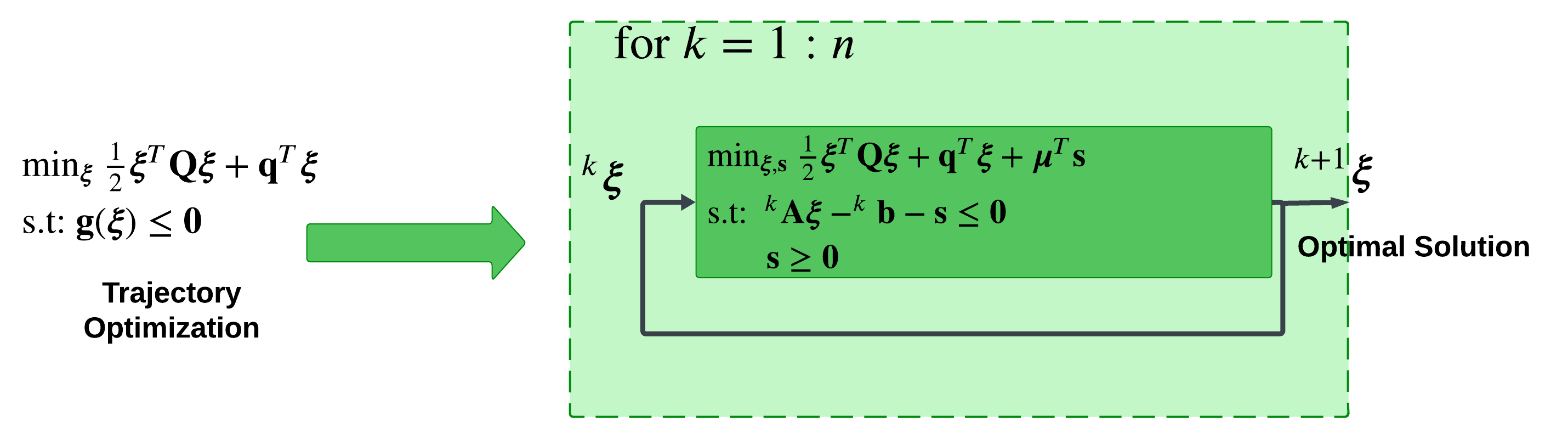}
    \caption[Convex-Concave Procedure ]{\hspace{-0.13cm}\small{Trajectory optimization problem is reduced to solving \gls{qp} problem.}}
    \label{sqp_d}
\end{figure}

\noindent As can be seen, for our specific trajectory optimization problem, the \gls{ccp} approximation \eqref{sqp_1}-\eqref{sqp_3} has essentially reduced to a Quadratic Program (QP). Figure \ref{sqp_d} shows a graphical representation of the \gls{ccp} process.


\vspace{0.45cm}
\noindent\textbf{CCP method limitations:} While \gls{ccp} has proven effective, it faces critical limitations, particularly in cluttered and dynamic environments. The first limitation stems from the necessity of providing a collision-free initial trajectory guess, a condition challenging to meet in cluttered and dynamic scenarios. The addition of slack variables solves this problem at the cost of increased computation time \cite{lipp2016variations}. The second limitation arises from the need to solve a constrained optimization problem at each iteration, making real-time implementation impractical for highly cluttered environments. Finally, the affine approximation is conservative and removes a large part of obstacle-free space from the feasible region of the optimization \cite{lipp2016variations}.

\vspace{0.45cm}
\noindent \textbf{Existing works:} \gls{ccp} is extensively used for trajectory optimization in robotics, often by a different name \gls{scp}. For example, \cite{chen2015decoupled} uses \gls{scp} in their work on decoupled multiagent path planning via incremental sequential convex programming. Similarly, \cite{augugliaro2012generation} applies \gls{scp} for the generation of collision-free trajectories for a quadrocopter fleet. Moreover, \cite{virgili2018recursively} employs this method in a recursively feasible and convergent sequential convex programming procedure to solve non-convex problems with linear equality constraints. These references highlight the effectiveness of \gls{ccp}/\gls{scp} in addressing complex optimization problems in robotics. However, as mentioned before, the existing approaches of \gls{ccp}/\gls{scp} do not show good performance in highly-cluttered or dynamic environments. For example, the approach of \cite{augugliaro2012generation} is restricted to a very small swarm size. In contrast, the optimizer introduced in this thesis is much more scalable for larger swarms and can also allow for navigation over LiDAR point clouds by treating them as point obstacles. Such results are not possible with \gls{ccp}/\gls{scp}.

Additionally, it is worth noting that the \gls{scp} methods inherit the limitations of \gls{ccp} method such as the potential for local optima and the requirement for convexity in the problem formulation.



\subsection{Sampling-based Optimizers}
Sampling-based optimizers function by iteratively sampling in the space of trajectories (or control inputs) to produce potential solutions and refining them across multiple iterations \cite{bharadhwaj2020model, zhang2022simple, jankowski2023vp,pinneri2021sample,huang2021cem,mottahedi2022constrained,homem2014monte}. Each iteration involves randomly selecting points or configurations within the search space and evaluating their performance based on a specified objective function. Through this iterative process of sampling and refinement, the optimizer endeavors to converge towards a solution. 

In the following, I will review some \gls{sota} sampling-based optimizers used for comparisons in this thesis. 

\subsubsection{Cross-Entropy Method (CEM)}
One of the common sampling-based methods is the Cross-Entropy Method \gls{cem} \cite{bharadhwaj2020model, zhang2022simple,pinneri2021sample,stulp2012path}. This method is used to tackle optimization problems, especially in scenarios where conventional deterministic approaches encounter challenges, such as high-dimensional or non-convex optimization problems. To understand how the \gls{cem} method solves our trajectory optimization problem  \eqref{g_1}-\eqref{g_3}, we follow the following steps  

\begin{enumerate}
    \item \textbf{Initialization}: Initialize the number of samples , $N_{b}$, and distribution parameters including mean, $^{l}\boldsymbol{\mu}$, and covariance $^{l}\boldsymbol{\Sigma}$ at $l=1$.
    \item \textbf{Sample Generation}: Generate $N_{b}$ samples $\boldsymbol{\xi}_{1},...,\boldsymbol{\xi}_{N_{b}}$ from Gaussian distribution $\mathcal{N}(^{l}\boldsymbol{\mu},^{l}\boldsymbol{\Sigma} )$, where $\mathcal{N}$ is a normal distribution with a specific mean and a standard deviation that characterizes the spread of the distribution.
    \item \textbf{Evaluation:} Evaluate each sampled solution by computing the objective function and checking whether it satisfies the inequality constraint. We utilize linear penalty to evaluate samples. Thus, the evaluation can be obtained through computing. 
    
    \begin{align}
    c_{ev,i} = \frac{1}{2}\boldsymbol{\xi}^{T}_{i}\mathbf{Q}\boldsymbol{\xi}_{i}+\mathbf{q}^T\boldsymbol{\xi}_{i}+\sum_{j}\max(\mathbf{0},\mathbf{g}_{j}(\boldsymbol{\xi}_{i})),~~~i =1,..., N_{b}
    \end{align}
    \item \textbf{Selection:} Choose top $N_{elite}$ samples from $c_{ev,i}$.
    \item \textbf{Parameter Update:} Update the mean, $^{l+1}\boldsymbol{\mu}$, and covariance of the probability distribution, $^{l+1}\boldsymbol{\Sigma}$, using 
    
        \begin{align}
         &^{l+1}\boldsymbol{\mu} = \frac{1}{N_{elite}}
         \sum\limits_{m\in \mathcal{C} } \boldsymbol{\xi}_{m}, \label{update_mean_cem} \\
        &^{l+1}\boldsymbol{\Sigma}=\frac{1}{N_{elite}}\sum_{m\in \mathcal{C}}(\boldsymbol{\xi}_{m}-^{l+1}\boldsymbol{\mu})(\boldsymbol{\xi}_{m}-^{l+1}\boldsymbol{\mu})^{T},
         \label{update_covariance_cem}
        \end{align}
    \noindent where set $\mathcal{C}$ consists of the top $N_{elite}$ samples. 
    \item \textbf{Termination Criterion:} Repeat the iteration process until a termination criterion, reaching a maximum number of iterations, is met.
\end{enumerate}

\vspace{0.45cm}
\noindent\textbf{\gls{cem} limitations:}
There are two main issues with \gls{cem} methods. Firstly, \gls{cem} requires considering a large number of samples to ensure finding an optimal solution \cite{botev2013cross}. This consideration makes \gls{cem} computationally heavy for large-scale problems with a large number of variables. Secondly, the performance of \gls{cem} heavily depends on the initial distribution. If all the samples fall into infeasible regions, \gls{cem} may not be able to find a feasible solution \cite{botev2013cross}.

\subsubsection{Covariance Matrix Adaptation Evolution Strategy (CMA-ES)}
By just changing the distribution update rules \eqref{update_mean_cem}-\eqref{update_covariance_cem}, it is possible to obtain different variants of sampling-based optimizers. One such method is \gls{cmeas} \cite{auger2012tutorial}. Its overall process is the same as \gls{cem}. In the context of this thesis, it begins by creating a set of potential solutions, which in this case are various possible robot trajectories. These trajectories are assessed using an objective function that considers factors such as distance to the target, trajectory smoothness and avoidance of collisions. The top-performing trajectories are then selected to generate a new set of potential trajectories for the next iteration. This is achieved by sampling from a multivariate normal distribution, with the mean and covariance matrix of the distribution updated based on the successful trajectories from the previous iteration. This procedure is repeated until a satisfactory trajectory is identified \cite{nishida2018psa}.

One of the recent works that uses \gls{cmeas} method is \gls{vpsto}. This method uses \gls{cmeas} to optimize the trajectories, defining velocity and acceleration limits and internally constraining the solution to those. \gls{vpsto} is one of the recent \gls{cmeas}-based methods designed to optimize robot behavior in complex dynamic environments \cite{jankowski2023vp}.

\vspace{0.45cm}
\noindent \textbf{\gls{cmeas} limitations:} 
One of the main \gls{cmeas} limitations is its dependency on the quality of chosen features or the underlying parametric function space. The quality of the solution depends heavily on the selection of these parameters \cite{auger2012tutorial}. For example, in \gls{vpsto} these parameters include the number of via-points, the selection of via-points and trade-off weights in the cost function. Another limitation is the high computational time. \gls{cmeas} uses high-dimensional trajectory representations, which can be computationally expensive and inefficient, limiting the speed at which the system can react to changes in the environment. Also, it can suffer from local optima, where it may get stuck in a suboptimal solution \cite{nishida2018psa}. 

    \chapter{Paper I: A Novel Trajectory Optimization Algorithm}\label{papper_1}

\section{Overview of the Main Algorithmic Results}
In this chapter, a novel algorithm \cite{rastgar2020novel} for solving the optimization problem \eqref{g_1}-\eqref{g_3} is introduced. At a broad level, the main features of the proposed approach can be described as follows:

\begin{figure}[h]
    \centering
\includegraphics[scale=0.3]{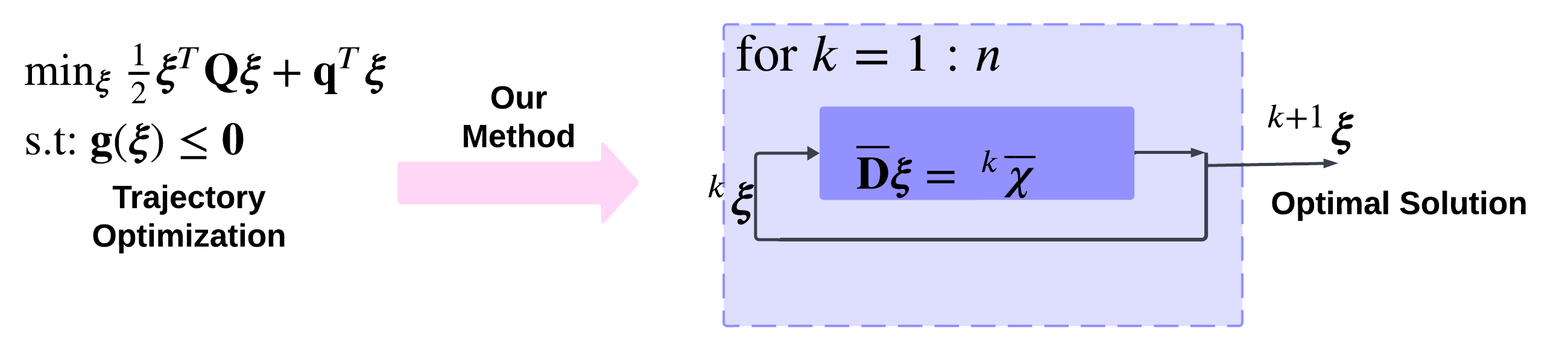}
    \caption[Our optimizer schematic]{\small{The trajectory optimization problem can be reduced to solving a system of linear equations where $\overline{\mathbf{D}}$ and $^{k}\overline{\boldsymbol{\chi}}$ represent a fixed matrix and changing vector during different iterations, respectively}}
    \label{paper_1}
\end{figure}

\begin{itemize}
    \item I show that the solving \eqref{g_1}-\eqref{g_3} can be reduced to solving a system of linear equations \eqref{lin_fixed_sol} with matrix $\overline{\mathbf{D}}$ being fixed across all iterations and vector $ \overline{\boldsymbol{\chi}}$ (see Figure \ref{paper_1}). Later, it will be explained how $\overline{\mathbf{D}}$ and $ \overline{\boldsymbol{\chi}}$ are derived latter.
    
    \begin{align}
\overline{\mathbf{D}}\boldsymbol{\xi}= \hspace{0.1cm}^{k}\overline{\boldsymbol{\chi}}    
   \label{lin_fixed_sol}
    \end{align}
    
   Since the matrix $\overline{\mathbf{D}}$ is fixed across all iterations  \eqref{lin_fixed_sol}, I can show that: 
    
    \begin{enumerate}
        \item The factorization/inverse of $\overline{\mathbf{D}}$ can be computed once and used across all iterations. 
        \item The size of matrix $\overline{\mathbf{D}}$ does not change with the number of constraints and only depends on the planning horizon. Thus, an increase in the number of obstacles does not affect the computation cost for factorization of $\overline{\mathbf{D}}$.
        \item I further show that $\overline{\mathbf{D}}$ has block-diagonal structure. Thus, the computation along each motion axis can be decoupled in the following manner:
        
        \begin{align}
            \overline{\mathbf{D}}_x \boldsymbol{\xi}_x = {^k}\overline{\boldsymbol{\chi}}_x, \qquad  \overline{\mathbf{D}}_y \boldsymbol{\xi}_y = {^k}\overline{\boldsymbol{\chi}}_y,  \qquad \overline{\mathbf{D}}_z \boldsymbol{\xi}_z = {^k}\overline{\boldsymbol{\chi}}_z \label{simple}
        \end{align}
    \end{enumerate}

  \item To obtain the computational structure of the form  \eqref{lin_fixed_sol}, I present a novel reformulation of the quadratic collision avoidance constraints. I show that this new reformulation has a multi-convex structure that can be leveraged through mathematical concepts such as \gls{am} and the augmented Lagrangian method.
\end{itemize}

 In the next sections, I will outline the advantages of the proposed work over \gls{sota} method and explain the main results in detail.

\vspace{0.2cm}
\section{Advantages of the proposed Approach Over SOTA} 

\begin{itemize}
\item \textbf{Efficient Computational Complexity:} The per-iteration computational complexity of the proposed optimizer is significantly lower than \gls{sota} approaches like \gls{ccp} \cite{lipp2016variations} (refer to Figure\ref{comptime_scaling}). I show while the solution quality of our optimizer is competitive with \gls{ccp}, it can be several orders of magnitude faster.    
\item \textbf{Lower Computation time:} The proposed optimizer offers the possibility of caching the matrix factorization part and thus reducing the entire computation to computing matrix-vector products or evaluating some symbolic expressions.
\end{itemize}



\section{Main Algorithmic Results} \label{main_result_1}

In this section, I present the main theoretical details of the proposed optimizer. The discussion is initiated by providing the context and rationale behind the proposed novel collision avoidance model. Following this, I delve into the details of how this model is leveraged within the optimization problem.

\begin{figure}[h]
    \centering
    \includegraphics[scale=0.98]{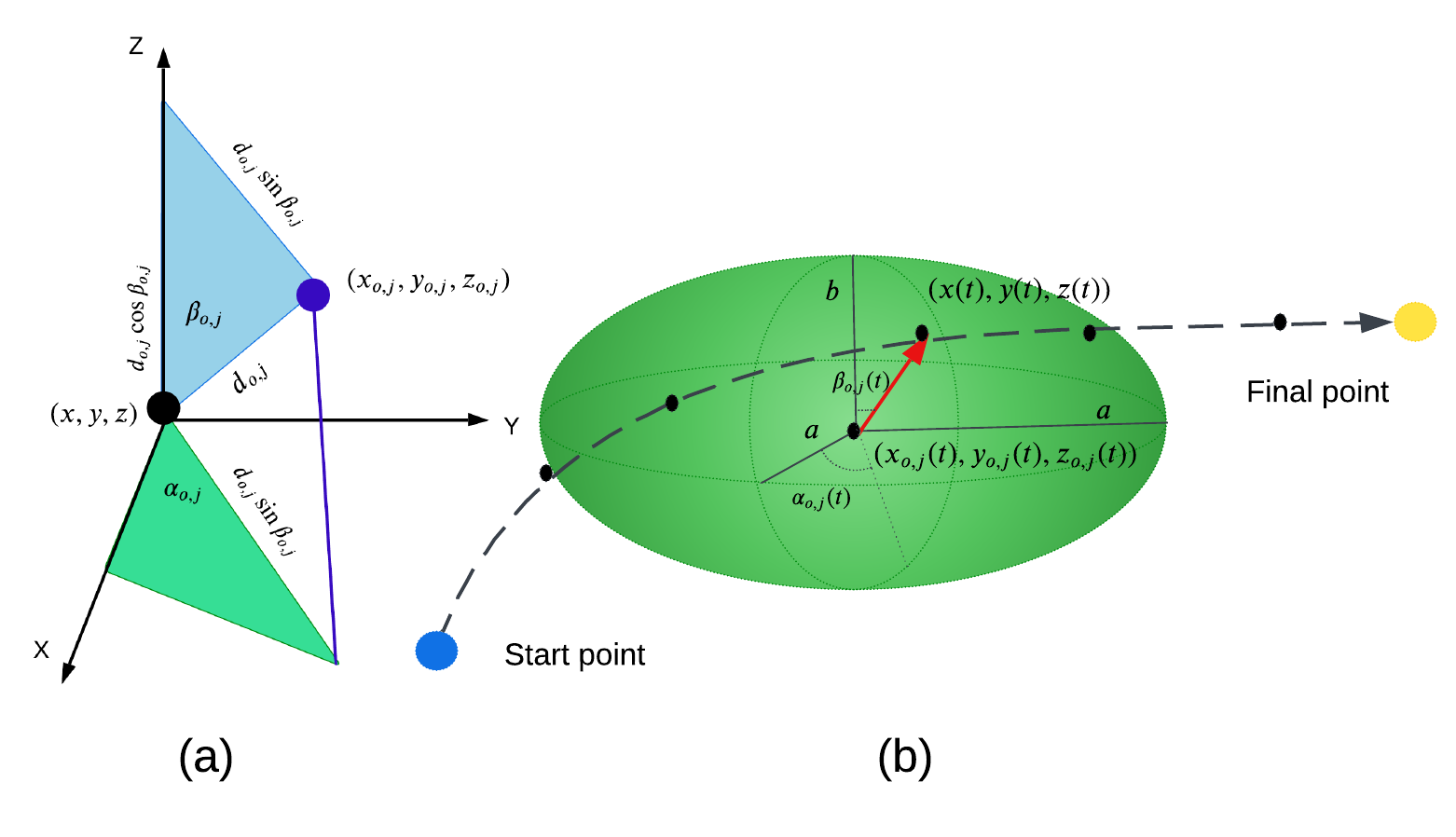}
    \vspace{-0.6cm}
    \caption[Intuition behind our collision avoidance model]{\small{(a) The polar/spherical relationship between the positions of the robot and the obstacle can be derived through trigonometry. The vector $d_{o,j}(t)$ represents the line-of-sight distance, representing the distance between the centers of the robot and the obstacle. The polar angle, $\beta_{o,j}(t)$, signifies the angle that $d_{o,j}(t)$ makes with the z-axis. Additionally, the azimuth angle, $\alpha_{o,j}(t)$, serves as the normal polar/spherical coordinate in the x-y plane. Together, these parameters provide a comprehensive polar/spherical representation capturing the geometric relationship between the robot and obstacle positions (b) Intuition behind the proposed collision avoidance model: the point $\hspace{-0.1cm}(x(\hspace{-0.02cm}t\hspace{-0.02cm}),y(\hspace{-0.02cm}t\hspace{-0.02cm}),z(\hspace{-0.02cm}t\hspace{-0.02cm})\hspace{-0.02cm})$ that is in collision with the obstacle needs to be pushed away from the center of the obstacle along the directions $\alpha_{o, j}(t), \beta_{o, j}(t)$.}
}
\vspace{-0.3cm}
    \label{polar}
\end{figure}

\vspace{0.45cm}
\noindent\textbf{Reformulating collision avoidance constraint:} I adopt polar/ spherical  representation to reformulate the collision avoidance constraints \eqref{g_3} as:

\small
\begin{subequations}
\begin{align}
&\colorbox{my_green}{$x(t)-x_{o,j}(t)\hspace{-0.1cm}- ad_{o,j}(t)\hspace{-0.02cm}\cos{\alpha_{o,j}}(t)\hspace{-0.01cm}\sin{\beta_{o,j}}(t)=0$} \label{collision_x}\\
&\colorbox{my_pink}{$y(t)-y_{o,j}(t) 
-\hspace{-0.08cm}ad_{o,j}(t)\sin{\alpha_{o,j}(t)}\sin{\beta_{o,j}}(t)=0$}\\
&\colorbox{my_blue}{$z(t)  -z_{o,j}(t) -bd_{o,j}(t)\cos{\beta_{o,j}(t)}=0$}
\\
&d_{o,j}(t) \geq 1, \forall t, 
\label{collision_d} 
\end{align}
\end{subequations}
\normalsize

\noindent where $d_{o,j}(t)$ is the distance between the robot center and $j^{th}$ obstacle center. Also, $\alpha_{o,j}(t)$ and $\beta_{o,j}(t)$ are angles between the robot and obstacle center (see Figure \ref{polar}(a)). These variables, derived from the polar/spherical representation, play a key role in guiding the optimization problem to avoid collisions effectively. To understand it better, consider a scenario where a point $(x(t), y(t), z(t))$ resides within an obstacle (Figure \ref{polar}(b)). The collision can be avoided if we push away the considered point from the center of the obstacle, $(x_{o, j}(t), y_{o, j}(t), z_{o, j}(t))$ along the directions of $\alpha_{o,j}(t)$ and $\beta_{o,j}(t)$. The parameter $d_{o,j}(t)$ tells us how much the point needs to move away from the obstacle's center. We note that $d_{o,j}(t)$ has an analytical form as \eqref{d_max}. 

\small
\begin{align}
    d_{o,j}(t) = \max(1, \sqrt{\frac{(x(t)-x_{o,j}(t))^{2}}{a^2}+ \frac{(y(t)-y_{o,j}(t))^{2}}{a^2}+ \frac{(z(t)-z_{o,j}(t))^{2}}{b^2}}). \label{d_max}
\end{align}
\normalsize
\vspace{-0.2cm}

\begin{remark}
    For clarity and to facilitate the tracking of changes in the optimization problem, each term of the optimization problem is highlighted with a specific color.
\end{remark}


\vspace{0.45cm}
\noindent \textbf{Reformulating Trajectory Optimization Problem:} By considering the new formulation of collision avoidance constraints \eqref{collision_x}-\eqref{collision_d}, the original optimization problem \eqref{g_1}-\eqref{g_3} can be rephrased in the following manner.

\small
\begin{subequations}
    \begin{align} 
    &\min_{\scalebox{0.7}{$\begin{matrix} x_(t), y(t),z(t),c_{\alpha,j}(t),\\s_{\alpha,j}(t),c_{\beta,j}(t),
 s_{\beta,j}(t),\\ d_{o,j}(t),\alpha_{o,j}(t),\beta_{o,j}(t)   \end{matrix}$}}c_{x}(x^{(q)}(t))+c_{y}(y^{(q)}(t))+ c_{z}(z^{(q)}(t)) \label{cost_2}\\
 &\text{s.t:} \nonumber \\
  &\colorbox{my_orange}{$   c_{\alpha,j}(t) = \cos{\alpha_{o,j}(t)}$},~  \colorbox{my_yellow}{$
     s_{\alpha,j}(t) = \sin{\alpha_{o,j}(t)} $}\label{new_var_1}\\
   &\colorbox{my_purple}{$  c_{\beta,j}(t) = \cos{\beta_{o,j}(t)}$}, \colorbox{my_grey}{$
     s_{\beta,j}(t) = \sin{\beta_{o,j}(t)} $}\label{new_var_2}\\
    &d_{o,j}(t) \geq 1\\
&\colorbox{my_green}{$x(t)-x_{o,j}(t)- ad_{o,j}c_{\alpha,j}(t)s_{\beta,j}(t)=0$} \label{eq1} \\
&\colorbox{my_pink}{$y(t)-y_{o,j}(t)
-ad_{o,j}s_{\alpha,j}(t)s_{\beta,j}(t)=0 $} \label{eq2}\\
&\colorbox{my_blue}{$z(t)-z_{o,j}(t) -bd_{o,j}c_{\beta,j}(t)=0,$} 
\label{collision_proposed_1} 
    \end{align}
\end{subequations}
\normalsize

\noindent where $c_{x}(x^{(q)}(t))$, $c_{y}(y^{(q)}(t))$, and $c_{z}(z^{(q)}(t))$ represent the quadratic cost functions along each motion axis. I have introduced additional variables $c_{\alpha,j}, s_{\alpha,j},c_{\beta,j}$ and $s_{\beta,j}$. These new variables act as a copy of sine and cosine of angles, $\alpha_{o,j}(t) $ and $ \beta_{o,j}(t) $ in the collision avoidance constraints \eqref{collision_x}-\eqref{collision_d}. Moreover, new equality constraints \eqref{new_var_1}-\eqref{new_var_2} have introduced to maintain the relationship between $c_{\alpha,j},s_{\alpha,j},c_{\beta,j},s_{\beta,j}$ and $\cos{\alpha_{o,j}(t)}, \sin{\alpha_{o,j}(t)},
\cos{\beta{o,j}(t)},
\sin{\beta{o,j}(t)}$, respectively.   

I now relax all the equality constraints in optimization \eqref{new_var_1}-\eqref{collision_proposed_1} using augmented Lagrangian method as

\begin{align}
&\mathcal{L}\Big(x(t),y(t),z(t), c_{\alpha,j}(t),s_{\alpha,j}(t),c_{\beta,j}(t),
 s_{\beta,j}(t), d_{o,j}(t),\alpha_{o,j}(t),\beta_{o,j}(t)\Big ) \nonumber \\ &=  
     c_{x}(x^{(q)}(t))+c_{y}(y^{(q)}(t))+c_{z}(z^{(q)}(t)) \nonumber \\
&~~+\sum_{t=0}^{t=n_{p}}\sum_{j=1}^{j=n_{o}} \Big( \lambda_{x,j}(t)(\colorbox{my_green}{$x(t)-x_{o,j}(t)- ad_{o,j}c_{\alpha,j}(t)s_{\beta,j}(t)$}) \nonumber \\&~~+ \frac{\rho_{o}}{2}(\colorbox{my_green}{$x(t)-x_{o,j}(t)- ad_{o,j}c_{\alpha,j}(t)s_{\beta,j}(t)$})^{2} \Big)\nonumber \\
&~~+\sum_{t=0}^{t=n_{p}}\hspace{0.1cm}\sum_{j=1}^{j=n_{o}}\Big(\lambda_{y,j}(t)(\colorbox{my_pink}{$y(t)-y_{o,j}(t)
-ad_{o,j}s_{\alpha,j}(t)s_{\beta,j}(t) $}) \nonumber \\
&~~+ \frac{\rho_{o}}{2}(\colorbox{my_pink}{$y(t)-y_{o,j}(t)
-ad_{o,j}s_{\alpha,j}(t)s_{\beta,j}(t) $}) \Big)
\nonumber \\
&~~+\sum_{t=0}^{t=n_{p}}\sum_{j=1}^{j=n_{o}}\Big(\lambda_{z,j}(t)(\colorbox{my_blue}{$z(t)-z_{o,j}(t) -bd_{o,j}c_{\beta,j}(t)$} ) \nonumber \\
&~~+ \frac{\rho_{o}}{2}(\colorbox{my_blue}{$z(t)-z_{o,j}(t) -bd_{o,j}c_{\beta,j}(t)$} )\Big) \nonumber \\
&~~+\hspace{-0.22cm}\sum_{t=0}^{t=n_{p}}\hspace{-0.1cm}\sum_{j=1}^{j=n_{o}}\Big(\hspace{-0.06cm}\frac{\rho}{2}(\colorbox{my_orange}{$\hspace{-0.05cm}c_{\alpha,j}\hspace{-0.04cm}(\hspace{-0.03cm}t\hspace{-0.03cm})\hspace{-0.1cm}- \hspace{-0.07cm}\cos{\alpha_{o,j}(\hspace{-0.03cm}t\hspace{-0.03cm})}$}\hspace{-0.2cm}+\hspace{-0.1cm}\frac{\lambda_{c_{\alpha,j}}(\hspace{-0.03cm}t\hspace{-0.03cm})}{\rho}\hspace{-0.07cm})^{2}\hspace{-0.2cm}+\hspace{-0.12cm}\frac{\rho}{2}(\colorbox{my_yellow}{$s_{\alpha,j}(\hspace{-0.03cm}t\hspace{-0.03cm})\hspace{-0.1cm}- \hspace{-0.05cm}\sin{\alpha_{o,j}(\hspace{-0.03cm}t\hspace{-0.03cm})}$}\hspace{-0.2cm}+\hspace{-0.1cm}\frac{\lambda_{s_{\alpha,j}}\hspace{-0.04cm}(\hspace{-0.03cm}t\hspace{-0.03cm})}{\rho}\hspace{-0.07cm})^{2}\hspace{-0.1cm}\Big) \nonumber 
\\
&~~+\hspace{-0.22cm}\sum_{t=0}^{t=n_{p}}\hspace{-0.1cm}\sum_{j=1}^{j=n_{o}}\Big(\hspace{-0.06cm}
\frac{\rho}{2}(\colorbox{my_purple}{$\hspace{-0.05cm}c_{\beta,j}\hspace{-0.04cm}(\hspace{-0.03cm}t\hspace{-0.03cm})\hspace{-0.1cm}- \hspace{-0.07cm}\cos{\beta_{o,j}(\hspace{-0.03cm}t\hspace{-0.03cm})}$}\hspace{-0.2cm}+ \hspace{-0.12cm}\frac{\lambda_{c_{\beta,j}}(\hspace{-0.03cm}t\hspace{-0.03cm})}{\rho})^{2}\hspace{-0.2cm}+\hspace{-0.12cm}\frac{\rho}{2}(\colorbox{my_grey}{$s_{\beta,j}(\hspace{-0.03cm}t\hspace{-0.03cm})\hspace{-0.1cm}-\hspace{-0.05cm} \sin{\beta_{o,j}(\hspace{-0.03cm}t\hspace{-0.03cm})}$}\hspace{-0.2cm}+\hspace{-0.1cm}\frac{\lambda_{s_{\beta,j}}\hspace{-0.04cm}(\hspace{-0.03cm}t\hspace{-0.03cm})}{\rho})^{2}\hspace{-0.1cm}\Big) 
\label{aug_2}
\end{align}

\noindent Thus, the trajectory optimization can be written as 

\vspace{-0.4cm}
\small
\begin{align}
&\min_{\hspace{1cm}\scalebox{0.7}{$\begin{matrix} x(t), y(t),z(t),\\c_{\alpha,j}(t),s_{\alpha,j}(t),\\c_{\beta,j}(t),
 s_{\beta,j}(t), \\d_{o,j}(t),\alpha_{o,j}\\(t),\beta_{o,j}(t)   \end{matrix}$}} \hspace{-0.4cm}\mathcal{L}\Big(x(t),y(t),z(t), c_{\alpha,j}(t),s_{\alpha,j}(t),c_{\beta,j}(t),
 s_{\beta,j}(t), d_{o,j}(t),\alpha_{o,j}(t),\beta_{o,j}(t)\Big)
   \label{cost_aug}\\
&\hspace{1cm}d_{o,j}(t) \geq 1, \label{ineq_1}
    \end{align}
\normalsize

\noindent where $n_{p}$ and $n_{o}$ stands for the number of planning steps and obstacles, respectively. The parameters $\hspace{-0.1cm}\lambda_{x,j}(\hspace{-0.02cm}t\hspace{-0.02cm}), \hspace{-0.03cm}\lambda_{y,j}(\hspace{-0.02cm}t\hspace{-0.02cm}),\hspace{-0.03cm}\lambda_{z,j}(\hspace{-0.02cm}t\hspace{-0.02cm}), \hspace{-0.03cm}\lambda_{c_{\alpha,j}}\hspace{-0.03cm}(\hspace{-0.02cm}t\hspace{-0.02cm}), \hspace{-0.03cm}\lambda_{s_{\alpha,j}}\hspace{-0.03cm}(\hspace{-0.02cm}t\hspace{-0.02cm}),\hspace{-0.03cm}\lambda_{c_{\beta,j}}(\hspace{-0.02cm}t\hspace{-0.02cm})$ and $\lambda_{s_{\beta,j}}\hspace{-0.05cm}(\hspace{-0.02cm}t\hspace{-0.02cm})$ are Lagrange multipliers. The $\rho_{o}$ and $\rho$ are scalar. As can be seen, we relaxed the equality constraints in \eqref{eq1}-\eqref{collision_proposed_1} and transferred them~into cost function \eqref{cost_aug} using a combination of quadratic penalties and linear terms multiplied with Lagrange multipliers. Along similar lines, we also relaxed the equality constraints in \eqref{new_var_1}-\eqref{new_var_2} as quadratic penalties in~\eqref{aug_2}.
     
On initial inspection, the formulation \eqref{cost_aug}-\eqref{ineq_1} may seem like a typical non-linear programming problem. However, upon closer examination, we recognize its multi-convex structure within the space $(x(t),y(t), z(t))$, $d_{o,j}(t)$, $(c_{\alpha,j}(t), s_{\alpha,j}(t))$, and $(c_{\beta,j}(t),s_{\beta,j}(t))$. That is,

\begin{itemize}
    \item If we consider the optimization variables $d_{o,j}(t)$, $(c_{\alpha,j}(t), s_{\alpha,j}(t))$, and $(c_{\beta,j}(t),s_{\beta,j}(t))$ fixed, then the optimization problem \eqref{cost_aug}-\eqref{ineq_1} is convex in terms of $(x(t),y(t), z(t))$.
    \item If we consider the optimization variables, $(x(t),y(t), z(t))$, $d_{o,j}(t)$, and $(c_{\beta,j}(t),s_{\beta,j}(t))$ fixed, then the optimization problem \eqref{cost_aug}-\eqref{ineq_1} is convex in terms of $(c_{\alpha,j}(t), s_{\alpha,j}(t))$. 
    \item Similarly, for a given $(x(t),y(t), z(t))$, $d_{o,j}(t)$, and $(c_{\alpha,j}(t), s_{\alpha,j}(t))$, then the optimization problem \eqref{cost_aug}-\eqref{ineq_1} is convex in terms of $(c_{\beta,j}(t),s_{\beta,j}(t))$.
\end{itemize}

This multi-convex structure enables us to use techniques such as \gls{am} to solve the trajectory optimization problem effectively. Algorithm \ref{alg_1} outlines the step-by-step process of solving \eqref{cost_aug}-\eqref{ineq_1}. We comprehensively analyze each stage of the Algorithm \ref{alg_1} in the subsequent paragraphs

\vspace{0.45cm}
\noindent \textbf{Analysis and Description of Algorithm \ref{alg_1}:} Now, the details of the proposed algorithm can be explained as follows:

\begin{itemize}
    \item \textbf{Lines 1-2:} The algorithm begins with initializing $^{k}d_{o,j}(t)$, $^{k}\alpha_{o,j}(t)$, and $^{k}\beta_{o,j}(t)$, and subsequently calculating $^{k}c_{\alpha,j}(t)$, $^{k}s_{\alpha,j}(t)$, $^{k}c_{\beta,j}(t)$, and $^{k}s_{\beta,j}(t)$ at $k=0$.
    \item \textbf{Line 3:} Following this, we compute the optimization variables $^{k+1}x(t),^{k+1}y(t)$ and  $^{k+1}z(t)$. Thus, first, we inspect equations \eqref{cost_aug}-\eqref{ineq_1} and identify terms associated with $x(t),y(t)$, and $z(t)$ and rewrite the trajectory optimization as \eqref{com_x}-\eqref{com_z}. Remarkably, given values of $^{k}d_{o,j}(t)$, $^{k}\alpha_{o,j}(t)$, and $^{k}\beta_{o,j}(t)$, \eqref{com_x}-\eqref{com_z} are decoupled from each other, as they involve distinct terms (illustrated in different colors). So they can be solved in parallel. To solve trajectory optimization, \eqref{com_x}, we parametrize $x(t)$ using \eqref{parameter} and assume that for the $k^{th}$ iteration, the first term in \eqref{com_x} takes the following form 

\vspace{-0.3cm}
\small
\begin{align}
c_{x}(x^{(q)}(t))= \frac{1}{2}\boldsymbol{\xi}_x^T\mathbf{Q}_x \boldsymbol{\xi}_x+\hspace{0.1cm}^{k}\mathbf{q}_x^T \boldsymbol{\xi}_x, 
\label{quad_form}
\end{align}
\normalsize

\noindent for some constant positive definite matrix $\mathbf{Q}_x$, and vector $\mathbf{q}_x$. 
The exact expression for these depends on the definition of $c_{x}(x^{(q)}(t))$, and we discuss some possible choices in implementation details. The second term of \eqref{com_x} also can be defined as:

\small
\begin{align}
\sum_{j=1}^{n_{o}}\hspace{-0.1cm}\Big(^{k}\boldsymbol{\lambda}_{x,j}(\mathbf{P} \boldsymbol{\xi}_x\hspace{-0.2cm}-\hspace{-0.1cm}\mathbf{x}_{o,j}\hspace{-0.1cm}-\hspace{-0.1cm}a\hspace{0.03cm}^{k}\mathbf{d}_{o,j}\hspace{0.03cm}^{k}\mathbf{c}_{\alpha,j}\hspace{0.03cm}^{k}\mathbf{s}_{\beta,j})\hspace{-0.1cm}+\hspace{-0.1cm}\frac{\rho_{o}}{2}(\mathbf{P}\boldsymbol{\xi}_x \hspace{-0.2cm}- \hspace{-0.1cm}\mathbf{x}_{o,j} \hspace{-0.1cm}-\hspace{-0.1cm}a\hspace{0.03cm}^{k}\mathbf{d}_{o,j}\hspace{0.03cm}^{k}\mathbf{c}_{\alpha,j}\hspace{0.03cm}^{k}\mathbf{s}_{\beta,j})^{2} \Big)
\end{align}
\normalsize

\noindent where $\boldsymbol{\lambda}_{x,j}, \mathbf{x}_{o,j}, \mathbf{d}_{o,j},\mathbf{c}_{\alpha,j}$ and $\mathbf{s}_{\beta,j}$ are formed by stacking $\lambda_{x,j}(t), x_{o,j}(t), d_{o,j}(t),c_{\alpha,j}(t)$ and $s_{\beta,j}(t)$ at different time steps. Additionally, using some simplifications, the quadratic programming problem \eqref{com_x} can be reduced to solving a set of linear equations as

\vspace{-0.2cm}
\small
\begin{align}
\overbrace{(\mathbf{Q}_x+\rho_{o}n_{o}\mathbf{P}^T\mathbf{P})}^{\overline{\mathbf{D}}_{x}}\boldsymbol{\xi}_x\hspace{-0.1cm}=\overbrace{\hspace{-0.1cm}-(\hspace{0.03cm}^{k}\mathbf{q}_x\hspace{-0.1cm}+\hspace{-0.1cm}\sum_{j=1}^{n_{o}}\mathbf{P}^T \hspace{0.03cm}^{k}\boldsymbol{\lambda}_{x,j}\hspace{-0.1cm}-\hspace{-0.1cm}\rho_{o}\mathbf{P}^T(\mathbf{x}_{o,j}\hspace{-0.1cm}+\hspace{-0.1cm}a\hspace{0.03cm}^{k}\mathbf{d}_{o,j}\hspace{0.03cm}^{k}\mathbf{c}_{\alpha,j}\hspace{0.03cm}^{k}\mathbf{s}_{\beta,j}))}^{\overline{\boldsymbol{\chi}}_{x}}.
\label{step_x_2d_simplify}
\end{align}
\normalsize

\noindent As can be seen, the optimization problem is reduced to the structure of \eqref{simple} with a fixed matrix across all the iterations. Similarly, we solve \eqref{com_y} and \eqref{com_z} to compute $y(t)$ and $z(t)$. 

\item \textbf{Line 4:} In this stage, we derive the optimization variables $^{k+1}c_{\alpha,j}(t)$ and $^{k+1}s_{\alpha,j}(t)$. To compute these variables, we look at equations \eqref{cost_aug}-\eqref{ineq_1} and identify terms associated with $c_{\alpha,j}(t)$ and $s_{\alpha,j}(t)$. The trajectory optimization is then reformulated as \eqref{com_c_alpha}-\eqref{com_s_alpha}. Importantly, with given values of $^{k}d_{o,j}(t)$, $^{k+1}x(t)$, $^{k+1}y(t)$, $^{k+1}z(t)$, $^{k}\alpha_{o,j}(t)$ and $^{k}\beta_{o,j}(t)$, the trajectory optimization problems \eqref{com_c_alpha}-\eqref{com_s_alpha} are independent, featuring distinct terms (color-coded for clarity). Thus, they are thus amenable to parallel computation. Crucially, for a specific obstacle index $j$, $^{k+1}c_{\alpha,j}(t)$ and $^{k+1}s_{\alpha,j}(t)$ are temporally uncorrelated. Similar to line 3, we stack our optimization variables at different time instances. Moreover, we observe the decoupling of optimization variables across various obstacles. Consequently, optimization \eqref{com_c_alpha}-\eqref{com_s_alpha} can be decomposed into $n_{o}\times n_{p}$ parallel optimizations, each entailing the minimization of a single-variable quadratic function. The solutions for each can be derived symbolically, further enhancing the computational efficiency of the algorithm.

\item \textbf{Line 5:} While optimization \eqref{com_alpha} poses a non-convex challenge, an approximate solution can be derived through simple geometric intuition. Illustrated in Figure \ref{polar} (b), each set of feasible $x(t)- x_{o,j}(t)$, $y(t)- y_{o,j}(t)$, and $z(t)- z_{o,j}(t)$ forms an ellipsoid centered at the origin with dimensions $(ad_{o,j}(t),ad_{o,j}(t),bd_{o,j}(t))$. Consequently, \eqref{com_alpha} can be viewed and obtained as a projection of $x(t)- x_{o,j}(t)$ and $y(t)- y_{o,j}(t)$ onto an axis-aligned ellipsoid centered at the origin. 

\item \textbf{Line 6:} In this phase, we determine the optimization variables $^{k+1}c_{\beta,j}(t)$ and $^{k+1}s_{\beta,j}(t)$. To achieve this, we meticulously examine equations \eqref{cost_aug}-\eqref{ineq_1}, isolating terms associated with $c_{\beta,j}(t)$ and $s_{\beta,j}(t)$. The trajectory optimization is then reconfigured as \eqref{com_c_alpha}-\eqref{com_s_alpha}. Importantly, given values of $^{k}d_{o,j}(t)$, $^{k+1}x(t)$, $^{k+1}y(t)$, and $^{k+1}\alpha_{o,j}(t)$, the trajectory optimization problems \eqref{com_c_beta}-\eqref{com_s_beta} are independent, featuring distinct terms (color-coded for clarity). Hence, they lend themselves to parallel computation.  Similar to line 4, our optimization problems \eqref{com_c_beta}-\eqref{com_s_beta} can be decomposed into $n_{o}\times n_{p}$ parallel optimizations, each entailing the minimization of a single-variable quadratic function. 

\item \textbf{Line 7:} Similar to the intuition applied in line 5, optimization problem \eqref{com_beta} can be understood and derived as a projection of $z(t)- z_{o,j}(t)$ and $y(t)- y_{o,j}(t)$ onto an axis-aligned ellipsoid centered at the origin.

\item \textbf{Line 8:} We compute $^{k+1}d_{o,j}(t)$ using \eqref{d_max}.

\item \textbf{Line 9:} In this step, we update the Lagrange multipliers based on \eqref{lamda_x}-\eqref{lamda_s_beta}. The rules for these updates are adopted from \cite{boyd2011distributed}. Additionally, in each iteration, we increment the weights of the quadratic penalties, $\rho$ and $\rho_{o}$, if the residuals do not fall below the specified threshold.

\vspace{-0.2cm}
\small
\begin{subequations}
    \begin{align}
&\hspace{-0.33cm}^{k+1}\lambda_{x\hspace{-0.02cm},\hspace{-0.02cm}j}\hspace{-0.02cm}(\hspace{-0.035cm}t\hspace{-0.035cm})\hspace{-0.1cm}=\hspace{-0.08cm}^{k}\lambda_{x\hspace{-0.02cm},\hspace{-0.02cm}j}(\hspace{-0.035cm}t\hspace{-0.035cm})\hspace{-0.1cm}+\hspace{-0.1cm}\rho_{o}
  (\colorbox{my_green}{$\hspace{-0.1cm}^{k+1}x(\hspace{-0.03cm}t\hspace{-0.03cm})\hspace{-0.1cm}-\hspace{-0.1cm}x_{o\hspace{-0.02cm},\hspace{-0.02cm}j}(\hspace{-0.03cm}t\hspace{-0.03cm})\hspace{-0.1cm}-\hspace{-0.1cm}a\hspace{0.03cm}^{k+1}d_{o\hspace{-0.02cm},\hspace{-0.02cm}j}(\hspace{-0.03cm}t\hspace{-0.03cm})\hspace{-0.05cm}^{k+1}c_{\alpha\hspace{-0.02cm},\hspace{-0.02cm}j}(\hspace{-0.02cm}t\hspace{-0.03cm})^{k+1}s_{\beta,j}(\hspace{-0.02cm}t\hspace{-0.03cm}))$}\label{lamda_x} \\
&\hspace{-0.33cm}^{k+1}\lambda_{y\hspace{-0.02cm},\hspace{-0.02cm}j}(\hspace{-0.035cm}t\hspace{-0.035cm})\hspace{-0.08cm}= \hspace{-0.08cm}^{k}\lambda_{y\hspace{-0.02cm},\hspace{-0.02cm}j}(\hspace{-0.035cm}t\hspace{-0.035cm})\hspace{-0.08cm}+\hspace{-0.08cm}\rho_{o} (\colorbox{my_pink}{$\hspace{-0.1cm}^{k+1}y(\hspace{-0.03cm}t\hspace{-0.03cm})\hspace{-0.1cm}-\hspace{-0.1cm}y_{o,j}(\hspace{-0.03cm}t\hspace{-0.03cm})\hspace{-0.1cm}-\hspace{-0.1cm}a\hspace{0.03cm}^{k+1}d_{o\hspace{-0.02cm},\hspace{-0.02cm}j}(\hspace{-0.03cm}t\hspace{-0.03cm})\hspace{-0.05cm}^{k+1}s_{\alpha\hspace{-0.02cm},\hspace{-0.02cm}j}(\hspace{-0.02cm}t\hspace{-0.03cm})^{k+1}s_{\beta,j}(\hspace{-0.02cm}t\hspace{-0.03cm}))\hspace{-0.1cm}$}
  \label{lamda_y}\\&\hspace{-0.33cm}^{k+1}\lambda_{z\hspace{-0.02cm},\hspace{-0.02cm}j} (\hspace{-0.035cm}t\hspace{-0.035cm})\hspace{-0.08cm}= \hspace{-0.08cm}^{k}\lambda_{z\hspace{-0.02cm},\hspace{-0.02cm}j}(\hspace{-0.035cm}t\hspace{-0.035cm})\hspace{-0.08cm}+\hspace{-0.08cm}\rho_{o}(\colorbox{my_blue}{$\hspace{-0.1cm}^{k+1}z(\hspace{-0.03cm}t\hspace{-0.03cm})\hspace{-0.1cm}-\hspace{-0.1cm}z_{o,j}(\hspace{-0.03cm}t\hspace{-0.03cm})\hspace{-0.1cm}-\hspace{-0.1cm}b\hspace{0.03cm}^{k+1}d_{o,j}(\hspace{-0.03cm}t\hspace{-0.03cm})\hspace{-0.05cm}^{k+1}c_{\beta,j}(\hspace{-0.02cm}t\hspace{-0.03cm}))$} \label{lamda_z}\\
&\hspace{-0.33cm}^{k+1}\lambda_{c_{\alpha,\hspace{-0.02cm}j}}(\hspace{-0.035cm}t\hspace{-0.035cm})\hspace{-0.08cm}=\hspace{-0.08cm}
^{k}\lambda_{c_{\alpha,\hspace{-0.02cm}j}}(\hspace{-0.035cm}t\hspace{-0.035cm})\hspace{-0.08cm}+\hspace{-0.08cm}\rho (\colorbox{my_orange}{$   c_{\alpha,j}(t) - \cos{\alpha_{o,j}(t)})$} \label{lamda_c_alpha}\\
&\hspace{-0.33cm}^{k+1}\lambda_{s_{\alpha,\hspace{-0.02cm}j}}(\hspace{-0.035cm}t\hspace{-0.035cm})\hspace{-0.08cm}=\hspace{-0.08cm}
^{k}\lambda_{s_{\alpha,\hspace{-0.02cm}j}}(\hspace{-0.035cm}t\hspace{-0.035cm})\hspace{-0.08cm}+\hspace{-0.08cm}\rho( \colorbox{my_yellow}{$
     s_{\alpha,j}(t) - \sin{\alpha_{o,j}(t)}) $}\label{lamda_s_alpha}\\
&\hspace{-0.33cm}^{k+1}\lambda_{c_{\beta,\hspace{-0.02cm}j}}(\hspace{-0.035cm}t\hspace{-0.035cm})\hspace{-0.08cm}=\hspace{-0.08cm}
^{k}\lambda_{c_{\beta,\hspace{-0.02cm}j}}(\hspace{-0.035cm}t\hspace{-0.035cm})\hspace{-0.08cm}+\hspace{-0.08cm}\rho \colorbox{my_purple}{$ ( c_{\beta,j}(t) - \cos{\beta_{o,j}(t)})$} \label{lamda_c_beta}\\ 
&\hspace{-0.33cm}^{k+1}\lambda_{s_{\beta,\hspace{-0.02cm}j}}(\hspace{-0.035cm}t\hspace{-0.035cm})\hspace{-0.08cm}=\hspace{-0.08cm}
^{k}\lambda_{s_{\beta,\hspace{-0.02cm}j}}(\hspace{-0.035cm}t\hspace{-0.035cm})\hspace{-0.08cm}+\hspace{-0.08cm}\rho\colorbox{my_grey}{$(
     s_{\beta,j}(t) - \sin{\beta_{o,j}(t)}). $} \label{lamda_s_beta}
\end{align}
\end{subequations}
\normalsize
\end{itemize}

\vspace{-0.5cm}
\begin{algorithm}
\DontPrintSemicolon
\SetAlgoLined
\SetNoFillComment
\scriptsize
\caption{\scriptsize Alternating Minimization for Solving \eqref{cost_aug}-\eqref{ineq_1}  }\label{alg_1}
\KwInitialization{Initiate $^{k}d_{o,j}(t)$, $^{k}\alpha_{o,j}(t)$, $^{k}\beta_{o,j}(t)$  }
\While{$k \leq maxiter$}{

\vspace{-0.07cm}
\hspace{-0.15cm}Compute \hspace{-0.05cm}\colorbox{my_orange}{\hspace{-0.1cm}$^{k}\hspace{-0.05cm}c_{\hspace{-0.031cm}\alpha\hspace{-0.021cm},\hspace{-0.031cm}j}\hspace{-0.021cm}(\hspace{-0.03cm}t\hspace{-0.03cm})\hspace{-0.05cm}=\hspace{-0.05cm}\cos{^{k}\alpha_{\hspace{-0.031cm}o\hspace{-0.031cm},\hspace{-0.05cm}j}\hspace{-0.051cm}(\hspace{-0.03cm}t\hspace{-0.03cm})}\hspace{-0.041cm}\hspace{-0.05cm}$},\colorbox{my_yellow}{\hspace{-0.1cm}$
^{k}s_{\hspace{-0.04cm}\alpha\hspace{-0.021cm},\hspace{-0.031cm}j}\hspace{-0.07cm}(\hspace{-0.03cm}t\hspace{-0.03cm})\hspace{-0.07cm}=\hspace{-0.07cm}\sin{^{k}\hspace{-0.05cm}\alpha_{\hspace{-0.03cm}o\hspace{-0.02cm},\hspace{-0.035cm}j}\hspace{-0.02cm}(\hspace{-0.03cm}t\hspace{-0.03cm})}\hspace{-0.05cm},\hspace{-0.2cm}$}
   \colorbox{my_purple}{$\hspace{-0.1cm}^{k}c_{\beta,\hspace{-0.02cm}j}(\hspace{-0.03cm}t\hspace{-0.03cm})\hspace{-0.08cm}=\hspace{-0.08cm}\cos{\hspace{-0.02cm}^{k}\hspace{-0.035cm}\beta_{\hspace{-0.03cm}o\hspace{-0.02cm},\hspace{-0.02cm}j}\hspace{-0.06cm}(\hspace{-0.03cm}t\hspace{-0.03cm})}$\hspace{-0.05cm}},\colorbox{my_grey}{\hspace{-0.12cm}$
^{k}\hspace{-0.021cm}s_{\beta\hspace{-0.021cm},\hspace{-0.021cm}j}\hspace{-0.011cm}(\hspace{-0.03cm}t\hspace{-0.03cm})\hspace{-0.05cm}=\hspace{-0.08cm}\sin{^{k}\hspace{-0.02cm}\beta_{o,j}\hspace{-0.02cm}(\hspace{-0.03cm}t\hspace{-0.03cm})}\hspace{-0.19cm}$}

\hspace{-0.15cm}Compute $^{k+1}x(t)$,$^{k+1}y(t)$ and $^{k+1}z(t)$ 
\vspace{-0.4cm}
\begin{subequations}
    \begin{align}
\hspace{-0.7cm}^{k+1}x(\hspace{-0.02cm}t\hspace{-0.02cm}) &= \arg \min_{x(t)}c_{x}(x^{(q)}(t))\hspace{-0.07cm}+\hspace{-0.07cm}\sum_{t=0}^{t=n_{p}}\sum_{j=1}^{j=n_{o}}\Big(\hspace{0.05cm}^{k}\lambda_{x,j}(t)
 (\colorbox{my_green}{$\hspace{-0.1cm}x(\hspace{-0.03cm}t\hspace{-0.03cm})\hspace{-0.1cm}-\hspace{-0.1cm}x_{o,j}(\hspace{-0.03cm}t\hspace{-0.03cm}) -
  a^{k}d_{o,j}(\hspace{-0.03cm}t\hspace{-0.03cm})
 $} \nonumber \\ &~~\colorbox{my_green}{$\times\hspace{0.03cm}^{k}c_{\alpha,j}(\hspace{-0.03cm}t\hspace{-0.03cm})\hspace{0.03cm}^{k}s_{\beta,j}(\hspace{-0.03cm}t\hspace{-0.03cm})\hspace{-0.04cm}$\hspace{-0.07cm}})\hspace{-0.07cm}+\hspace{-0.1cm} \frac{\rho_{o}}{2}(\colorbox{my_green}{$\hspace{-0.1cm}x(\hspace{-0.03cm}t\hspace{-0.03cm})\hspace{-0.05cm}-\hspace{-0.05cm}x_{o,j}(\hspace{-0.03cm}t\hspace{-0.03cm})\hspace{-0.05cm}-\hspace{-0.05cm}a\hspace{0.03cm}^{k}d_{o,j}(\hspace{-0.03cm}t\hspace{-0.03cm})\hspace{0.05cm}^{k}c_{\alpha,j}(\hspace{-0.02cm}t\hspace{-0.03cm}) \hspace{0.03cm}^{k}s_{\beta,j}(\hspace{-0.03cm}t\hspace{-0.03cm})$\hspace{-0.05cm}})^{2}\hspace{-0.05cm}\Big) \label{com_x}\\
   \hspace{-0.7cm}^{k+1}y(t)
   \hspace{-0.1cm} &=\arg \min_{y(\hspace{-0.03cm}t\hspace{-0.03cm})}c_{y}(y^{(q)}(t))\hspace{-0.07cm}+\sum_{t=0}^{t=n_{p}}\sum_{j=1}^{j=n_{o}}\Big(\hspace{0.05cm}^{k}\lambda_{y,j}(t)(\colorbox{my_pink}{$y(\hspace{-0.03cm}t\hspace{-0.03cm})\hspace{-0.05cm}-\hspace{-0.05cm}y_{o,j}(\hspace{-0.03cm}t\hspace{-0.03cm}) -a\hspace{0.03cm}^{k}d_{o,j}(\hspace{-0.03cm}t\hspace{-0.03cm}) 
    \hspace{-0.1cm}$}\nonumber \\&~~\colorbox{my_pink}{$
\times \hspace{0.03cm}^{k}s_{\alpha,j}(\hspace{-0.03cm}t\hspace{-0.03cm}) \hspace{0.03cm}^{k}s_{\beta,j}(\hspace{-0.03cm}t\hspace{-0.03cm})\hspace{-0.05cm}$}\hspace{-0.03cm}) 
\hspace{-0.05cm}+\hspace{-0.1cm}\frac{\rho_{o}}{2}(\colorbox{my_pink}{$\hspace{-0.05cm}y(\hspace{-0.03cm}t\hspace{-0.03cm})\hspace{-0.07cm}-\hspace{-0.07cm}y_{o,j}(\hspace{-0.03cm}t\hspace{-0.03cm})
\hspace{-0.07cm}-\hspace{-0.07cm}a\hspace{0.03cm}^{k}d_{o,j}(\hspace{-0.03cm}t\hspace{-0.03cm})\hspace{0.03cm}^{k}s_{\alpha,j}(\hspace{-0.03cm}t\hspace{-0.03cm})\hspace{0.03cm}^{k}s_{\beta,j}(\hspace{-0.03cm}t\hspace{-0.03cm})\hspace{-0.05cm}$}\hspace{-0.03cm})^2\hspace{-0.03cm}\Big) \label{com_y}\\
\hspace{-0.7cm}^{k+1}z(t)\hspace{-0.07cm}&=\hspace{-0.07cm}\arg \min_{z(t)}c_{z}(z^{(q)}(t))+\hspace{-0.2cm}\sum_{t=0}^{t=n_{p}}\sum_{j=1}^{j=n_{o}}\hspace{-0.1cm}\Big(\hspace{0.05cm}^{k}\lambda_{z,j}(\hspace{-0.03cm}t\hspace{-0.03cm})(\colorbox{my_blue}{$\hspace{-0.05cm}z(\hspace{-0.03cm}t\hspace{-0.03cm})-z_{o,j}(\hspace{-0.03cm}t\hspace{-0.03cm})-b\hspace{0.05cm}^{k}d_{o,j}(\hspace{-0.03cm}t\hspace{-0.03cm}) $} \nonumber \\
&~~\colorbox{my_blue}{$\times\hspace{0.05cm}^{k}c_{\beta,j}(t)$} ) 
+ \frac{\rho_{o}}{2}(\colorbox{my_blue}{$z(t)-z_{o,j}(t) -b\hspace{0.05cm}^{k}d_{o,j}\hspace{0.05cm}^{k}c_{\beta,j}(t)$} )^{2}\Big) \label{com_z}
\end{align}
\end{subequations}
\vspace{-0.45cm}

\hspace{-0.15cm}Compute $\hspace{0.05cm}^{k+1}c_{\alpha,j}(t)$,$\hspace{0.05cm}^{k+1}s_{\alpha,j}(t)$
\vspace{-0.4cm}
\begin{subequations}
    \begin{align}
\hspace{-0.4cm}^{k+1}c_{\alpha,j}\hspace{-0.06cm}(t)&=\arg\min_{c_{\alpha,j}(t)}\sum_{t}\sum_{j}\Big(\frac{\rho}{2}(\colorbox{my_orange}{$c_{\alpha,j}\hspace{-0.04cm}(t)- \cos{^{k}\alpha_{o,j}(t)}$} +\frac{\lambda_{c_{\alpha,j}}(t)}{\rho})^{2}\nonumber \\
&~~+\hspace{0.1cm}^{k}\lambda_{x,j}(t)(\colorbox{my_green}{$^{k+1}x(\hspace{-0.04cm}t\hspace{-0.04cm})-\hspace{-0.05cm}x_{o,j}\hspace{-0.03cm}(\hspace{-0.04cm}t\hspace{-0.04cm})\hspace{-0.07cm} $} \colorbox{my_green}{$-\hspace{-0.07cm}
  a^{k}d_{o,j}(t)\hspace{-0.03cm}c_{\alpha,j}(\hspace{-0.03cm}t\hspace{-0.03cm})\hspace{0.03cm}^{k}s_{\beta,j}\hspace{-0.03cm}(t\hspace{-0.03cm})\hspace{-0.04cm}$})
  \nonumber \\
&~~ +\frac{\rho_{o}}{2}(\colorbox{my_green}{$^{k+1}x(t)-x_{o,j}(t)-a\hspace{0.03cm}^{k}d_{o,j}(t)c_{\alpha,j}(t) \hspace{0.06cm}^{k}s_{\beta,j}(t)$})^{2}\Big) \label{com_c_alpha}\\
\hspace{-0.4cm}^{k+1}s_{\alpha,j}(t)&=\arg\min_{s_{\alpha,j}(t)}\sum_{t}\sum_{j}\Big(\frac{\rho}{2}(\colorbox{my_yellow}{$s_{\alpha,j}(t)- \sin{^{k}\alpha_{o,j}(t)}$}+\frac{\lambda_{s_{\alpha,j}}(t)}{\rho})^{2} \nonumber \\
&~~+\hspace{0.1cm}^{k}\lambda_{y,j}(t)(\colorbox{my_pink}{$^{k+1}y(t)-y_{o,j}\hspace{-0.03cm}(t)-\hspace{-0.07cm}
  a^{k}d_{o,j}(t)s_{\alpha,j}(t)^{k}s_{\beta,j}(t)\hspace{-0.04cm}$})
  \nonumber \\&~~+ \frac{\rho_{o}}{2}(\colorbox{my_pink}{$^{k+1}y(\hspace{-0.03cm}t\hspace{-0.03cm})-y_{o,j}(\hspace{-0.03cm}t\hspace{-0.03cm})-a\hspace{0.03cm}^{k}d_{o,j}(\hspace{-0.03cm}t\hspace{-0.03cm})s_{\alpha,j}(t\hspace{-0.03cm}) \hspace{0.03cm}^{k}s_{\beta,j}\hspace{-0.03cm}(\hspace{-0.03cm}t\hspace{-0.03cm})$})^{2}\Big) \label{com_s_alpha}
\end{align}
\end{subequations}
\vspace{-0.55cm}

\hspace{-0.15cm}Compute $\hspace{0.05cm}^{k+1}\alpha_{o,j}(t)$
\vspace{-0.5cm}
\begin{align}
    \hspace{0.05cm}^{k+1}\alpha_{o,j}(t) = \arg \min_{\alpha_{o,j}(t)}\sum_{t}\hspace{-0.08cm}\sum_{j}(\alpha_{o,j}(t) - \arctan 2\frac{^{k+1}s_{\alpha,j}(t)}{^{k+1}c_{\alpha,j}(t)})^2 \label{com_alpha}
\end{align}
\vspace{-0.6cm}

\hspace{-0.15cm}Compute $\hspace{0.05cm}^{k+1}c_{\beta,j}(t)$,$\hspace{0.05cm}^{k+1}s_{\beta,j}(t)$
\vspace{-0.4cm}
\begin{subequations}
    \begin{align}
\hspace{-0.4cm}^{k+1}c_{\beta,j}(t)&=\arg\min_{c_{\beta,j}(t)}\sum_{t}\sum_{j}\Big(\frac{\rho}{2}(\colorbox{my_purple}{$c_{\beta,j}(t)- \cos{^{k}\beta_{o,j}(t)}$}+\frac{\lambda_{c_{\beta,j}}(t)}{\rho})^{2} \nonumber \\&~~+\hspace{0.1cm}^{k}\lambda_{z,j}(t)(\colorbox{my_blue}{$^{k+1}z(t)-z_{o,j}(t)-
  b^{k}d_{o,j}(t)c_{\beta,j}(t)$})
  \nonumber \\&~~+ \frac{\rho_{o}}{2}(\colorbox{my_blue}{$^{k+1}z(t)-z_{o,j}(t)-b\hspace{0.05cm}^{k}d_{o,j}(t)c_{\beta,j}(t)$})^{2}\Big) \label{com_c_beta}\\
  \hspace{-0.85cm}
\hspace{-0.4cm}^{k+1}s_{\beta,j}(t)&=\arg\min_{s_{\beta,j}(t)}\sum_{t}\sum_{j}\Big(\frac{\rho}{2}(\hspace{-0.01cm}\colorbox{my_grey}{$s_{\beta,j}(t)- \sin{^{k}\beta_{o,j}\hspace{-0.05cm}(t)}$}\hspace{-0.2cm}\frac{\lambda_{s_{\beta,j}}(t)}{\rho})^{2}+ \nonumber \\&~~+\hspace{0.05cm}^{k}\lambda_{y,j}(t)(\colorbox{my_pink}{$^{k+1}y(t)-y_{o,j}(t) $}
\colorbox{my_pink}{$-
  a^{k}d_{o,j}(t)^{k+1}s_{\alpha,j}\hspace{-0.03cm}(t)s_{\beta,j}(\hspace{-0.03cm}t)\hspace{-0.04cm}$}) \nonumber \\&~~+\frac{\rho_{o}}{2}\hspace{-0.05cm}(\colorbox{my_pink}{$^{k+1}y(t\hspace{-0.03cm})-y_{o,j}(t)-a^{k}d_{o,j}(t)^{k+1}s_{\alpha,j}(t) s_{\beta,j}(t)$})^{2}\Big) \label{com_s_beta}
\end{align}
\end{subequations}
\vspace{-0.6cm}

\hspace{-0.15cm}Compute $\hspace{0.05cm}^{k+1}\beta_{o,j}(t)$
\vspace{-0.5cm}
\begin{align}
    \hspace{0.05cm}^{k+1}\beta_{o,j}(t) = \arg \min_{\beta_{o,j}(t)}\sum_{t}\hspace{-0.08cm}\sum_{j}(\beta_{o,j}(t) - \arctan 2\frac{^{k+1}s_{\beta,j}(t)}{^{k+1}c_{\beta,j}(t)}^2
    \label{com_beta}
\end{align}
\vspace{-0.5cm}

\hspace{-0.15cm}Compute $\hspace{0.05cm}^{k+1}d_{o,j}(t)$ through \eqref{d_max} using updated $(x(t),y(t),z(t))$

\hspace{-0.15cm}Update $\lambda_{x,j}(t), \lambda_{y,j}(t), \lambda_{z,j}, \lambda_{c_{\alpha,j}}(t), \lambda_{s_{\alpha,j}}(t),\lambda_{c_{\beta,j}}(t)$ and $\lambda_{s_{\beta,j}}(t)$ at $k+1$
}
\end{algorithm}
\normalsize
\section{Validation and Benchmarking}
\noindent\textbf{Implementation Details:} 
We implemented Algorithm \ref{alg_1} in Python, utilizing the Numpy \cite{oliphant2006guide} libraries, and incorporated CVXOPT \cite{vandenberghe2010cvxopt} to solve the \gls{qp}s within each iteration of the \gls{ccp}. The execution of all benchmarks took place on a laptop with a 2.60 GHz processor and 32 GB RAM. To enhance the computational efficiency of the \gls{ccp}, we adopted the heuristic proposed in \cite{lipp2016variations}, which involves considering only $6-8\%$ of the total number of collision avoidance constraints at each iteration. It is crucial to note that this heuristic's effectiveness is highly dependent on problem parameters, and determining the specific percentage involved multiple trial and error iterations. The following cost function was employed in our analysis.

\begin{align}
c_{x}(x^{(q)}(t)) = \sum_{t=0}^{n_{p}} w_{1}\ddot{x}(t)^2+w_{2}(x(t)-x_{des}(t))^{2}
\label{cost_sim}\\
c_{y}(y^{(q)}(t)) = \sum_{t=0}^{n_{p}} w_{1}\ddot{y}(t)^2+w_{2}(y(t)-y_{des}(t))^{2} \label{cost_sim_2}\\
c_{z}(z^{(q)}(t)) = \sum_{t=0}^{n_{p}} w_{1}\ddot{z}(t)^2+w_{2}(z(t)-z_{des}(t))^{2}, \label{cost_sim_3}
\end{align}

\noindent where each cost function comprises two terms: the first term ensures smoothness, while the second term relates to the tracking of the desired trajectory $(x_{des}(t), y_{des}(t), z_{des}(t))$. The weights $w_{1}$ and $w_{2}$ enable a trade-off between different components of the cost function. We employ the following metrics to evaluate our proposed method
\begin{itemize}
     \item Smoothness cost: This metric shows the acceleration values and it can be defined as 
     
     \begin{align}
    \sum_{t=0}^{n_{p}}\Big(\ddot{x}(t)^2\hspace{-0.1cm}+\hspace{-0.1cm}\ddot{y}(t)^2\hspace{-0.1cm}+\hspace{-0.1cm}\ddot{z}(t)^2\Big). \label{smoothness_costt} 
     \end{align}
     
      \item Tracking cost: This metric shows how well our optimizer follows a desired trajectory and it can be defined as
       \begin{align}
     \sum_{t=0}^{n_{p}}\Big((x(t)-x_{des}(t))^{2}+(y(t)-y_{des}(t))^{2}+(z(t)-z_{des}(t))^{2}\Big).\label{tracking_costt} 
     \end{align}
     
    \item Computation time: This metric shows how long it takes for our proposed optimizer to find a collision-free trajectory.
    \item Scalability: Computation time changes by increasing the number of obstacles and for a fixed number of iterations. 
\end{itemize}
\subsection{Benchmarks and Qualitative Results}
\begin{figure*}[t]
  \centering  
\subfigure[]{
    \includegraphics[width= 6.2cm, height=5.2cm] {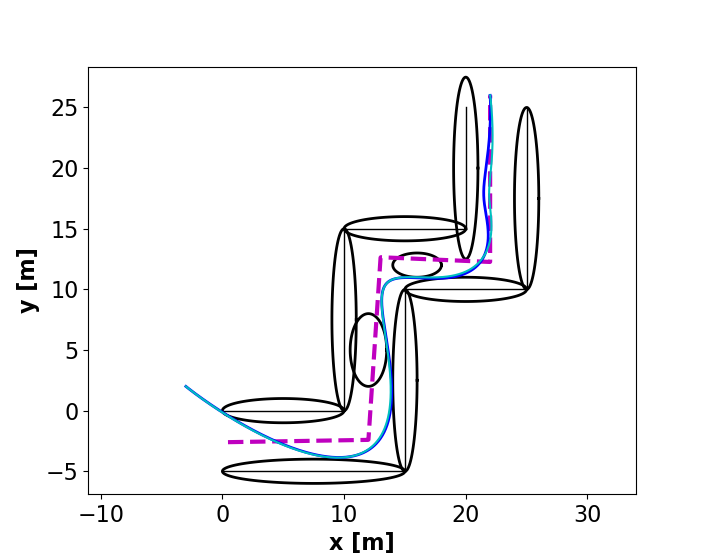}
    \label{traj_1_static}
   }\hspace{-0.7cm}
\subfigure[]{
    \includegraphics[width= 6.2cm, height=5.2cm] {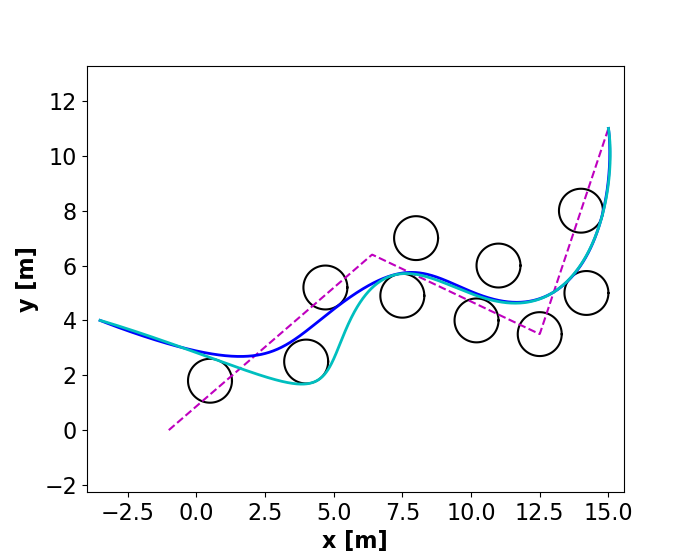}
    \label{traj_2_static}
   }                 
\caption[Static benchmark of Algorithm \ref{alg_1}]{\small{(a) and (b): Static obstacle benchmarks depicting a narrow corridor-like scene and an environment with randomly placed obstacles. The paths obtained with our proposed optimizer are marked in blue, while the \gls{ccp} approach paths are marked in cyan. The desired trajectory to be tracked is indicated in magenta.}}
\end{figure*}

\begin{figure*}[t]
  \centering  
  \subfigure[]{
    \includegraphics[width= 6.2cm, height=5cm] {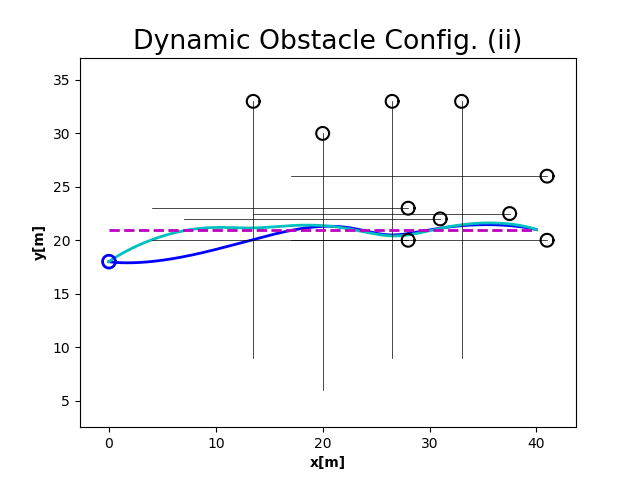}
    \label{traj_1_dyn}
   }\hspace{-0.7cm}
\subfigure[]{
    \includegraphics[width= 6.2cm, height=5cm] {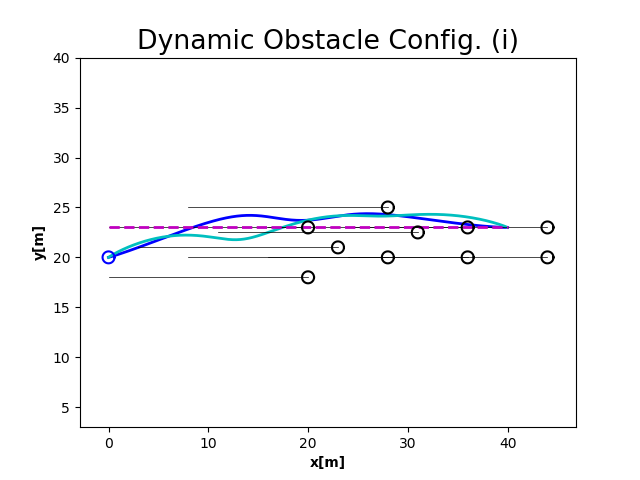}
    \label{traj_2_dyn}
   }
\caption[Dynamic benchmark of Algorithm \ref{alg_1}]{\small{(a) and (b): Dynamic obstacle benchmarks depicting environments where obstacles are moving in opposite and perpendicular directions relative to the agent. The paths obtained with our proposed optimizer are marked in blue, while the \gls{ccp} approach paths are marked in cyan. The desired trajectory to be tracked is indicated in magenta.}}
\end{figure*}

\begin{figure}[!h]
  \centering  
    \includegraphics[scale=0.31] {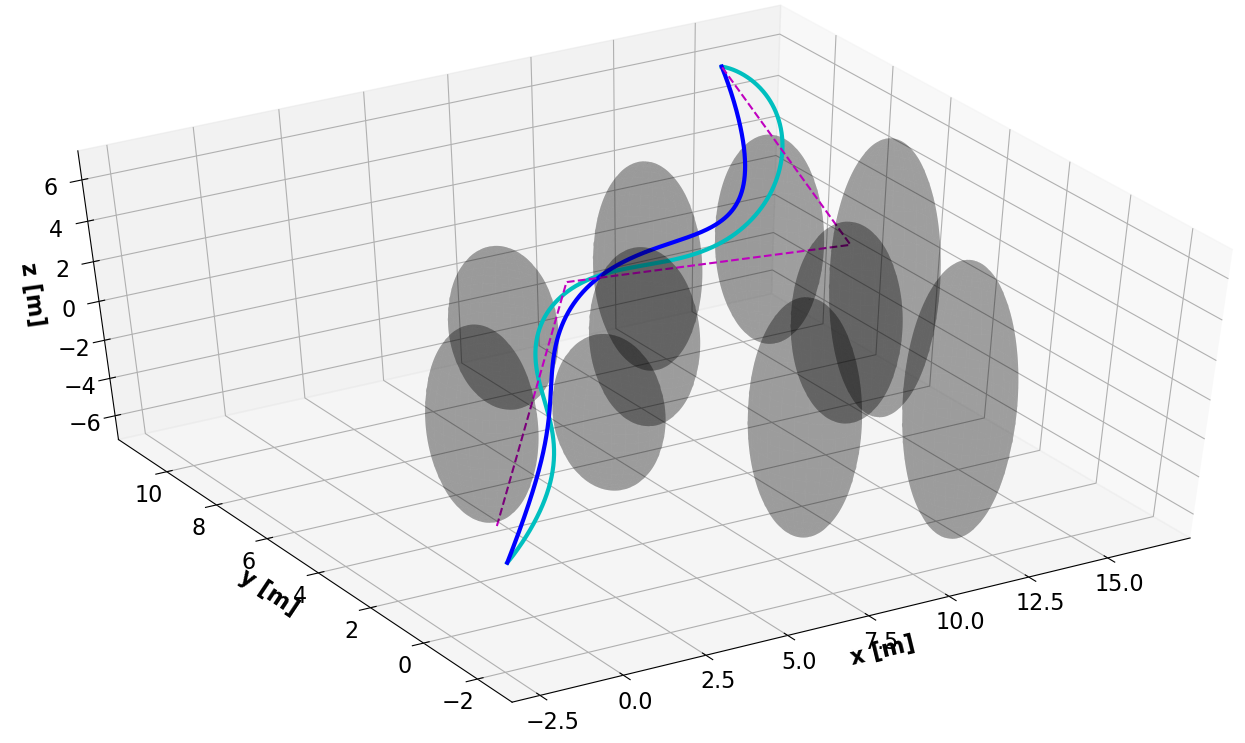}
\caption[3D benchmark of Algorithm \ref{alg_1}]{\small{3D obstacle benchmark. The paths obtained with our proposed optimizer are marked in blue, while the \gls{ccp} approach paths are marked in cyan. The desired trajectory to be tracked is indicated in magenta.}}
 \label{fig_3d_q}
\end{figure}

Three benchmarks, including scenarios with 2D static obstacles (Figures \ref{traj_1_static} and \ref{traj_2_static}, for a narrow corridor-like scene and an environment with randomly placed obstacles), 2D dynamic obstacles (Figures \ref{traj_1_dyn} and \ref{traj_2_dyn}) and 3D collision avoidance (Figure \ref{fig_3d_q}) are considered. For each benchmark, we considered $n_{o}=10$ obstacles and generated 15 different problem instances by varying the initial position and velocity for a given final state. The planning time interval ranged from $15s$ to $60s$, depending on the start and goal positions, and was discretized into $n_{p}=1000$ steps. Thus, the total number of collision avoidance constraints across all benchmarks was $10,000$.

\subsection{ Convergence Validation}
\begin{figure}[t]
  \centering  
    \includegraphics[scale=0.57] {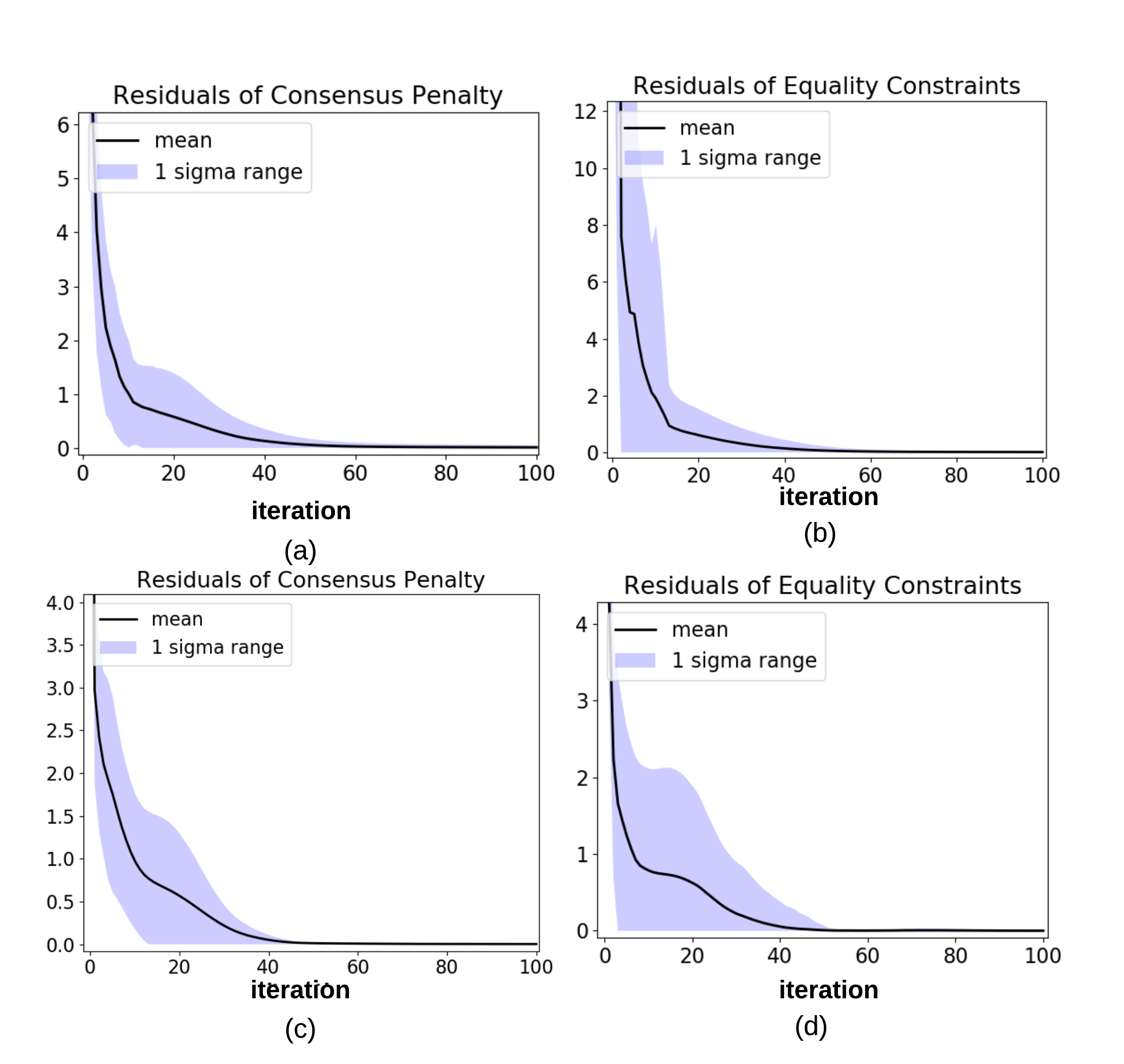}
\caption[Convergence validation]{\small{The general trend of residuals of equality constraints \eqref{collision_proposed_1} and consensus term \eqref{new_var_1} with iterations for 2d static and dynamic obstacles}}
 \label{fig_3d}
\end{figure}

A key validation for our optimizer, Algorithm \ref{alg_1} is the decrease in different residuals over iteration. Figure \ref{fig_3d} shows this trend for equality constraints \eqref{collision_proposed_1} and \eqref{new_var_1}. As can be seen, on average, around 100 iterations suffice to achieve a residual on the order of $10^{-3}$.

\subsection{Quantitive Results}
\begin{figure}[!h]
  \centering  
    \includegraphics[scale=0.4] {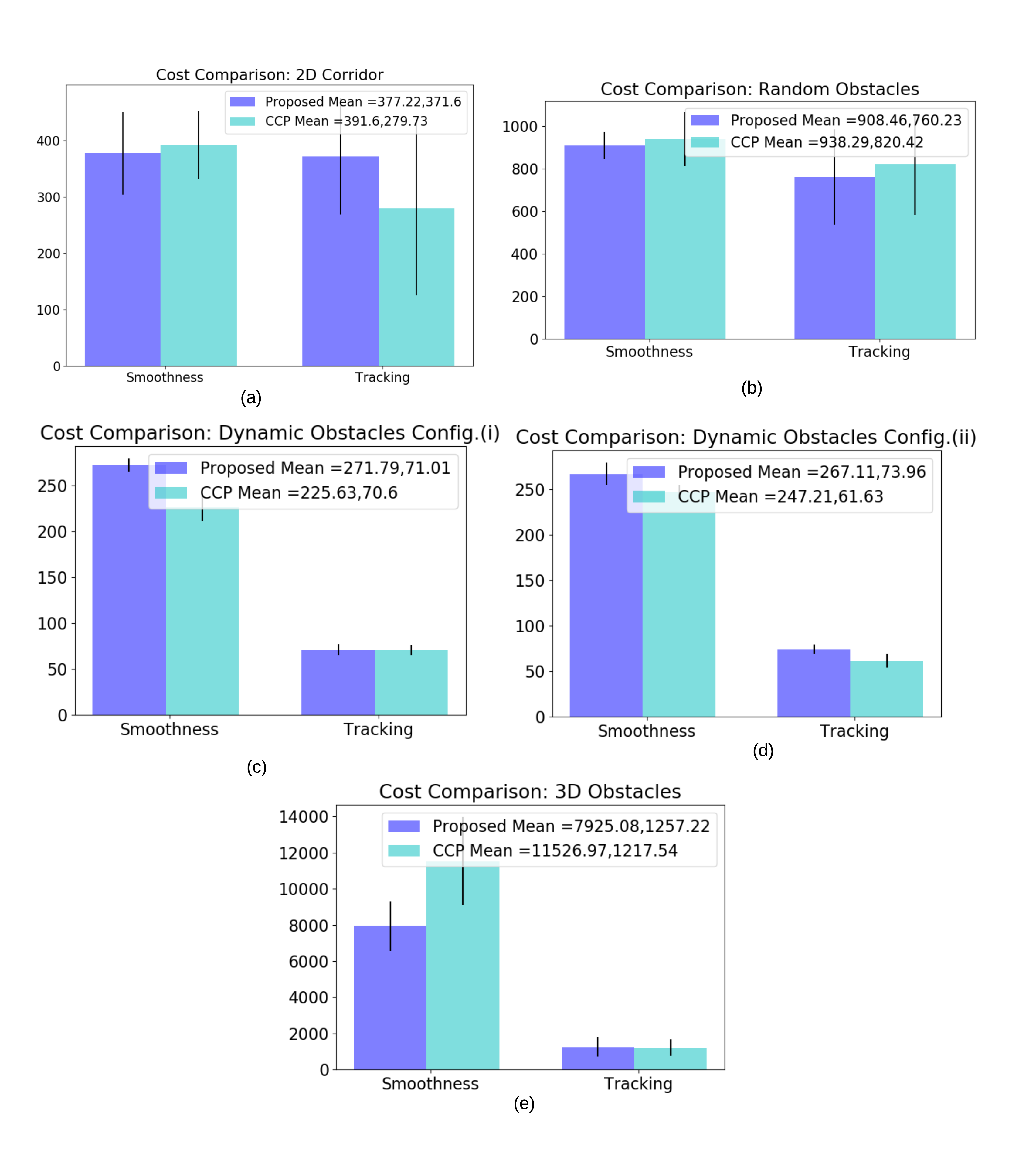}
\caption[The optimal cost comparisons between Algorithm \ref{alg_1} and CCP]{\small{The optimal cost statistics for static 2D obstacles (a)-(b), dynamic 2D obstacles (c)-(d) and 3D environments (e).} } \label{optimal_cost}
\end{figure}

\noindent\textbf{Optimal Cost Analysis:}
Figure \ref{optimal_cost} presents the statistical analysis of the optimal costs, including smoothness and tracking across various benchmarks. A diverse trend is observed in the 2D benchmarks. However, upon averaging the costs across all instances, both the proposed optimizer and \gls{ccp} demonstrate very similar smoothness and tracking costs, with \gls{ccp} showing a marginal superiority. Notably, for the 3D obstacle benchmark in Figure \ref{optimal_cost}, the proposed optimizer consistently achieves solutions with significantly lower smoothness costs.

\begin{figure}[h]
 \centering  
    \includegraphics[scale=0.57] {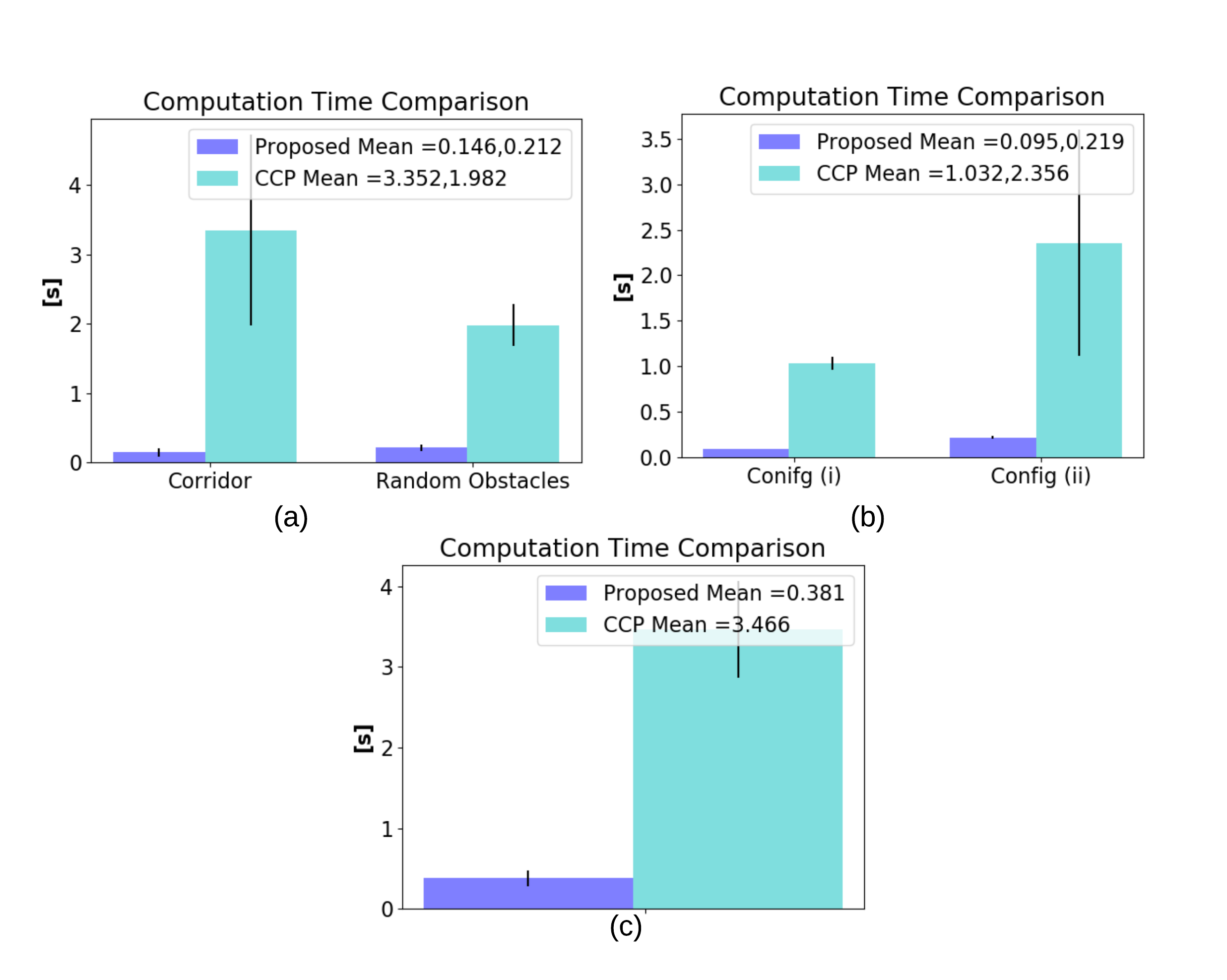}
    \vspace{-0.5cm}
\caption[Computation time comparison between Algorithm 1 and \ref{alg_1}]{\small{The computation time comparison for static 2D obstacles (a), dynamic 2D obstacles (b) and 3D environments (c).}} \label{cost_com}
\end{figure}

\noindent \textbf{Computation Time Comparison:}
We now introduce one of the key results of this paper. Figs. \ref{cost_com} presents the statistical analysis of computation times across different benchmarks. Notably, for the 2D static obstacle benchmark (see Figure \ref{cost_com}(a)), the proposed optimizer achieves an average speed-up of up to two orders of magnitude compared to \gls{ccp}. Furthermore, the computation times for \gls{ccp} exhibit high variance, suggesting that the worst-case difference in computation time could be even more pronounced. The proposed optimizer demonstrates a similar speed-up in both the dynamic obstacle (see Figure \ref{cost_com}(b)) and 3D benchmarks (see Figure \ref{cost_com}(c)).

\begin{figure}[!h]
  \centering  
    \includegraphics[scale=0.37]{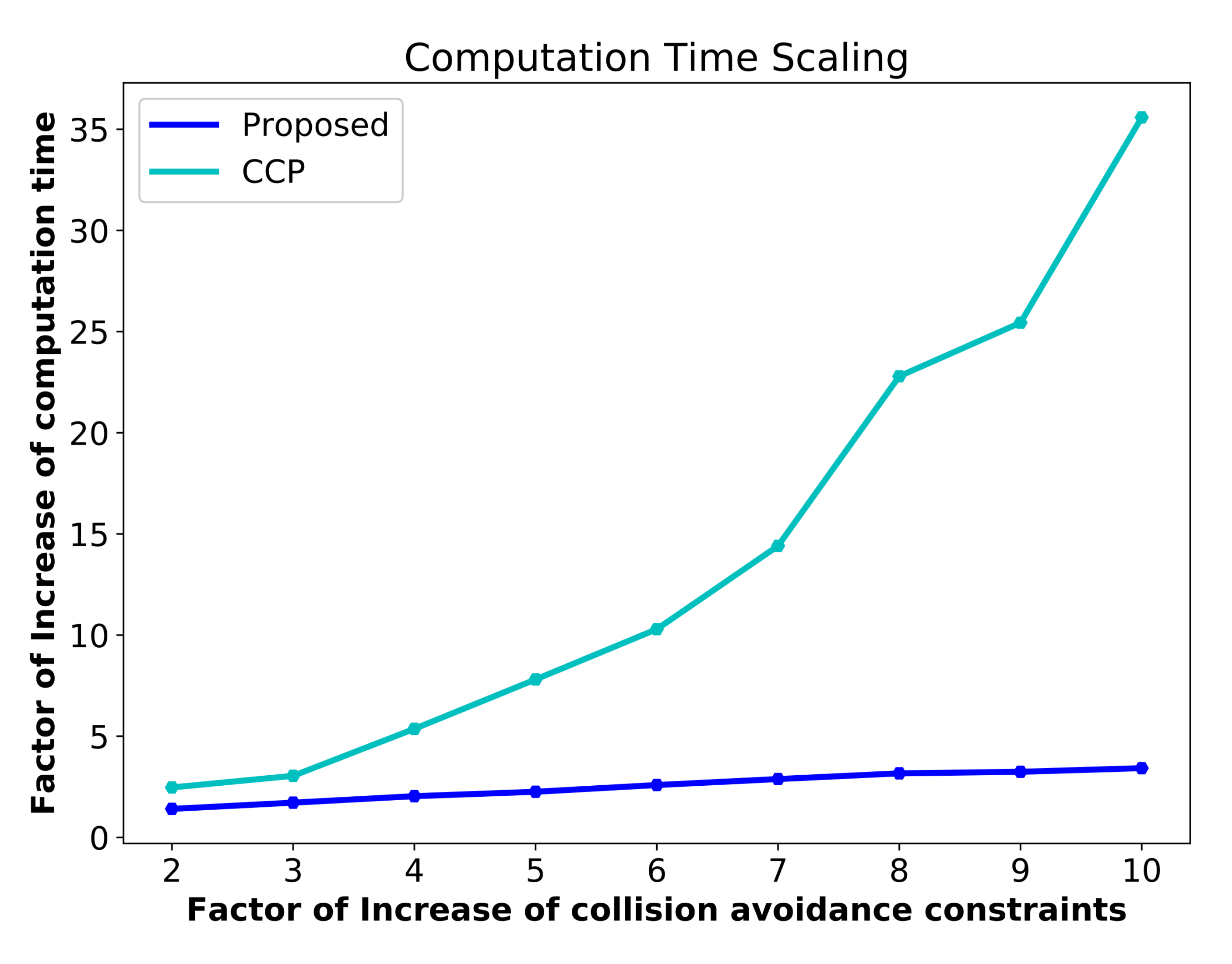}
    \vspace{-0.3cm}
\caption[Comparison of computation time scaling between our method and CCP]{\small{Scaling of computation time with the number of collision avoidance constraints: \gls{ccp} has quadratic scaling, while our optimizer shows sub-linear growth.}}
\label{comptime_scaling}
\end{figure}

\noindent\textbf{Computation Time Scaling:}
Figure \ref{comptime_scaling} reveals the second important result in this paper. It illustrates how computation time varies with an increase in the number of collision avoidance constraints. \gls{ccp} exhibits almost quadratic scaling, consistent with a similar observation presented in \cite{augugliaro2012generation} (see Figure 3 in \cite{augugliaro2012generation}). In contrast, the proposed optimizer displays sub-linear growth in computation time. This nice characteristic stems from the fact that the computation cost of the left-hand side of \eqref{step_x_2d_simplify} does not depend on the number of obstacles.


\subsection{Real-world Demonstration}
\begin{figure}[!h]
  \centering  
    \includegraphics[scale=0.47]{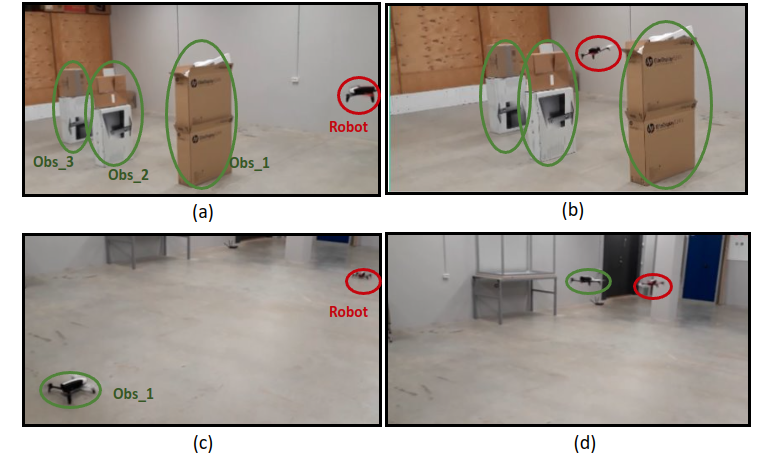}
\caption[Real-world experiment]{\small{Snapshots of real-world experiments utilizing our optimizer in both a static (a-b) and dynamic (c-d) environment. Obstacles are marked in green, and the robot is shown in red. In (c-d), the dynamic obstacle is represented by a moving robot.}}
\label{real_world_1}
\end{figure}

We demonstrated some snapshots from real-world experiments using Parrot Bebop robot \footnote{https://www.youtube.com/watch?v=_HX0fErJzQo} in Figure \ref{real_world_1}. These snapshots were obtained from the qualitative results of our proposed optimizer over two configurations, including one static and one dynamic environment. 

\section{Connections to the Rest of the Thesis}

This work serves as the foundational cornerstone for all subsequent research presented in this body of work. The conceptualization and modeling of collision avoidance constraints,  as well as additional constraints introduced in subsequent papers, draw inspiration from the intuition developed in this paper. Furthermore, the algorithms presented in our subsequent works extensively leverage the principles of the \gls{am} method, building upon the concepts and analyses outlined in this foundational work.  Key methodologies such as parametrized trajectory optimization and the augmented Lagrangian method, outlined in this foundational work, form an integral part of the analytical framework adopted in subsequent studies.

    \chapter{Paper II: Batch Trajectory Optimization Algorithm}\label{paper_2}
\section{Context} 
\begin{figure}[!h]
    \centering
    \includegraphics[scale=0.46]{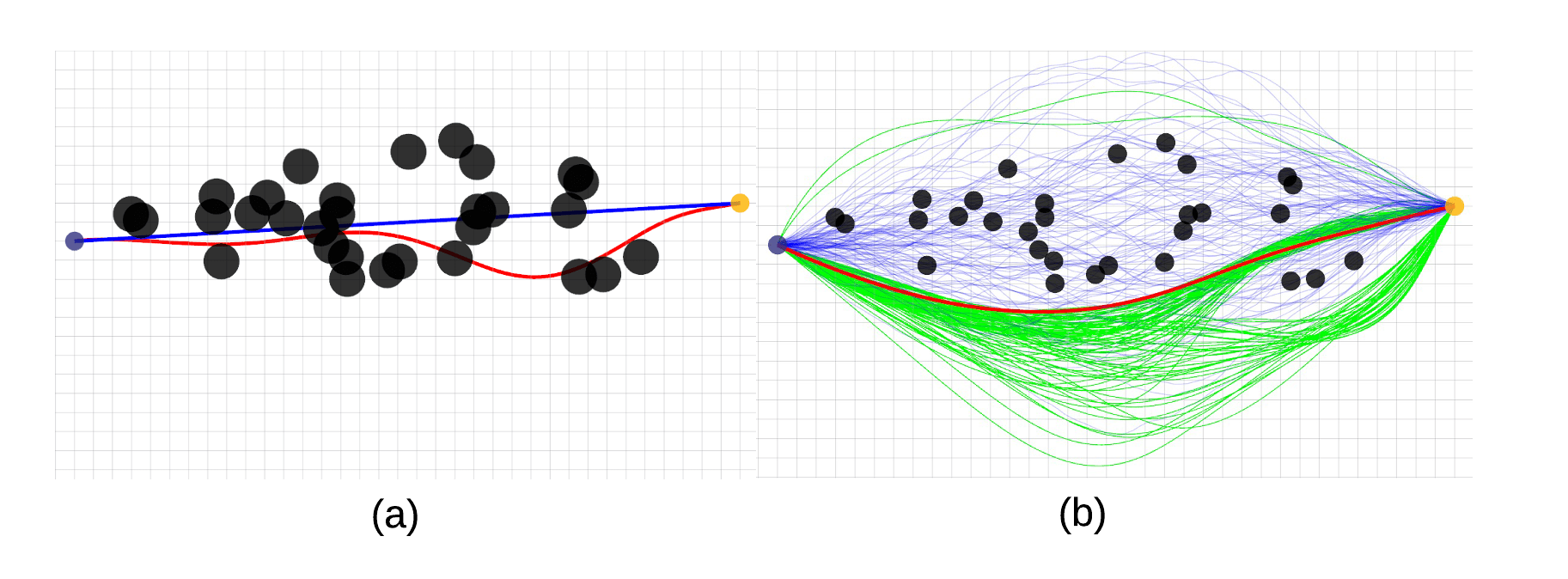}
    \vspace{-0.5cm}
    \caption[Naive initialization vs. batch initialization]{\small{(a): Naive initialization (e.g., straight blue line) may lead the trajectory optimizer to unsafe local minima. Our batch setting allows us to run hundreds of different instances of the problem in real-time, obtained by different initializations of the problem. \textcolor{black}{In (b)}, the blue trajectories represent initialization samples drawn from a Gaussian distribution \cite{kalakrishnan2011stomp}. After a few iterations, our batch optimizer returns a distribution of locally optimal trajectories (green) residing in different homotopies (best cost trajectory: red)}}
    \label{batch_inital}
\end{figure}

As discussed in Chapters \ref{literature} and \ref{intro}, a significant challenge in trajectory optimization problems is selecting an appropriate initial guess. A poor initial guess may cause the optimizer to run for a long time without converging to a solution or converging to a bad solution (see Figure \ref{batch_inital}(a)). Therefore, the focus of this chapter is to tackle this issue by considering a rather simple idea. I can initialize a trajectory from multiple initial guesses, which, for example, could be drawn from a distribution. This will lead to a distribution of locally optimal trajectories, as shown in Figure \ref{batch_inital}(b). I can then choose one of them based on their associated optimal cost value. Although, simple, the conventional wisdom suggests that this idea is unlikely to be useful as the computational cost of running several hundred trajectory optimizations could be prohibitive. This chapter essentially challenges this conventional wisdom for a class of optimization problems that cover autonomous navigation of a rectangular-shaped robot in cluttered and dynamic environments. Specifically, I propose a batch-trajectory optimizer that leverages \gls{gpu} parallelization to runs hundreds of different instances of the considered trajectory optimization problem in real-time. 

\section{Problem Formulation} 
I am interested in solving a batch of trajectory optimization for autonomous navigation. In this chapter, a slightly different variant of the problem is discussed. Specifically, I consider a rectangular shaped robot for which collision avoidance also depends on the orientation. The mathematical problem is given by:


\small
\begin{subequations}
\begin{align}
   &\min_{x_{i}(t), y_{i}(t), \psi_{i}(t)}  \sum_{t} \hspace{-0.12cm} \Big(\ddot{x}_{i}^2(t)\hspace{-0.1cm}+\hspace{-0.1cm}\ddot{y}_{i}^2(t) \hspace{-0.1cm}+ \hspace{-0.1cm}\ddot{\psi}_{i}^2(t) \hspace{-0.1cm} + \hspace{-0.1cm}(x_{i}(t) \hspace{-0.1cm}-\hspace{-0.1cm}x_{des}(t))^2
   \hspace{-0.1cm}+\hspace{-0.1cm}
   (y_{i}(t)\hspace{-0.1cm}-\hspace{-0.1cm}y_{des}(t))^2\Big)\hspace{-0.05cm},\hspace{-0.05cm}1\hspace{-0.05cm}\leq\hspace{-0.05cm}\hspace{-0.04cm}i\leq\hspace{-0.1cm}N_{b}  \label{acc_cost}\\ 
  & \text{s.t.:} \nonumber \\
   &(x_{i}(t_{0}), y_{i}(t_{0}), x_{i}(t_{f}), y_{i}(t_{f})) = \mathbf{b}\label{eq_multiagent_1}\\
   &(\psi_{i}(t_{0}), \psi_{i}(t_{f})) = \mathbf{b}_{\psi}\label{eq1_multiagent_1}\\
    &\dot{x}_{i}^{2}(t) + \dot{y}_{i}^{2}(t)  \leq v^{2}_{max}, ~~~
    \ddot{x}_{i}^{2}(t) + \ddot{y}_{i}^{2}(t) \leq a^{2}_{max} \label{acc_constraint}\\
  & 
    -\frac{\hspace{-0.07cm}(x_{i}(t)\hspace{-0.07cm}+\hspace{-0.07cm}r_{m}\hspace{-0.04cm}\cos{\psi_{i}(t)}\hspace{-0.095cm}-
 \hspace{-0.095cm}x_{o,j}(t)\hspace{-0.05cm})^{2}}{a^2}\hspace{-0.062cm} - \hspace{-0.062cm}
 \frac{\hspace{-0.072cm}(y_{i}(t) \hspace{-0.075cm}+\hspace{-0.075cm}r_{m}\sin{\psi_{i}\hspace{-0.04cm}(t)}
 \hspace{-0.075cm}-\hspace{-0.065cm}y_{o,j}(t)\hspace{-0.06cm})^{2}}{b^2}
  \hspace{-0.1cm}+\hspace{-0.1cm}1\hspace{-0.1cm}\leq \hspace{-0.1cm}0, 1\hspace{-0.05cm}\leq \hspace{-0.05cm}m\leq n_{c} \label{coll_multiagent}
\end{align}
\end{subequations}
\normalsize

\begin{figure}[!h]
    \centering
    \includegraphics[scale=0.32]{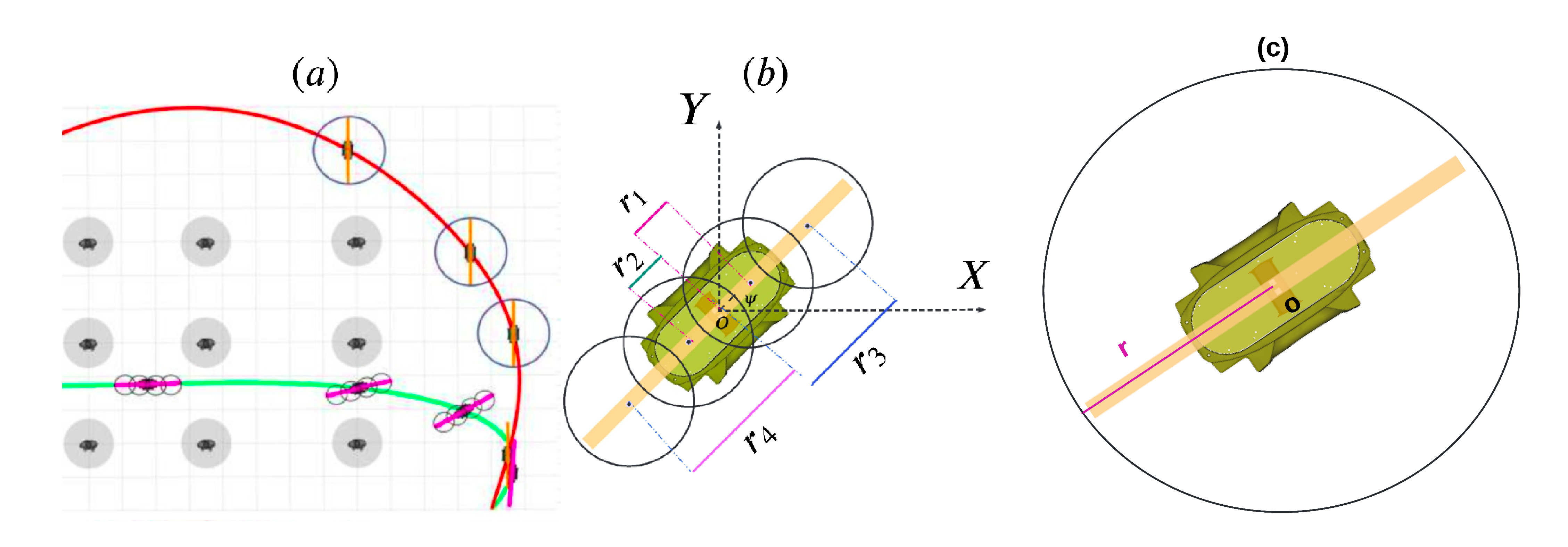}
    \caption[Modelling a robot with a rectangular footprint]{\small{(a): Robots with rectangular footprints can be modeled in two ways: a combination of circles (utilized in this work) and a single circle \cite{schwarting2017parallel}. Using a combination of overlapping circles, as depicted, enables the incorporation of rotational motions, facilitating better maneuverability in confined spaces. In contrast, the circular footprints may be overly conservative, potentially compelling the robot to take larger detours.
    (b) A rectangular footprint with center $O(t) = (x_{i}(t), y_{i}(t))$ and heading angle $\psi_{i}(t)$ is represented through combinations of four circles. (c) The robot is modeled as a single circle.}}
    \label{multi_circle}
    \end{figure}

\noindent Where the objective of the cost function \eqref{acc_cost} is to minimize the sum of squared linear and angular accelerations, as well as the tracking error from a desired position trajectory $(x_{\text{des}}(t), y_{\text{des}}(t))$ at different time instants. Here, $\psi_{i}(t)$ represents the heading angle of the robot. The vectors $\mathbf{b}$ and $\mathbf{b}_{\psi}$ in  \eqref{eq_multiagent_1} and \eqref{eq1_multiagent_1} represent the initial and final values of boundary conditions on linear and angular positions and their derivatives. We assume $v_{\text{max}}$ and $a_{\text{max}}$ as the maximum velocity and acceleration, inequality \eqref{acc_constraint} sets bounds on the total velocity and acceleration. Inequality \eqref{coll_multiagent} introduces collision avoidance constraints. The obstacle locations are defined by $(x_{o,j}(t), y_{o,j}(t))$, assumed to be axis-aligned ellipses with dimensions $(a, b)$. We consider the robot to have rectangular footprints, which we can represent as a collection of $n_{c}$ overlapping circles \cite{schwarting2017parallel}, \cite{singh2020bi}, each positioned at coordinates $(\pm r_{m}, 0)$  in the local frame. Thus, \eqref{coll_multiagent} ensures that the $m^{th}$ circle of the footprint does not overlap with the $j^{th}$ elliptical obstacle, as illustrated in Figure \ref{multi_circle}(b).

\section{Overview of the Main Algorithmic Results}

\begin{figure}[!h]
    \centering
    \includegraphics[scale=0.6]{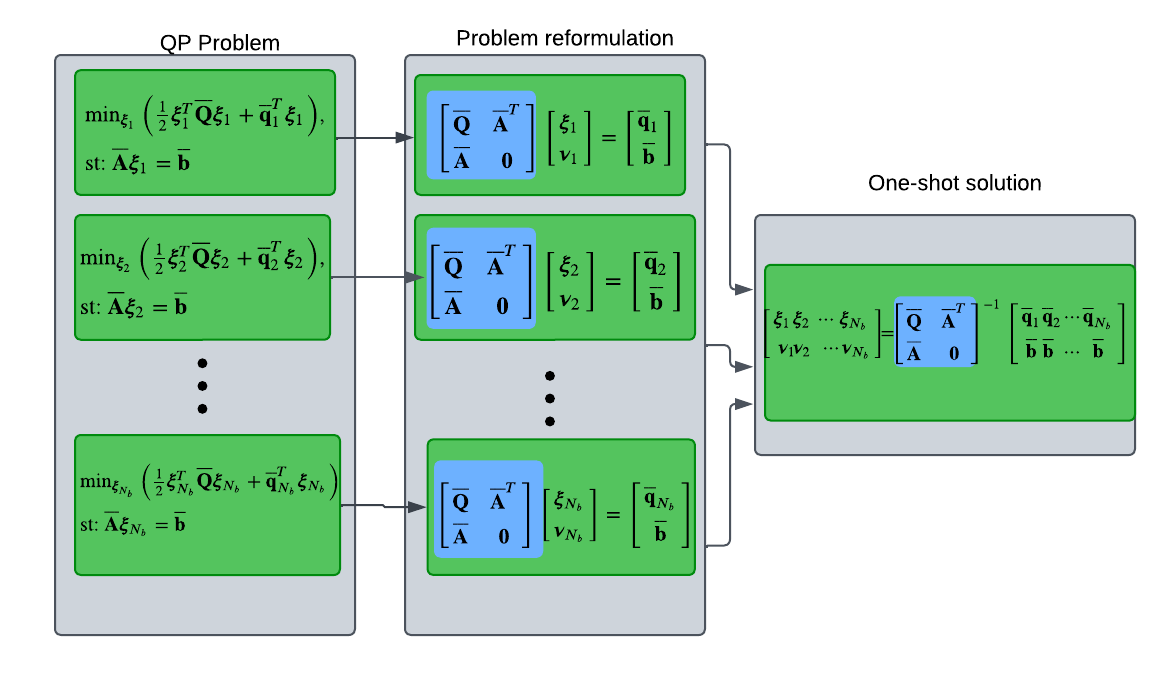}
    \vspace{-0.3cm}
    \caption[Visualization of the Main Idea of Algorithm \ref{alg_2}]{\small{Visualization of the Main Idea: Schematic representation illustrating the structure and relationships of a set of equality-constrained \gls{qp}s. The diagram demonstrates how varying vectors among instances, denoted as $\overline{\mathbf{q}}_{i}$, undergo a reduction process, revealing insights into computational simplifications. Each green box represents an individual \gls{qp} instance. Notably, the matrix inside the blue box remains constant across all \gls{qp} instances, enabling a one-time computation for subsequent matrix-vector productions.}}
    \label{qp_diagram}
    \end{figure}

Figure \ref{qp_diagram} and \ref{overview_4} illustrate how our approach addresses trajectory optimization for various initializations across \gls{gpu}s. It is also shown in Figure \ref{overview_4} how the off-the-shelf optimizers handle trajectory optimization problems using multiple initializations.  

    I show that the main computation associated with solving \eqref{acc_cost} - \eqref{coll_multiagent} can be reduced to solving a set of equality-constrained \gls{qp}s \eqref{over_1}, characterized by a distinctive structure where only the vector $\overline{\mathbf{q}}_{i}$ varies among the instances of the problem and also across the iterations of each problem's solution process.

\small
\begin{align}
\min_{\boldsymbol{\xi}_{i}} \Big(\frac{1}{2} \boldsymbol{\xi}^{T}_{i}\overline{\mathbf{Q}} \boldsymbol{\xi}_{i} + \hspace{0.1cm}^{k}\overline{\mathbf{q}}^{T}_{i} \boldsymbol{\xi}_{i}\Big), \qquad
 \text{s.t.: } \overline{\mathbf{A}} \boldsymbol{\xi}_{i} = \overline{\mathbf{b}},~~~~ i\in[1,N_{b}] \label{over_1}
\end{align}
\normalsize
    
Based on our discussion in Section \ref{qp_definition}, the $i^{th}$ optimization problem \eqref{over_1} can be reduced to a set of linear equations as

\small
\begin{align}
   \overbrace{ \begin{bmatrix}
        \overline{\mathbf{Q}} & \overline{\mathbf{A}}^{T} \\ 
        \overline{\mathbf{A}} & \mathbf{0}
    \end{bmatrix}}^{\mathbf{\bar{D}}} \begin{bmatrix}
        \boldsymbol{\xi}_{i}\\ \boldsymbol{\nu}_{i}
    \end{bmatrix} = \overbrace{\begin{bmatrix}
       ^{k} \overline{\mathbf{q}}_{i} \\ \overline{\mathbf{b}}
    \end{bmatrix}}^{\hspace{0.1cm}^{k}\bar{\boldsymbol{\chi}}_{i}} \label{over_2}
\end{align}
\normalsize

\noindent where $\boldsymbol{\nu}_{i}$ is dual optimization variable. Since the matrix in \eqref{over_2} is constant for different batch instances, and also all batches are independent of each other, I can compute solution across all the batches at a given iteration $k$ in one-shot as follows:

\small
\begin{align}
\begin{bmatrix} 
\begin{array}{@{}c|c|cc@{}}
\boldsymbol{\xi}_{1} &...& \boldsymbol{\xi}_{N_{b}} \\ \boldsymbol{\nu}_{1} &...& \boldsymbol{\nu}_{N_{b}}
\end{array}
\end{bmatrix} =
    \overbrace{(\begin{bmatrix}
        \overline{\mathbf{Q}} & \overline{\mathbf{A}}^{T} \\ 
        \overline{\mathbf{A}} & \mathbf{0}
    \end{bmatrix}}^{constant})^{-1}
\begin{bmatrix}
\begin{array}{@{}c|c|cc@{}}
    ^{k}\overline{\mathbf{q}}_{1} & ... & ^{k}\overline{\mathbf{q}}_{N_{b}}  \\
    \overline{\mathbf{b}} & ... & \overline{\mathbf{b}}
    \end{array}\end{bmatrix},
    \label{over_3}
\end{align}
\normalsize

\noindent where $|$ implies that the columns are stacked horizontally. As can be seen, the optimization problem is reduced to the Fig \ref{overview_4} form. Additionally, for providing a graphical representation of the concepts discussed, I visualize each of \gls{qp}s for all initial guesses in Figure \ref{qp_diagram}. 

\noindent 

The main feature of this reduced problem is the matrix $\overline{\mathbf{D}}$ is fixed for both iterations and all initial guesses. As a result:

    \begin{itemize}
        \item The factorization/inverse of $\overline{\mathbf{D}}$ can be computed once and used across all iterations and all batches. 
        \item The size of matrix $\overline{\mathbf{D}}$ does not change with the number of constraints and batches and depends on the planning horizon. Thus, by increasing the number of obstacles or batches, the computation cost for factorization of $\overline{\mathbf{D}}$ is fixed.
        \item The structure of the reduced problem is just a matrix-vector product that is compatible with \gls{gpu} architecture. Thus, all the trajectories can be computed in parallel over \gls{gpu}  
        \item Computing trajectories over \gls{gpu} accelerates computations and provides real-time solutions. 
    \end{itemize}

In addition to the main algorithmic features of this work, I utilized a combination of circles to model the robot footprint instead of a single circle. This modeling: 
\begin{itemize}
    \item Improves maneuverability, especially in narrow spaces. For instance, as shown in Figure \ref{multi_circle}, representing the robot with multiple circles provides a less conservative space compared to using just one circle.
    \item Provides information about the heading angle of the robot.
\end{itemize}  

In the next sections, I will first outline the existing works and the advantages of our work over \gls{sota} methods. Then, I explain the main results in detail and validate the proposed method through several benchmarks.

\section{Connections to Existing Works on Batch Trajectory Optimization}

\begin{figure}[!h]
    \hspace{-0.7cm}
    \includegraphics[scale=0.25]{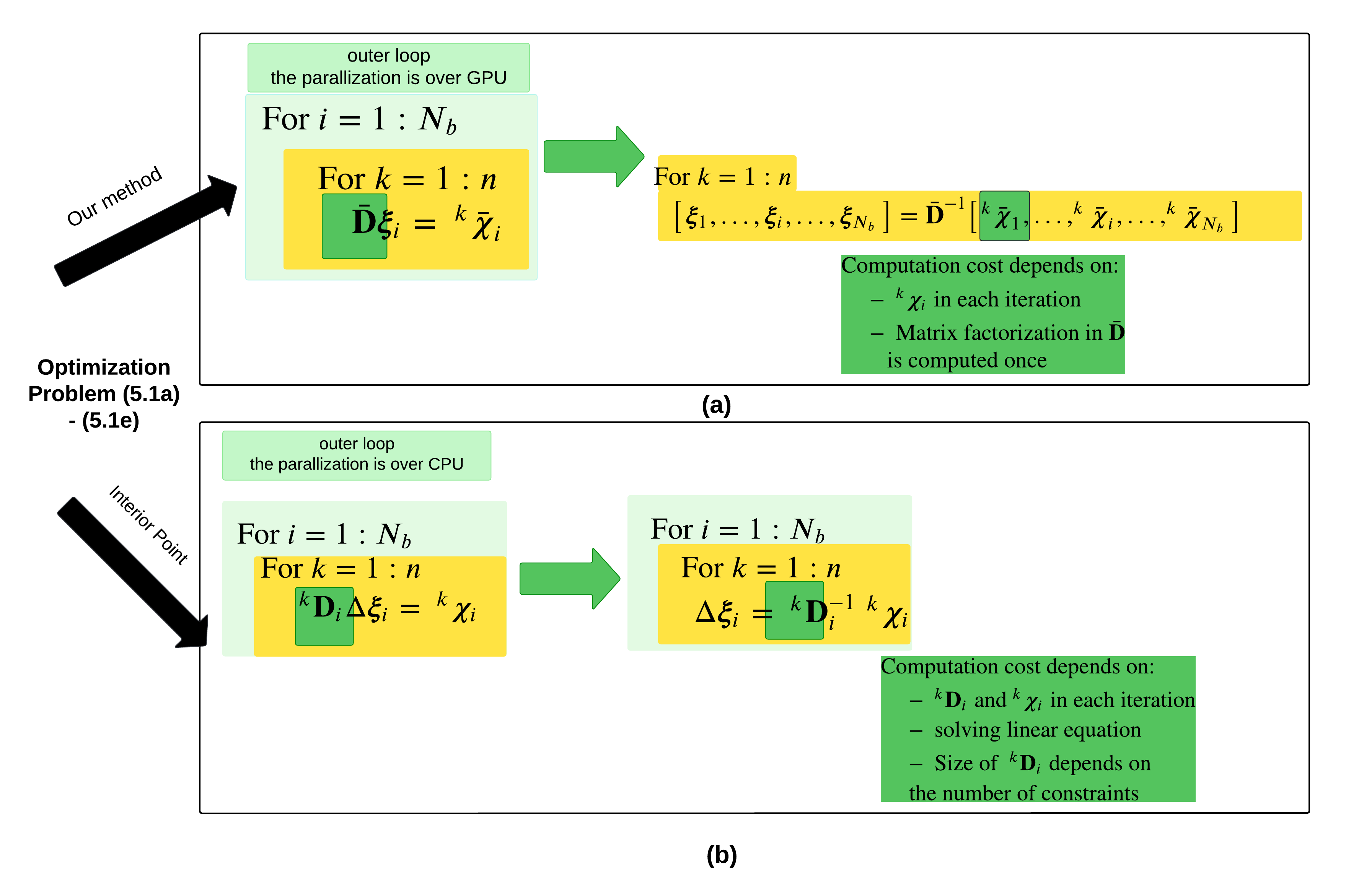}
    \caption[Overview of batch trajectory optimization]{\small{Overview of batch trajectory optimization vs Interior-Point method}}
    \label{overview_4}
    \end{figure}

In this section, I explain how the vectorized structure in our proposed optimizer helps vis-a-vis a baseline approach. It is possible to use any off-the-shelf optimizer across parallel CPU threads in order to run it from different intializations. For example, if one uses an interior-point method, then the parallel/batch trajectory optimization takes the form of the pipeline shown in Figure \ref{overview_4} (b). As can be seen, each initialization would lead to a different set of matrix $\mathbf{D}_i$, which also changes across each iteration $k$. The outer loop can be parallelized across CPU threads/cores. However, the number of threads in CPUs is limited.

In contrast, for the proposed approach shown in Figure \ref{overview_4} (a), I solve linear equations for which the matrix part does not change either across iterations or across different initialization batches. This allows for much better parallelization opportunities across GPU cores that are much larger in number than \gls{cpu} threads. In fact, \gls{gpu} parallelization of computational that can be inherently vectorized across batch instances forms the core of the modern deep learning algorithms.

An alternated competing approach was proposed in \cite{de2005tutorial} that combined \gls{gd} with the Cross-Entropy Method \cite{botev2013cross}. This approach benefits from the fact that GD is easily batchable/parallelizable across GPU cores. Our batch optimizer can enable methods such as the one proposed in \cite{botev2013cross} to utilize a more powerful optimizer than \gls{gd}, thereby enhancing the overall performance. However, it is important to note that \gls{gd} method still suffers from the limitations mentioned in section \ref{gd_methods}, especially for high-dimensional problems.

\section{Advantages Over SOTA Methods in Navigtion Performance}



The proposed work advances the \gls{sota} in several key aspects. Firstly, the parallel initialization naturally identifies a range of locally optimal trajectories within various homotopies. Secondly, I enhance navigation quality, including success rates and tracking performance, compared to the baseline approach, which relies on computing a single locally optimal trajectory at each control loop. Lastly, I demonstrate that when initialized with trajectory samples from a Gaussian distribution, the proposed batch optimizer surpasses the performance of the \gls{sota} \gls{cem} \cite{rubinstein1997optimization, de2005tutorial} in terms of solution quality.

\section{Main Results}\label{main_results_2}

\noindent \textbf{Reformulated constraints:} To reach this aim, I extend the polar/spherical representation of collision avoidance constraints discussed in Section \ref{main_result_1}, and reformulate \eqref{coll_multiagent} in the form $\mathbf{f}_c = \mathbf{0}$, where

\small
\begin{align}
\hspace{-0.2cm}\colorbox{my_green}{$ \mathbf{f}_{c}(x_{i}(t),y_{i}(t),\psi_{i}(t))\hspace{-0.1cm}=\hspace{-0.1cm}
\hspace{-0.06cm}\left \{ \hspace{-0.23cm}\begin{array}{lcr}
x_{i}(t) \hspace{-0.04cm}+ \hspace{-0.05cm}r_{m} \cos{\psi_{i}}(t)\hspace{-0.07cm}-\hspace{-0.08cm}x_{o,j}(t) \hspace{-0.07cm}-ad_{mj,i}(t)\cos{\alpha_{mj,i}}(t) \\
y_{i}(t)\hspace{-0.04cm} + \hspace{-0.05cm}r_{m} \sin{\psi_{i}}(t)\hspace{-0.07cm}-\hspace{-0.07cm}y_{o,j}(t) 
\hspace{-0.08cm}-bd_{mj,i}(t)\sin{\alpha_{mj,i}(t)}
\end{array} \hspace{-0.25cm} \right \}\hspace{-0.10cm}, $}\nonumber \\  d_{mj,i}(t)\hspace{-0.10cm} \geq \hspace{-0.07cm}1 
\label{collision_1} 
\end{align}
\normalsize

\noindent where $d_{mj,i}(t)$ and $\alpha_{mj,i}(t)$ are the line-of-sight distance and angle between the $m^{th}$ circle of the robot and $j^{th}$ obstacle. As can be seen, \eqref{collision_1} exhibits convexity within the space of $(x_{i}(t), y_{i}(t))$ and $ (\cos{\psi}_{i}(t), \sin{\psi}_{i}(t))$. It should be mentioned that this convex characteristic later becomes advantageous in leveraging the proposed optimization problem. 

Similarly, I reformulate the velocity and acceleration constraints in the form
$ \mathbf{f}_{v} =\mathbf{0} $ and $\mathbf{f}_{a} = \mathbf{0}$, where 

\vspace{-0.22cm}
\small
\begin{subequations} 
\begin{align}
 \colorbox{my_pink}{$ \mathbf{f}_{v} =\left \{ \hspace{-0.20cm} \begin{array}{lcr}
 \dot{x}_{i}(t) - d_{v,i}(t)v_{max}\cos{\alpha_{v,i}(t)}  \\
 \dot{y}_{i}(t) - d_{v,i}(t)v_{max}\sin{\alpha_{v,i}(t)} 
\end{array} \hspace{-0.20cm} \right \} $}
,d_{v,i}(t) \leq 1 , ~  \forall t,i  \label{fv} \\
\colorbox{my_blue}{$ \mathbf{f}_{a} = \left \{ \hspace{-0.20cm}\begin{array}{lcr}
\ddot{x}_{i}(t) - d_{a,i}(t)a_{max}\cos{\alpha_{a,i}(t)} \\
\ddot{y}_{i}(t) - d_{a,i}(t)a_{max}\sin{\alpha_{a,i}(t)}
\end{array} \hspace{-0.20cm} \right \}$},  d_{a,i}(t) \leq 1 , ~\forall t,i \label{fa}
\end{align}
\end{subequations}
 \normalsize

\noindent where $d_{v,i},d_{a,i},\alpha_{v,i}(t)$ and $\alpha_{a,i}(t)$ are variables based on polar/spherical representation. Similar to \eqref{collision_1}, \eqref{fv} and \eqref{fa} are also convex in the space of $(\sin{\alpha_{v,i}(t)},\cos{\alpha_{v,i}(t)})$, $d_{v,i}$, $(\sin{\alpha_{a,i}(t)},\cos{\alpha_{a,i}(t)})$ and $d_{a,i}$.

The insights gained from the convex features of \eqref{collision_1}-\eqref{fa} lead us to reformulate the trajectory optimization problem \eqref{acc_cost}-\eqref{coll_multiagent} as follows:

\vspace{-0.22cm}
\small
\begin{subequations}
\begin{align}
&\min_{\scalebox{0.7}{$\begin{matrix} x_{i}(t), y_{i}(t), \psi_{i}(t),\\c_{\psi,i}(t),s_{\psi,i}(t),d_{v,i}(t),
 d_{a,i}(t), \\d_{mj,i}(t),\alpha_{v,i}(t),\alpha_{a,i}(t),\alpha_{mj,i}(t)    \end{matrix}$}}
\hspace{-0.9cm}\sum_{t}\hspace{-0.1cm}\Big(\ddot{x}^2_{i}(t)\hspace{-0.1cm}+\hspace{-0.1cm}\ddot{y}^2_{i}(t) +\hspace{-0.1cm}\ddot{\psi}_{i}^2(t) \hspace{-0.1cm} + \hspace{-0.1cm}(x_{i}(t) \hspace{-0.1cm}-\hspace{-0.1cm}x_{des}(t))^2
   \hspace{-0.1cm}+\hspace{-0.1cm}
   (y_{i}(t)\hspace{-0.1cm}-\hspace{-0.1cm}y_{des}(t))^2\Big)   \label{acc_cost1}\\
   &\text{s.t.:} \nonumber \\
& (x_{i}(t_{0}), y_{i}(t_{0}), x_{i}(t_{f}), y_{i}(t_{f})) = \mathbf{b}\label{eq_multiagent}\\
 &  (\psi_{i}(t_{0}), \psi_{i}(t_{f})) = \mathbf{b}_{\psi}\label{eq1_multiagent}\\
&\colorbox{my_pink}{$ \mathbf{f}_{v}(t) = \mathbf{0}, $} \qquad d_{v,i}(t)  \leq 1,   \forall t,i \label{cond1}  \\
&\colorbox{my_blue}{$ \mathbf{f}_{a}(t) = \mathbf{0}$},  \qquad d_{a,i}(t) \leq 1, \forall t,i \label{fa_fv1} \\
&\colorbox{my_yellow}{$  c_{\psi,i}(t) = \cos{\psi_{i}(t)}$} \label{cos_ref}\\ 
& \colorbox{my_purple}{$ s_{\psi,i}(t) =  \sin{\psi_{i}(t)}$} \label{sin_ref}\\
& \colorbox{my_green}{$ 
\mathbf{f}_{c} \hspace{-0.1cm}: \hspace{-0.1cm}\left \{ \begin{array}{lcr}\hspace{-0.2cm}
x_{i}(t) + r_{m} c_{\psi,i}(t)-x_{o,j}(t) -ad_{mj,i}(t)\cos{\alpha_{mj,i}}(t) =0\\
\hspace{-0.2cm}y_{i}(t) + r_{m} s_{\psi,i}(t)-y_{o,j}(t) 
-bd_{mj,i}(t)\sin{\alpha_{lj,i}(t)} =0
\end{array} \right \} $}\hspace{-0.2cm}\label{coll_con}
 \\ &d_{mj,i}(t) \geq 1, \label{fc1}
\end{align}
\end{subequations}
\normalsize

\noindent where two new slack variables, $c_{\psi,i}(t)$ and $s_{\psi,i}(t)$ are simply the copy of $\cos{\psi_{i}(t)}$ and $\sin{\psi_{i}(t)}$. Our main trick is to treat $c_{\psi,i}(t)$ and $s_{\psi,i}(t)$
as independent variables and somehow ensure that when the optimization converges, they indeed resemble the cosine and sine of $\psi_{i}(t)$. 

The above optimization is defined in terms of time-dependent functions. To ensure smoothness in trajectories,  using \eqref{parameter}, I parameterize the optimization variables, $x_{i}(t), y_{i}(t), z_{i}(t), \psi_{i}(t), c_{\psi,i}(t)$ and $s_{\psi,i}(t)$. In our parametrization, $\boldsymbol{\xi}_{x,i}, \boldsymbol{\xi}_{y,i}, \boldsymbol{\xi}_{\psi,i}, \boldsymbol{\xi}_{c,i}$ and $\boldsymbol{\xi}_{s,i}$ represent the coefficients of the polynomial. 



Now, using \eqref{parameter} and some simplifications, the problem \eqref{acc_cost1}-\eqref{fc1} can be formulated as 

\small
\begin{subequations}
\begin{align}
 &\min_{\hspace{0.7cm}\boldsymbol{\xi}_{i}, \boldsymbol{\alpha}_{i}, \mathbf{d}_{i}, \boldsymbol{\xi}_{\psi,i}} \hspace{-0.1cm}\Big(\frac{1}{2}\boldsymbol{\xi}_{i}^T\mathbf{Q}\boldsymbol{\xi}_{i}  + \mathbf{q}^T\boldsymbol{\xi}_{i}+ \frac{1}{2}\boldsymbol{\xi}_{\psi,i}^{T}\ddot{\mathbf{P}}^{T}\ddot{\mathbf{P}}\boldsymbol{\xi}_{\psi,i} \Big)\label{reform_cost}  \\
 &\hspace{0.7cm}\text{s.t.:} \nonumber \\
 &\hspace{0.7cm} \mathbf{A}\boldsymbol{\xi}_{i} = \mathbf{b}, ~~~ \mathbf{A}_{\psi}\boldsymbol{\xi}_{\psi,i} = \mathbf{b}_{\psi}   \label{reform_eq}\\
&\hspace{0.7cm}  \mathbf{F}\boldsymbol{\xi}_{i} = \mathbf{g}_{i}(\boldsymbol{\alpha}_{i},\boldsymbol{\xi}_{\psi,i},\mathbf{d}_{i} ) \label{reform_bound}\\
 &\hspace{0.7cm}\mathbf{d}_{min}\leq \mathbf{d}_{i} \leq \mathbf{d}_{max}. \label{d_bound}
\end{align}
\end{subequations}
\normalsize

\noindent where, $\boldsymbol{\xi}_{i} =(\boldsymbol{\xi}_{x,i}, \boldsymbol{\xi}_{c,i}, \boldsymbol{\xi}_{y,i}, \boldsymbol{\xi}_{s,i}) $, $\boldsymbol{\xi}_{\psi,i}$, 
$\boldsymbol{\alpha}_{i} = (\boldsymbol{\alpha}_{mj,i}, \hspace{0.1cm} \boldsymbol{\alpha}_{v,i}, \hspace{0.1cm} \boldsymbol{\alpha}_{a,i})$ and $\mathbf{d}_{i} = (\mathbf{d}_{mj,i},\mathbf{d}_{v,i},\mathbf{d}_{a,i})$ are optimization variables to be obtained. The matrix $\mathbf{A}$ is generated by stacking the first and last row of $\mathbf{P}$ and their derivations corresponding to equality boundaries. Similarly, the matrix $\mathbf{A}_{\psi}$ is generated by stacking the first and last row of $\mathbf{P}$. The constant vectors, $\mathbf{d}_{min}$ and $\mathbf{d}_{max}$ are formed by stacking the lower $([1,0,0])$ and upper bounds $([\infty,1,1])$ of $\mathbf{d}_{mj,i}, \mathbf{d}_{v,i}, \mathbf{d}_{a,i}$. The matrix and vector $\mathbf{Q}$ and $\mathbf{q}$ are used to convert the acceleration and tracking cost in \eqref{acc_cost1} into \gls{qp} problem and can be defined as

\small
\begin{align}
\mathbf{Q} = \begin{bmatrix}
\ddot{\mathbf{P}}^T\ddot{\mathbf{P}}+\mathbf{P}^{T}\mathbf{P}&&&\\
& 0&&\\
&&\ddot{\mathbf{P}}^T\ddot{\mathbf{P}}+\mathbf{P}^{T}\mathbf{P}&\\
&&&0
\end{bmatrix} ~~~ \mathbf{q} = \begin{bmatrix}
-\mathbf{P}^T\mathbf{x}_{des}\\
\mathbf{0}\\
-\mathbf{P}^T\mathbf{y}_{des}\\
\mathbf{0}\\
\end{bmatrix}
\label{Q_matrix}
\end{align}
\normalsize

\noindent where $\mathbf{x}_{des}$ and $\mathbf{y}_{des}$ are obtained by stacking $x_{des}(t)$ and $y_{des}(t)$ at different time instances. Also, the matrix $\mathbf{F}$ and vector $\mathbf{g}_{i}$  in \eqref{reform_bound} are obtained by rewriting constraints  \eqref{cond1} - \eqref{coll_con} in the following manner.

\small
\begin{align}
\overbrace{ \begin{bmatrix} 
\colorbox{my_pink}{$\begin{bmatrix}
\dot{\mathbf{P}}  &  \mathbf{0}  \\
\end{bmatrix} $}& \mathbf{0}   \\
\colorbox{my_blue}{$
\begin{bmatrix}
\ddot{\mathbf{P}}  &  \mathbf{0}  \\
\end{bmatrix}$} & \mathbf{0}    \\
  \colorbox{my_green}{$\mathbf{A}_{o}$} & \mathbf{0} \\
\colorbox{my_yellow}{$\begin{bmatrix} \mathbf{0} & \mathbf{P}
\end{bmatrix} $}& \mathbf{0} \\
\mathbf{0} & \colorbox{my_pink}{$ \begin{bmatrix}
\dot{\mathbf{P}}  &  \mathbf{0}  \\
\end{bmatrix} $} \\
\mathbf{0} &  \colorbox{my_blue}{$\begin{bmatrix}
\ddot{\mathbf{P}}  &  \mathbf{0}  \\
\end{bmatrix} $}\\
\mathbf{0} & \colorbox{my_green}{$\mathbf{A}_{o}$} \\
\mathbf{0} & \colorbox{my_purple}{$\begin{bmatrix} \mathbf{0} & \mathbf{P}
\end{bmatrix}$}
\end{bmatrix}}^{\mathbf{F}}
\begin{bmatrix}   \boldsymbol{\xi}_{x,i} \\
\boldsymbol{\xi}_{c,i} \\ 
\boldsymbol{\xi}_{x,i} \\
\boldsymbol{\xi}_{s,i}
\end{bmatrix} =
\overbrace{\begin{bmatrix}
\colorbox{my_pink}{$\mathbf{d}_{v,i}\cos(\boldsymbol{\alpha}_{v,i})$} \\
\colorbox{my_blue}{$\mathbf{d}_{a,i}\cos(\boldsymbol{\alpha}_{a,i}) $}\\
\colorbox{my_green}{$\mathbf{b}_{o_{1,i}}(\mathbf{d}_{mj,i},\boldsymbol{\alpha}_{mj,i})$} \\
\colorbox{my_yellow}{$\cos{\boldsymbol{\psi}_{i}}$} \\
\colorbox{my_pink}{$\mathbf{d}_{v,i}\sin(\boldsymbol{\alpha}_{v,i})$} \\
\colorbox{my_blue}{$\mathbf{d}_{a,i}\sin(\boldsymbol{\alpha}_{a,i})$} \\
 \colorbox{my_green}{$\mathbf{b}_{o_{2,i}}(\mathbf{d}_{mj,i},\boldsymbol{\alpha}_{mj,i})$} \\
 \colorbox{my_purple}{$\sin{\boldsymbol{\psi}_{i}} $}
\end{bmatrix}}^{\mathbf{g}_{i}}
\end{align}

\noindent where 

\small
\begin{align}
\mathbf{b}_{o_{1,i}}&=\colorbox{my_green}{$\begin{bmatrix}
\mathbf{x}_{o,j} \\ \vdots  \\ \mathbf{x}_{o,j} 
\end{bmatrix}
+ a \begin{bmatrix}
\mathbf{d}_{1j,i} \hspace{0.1 cm}\cos{\boldsymbol{\alpha}_{1j,i}} \\
\vdots\\
\mathbf{d}_{mj,i} \hspace{0.1 cm}\cos{\boldsymbol{\alpha}_{mj,l}}
\end{bmatrix}$} , 
\colorbox{my_green}{$\mathbf{A}_{o}\hspace{-0.07cm} = \hspace{-0.07cm}\begin{bmatrix}
\mathbf{P}  &  r_{1}\mathbf{P}  \\
\vdots  & \vdots  \\
\mathbf{P}  & r_{m}\mathbf{P}\\
\end{bmatrix}$},
 \nonumber \\
\mathbf{b}_{o_{2,l}} &= \colorbox{my_green}{$\begin{bmatrix} 
\mathbf{y}_{o,j} \\ \vdots  \\ \mathbf{y}_{o,j} 
\end{bmatrix}
+ b \begin{bmatrix}
\mathbf{d}_{1j,i}\sin{\boldsymbol{\alpha}_{1j,i}} \\
\vdots\\
\mathbf{d}_{mj,i}\sin{\boldsymbol{\alpha}_{mj,i}}
\end{bmatrix}$}
\end{align}
\normalsize

\noindent and $\mathbf{d}_{v,i}$, $\mathbf{d}_{a,i}$, $\mathbf{d}_{mj,i}$, are constructed by stacking $d_{v,i}(t)$, $d_{a,i}(t)$, and $d_{mj,i}(t)$ at different time instances. Similar derivation is used for generating $\mathbf{x}_{o,j}, \mathbf{y}_{o,j}$, $\boldsymbol{\alpha}_{v,i}$, $\boldsymbol{\alpha}_{a,i}$, $\boldsymbol{\alpha}_{mj,i}$ and $\boldsymbol{\psi}_{i}$ as well. 

To show how the proposed reformulation \eqref{reform_cost}-\eqref{d_bound} offers a computational advantage over \eqref{acc_cost}-\eqref{coll_multiagent}, an additional layer of simplification is necessary. So, using the augmented Lagrangian method, I relax the non-convex equality constraints \eqref{reform_bound} as $l_2$ penalties.

\small
\begin{align}
&\min_{\boldsymbol{\xi}_{i}, \boldsymbol{\alpha}_{i}, \mathbf{d}_{i}, \boldsymbol{\xi}_{\psi,i}} 
 \Big(\frac{1}{2}\boldsymbol{\xi}_{i}^T\mathbf{Q}\boldsymbol{\xi}_{i}+\mathbf{q}^T\boldsymbol{\xi}_{i}
 +\frac{1}{2}\boldsymbol{\xi}_{\psi,i}^T\ddot{\mathbf{P}}^{T}\ddot{\mathbf{P}}\boldsymbol{\xi}_{\psi,i} 
 - \langle\boldsymbol{\lambda}_{i}, \boldsymbol{\xi}_{i}\rangle 
  - \langle\boldsymbol{\lambda}_{\psi,i}, \boldsymbol{\xi}_{\psi,i}\rangle\nonumber \\
 &\hspace{0.8cm}+\frac{\rho}{2}\left\Vert \mathbf{F} \boldsymbol{\xi}_{i}  
 -\mathbf{g}(\boldsymbol{\xi}_{\psi,i},\boldsymbol{\alpha}_{i},\mathbf{d}_{i} ) \right \Vert_2^2 \Big),
 \label{reform_formulation}
 \end{align}
\normalsize

\noindent where the vectors $\boldsymbol{\lambda}_{i}$ and $\boldsymbol{\lambda}_{\psi,i}$ are known as the Lagrange multipliers and are crucial for ensuring that the $l_2$ penalties of the equality constraints are driven to zero.

Upon careful examination of \eqref{reform_formulation}, it becomes apparent that:

\begin{itemize}
    \item For a given $\boldsymbol{\xi}_{\psi,i},\boldsymbol{\alpha}_{i}$, and $\mathbf{d}_{i}$, \eqref{reform_formulation} is convex in the space of $\boldsymbol{\xi}_{i}$.
    \item  For a given $\boldsymbol{\xi}_{i},\boldsymbol{\alpha}_{i}$, and $\mathbf{d}_{i}$, \eqref{reform_formulation} is non-convex in terms of $\boldsymbol{\xi}_{\psi,i}$, but it can be replaced with a convex surrogate from \cite{singh2020bi}.
    \item For a given $\boldsymbol{\xi}_{\psi,i},\boldsymbol{\xi}_{i}$, and $\mathbf{d}_{i}$, \eqref{reform_formulation} is solvable in a closed-form for $\boldsymbol{\alpha}_{i}$.
    \item For a given $\boldsymbol{\xi}_{\psi,i},\boldsymbol{\alpha}_{i}$, and $\boldsymbol{\xi}_{i}$, \eqref{reform_formulation} is convex in the space of $\mathbf{d}_{i}$ and has a closed-form solution.
\end{itemize}

Motivated by the above discussion, I adopt an \gls{am} approach for minimizing \eqref{reform_formulation}. At each step of AM, I only optimize over only one of the optimization variables and the rest of them are considered fixed. Algorithm \ref{alg_2} explains the steps in solving \eqref{reform_formulation}.

\begin{algorithm}[!h]
\DontPrintSemicolon
\SetAlgoLined
\SetNoFillComment
\scriptsize
\caption{\scriptsize Proposed Batch Optimizer Algorithm for the $i^{th}$ agent }\label{alg_2}
\KwInitialization{Initiate $^{k}\mathbf{d}_{i}$, $^{k}\boldsymbol{\alpha}_{i}$, $^{k}\boldsymbol{\xi}_{\psi,i}$ at $k=0$  }
\While{$k \leq maxiter$}{
\hspace{-0.15cm}Update ${^{k+1}}\boldsymbol{\xi}_{i}$ through 
\vspace{-0.3cm}
\begin{align}
\hspace{-0.5cm}{^{k+1}}\boldsymbol{\xi}_{i}\hspace{-0.1cm}=\hspace{-0.1cm}\min_{\boldsymbol{\xi}_{i}}\hspace{-0.07cm}\Big(\frac{1}{2}\boldsymbol{\xi}_{i}^T\mathbf{Q}\boldsymbol{\xi}_{i} 
 \hspace{-0.1cm}-\hspace{-0.1cm}\langle {^{k}}\boldsymbol{\lambda}_{i}, \boldsymbol{\xi}_{i}\rangle 
\hspace{-0.1cm} +\hspace{-0.1cm}\frac{\rho}{2}\left\Vert \mathbf{F}\boldsymbol{\xi}_{i}  
 \hspace{-0.1cm}-\hspace{-0.1cm}\mathbf{g}({^{k}}\boldsymbol{\xi}_{\psi,i},{^{k}}\mathbf{d}_{i}, {^{k}}\boldsymbol{\alpha}_{i}) \right \Vert_2^2 
  \Big )
  \label{zeta_1_sepration_1}, \text{s.t.:}~\mathbf{A}\boldsymbol{\xi}_{i}\hspace{-0.1cm}= \hspace{-0.1cm}\mathbf{b}
 \end{align}

\hspace{-0.15cm}Update ${^{k+1}}\boldsymbol{\xi}_{\psi,i}$ through
\vspace{-0.3cm}
\begin{align}
{^{k+1}}{\boldsymbol{\xi}_{\psi,i}} &= \min_{\boldsymbol{\xi}_{\psi,i}}\Big(\frac{1}{2}
\boldsymbol{\xi}_{\psi,i}^T\ddot{\mathbf{P}}^{T}\ddot{\mathbf{P}}\boldsymbol{\xi}_{\psi,i}
 - \langle^{k}\boldsymbol{\lambda}_{\psi,i},
\boldsymbol{\xi}_{\psi,i}\rangle  
 \nonumber \\&~~+ \frac{\rho}{2}\left\Vert \mathbf{F}^{k+1}\boldsymbol{\xi}_{i}\hspace{-0.1cm} - \hspace{-0.05cm}
 \mathbf{g}\hspace{-0.02cm}(\boldsymbol{\xi}_{\psi,i},\hspace{-0.14cm}{^{k}}\boldsymbol{\alpha}_{i},\hspace{-0.14cm}{^{k}}\mathbf{d}_{i},\hspace{-0.14cm})\right \Vert_2^2
 \Big), \text{s.t.: }\mathbf{A}_{\psi}\boldsymbol{\xi}_{\psi,i} \hspace{-0.1cm}=\hspace{-0.07cm} \mathbf{b}_{\psi} \label{compute_psi_tilta}
\end{align}

\hspace{-0.15cm}Update ${^{k+1}}\boldsymbol{\alpha}_{i}$ through
\vspace{-0.3cm}
\begin{align}
\hspace{-0.7cm}{^{k+1}}\boldsymbol{\alpha}_{i} \hspace{-0.1cm} = \hspace{-0.05cm}\min_{\boldsymbol{\alpha}_{i}}\hspace{-0.05cm}\Big(\hspace{-0.05cm} \frac{\rho}{2}\left\Vert \mathbf{F}^{k+1} \boldsymbol{\xi}_{i}  
 \hspace{-0.1cm}-\hspace{-0.05cm}\mathbf{g}(^{k+1}\boldsymbol{\xi}_{\psi,i}, \boldsymbol{\alpha}_{i},\hspace{-0.13cm}^{k} \mathbf{d}_{i} ) \right \Vert_2^2 \Big) 
 \label{compute_ze_3}
\end{align}

\hspace{-0.15cm}Update ${^{k+1}}\mathbf{d}_{i}$ through
\vspace{-0.3cm}
\begin{align}
\hspace{-0.65cm}{^{k+1}}\mathbf{d}_{i} \hspace{-0.1cm}= \hspace{-0.05cm}\min_{\mathbf{d}_{i}}\Big(\hspace{-0.05cm} \frac{\rho}{2}\left\Vert \mathbf{F}^{k+1} \boldsymbol{\xi}_{i}  \hspace{-0.05cm}
 -
 \hspace{-0.05cm}\mathbf{g}(^{k+1}\boldsymbol{\xi}_{\psi,i}, \hspace{-0.2cm}^{k+1}\boldsymbol{\alpha}_{i}, \mathbf{d}_{i}) \right \Vert_2^2 \Big)
 \label{compute_ze_4}
\end{align}

\hspace{-0.15cm}Update Lagrange multiplier coefficient through
\vspace{-0.3cm}
\begin{align}
&{^{k+1}}\boldsymbol{\lambda}_{i}\hspace{-0.09cm}=\hspace{-0.09cm}{^{k}}\boldsymbol{\lambda}_{i} \hspace{-0.09cm}- \hspace{-0.09cm}\rho( \mathbf{F}^{k+1}\boldsymbol{\xi}_{i} 
 -\hspace{-0.09cm}\mathbf{g}(^{k+1}\boldsymbol{\xi}_{\psi,i} , ^{k+1}\boldsymbol{\alpha}_{i},^{k+1}\mathbf{d}_{i} )\hspace{-0.04cm})\mathbf{F}
\label{update_lambda2_1} \\ 
&{^{k+1}}\boldsymbol{\lambda}_{\psi,i}\hspace{-0.09cm}=\hspace{-0.09cm}{^{k}}\boldsymbol{\lambda}_{\psi,i} \hspace{-0.09cm}- \hspace{-0.09cm}\rho_{\psi}( \mathbf{F}^{k+1}\boldsymbol{\xi}_{i} 
 -\hspace{-0.09cm}\mathbf{g}(^{k+1}\boldsymbol{\xi}_{\psi,i} , ^{k+1}\boldsymbol{\alpha}_{i},^{k+1}\mathbf{d}_{i} )\hspace{-0.04cm})\mathbf{F}
\label{update_lambda2_2} 
\end{align}
}
\noindent \textbf{Return} ${^{k+1}}\boldsymbol{\xi}_{i}, {^{k+1}}\boldsymbol{\xi}_{\psi,i}, {^{k+1}}\boldsymbol{\alpha}_{i}, {^{k+1}}\mathbf{d}_{i}$
\end{algorithm}

\noindent \textbf{Analysis and Description of Algorithm \ref{alg_2}:} I provide a detailed breakdown of each line in Algorithm \ref{alg_2}.

\noindent \textbf{Line 2:} A glance at \eqref{reform_formulation} reveals that \eqref{reform_formulation} has the same structure as \eqref{over_1} where

\vspace{-0.6cm}
\small
\begin{align}
&\overline{\mathbf{Q}}= \hspace{-0.05cm} \mathbf{Q} + \rho \mathbf{F}^{T}\mathbf{F},~~
\nonumber \\
&\overline{\mathbf{q}}_{i} = \hspace{-0.05cm} -{^k}\boldsymbol{\lambda}_{i}-\mathbf{q}
-( \rho_{o}\mathbf{F}^{T}\mathbf{g}({^k}\boldsymbol{\xi}_{\psi,i},{^k}\boldsymbol{\alpha}_{i},{^k} \mathbf{d}_{i}))^{T}
\label{zeta_1_sepration_3}
 \end{align}
\normalsize

\noindent Thus the solution of \eqref{zeta_1_sepration_1} over all the batches can be computed in one-shot using \eqref{over_3}.

\noindent \textbf{Line 3:}
In this step, I calculate the optimization variable $\boldsymbol{\xi}_{\psi,i}$. To compute this variable, I inspect \eqref{reform_formulation} and determine terms associated with $\boldsymbol{\xi}_{\psi,i}$. Given $^{k+1}\boldsymbol{\xi}{i}$, $^{k}\boldsymbol{\alpha}{i}$, and $^{k}\mathbf{d}_{i}$, trajectory optimization \eqref{reform_formulation} can be expressed as \eqref{compute_psi_tilta}. The optimization problem \eqref{compute_psi_tilta} is then simplified as

\small
\begin{align}
\hspace{-0.2cm}{^{k+1}}{\boldsymbol{\xi}_{\psi,i}}\hspace{-0.1cm}= \hspace{-0.1cm}\min_{\boldsymbol{\xi}_{\psi,i}}\Big( \frac{1}{2}
\boldsymbol{\xi}_{\psi,i}^T\ddot{\mathbf{P}}^{T}\ddot{\mathbf{P}}\boldsymbol{\xi}_{\psi,i}
 \hspace{-0.1cm}-\hspace{-0.1cm}\langle ^{k}\boldsymbol{\lambda}_{\psi,i},
\boldsymbol{\xi}_{\psi,i}\rangle 
 \hspace{-0.1cm}+\hspace{-0.1cm}\frac{\rho_{o}}{2}\left\Vert  \begin{bmatrix}
   {^{k+1}}\boldsymbol{c}_{\psi,i} \\ {^{k+1}}\boldsymbol{s}_{\psi,i}
 \end{bmatrix} \hspace{-0.1cm}-\hspace{-0.1cm}\begin{bmatrix}
    \cos{\mathbf{P}\boldsymbol{\xi}_{\psi,i}}\\
    \sin{\mathbf{P}\boldsymbol{\xi}_{\psi,i}}
 \end{bmatrix}\right \Vert_2^2
 \Big) \label{compute_psi_tilta1}
\end{align}
\normalsize

\noindent Here, $\boldsymbol{c}_{\psi,i}$ and $\boldsymbol{s}_{\psi,i}$ are formed by aggregating the obtained slack variables, $c_{\psi,i}(t)$ and ${s}_{\psi,i}(t)$, across different time instances. Subsequently, by substituting the third term in \eqref{compute_psi_tilta1} with a convex surrogate \cite{singh2020bi}, $\boldsymbol{\xi}_{\psi,i}$ and consequently $\boldsymbol{\psi}_{i}$ can be determined by solving

\small
\begin{align}
{^{k+1}}{\boldsymbol{\xi}_{\psi,i}} &= \min_{\boldsymbol{\xi}_{\psi,i}}\Big( \frac{1}{2}
\boldsymbol{\xi}_{\psi,i}^T\ddot{\mathbf{P}}^{T}\ddot{\mathbf{P}}\boldsymbol{\xi}_{\psi,i}
 - \langle ^{k}\boldsymbol{\lambda}_{\psi,i},
\boldsymbol{\xi}_{\psi,i}\rangle 
 \nonumber \\&~~
 + \frac{\rho_{o}}{2}
\left\Vert
\arctan2({^{k+1}}\boldsymbol{s}_{\psi,i},{^{k+1}}\boldsymbol{c}_{\psi,i}) - \mathbf{P}\boldsymbol{\xi}_{\psi,i}
\right \Vert_2^2 \Big),~ \text{s.t.:} ~\mathbf{A}\boldsymbol{\xi}_{\psi,i} = \mathbf{b}_{\psi}.
\label{compute_convex_surrogate}
\end{align}
\normalsize 

\noindent where the optimization problem \eqref{compute_convex_surrogate} takes the form of a \gls{qp}, allowing for the efficient computation of its batch solution in one shot using \eqref{over_3}.

\noindent \textbf{Line 4:}
In this stage, the objective is to compute ${^{k+1}}\boldsymbol{\alpha}_{i}$ using \eqref{reform_formulation}. To achieve this, I identify the terms associated with ${^{k+1}}\boldsymbol{\alpha}_{i}$, update $\boldsymbol{\xi}_{i}$ and $\boldsymbol{\xi}_{\psi,i}$, and then simplify the proposed optimization problem as shown in \eqref{compute_ze_3}. ${^{k+1}}\boldsymbol{\alpha}_{i}$ comprises three variables: $\boldsymbol{\alpha}_{mj,i}$, $\boldsymbol{\alpha}_{v,i}$, and $\boldsymbol{\alpha}_{a,i}$, each of which is independent. Consequently, I can decompose \eqref{compute_ze_3} into three parallel optimization problems as

\small
\begin{subequations}
\begin{align}
{^{k+1}}\boldsymbol{\alpha}_{mj,i} &=   \min_{\boldsymbol{\alpha}_{mj,i}} 
\hspace{-0.08cm}\frac{\rho_{o}}{2}
\times\left\Vert  \begin{matrix}
 \overbrace{  
   {^{k+1}}\mathbf{x}_{i}+ r_{m} \cos{{^{k+1}}\boldsymbol{\psi}_{i}} -  \mathbf{x}_{o,j}
}^{{^{k+1}}\Tilde{\mathbf{x}}_{i}}
 -a{^{k}} \mathbf{d}_{mj,i} \cos{\boldsymbol{\alpha}_{mj,i}} 
\\
 \underbrace{
   {^{k+1}}\mathbf{y}_{i} + r_{m}\sin{{^{k+1}}\boldsymbol{\psi}_{i}} -  \mathbf{y}_{o,j}
 }_{{^{k+1}}\Tilde{\mathbf{y}}_{i}}
 -b{^{k}}\mathbf{d}_{mj,i}\sin{\boldsymbol{\alpha}_{mj,i}}
 \end{matrix}
 \right \Vert_2^2 \hspace{-0.2cm}
 \label{compute_zeta_33}  \\
{^{k+1}}\boldsymbol{\alpha}_{v,i} &= \min_{\boldsymbol{\alpha}_{v,i}}\frac{\rho_{o}}{2} \left\Vert
\begin{matrix}
{^{k+1}}\dot{\mathbf{x}}_{i}-{^{k}}\mathbf{d}_{v,i}\cos{\boldsymbol{\alpha}_{v,i}}
\\
 {^{k+1}}\dot{\mathbf{y}}_{i}-{^{k}}\mathbf{d}_{v,i}\sin{\boldsymbol{\alpha}_{v,i}}
 \end{matrix}
\right \Vert_2^2 \label{compute_av1}
\\
{^{k+1}}\boldsymbol{\alpha}_{a,i}  &= \min_{\boldsymbol{\alpha}_{v,i}}
\frac{\rho_{o}}{2} \left\Vert
\begin{matrix}
{^{k+1}}\ddot{\mathbf{x}}_{i}-{^{k}}\mathbf{d}_{a,i}\cos{\boldsymbol{\alpha}_{a,i}}
\\
 {^{k+1}}\ddot{\mathbf{y}}_{i}-{^{k}}\mathbf{d}_{a,i}\sin{\boldsymbol{\alpha}_{a,i}}
 \end{matrix}
\right \Vert_2^2 
\label{compute_acc}
\end{align}
\end{subequations}
\normalsize

Similar to my prior work, although \eqref{compute_zeta_33}-\eqref{compute_acc} exhibit non-convexity, their solutions can be readily computed through geometric reasoning.  It is noteworthy that each element of $\boldsymbol{\alpha}_{mj,i}$, $\boldsymbol{\alpha}_{v, i}$, and $\boldsymbol{\alpha}_{a,i}$ is independent of the others for a given position trajectory.
Consequently, \eqref{compute_zeta_33} can be interpreted as the projection of ${{^{k+1}}\Tilde{\mathbf{x}}_{i}}$ and ${{^{k+1}}\Tilde{\mathbf{y}}_{i}}$ onto an axis-aligned ellipse centered at the origin. Similarly, \eqref{compute_av1} and \eqref{compute_acc} can be viewed as the projection of (${^{k+1}}\dot{\mathbf{x}}_{i}$, ${^{k+1}}\dot{\mathbf{y}}_{i}$) and (${^{k+1}}\ddot{\mathbf{x}}_{i}$, ${^{k+1}}\ddot{\mathbf{y}}_{i}$) onto a circle centered at the origin with radii $\mathbf{d}_{v,i}\times v_{max}$ and $\mathbf{d}_{a,i} \times a_{max}$, respectively. Thus, the solutions of \eqref{compute_zeta_33}-\eqref{compute_acc} can be expressed as

\small
\begin{subequations}
\begin{align}
{^{k+1}}\boldsymbol{\alpha}_{mj,i} = \arctan2 ({^{k+1}}\Tilde{\mathbf{y}}_{i}, {^{k+1}}\Tilde{\mathbf{x}}_{i}).
\label{zeta_2}\\
{^{k+1}}\boldsymbol{\alpha}_{v,i} = \arctan2 ({^{k+1}}\dot{\mathbf{y}}_{i}, {^{k+1}}\dot{\mathbf{x}}_{i}).
\label{alpha_v}\\
{^{k+1}}\boldsymbol{\alpha}_{a,i} = \arctan2 ({^{k+1}}\ddot{\mathbf{y}}_{i}, {^{k+1}}\ddot{\mathbf{x}}_{i}).
\label{alpha_a}
 \end{align}
 \end{subequations}
 \normalsize

\noindent \textbf{Line 5:} I update the acquired optimization variables and reformulate the optimization problem \eqref{reform_formulation} in terms of ${^{k+1}}\mathbf{d}_{i}$ as shown in \eqref{compute_ze_4}. Following a similar procedure as in the previous step, the proposed optimization problem \eqref{compute_ze_4} can be reduced into three independent optimization problems:

\small
\begin{subequations}
    \begin{align}
{^{k+1}}\mathbf{d}_{mj,i} &= \min_{\mathbf{d}_{mj,i} \geq 1 }  \frac{\rho_{o}}{2}\left\Vert \begin{matrix}
{^{k+1}}\Tilde{\mathbf{x}}_{i}
- a\mathbf{d}_{mj,i} \cos{{^{k+1}}\boldsymbol{\alpha}_{mj,i}} \\
{^{k+1}}\Tilde{\mathbf{y}}_{i}
 -b\mathbf{d}_{mj,i} \sin{{^{k+1}}\boldsymbol{\alpha}_{mj,i}} \end{matrix}  \right \Vert_2^2
 \label{compute_d_mij} \\
 {^{k+1}}\mathbf{d}_{v,i} &= \max_{\mathbf{d}_{v,i}\leq 1}\frac{\rho_{o}}{2} \left\Vert \begin{matrix} {^{k+1}}\dot{\mathbf{x}}_{i}-\mathbf{d}_{v,i}\cos{{^{k+1}}\boldsymbol{\alpha}_{v,i}}\\
 {^{k+1}}\dot{\mathbf{y}}-\mathbf{d}_{v,i}\sin{{^{k+1}}\boldsymbol{\alpha}_{v,i}}
 \end{matrix}
\right \Vert_2^2  \label{compute_dv}\\
{^{k+1}}\mathbf{d}_{a,i} &=  \max_{\mathbf{d}_{a,i}\leq 1}\frac{\rho_{o}}{2} \left\Vert \begin{matrix} 
{^{k+1}}\ddot{\mathbf{x}}_{i}-\mathbf{d}_{a,i}\cos{{^{k+1}}\boldsymbol{\alpha}_{a,i}} \\
  {^{k+1}}\ddot{\mathbf{y}}_{i}-\mathbf{d}_{a,i}\sin{{^{k+1}}\boldsymbol{\alpha}_{a,i}}
  \end{matrix}
\right \Vert_2^2 \label{compute_da}
\end{align}
\end{subequations}
\normalsize

\noindent Since elements of $\mathbf{d}_{a,i}$ and $\mathbf{d}_{v,i}$ at different time instances are independent of each other, the problem can be reduced to $i \times n_{v}$ \gls{qp} parallel problems, allowing for simultaneous solution. In a similar way, I can reduce \eqref{compute_d_mij} problem into  $i\times m \times n_{v}$ \gls{qp}s. 

\noindent \textbf{Line 6:}
Lagrange multipliers are updated based on residuals \cite{goldstein2009split}.


\section{Validation and Benchmarking}
\noindent \textbf{Implementation Details:} I implemented both the proposed algorithm and \gls{cem} in Python, utilizing the JAX \cite{bradbury2018jax} libraries as a \gls{gpu}-accelerated backend. Additionally, I developed a \gls{mpc} framework on the proposed batch optimizer, initializing Lagrange multipliers $\boldsymbol{\lambda}_{i}$ and $\boldsymbol{\lambda}_{\psi, i}$ with the solution from the preceding control loop. The \gls{mpc} executed within a time budget of $0.04s$, sufficient for ten iterations of the proposed optimizer with a batch size of 1000. I employ three metrics to evaluate the proposed method, each defined as: 
\begin{itemize}
     \item Success-Rate: Total evaluated problems ($20$) in benchmark divided by successful runs with no collisions.
    \item Tracking Error: the tracking error can be defined as 
    \begin{align}
        (x_i(t)-x_{des}(t))^2+(y_{i}(t)-y_{des}(t))^2,
    \end{align}
    \noindent where the desired trajectory is chosen as a straight line between the initial and final point with a constant velocity.
    \item Smoothness cost: acceleration value used in navigation.
\end{itemize}
Furthermore, the validation benchmarks used are outlined below.

\begin{itemize}
    \item Benchmark 1, Static Crowd: In this benchmark, I generated an environment including static human crowds with size 30.
    \item Benchmark 2, Same Direction Flow: In this benchmark, the human crowd is in motion in the same direction as the robot and with a max velocity of $0.3 m/s$. The robot follows a straight-line trajectory with a velocity of $1.0 m/s$ and needs to overtake the human crowds.
    \item Benchmark 3, Opposite Direction Flow:
    For this benchmark, the crowd is moving with a max velocity of $1.0 m/s$, and the robot is moving in the opposite direction tracking a straight-line trajectory with $1.0 m/s$ desired speed.
\end{itemize}

\subsection{Qualitive Results} 
\begin{figure}[!h]
\centering
    \includegraphics[scale=0.6]{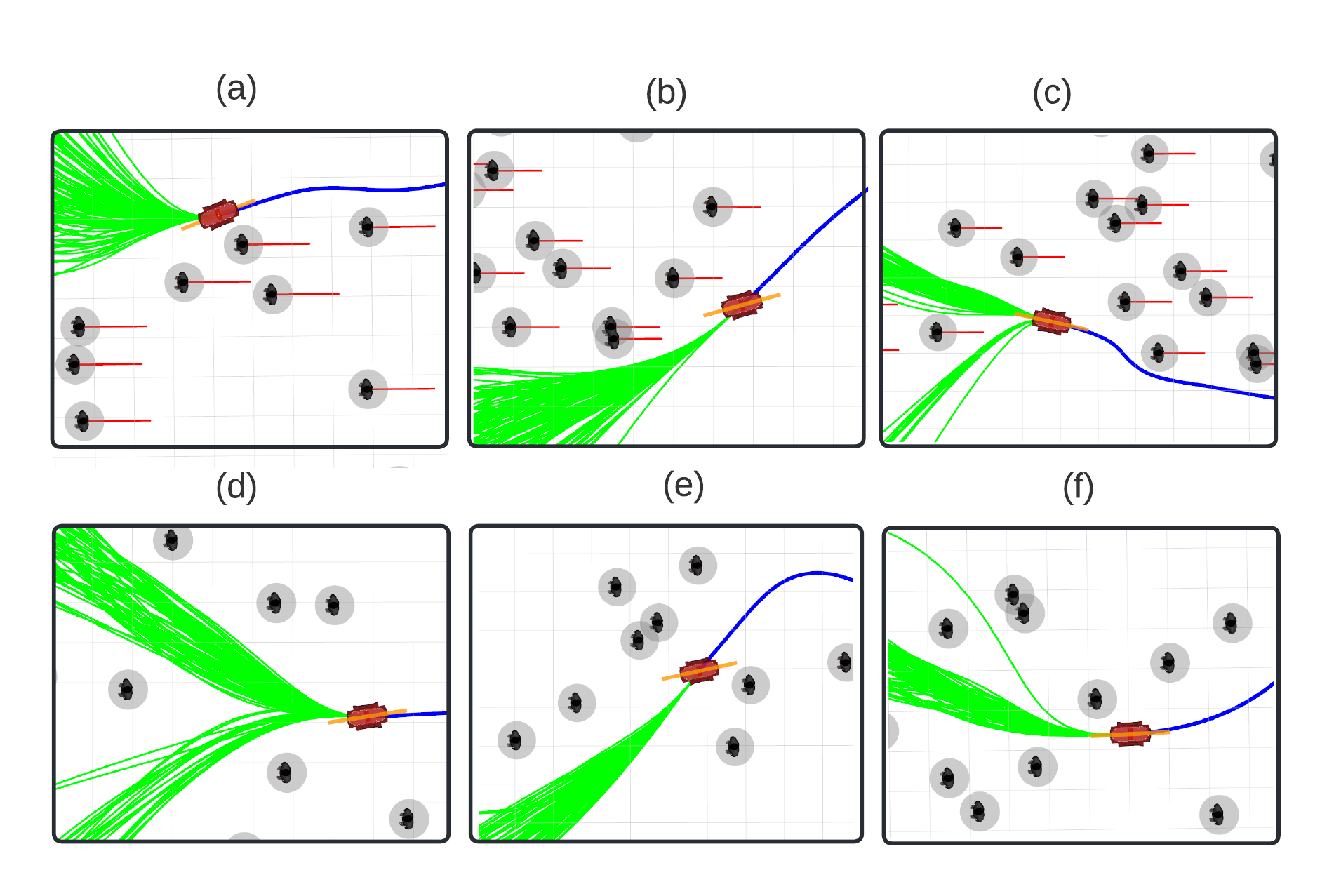}
    \caption[Qualitative result of MPC built on top of our batch optimizer]{\small{Qualitative result of \gls{mpc} built on top of our batch optimizer. The top and bottom rows show the navigation in Benchmarks 2 and 1, respectively. The green trajectories are the different locally optimal solutions obtained with our batch optimizer. The blue trajectories show the past positional traces of the robot. The trajectories in red show the predicted motion of the obstacles }}
    \label{modeling_snapshots}
\end{figure}

In Figure \ref{modeling_snapshots}, I illustrated the qualitative results of \gls{mpc} built on top of the proposed optimizer at three different snapshots for Benchmark 1 and Benchmark 2 (For better visualization, visit the youtube video \footnote{https://www.youtube.com/watch?v=ZlWJk-w03d8}). 

\subsection{Validating the Batch Optimizer}
\begin{figure}[h]
    \centering
    \includegraphics[scale=0.35]{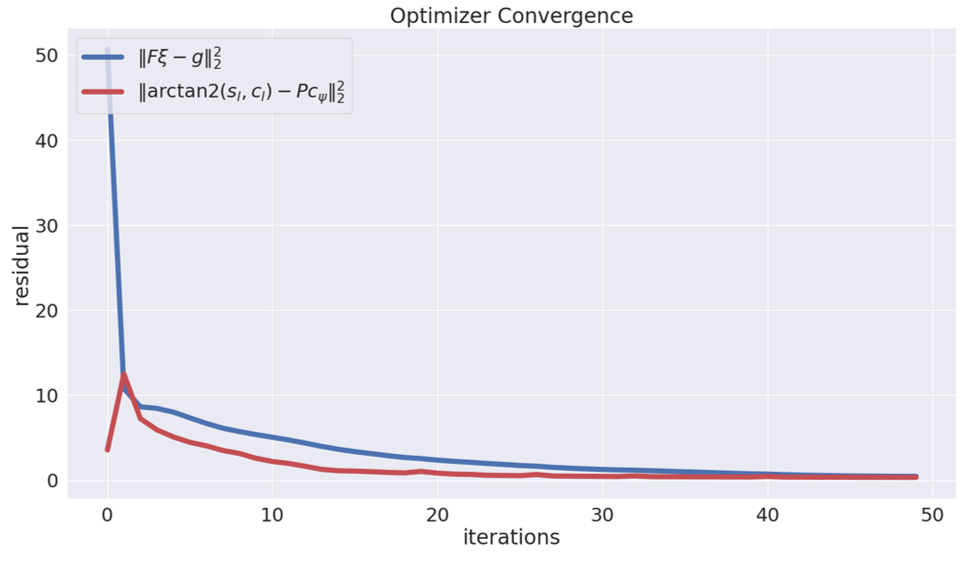}
    \caption[Batch optimizer residuals]{Validating optimizer convergence empirically. The residual constraints, $\Vert \mathbf{F}\boldsymbol{\xi}_{i}-\mathbf{g}_{i}\Vert$ and $\Vert \arctan2(s_{\psi,i},c_{\psi,i})-\mathbf{P}\boldsymbol{\xi}_{\psi,i}\Vert$ go to zero over iterations.}
    \label{convergence}
\end{figure}

\begin{figure}[!h]
\centering
    \includegraphics[scale=0.24]{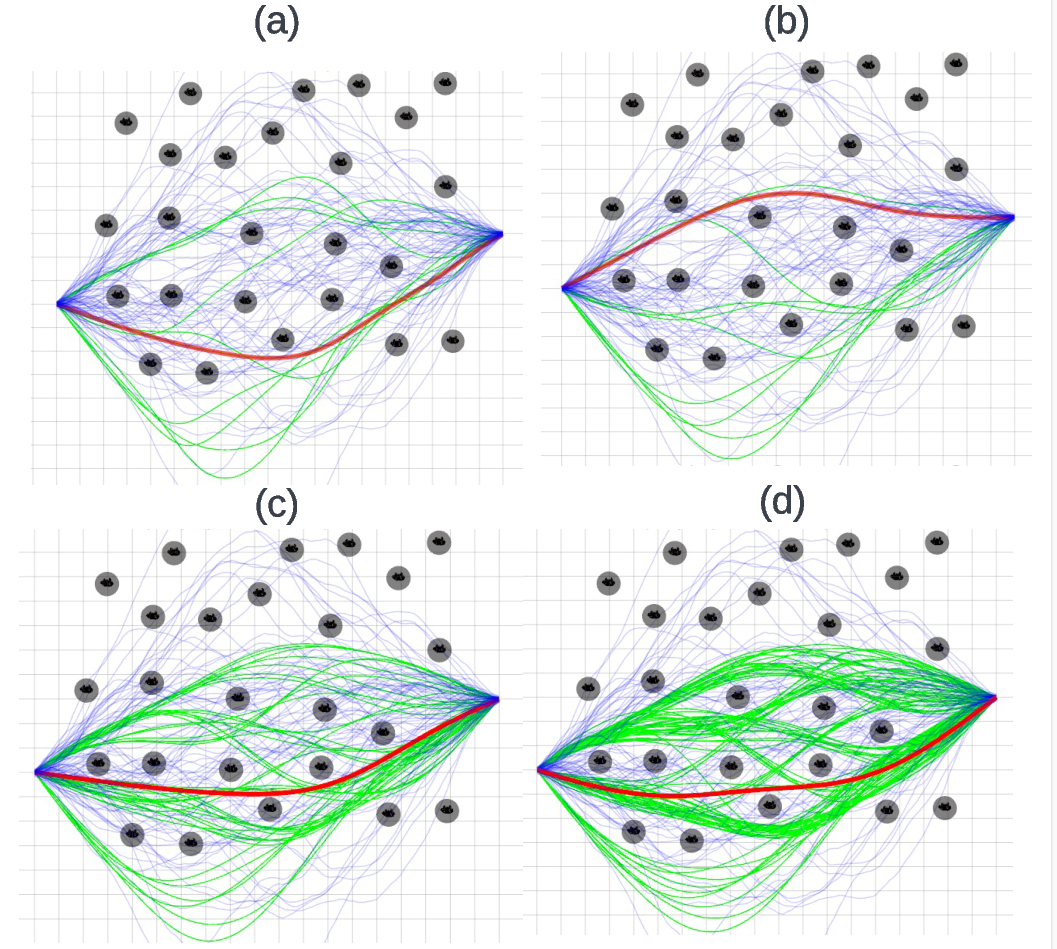}
    \caption[increasing the number of homotopies over iterations]{The number of feasible homotopies increases by iterations. Trajectories are plotted over 5, 20, 30, and 50 iterations in Figures (a), (b), (c), and (d), respectively. The initial guess trajectories, the best trajectory among batches, and feasible trajectories are in blue, red, and green, respectively. }
    \label{homotopy}
\end{figure}

\noindent Figure \ref{convergence} shows the changes of non-convex equality constraints residuals, \eqref{reform_eq}, and the convex surrogate residuals, introduced in \eqref{compute_convex_surrogate} for different iterations. Here, only the residual for the best trajectory of the batch is plotted. As can be seen, by increasing the number of iterations, the residuals approach zero. Thus, the kinematic and collision avoidance constraints are gradually satisfied. Also, another validation is provided in Figure \ref{homotopy}. In this Figure, as iteration progresses, more locally optimal trajectories residing in different homotopies are obtained.

\subsection{Quantitive Results}
\noindent \textbf{Comparison with Baseline \gls{mpc}:}
    In this step, the objective is to analyze how the navigation performance evolves with an increase in batch size. For comparison, I establish the \gls{mpc} setting with a batch size of one as the baseline and subsequently increase the number of batches. It is important to note that any off-the-shelf optimizer can be employed in this comparison. I tried to use ACADO optimizer \cite{houska2011acado}, but it could not achieve reliable and real-time performance for the provided benchmarks.
    
    Table \ref{Table_batch_size} presents our results. Notably, the \gls{mpc} baseline exhibits a very low success rate, approximately one percent. However, with an increase in the number of batches, our success rate experiences a gradual ascent and reaches to 97\% success rate for 1000 batches. Additionally, a slight increase in tracking error and acceleration is observed, attributed to the necessity for our optimizer to take detours in order to furnish a collision-free trajectory.

\begin{table}[]
\scriptsize
\centering
\caption{\small{Performance Metrics with Respect to Batch Size }}
\label{Table_batch_size}
\begin{tabular}{|l|l|l|l|}
\hline
\rowcolor[HTML]{9698ED} 
Batch size                            & Success rate & \begin{tabular}[c]{@{}l@{}}Tracking error (m)\\ mean/max/min\end{tabular} & \begin{tabular}[c]{@{}l@{}}Acceleration ($m/s^{2}$)\\ mean/max/min\end{tabular} \\ \hline
\rowcolor[HTML]{FFFFFF} 
\cellcolor[HTML]{009901}1 (Baseline)  & 0.016        & 3.19/4.78/1.69                                       & 0.097/0.29/0.0003                 \\ \hline
\cellcolor[HTML]{009901}200           & 0.8          & 3.15/4.89/1.51                                         & 0.15/0.37/0.0006                 \\ \hline
\cellcolor[HTML]{009901}400           & 0.88         & 3.10/4.77/1.32                                        & 0.139/0.32/0.0005             \\ \hline
\cellcolor[HTML]{009901}600           & 0.9          & 3.09/4.74/1.39                                         & 0.139/0.32/0.0005               \\ \hline
\cellcolor[HTML]{009901}800           & 0.95         & 3.07/4.75/1.45                                       & 0.136/0.31/0.004               \\ \hline
\cellcolor[HTML]{009901}\textbf{1000} & \textbf{100} & \textbf{3.06/4.65/1.56}                                                   & \textbf{0.166/0.31/0.03}        \\ \hline
\end{tabular}
\end{table}
\normalsize

\vspace{0.45cm}
   \noindent\textbf{Comparison with \gls{cem}:}
    I present the pivotal results of our paper by comparing our optimizer with the \gls{sota}, \gls{cem}, in an \gls{mpc} setting in Table \ref{table_com_cem}. To make the comparison, I consider 8k samples (8 times of the number of our samples) and heavily vectorized the cost evaluations in \gls{cem}. Through trial and error, I obtained the number of iterations of CEM that can be performed within $0.04s$ (the computation time of our MPC).

    \vspace{0.35cm}
    On \textbf{Benchmark 1}, our success rate is an impressive 100\%, surpassing \gls{cem} by 15\%. Furthermore, our average tracking error is 43\% lower than \gls{cem} on the same benchmark. For \textbf{Benchmark 2} and \textbf{Benchmark 3}, our optimizer's success rates are 20\% and 45\% higher than \gls{cem}, respectively. Similarly, our tracking error remains lower than \gls{cem} on these benchmarks. Notably, the \gls{cem} method generally deploys less acceleration on average compared to our proposed method. This arises from our method's dual focus on minimizing acceleration and addressing the need to navigate obstacles while following the desired trajectory. In more challenging scenarios, our method may utilize more acceleration to yield feasible trajectories, contributing to a higher success rate.

\begin{figure}[h]
\centering
    \includegraphics[scale=0.2]{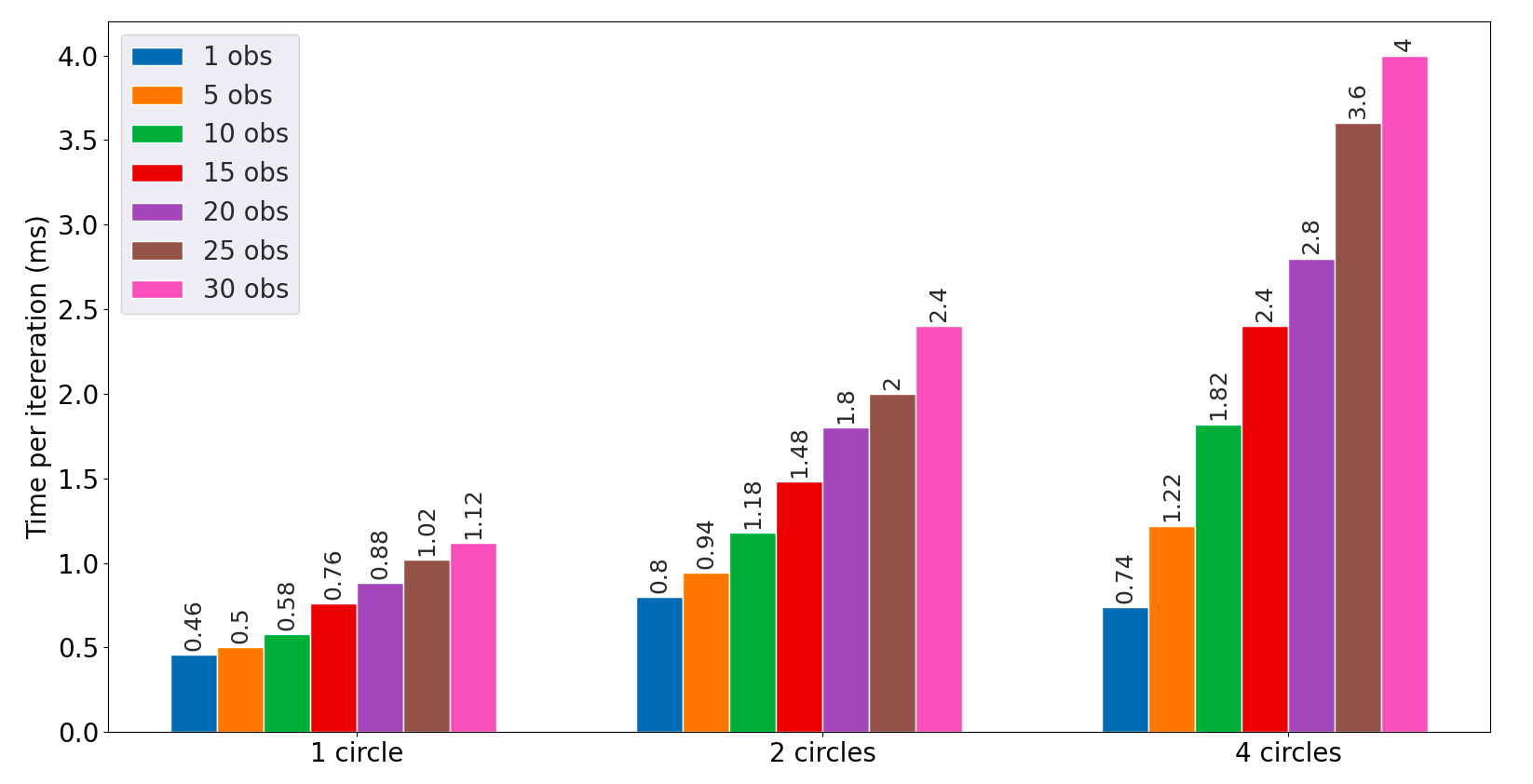}
    \caption[Computation time-scaling]{Time per iteration(ms) for batch size 1000. Solutions are typically obtained within $5-10$ iterations. The figure presents the scaling with respect to the number of circles used to approximate the footprint of the robot and the number of obstacles in the environment.}
    \label{modeling}
\end{figure}

\begin{table}[]
\scriptsize
\centering
\caption{\small{Comparison with \gls{cem}}}
\label{table_com_cem}
\begin{tabular}{|l|l|l|l|}
\hline
\rowcolor[HTML]{268A22} 
Method                   & Success rate   & \begin{tabular}[c]{@{}l@{}}Tracking error (m)\\ mean/max/min\end{tabular} & \begin{tabular}[c]{@{}l@{}}Acceleration ($m/s^{2}$)\\ mean/max/min\end{tabular} \\ \hline
\rowcolor[HTML]{C0C0C0} 
\textbf{Our Benchmark 1} & \textbf{100\%} & \textbf{2.44/5.44/0.079}                                                  & \textbf{0.13/0.32/0.0}             \\ \hline
\rowcolor[HTML]{C0C0C0} 
\textbf{Our Benchmark 2} & \textbf{95\%}  & \textbf{3.11/5.01/0.88}                                                   & \textbf{0.20/0.49/0.0}           \\ \hline
\rowcolor[HTML]{C0C0C0} 
\textbf{Our Benchmark 3} & \textbf{95\%}  & \textbf{2.74/4.82/0.07}                                                   & \textbf{0.17/0.45/0.0}          \\ \hline
\rowcolor[HTML]{FFA7A7} 
CEM Benchmark            & 85\%           & 3.5/5.22/1.94                                                             & 0.09/0.36/0.0                       \\ \hline
\rowcolor[HTML]{FFA7A7} 
CEM Benchmark            & 75\%           & 3.99/6.87/1.12                                                            & 0.09/0.28/0.0                    \\ \hline
\rowcolor[HTML]{FFA7A7} 
CEM Benchmark            & 40\%           & 3.25/4.05/2.56                                                            & 0.13/0.46/0.0                     \\ \hline
\end{tabular}
\end{table}
    \vspace{0.45cm}
    \noindent\textbf{Computation Time Scaling:} Figure \ref{modeling} visually presents the per-iteration computation time scaling of our batch optimizer for varying numbers of circles and obstacles. Notably, the plot indicates a (sub)linear scaling of per-iteration computation time in relation to the number of circles and obstacles. This trend can be attributed to the matrix algebra employed in our optimizer. Specifically, a linear increase in~the number of footprint circles or obstacles leads to a similar increase in the number of rows of matrices $\mathbf{F}$ and $\mathbf{g}_{i}$, while the number of columns remains constant. Consequently, the computation cost of obtaining $\mathbf{F}^T\mathbf{g}_{i}$ in \eqref{zeta_1_sepration_3} can be made approximately linear through suitable \gls{gpu} parallelization. Also, the matrices $\mathbf{F}^T\mathbf{F}$ need to be computed only once since they remain constant across iterations and batches. A similar linear complexity analysis applies to the solutions of (\ref{compute_psi_tilta}) as well, where solutions are available as symbolic formulae, ensuring linear complexity concerning the variables $n_{c}$ and $n_{o}$.

    Additionally, I assess the per-iteration computation time of our vectorized batch optimizer across \gls{gpu}s and multi-threaded \gls{cpu}s in Table. \ref{gpu-cpu}. The advantage of the latter lies in its ability to execute any off-the-shelf optimizer in parallel CPU threads without necessitating changes to the underlying matrix algebra. However, CPU parallelization introduces a scenario where each problem instantiation competes with others for computational resources, potentially slowing down the overall computation time. Our experiments revealed that a C++ version of our optimizer performed smoothly with a batch size of 5, achieving a per-iteration computation time of $0.0015s$, which was competitive with our \gls{gpu} implementation. Nevertheless, scaling beyond this batch size resulted in a significant slowdown. For instance, with a batch size of 6, the per-iteration CPU time increased to $0.03s$. Hence, for larger batch sizes, it became more practical to run sequential instantiations with mini-batches of 5. I note that a more rigorous implementation might improve batch equality-constrained \gls{qp} structure in some critical steps in the performance of multi-threaded CPU-based batch optimization. However, our current bench-marking establishes the computational benefit derived from several layers of reformulation that induced equality-constrained \gls{qp} structure in some critical steps of our batch optimizer and allowed for effortless \gls{gpu} acceleration.

\setlength{\arrayrulewidth}{1pt}
\begin{table}[]
\scriptsize
\centering
\caption{\small{Per-iteration comparison for \gls{gpu} vs multi-threaded \gls{cpu}}}
\label{gpu-cpu}
\begin{tabular}{|l|llllll|}
\hline
\rowcolor[HTML]{036400} 
\cellcolor[HTML]{FFFFFF}                   & \multicolumn{6}{c|}{\cellcolor[HTML]{036400}{\color[HTML]{FFFFFF} Batch size}}  \\ \cline{2-7} 
\rowcolor[HTML]{9AFF99} 
\multirow{-2}{*}{\cellcolor[HTML]{FFFFFF}} & \multicolumn{1}{l|}{\cellcolor[HTML]{9AFF99}5}               & \multicolumn{1}{l|}{\cellcolor[HTML]{9AFF99}200}             & \multicolumn{1}{l|}{\cellcolor[HTML]{9AFF99}400}             & \multicolumn{1}{l|}{\cellcolor[HTML]{9AFF99}600}             & \multicolumn{1}{l|}{\cellcolor[HTML]{9AFF99}800}             & 1000            \\ \hline
\rowcolor[HTML]{DAE8FC} 
GPU                                        & \multicolumn{1}{l|}{\cellcolor[HTML]{DAE8FC}0.0016}          & \multicolumn{1}{l|}{\cellcolor[HTML]{DAE8FC}\textbf{0.0017}} & \multicolumn{1}{l|}{\cellcolor[HTML]{DAE8FC}\textbf{0.0026}} & \multicolumn{1}{l|}{\cellcolor[HTML]{DAE8FC}\textbf{0.0033}} & \multicolumn{1}{l|}{\cellcolor[HTML]{DAE8FC}\textbf{0.0039}} & \textbf{0.0045} \\ \hline
\rowcolor[HTML]{FFCC67} 
CPU                                        & \multicolumn{1}{l|}{\cellcolor[HTML]{FFCC67}\textbf{0.0015}} & \multicolumn{1}{l|}{\cellcolor[HTML]{FFCC67}0.075}           & \multicolumn{1}{l|}{\cellcolor[HTML]{FFCC67}0.12}            & \multicolumn{1}{l|}{\cellcolor[HTML]{FFCC67}0.18}            & \multicolumn{1}{l|}{\cellcolor[HTML]{FFCC67}0.24}            & 0.3             \\ \hline
\end{tabular}
\end{table}

\section{Connection to the Rest of Thesis}

The concept of representing rectangular robots using multiple overlapping circles is drawn from the polar/spherical representation of collision avoidance constraints outlined in the earlier paper. Similarly, the approach to reformulating boundaries is inspired by previous work.

Moreover, the \gls{qp} structure proposed in Paper I acts as a foundation for the notion of formulating trajectory optimization for multiple initial guesses concurrently and redefining the problem. Additionally, Algorithm \ref{alg_2} serves as an extension of Algorithm \ref{alg_1} and employs analogous procedures to solve the trajectory optimization problem. By getting inspiration from batch trajectory optimization, the next paper is established. 

    \chapter{Paper III: Projection-based Trajectory Optimization}\label{paper_3}
\section{Context}



\begin{figure}[h]
    \centering
    \includegraphics[scale=0.81
    ]{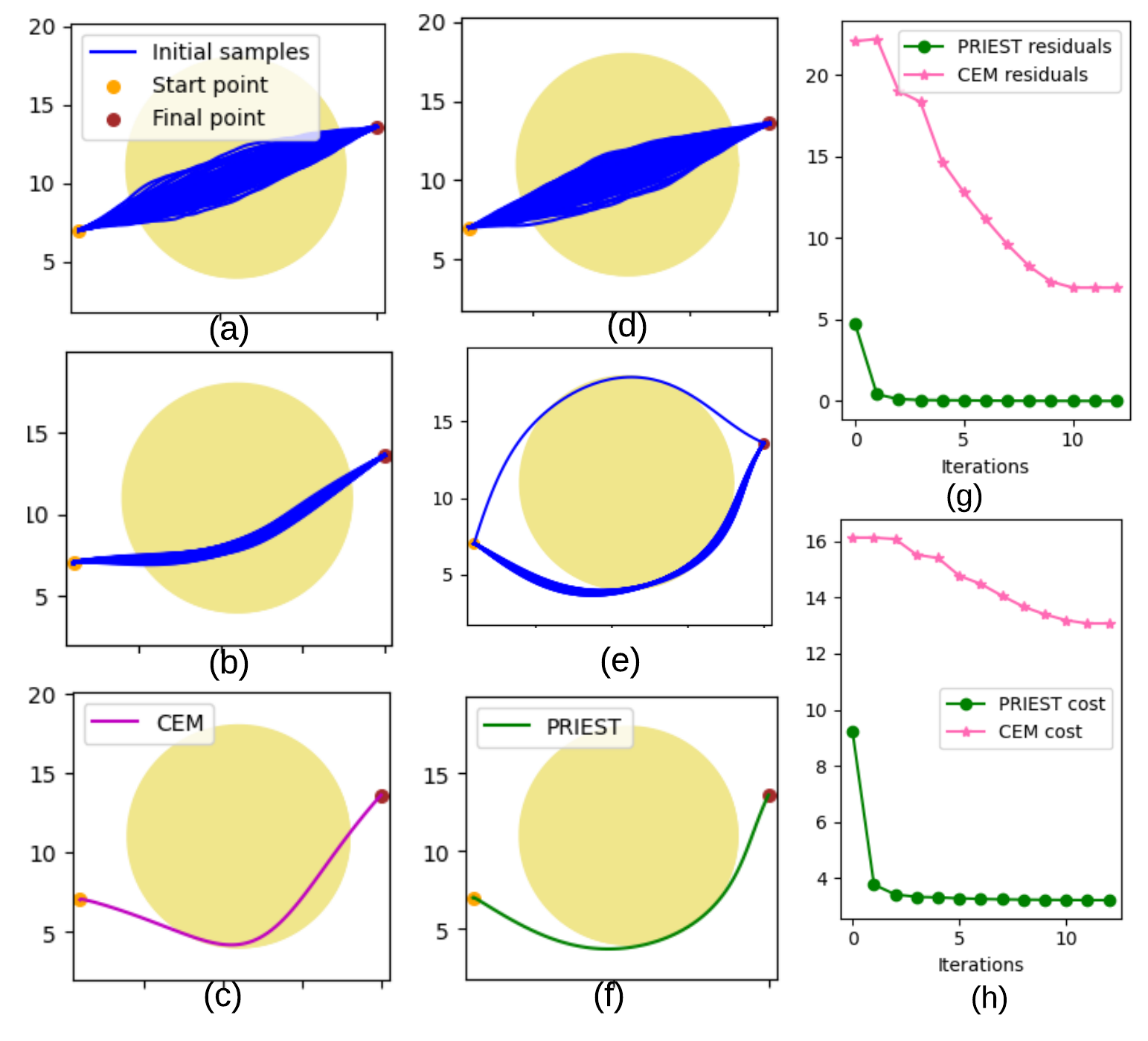}
    \caption[Simple comparison between CEM and PRIEST]{\small{A comparison between \gls{cem} (left column) and \gls{priest} (middle column). Figure (a)-(c) shows how \gls{cem} (or any typical sampling-based optimizer) struggles when all the sampled initial trajectories lie in the infeasible (high-cost) region. My approach, \gls{priest}, integrates a projection optimizer within any standard sampling-based approach that guides the samples toward feasible regions before evaluating their cost. Figure (g)-(h) presents changes in the cost function values, and constraint residuals change across both \gls{cem} and \gls{priest} iterations.}}
    \label{fig_teaser}
\end{figure}

As mentioned in previous chapters, gradient-based approaches and sampling-based planners are two classes of motion planning algorithms. Gradient-based approaches, \cite{metz2003rockit, vanroye2023fatrop}, rely on the differentiability of cost and constraint functions. These methods often require a well-initialized trajectory, posing challenges in rapidly changing environments. On the other hand, sampling-based planners, such as \gls{cem} \cite{bharadhwaj2020model} and CMA-ES \cite{jankowski2023vp}, explore the state-space through random sampling, allowing them to find locally optimal solutions without relying on differentiability. Despite their exploration capabilities, these optimizers face challenges when all sampled trajectories end up in the infeasible (high-cost) region (see Figure \ref{fig_teaser}(a-c)).
My main motivation for this chapter is to integrate the advantages of both sampling-based and gradient-based methods. 
In the next section, I will explain the contribution and main results of my work in detail.

\section{Overview of the Main Results}
\begin{figure}[!h]
    \centering
    \includegraphics[scale=0.75
    ]{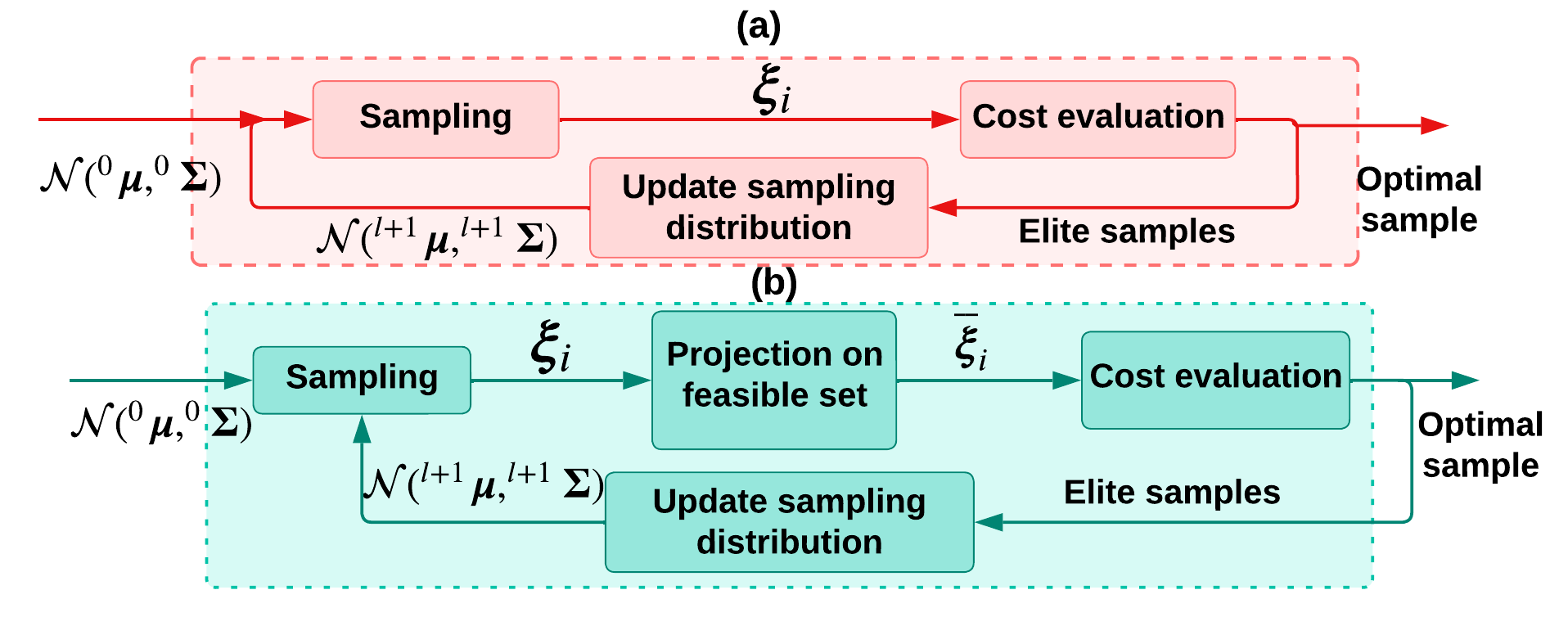}
    \caption{Comparison between a sampling-based optimizer (a) and PRIEST~(b).}
    \label{fig_overview}
\end{figure}

 Figure \ref{fig_overview} provides a visual overview of my proposed approach, highlighting its distinctive features compared to existing baselines. A pivotal distinction lies in the insertion of the projection optimizer between the sampling and cost evaluation blocks. This optimizer directs the sampling process toward feasible (low-cost) regions at each iteration. Thus, my approach, PRIEST, can handle pathological cases where all sampled trajectories are infeasible, e.g., due to violation of collision constraints (see Figure \ref{fig_teaser}(d-f)) 

The heart of \gls{priest} lies in an innovative optimizer with the distinct ability to take a set of trajectories, project each one onto the feasible set, and refine the sampling distribution. In this chapter, I illustrate how the proposed projection optimizer can be effectively parallelized and accelerated on \gls{gpu}s. The key to this achievement lies in the reformulation of underlying collision and kinematic constraints into polar/spherical form, coupled with an \gls{am} approach to tackle the resulting problem. Furthermore, my optimizer naturally integrates with decentralized variants of sampling-based optimizers \cite{zhang2022simple}, wherein multiple sampling distributions~are refined in parallel to enhance the optimality of the solution.
\section{Advantages Over SOTA Method}


   \gls{priest} shines when compared to existing approaches. It demonstrates superior performance in terms of success rate, time-to-reach the goal, computation time, etc. Notably, on the \gls{barn} dataset \cite{perille2020benchmarking}, \gls{priest} surpasses the \gls{ros} Navigation stack, boasting at least a 7\% increase in success rate and halving the travel time. On the same benchmarks,~our success rate is at least 35\% better than \gls{sota} local sampling-based optimizers like \gls{mppi} \cite{williams2017model} and log-MPPI \cite{mohamed2022autonomous}. Additionally, we consider a point-to-point navigation~task and compare \gls{priest} with the SOTA gradient-based solvers, ROCKIT \cite{metz2003rockit} (a collection of optimizers like IPOPT, ACADO, etc) and \gls{fatrop} \cite{vanroye2023fatrop}, and sampling-based methods \gls{cem} and \gls{vpsto} \cite{jankowski2023vp}. We show up to a $2$x improvement in success rate over these baselines. Finally, we show that \gls{priest} respectively has 17\% and 23\% higher success rates than the \gls{ros} Navigation stack and other \gls{sota} approaches in dynamic environments. 

\section{Problem Formulation}


\noindent \textbf{Trajectory Optimization:}
We are interested in solving the following 3D trajectory optimization:

\vspace{-0.45cm}
\small
\begin{subequations}
\begin{align}
   &\min_{x(t), y(t), z(t)} c_1(x^{(q)}(t), y^{(q)}(t), z^{(q)}(t))  
\label{acc_cost_priest}\\
&\text{s.t.:} \nonumber \\
&x^{(\hspace{-0.02cm}q\hspace{-0.02cm})}\hspace{-0.05cm}(\hspace{-0.02cm}t\hspace{-0.02cm}), \hspace{-0.05cm}y^{(\hspace{-0.02cm}q\hspace{-0.02cm})}\hspace{-0.05cm}(\hspace{-0.02cm}t\hspace{-0.02cm}), \hspace{-0.05cm}z^{(\hspace{-0.02cm}q\hspace{-0.02cm})}\hspace{-0.05cm}(\hspace{-0.02cm}t\hspace{-0.02cm})|_{t=t_{0}}\hspace{-0.05cm}=\hspace{-0.05cm}\mathbf{b}_{0},~
   x^{(\hspace{-0.02cm}q\hspace{-0.02cm})}\hspace{-0.05cm}(\hspace{-0.02cm}t\hspace{-0.02cm}),\hspace{-0.05cm}y^{(\hspace{-0.02cm}q\hspace{-0.02cm})}\hspace{-0.05cm}(\hspace{-0.02cm}t\hspace{-0.02cm}), \hspace{-0.05cm}z^{(\hspace{-0.02cm}q\hspace{-0.02cm})}\hspace{-0.05cm}(\hspace{-0.02cm}t\hspace{-0.02cm})|_{t=t_{f}}\hspace{-0.05cm}= \hspace{-0.05cm}\mathbf{b}_{f}\label{eq1_multiagent_1_priest}\\
    &\dot{x}^{2}(\hspace{-0.02cm}t\hspace{-0.02cm})\hspace{-0.05cm}+\hspace{-0.05cm} \dot{y}^{2}(\hspace{-0.02cm}t\hspace{-0.02cm})\hspace{-0.05cm}+ \hspace{-0.05cm}\dot{z}^{2}(\hspace{-0.02cm}t\hspace{-0.02cm}) \leq v^{2}_{max}, ~
    \ddot{x}^{2}(\hspace{-0.02cm}t\hspace{-0.02cm}) \hspace{-0.05cm}+\hspace{-0.05cm} \ddot{y}^{2}(\hspace{-0.02cm}t\hspace{-0.02cm}) \hspace{-0.05cm}+ \hspace{-0.05cm}\ddot{z}^{2}(\hspace{-0.02cm}t\hspace{-0.02cm}) \leq a^{2}_{max} \label{acc_constraint_priest}\\
    &s_{min} \leq (x(t), y(t), z(t)) \leq s_{max} \label{affine_ineq_priest}  \\
    &
    -\frac{\hspace{-0.07cm}(x(t)\hspace{-0.09cm}-
 \hspace{-0.06cm}x_{o, j}(t)\hspace{-0.01cm})^{2}}{a^2}\hspace{-0.09cm} - \hspace{-0.09cm}
 \frac{\hspace{-0.07cm}(y(t) \hspace{-0.075cm}-\hspace{-0.065cm}y_{o, j}(t)\hspace{-0.02cm})^{2}}{a^2} 
  \hspace{-0.09cm}-\hspace{-0.09cm}\frac{\hspace{-0.07cm}(z(t)\hspace{-0.09cm}- 
 \hspace{-0.08cm}z_{o, j}(t)\hspace{-0.01cm})^{2}}{b^2}\hspace{-0.06cm} \hspace{-0.06cm}
  + \hspace{-0.05cm}1\leq 0, \label{coll_multiagent_priest}
\end{align}
\end{subequations}
\normalsize

\noindent where the cost function $c_1(\cdot)$ is expressed in terms of $q^{th}$ derivatives of position-level trajectories, allowing for the inclusion of penalties related to accelerations, velocities, curvature, etc., where $q\hspace{-0.10cm}=\hspace{-0.10cm}{0,1,2}$. Additionally, we leverage differential flatness to enhance control costs within $c_1(\cdot)$. It should be mentioned that in my approach, the cost functions $c_1(\cdot)$ are not required to be convex, smooth, or even possess an analytical form. The vectors $\mathbf{b}_{0}$ and $\mathbf{b}_{f}$ in \eqref{eq1_multiagent_1_priest} denote the initial and final values of boundary conditions. The affine inequalities in \eqref{affine_ineq_priest} set bounds on the robot's workspace. Constraints on velocity and acceleration are imposed by \eqref{acc_constraint_priest}. Lastly, \eqref{coll_multiagent_priest} enforces collision avoidance, assuming obstacles are modeled as axis-aligned ellipsoids with dimensions $(a, a, b)$.

By adapting the parametrized optimization \eqref{parameter} and compact version of variables, we can reformulate \eqref{acc_cost_priest}-\eqref{coll_multiagent_priest} as

\vspace{-0.55cm}
\small
\begin{subequations}
\begin{align}   &\min_{\hspace{0.1cm}\boldsymbol{\xi}} c_1(\boldsymbol{\xi})  \label{cost_modify11} \\
&\text{s.t.:} \nonumber \\
    &\mathbf{A}\boldsymbol{\xi} = \mathbf{b}_{eq}\label{eq_modify11}\\
   &\mathbf{g}(\boldsymbol{\xi}) \leq \mathbf{0} \label{reform_bound},
   \vspace{-0.2cm}
\end{align}
\end{subequations}
\normalsize
\vspace{-0.5cm}

\noindent where $\boldsymbol{\xi} = \begin{bmatrix}
\boldsymbol{\xi}_{x}^{T}&\hspace{-0.3cm}\boldsymbol{\xi}_{y}^{T}&\hspace{-0.3cm}
\boldsymbol{\xi}_{z}^{T}
\end{bmatrix}^T$. With a slight abuse of notation, $c_1(\cdot)$ is now used to denote a cost function dependent on $\boldsymbol{\xi}$. The matrix $\mathbf{A}$ is block diagonal, where each block on the main diagonal consists of $ \begin{bmatrix}
\bold{P}_{0}&\hspace{-0.22cm}\dot{\bold{P}}_{0}&\hspace{-0.22cm}\ddot{\bold{P}}_{0}&\hspace{-0.22cm} \bold{P}_{-1}
\end{bmatrix}$. The subscripts $0$ and $-1$ signify the first and last row of the respective matrices, corresponding to the initial and final boundary constraints. The vector $\mathbf{b}_{eq}$ is simply the stack of $\mathbf{b}_{0}$ and $\mathbf{b}_{f}$. The function $\mathbf{g}$ contains all the inequality constraints \eqref{acc_constraint_priest}-\eqref{coll_multiagent_priest}.

\section{Main Results}


In this section, I introduce my main block, the projection optimizer, and subsequently, I detail its integration into a sampling-based optimizer.

\subsection{Projection Optimization}
I can demonstrate that for a specific class of constraint functions $\mathbf{g}$ involving quadratic and affine constraints, the optimization problem

\vspace{-0.5cm}
\small
\begin{subequations}
\begin{align}
     &\min_{\overline{\boldsymbol{\xi}}_{i}} \frac{1}{2}\Vert \overline{\boldsymbol{\xi}}_i-\boldsymbol{\xi}_{i}\Vert_2^2, ~ i=1,2,...,N_{b} 
    \label{projection_cost}\\
    & \text{s.t.: }\mathbf{A}\overline{\boldsymbol{\xi}}_{i} = \mathbf{b}_{eq}, \qquad \mathbf{g}(\overline{\boldsymbol{\xi}}_{i}) \leq  \mathbf{0}, \label{projection_const}
\end{align}
\end{subequations}
\normalsize

\noindent where the cost function \eqref{projection_cost} aims to minimally modify the $i^{th}$ sampled trajectory $\boldsymbol{\xi}_i$ to $\overline{\boldsymbol{\xi}}_{i}$ in order to satisfy the equality and inequality constraints, can be reduced to the fixed-point iteration of the following form

\vspace{-0.5cm}
\small
\begin{subequations}
    \begin{align}
    &{^{k+1}}\mathbf{e}_i, {^{k+1}}\boldsymbol{\lambda}_i = \mathbf{h} ({^k} \overline{\boldsymbol{\xi}}_{i}, {^k}\boldsymbol{\lambda}_i )\label{fixed_point_1}\\
&{^{k+1}}\overline{\boldsymbol{\xi}}_{i} =\arg\min_{\overline{\boldsymbol{\xi}}_{i}} \frac{1}{2}\Vert \overline{\boldsymbol{\xi}}_{i}-\boldsymbol{\xi}_i\Vert_2^2 +\frac{\rho}{2} \left\Vert \mathbf{F}\overline{\boldsymbol{\xi}}_{i} -{^{k+1}}\mathbf{e}_{i} \right\Vert_2^2-{^{k+1}}\boldsymbol{\lambda}_i^T\overline{\boldsymbol{\xi}}_{i} \label{fixed_point_22} \\  & \text{s.t.: }\mathbf{A}\overline{\boldsymbol{\xi}}_{i} = \mathbf{b}_{eq} \label{fixed_point_2}
\end{align}
\end{subequations}
\normalsize

\noindent The vector ${^{k+1}}\boldsymbol{\lambda}_i$ is the Lagrange multiplier at iteration $k+1$ of the projection optimization. Additionally, $\mathbf{F}$ and  $\mathbf{e}_{i}$ present a constant matrix and vector which will be defined later. The $\mathbf{h}$ is a closed-form analytical function. The primary computational challenge associated with projection optimization arises from solving the \gls{qp} \eqref{fixed_point_1}. However, due to the absence of inequality constraints in \eqref{fixed_point_22}-\eqref{fixed_point_2}, the \gls{qp} simplifies to an affine transformation, taking the following form:

\vspace{-0.48cm}
\small
\begin{subequations}
    \begin{align}
&({^{k+1}}\overline{\boldsymbol{\xi}}_{i}, {^{k+1}}\boldsymbol{\nu}_i) = \mathbf{M}\boldsymbol{\eta}({^k} \overline{\boldsymbol{\xi}}_{i})
    \label{affine_trans} \\
    &\mathbf{M} \hspace{-0.1cm}= \hspace{-0.1cm}\begin{bmatrix}
        \mathbf{I}+\rho\mathbf{F}^T\mathbf{F} & \mathbf{A}^{T} \\ 
        \mathbf{A} & \mathbf{0}
    \end{bmatrix}^{-1}\hspace{-0.2cm}, \boldsymbol{\eta} = \begin{bmatrix}
        -\rho\mathbf{F}^T {^{k+1}}\mathbf{e}_i+{^{k+1}}\boldsymbol{\lambda}_i+\boldsymbol{\xi}_i\\
        \mathbf{b}_{eq}
    \end{bmatrix} \label{matrix_vec}
\end{align}
\end{subequations}
\normalsize

\vspace{-0.13cm}
\noindent where $\boldsymbol{\nu}_i$ presents the dual variables regarding with the equality constraints. The derivation of the \eqref{fixed_point_22}-\eqref{fixed_point_2} is detailed below.

    \noindent\textbf{Reformulated constraints:}
     Considering the same motivation as Section \ref{main_result_1}, I reformulate the collision avoidance and boundary inequality constraints, \eqref{affine_ineq_priest} \eqref{coll_multiagent_priest}, $\mathbf{f}{_{o,j}}=0$, $\mathbf{f}{_{v}}=0$ and $\mathbf{f}{_{a}}=0$ as follows

\vspace{-0.3cm}
\small
\begin{subequations} 
\begin{align}
&\hspace{-0.2cm}\colorbox{my_green}{$\mathbf{f}_{o,j}=\hspace{-0.1cm}
\left \{ \begin{array}{lcr}
x(t)-x_{o,j}(t)-ad_{o,j}(t)\cos{\alpha_{o,j}}(t)\sin{\beta_{o,j}}(t) \\
y(t)-y_{o,j}(t)-bd_{o,j}(t)\sin{\alpha_{o,j}(t)}\sin{\beta_{o,j}}(t)\\
z(t)-z_{o,j}(t)-bd_{o,j}(t)\cos{\beta_{o,j}}
\end{array} \hspace{-0.25cm} \right \}\hspace{-0.10cm} $},  d_{o,j}(t) \geq 1 
\label{collision_priest} \\
& \colorbox{my_pink}{$ \mathbf{f}_{v} =\left \{ \hspace{-0.20cm} \begin{array}{lcr}
 \dot{x}(t) - d_{v}(t)v_{max}\cos{\alpha_{v}(t)}\sin{\beta_{v}(t)}  \\
 \dot{y}(t) - d_{v}(t)v_{max}\sin{\alpha_{v}(t)} \cos{\alpha_{v}(t)} \\
 \dot{z}(t) - d_{v}(t)v_{max}\cos{\beta_{v}(t)} 
\end{array} \hspace{-0.20cm} \right \} $}
,d_{v}(t) \leq 1 , ~  \forall t \label{fv_priest} \\
&\colorbox{my_blue}{$ \mathbf{f}_{a} = \left \{ \begin{array}{lcr}
\ddot{x}(t) - d_{a}(t)a_{max}\cos{\alpha_{a}(t)}\sin{\beta_{a}(t)} \\
\ddot{y}(t) - d_{a}(t)a_{max}\sin{\alpha_{a}(t)}\sin{\beta_{a}(t)}\\
\ddot{z}(t) - d_{a}(t)a_{max}\cos{\beta_{a}(t)} 
\end{array} \hspace{-0.20cm} \right \}$},  d_{a}(t) \leq 1 , ~\forall t \label{fa_priest}
\end{align}
\end{subequations}
 \normalsize
 
 \noindent where variables $d_{o,j}(t),\alpha_{o,j}(t),\beta_{o,j}(t),d_{v}(t),d_{a}(t),\alpha_{v}(t),\alpha_{a}(t)$, $\beta_{a}(t)$ and $\beta_{v}(t)$ are additional variables which will be computed along the projection part.

    \vspace{0.25cm}
 \noindent \textbf{Reformulated Problem:} Now, I leverage \eqref{collision_priest}-\eqref{fa_priest} and rewrite the projection problem \eqref{projection_cost}-\eqref{projection_const} as

\vspace{-0.4cm}
\small
\begin{subequations}
\begin{align} &\overline{\boldsymbol{{\xi}}}_{i} = \arg\min_{\overline{\boldsymbol{\xi}}_{i}} \frac{1}{2}\Vert \overline{\boldsymbol{\xi}}_{i}-\boldsymbol{\xi}_i\Vert_2^2 \label{cost_modify} \\ & \text{s.t.:} \nonumber \\&
    \mathbf{A}\overline{\boldsymbol{\xi}}_{i} = \mathbf{b}_{eq} \label{eq_modify_our}\\
   &\Tilde{\mathbf{F}}\overline{\boldsymbol{{\xi}}}_{i} = 
   \Tilde{\mathbf{e}}(\boldsymbol{\alpha}_{i}, \boldsymbol{\beta}_{i}, \mathbf{d}_{i})\label{reform_bound}\\&
 \mathbf{d}_{min} \leq \mathbf{d}_{i}\leq \mathbf{d}_{max}, \label{ine_1} \\&
 \mathbf{G}\overline{\boldsymbol{\xi}}_{i} \leq \boldsymbol{\tau} \label{bound_con}
\end{align}
\end{subequations}
\normalsize

\vspace{-0.2cm}
\noindent where $\hspace{-0.05cm}\boldsymbol{\alpha}_{i}, \boldsymbol{\beta}_{i}$ and $\hspace{-0.03cm}\mathbf{d}_{i}$ are the representation of \hspace{-0.2cm}$\begin{bmatrix}
      \boldsymbol{\alpha}_{o,i}^{T} &\hspace{-0.28cm}\boldsymbol{\alpha}_{v,i}^{T}&\hspace{-0.28cm}
      \boldsymbol{\alpha}_{a,i}^{T} 
  \end{bmatrix}^{T}\hspace{-0.22cm},\hspace{-0.1cm}\begin{bmatrix}
      \boldsymbol{\beta}_{o,i}^{T} &\hspace{-0.28cm}\boldsymbol{\beta}_{v,i}^{T} &
    \hspace{-0.28cm}  \boldsymbol{\beta}_{a,i}^{T} \hspace{-0.28cm}
  \end{bmatrix}^{T}\hspace{-0.3cm}$ and $
     \begin{bmatrix}
     \mathbf{d}^{T}_{o,i}&\hspace{-0.2cm}
     \mathbf{d}^{T}_{v,i} &\hspace{-0.2cm} \mathbf{d}^{T}_{a,i}
  \end{bmatrix}^{T}$\hspace{-0.1cm} respectively. The constant vector $\boldsymbol{\tau}$ is formed by stacking the $s_{min}$ and $s_{max}$ in appropriate form. The matrix $\mathbf{G}$ is formed by stacking $ -\bold{P}$ and $\bold{P}$ vertically. Similarly, $\mathbf{d}_{min}$, $\mathbf{d}_{max}$ are formed by stacking the lower ($[1, 0, 0]$), and upper bounds ($[\infty, 1, 1]$) of $\mathbf{d}_{o,i}, \mathbf{d}_{v,i}, \mathbf{d}_{a,i}$. Also, $ \Tilde{\mathbf{F}}$, and $\mathbf{e}$ are formed as

\vspace{-0.3cm}
  \small
  \begin{align}
      \tilde{\mathbf{F}} =\begin{bmatrix}
        \begin{bmatrix}
       \colorbox{my_green}{$ \mathbf{F}_{o}$} \\
       \colorbox{my_pink}{$ \dot{\bold{P}}$} \\
        \colorbox{my_blue}{$\ddot{\bold{P}}$}
    \end{bmatrix} \hspace{-0.25cm}&\hspace{-0.25cm} \mathbf{0} \hspace{-0.25cm}&\hspace{-0.25cm} \mathbf{0} \\ \mathbf{0}\hspace{-0.25cm} & \hspace{-0.25cm}\hspace{-0.15cm}\begin{bmatrix}
       \colorbox{my_green}{$ \mathbf{F}_{o} $}\\
        \colorbox{my_pink}{$\dot{\bold{P}} $}\\
        \colorbox{my_blue}{$\ddot{\bold{P}}$}
    \end{bmatrix} \hspace{-0.25cm}& \hspace{-0.25cm} \mathbf{0}\\
         \mathbf{0}\hspace{-0.25cm}& \hspace{-0.25cm}\mathbf{0} \hspace{-0.25cm}&\hspace{-0.25cm} \begin{bmatrix}
        \colorbox{my_green}{$\mathbf{F}_{o}$} \\
        \colorbox{my_pink}{$\dot{\bold{P}}$} \\
       \colorbox{my_blue}{$ \ddot{\bold{P}}$}
    \end{bmatrix}
    \end{bmatrix}, 
    \Tilde{\mathbf{e}} = \begin{bmatrix}
       \colorbox{my_green}{$ \mathbf{x}_{o} +
      a\mathbf{d}_{o,i}\cos{ \boldsymbol{\alpha}_{o,i}} \sin{ \boldsymbol{\beta}_{o,i}}$} \\ \colorbox{my_pink}{$\mathbf{d}_{v,i}v_{max}\cos{\boldsymbol{\alpha}_{v,i}}\sin{\boldsymbol{\beta}_{v,i}}$} \\
\colorbox{my_blue}{$\mathbf{d}_{a,i}a_{max}\cos{\boldsymbol{\alpha}_{a,i}}\sin{\boldsymbol{\beta}_{a,i}}$}\\
       \colorbox{my_green}{$ \mathbf{y}_{o} +
      a\mathbf{d}_{o,i}\sin{ \boldsymbol{\alpha}_{o,i}} \sin{ \boldsymbol{\beta}_{o,i}}$} \\
\colorbox{my_pink}{$\mathbf{d}_{v,i}v_{max}\sin{\boldsymbol{\alpha}_{v,i}}\sin{\boldsymbol{\beta}_{v,i}}$}\\
\colorbox{my_blue}{$\mathbf{d}_{a,i}a_{max}\sin{\boldsymbol{\alpha}_{a,i}}\sin{\boldsymbol{\beta}_{a,i}}$}\\
     \colorbox{my_green}{$ \mathbf{z}_{o} +
      b\hspace{0.1cm} \mathbf{d}_{o,i}\cos{ \boldsymbol{\beta}_{o,i}}$} \\
\colorbox{my_pink}{$\mathbf{d}_{v,i}v_{max}\cos{\boldsymbol{\beta}_{v,i}}$}\\
\colorbox{my_blue}{$\mathbf{d}_{a,i}a_{max}\cos{\boldsymbol{\beta}_{a,i}}$}
    \end{bmatrix}\hspace{-0.15cm},\label{f_e}
  \end{align} 
  \normalsize

\noindent where $\mathbf{F}_{o}$ is formed by stacking as many times as the number of obstacles. Also, $\mathbf{x}_{o}, \mathbf{y}_{o}, \mathbf{z}_{o}$ is obtained by stacking $x_{o,j}(t),y_{o,j}(t), z_{o,j}(t)$ at different time stamps and for all obstacles. 

\noindent \textbf{Solution Process:}
I relax the equality and affine constraints \eqref{reform_bound}-\eqref{bound_con} as $l_{2}$ penalties utilizing the augmented Lagrangian method

\vspace{-0.3cm}
\small
\begin{align}
\mathcal{L}&=\frac{1}{2}\left\Vert \overline{\boldsymbol{\xi}}_{i} - \boldsymbol{\xi}_{i}\right\Vert ^{2}_{2}- \langle\boldsymbol{\lambda}_{i}, \overline{\boldsymbol{\xi}}_{i}\rangle 
 +\frac{\rho}{2}\left\Vert \Tilde{\mathbf{F}}\overline{\boldsymbol{\xi}}_{i}  
 -\Tilde{\mathbf{e}} \right \Vert_2^2 
 +\frac{\rho}{2}\left\Vert \mathbf{G}\overline{\boldsymbol{\xi}}_{i}  
 -\boldsymbol{\tau}+\mathbf{s}_{i} \right \Vert_2^2  \nonumber \\
 &= \frac{1}{2}\left\Vert \overline{\boldsymbol{\xi}}_{i} - \boldsymbol{\xi}_{i}\right\Vert ^{2}_{2}- \langle\boldsymbol{\lambda}_{i}, \overline{\boldsymbol{\xi}}_{i}\rangle 
 +\frac{\rho}{2}\left\Vert \mathbf{F}\overline{\boldsymbol{\xi}}_{i}  
 -\mathbf{e} \right \Vert_2^2 
 \label{lagrange} 
 \end{align}
\normalsize

\noindent where,
$\hspace{-0.05cm}\mathbf{F} = \hspace{-0.1cm}\begin{bmatrix}
    \Tilde{\mathbf{F}} \\ \mathbf{G}
\end{bmatrix},
\mathbf{e} =\hspace{-0.1cm}\begin{bmatrix}
   \Tilde{\mathbf{e}} \\ \boldsymbol{\tau}\hspace{-0.1cm}-\mathbf{s}_{i}
\end{bmatrix}$. Also, $\boldsymbol{\lambda}_{i}$, $\rho$ and $\textbf{s}_{i}$ are Lagrange multiplier, scalar constant and slack variable. I reduce the problem \eqref{lagrange} subject to \eqref{eq_modify_our} using \gls{am} method \cite{jain2017non} as follows

\small
\begin{subequations}
\begin{align}
    ^{k+1}\boldsymbol{\alpha}_{i} &= \text{arg} \min_{\boldsymbol{\alpha}_{i}}\mathcal{L}(^{k}\overline{\boldsymbol{\xi}}_{i}, \boldsymbol{\alpha}_{i}, \hspace{0.1cm}^{k}\boldsymbol{\beta}_{i}, \hspace{0.1cm}^{k}\mathbf{d}_{i}, \hspace{0.1cm}^{k}\boldsymbol{\lambda}_{i},
    \hspace{0.1cm}^{k}\mathbf{s}_{i} ) \label{comp_alpha} \\
    ^{k+1}\boldsymbol{\beta}_{i} &= \text{arg} \min_{\boldsymbol{\beta}_{i}}\mathcal{L}(^{k}\overline{\boldsymbol{\xi}}_{i}, \hspace{0.1cm}^{k+1}\boldsymbol{\alpha}_{i}, \boldsymbol{\beta}_{i}, \hspace{0.1cm}^{k}\mathbf{d}_{i}, \hspace{0.1cm}^{k}\boldsymbol{\lambda}_{i},
    \hspace{0.1cm}^{k}\mathbf{s}_{i} ) \label{comp_beta}\\
    ^{k+1}\mathbf{d}_{i} &= \text{arg} \min_{\textbf{d}_{i}}\mathcal{L}(\hspace{0.1cm}^{k}\overline{\boldsymbol{\xi}}_{i}, \hspace{0.1cm}^{k+1}\boldsymbol{\alpha}_{i}, \hspace{0.1cm}^{k+1}\boldsymbol{\beta}_{i},\mathbf{d}_{i}, \hspace{0.1cm}^{k}\boldsymbol{\lambda}_{i},
    \hspace{0.1cm}^{k}\mathbf{s}_{i} )\label{comp_d}\\
     ^{k+1}\mathbf{s}_{i} &= \max (\mathbf{0}, -\textbf{G}\hspace{0.1cm}^{k}\overline{\boldsymbol{\xi}}_{i}+ \boldsymbol{\tau})\\
    \hspace{0.1cm}^{k+1}\boldsymbol{\lambda}_{i} &= ^{k}\boldsymbol{\lambda}_{i}
- \rho \mathbf{F}^{T}(\mathbf{F}\hspace{0.1cm} ^{k}\overline{\boldsymbol{\xi}}_{i} 
 -^{k}\Tilde{\mathbf{e}})\label{lag} \\
     ^{k+1}\mathbf{e}&=
     \begin{bmatrix}
        \Tilde{\mathbf{e}}(\hspace{0.1cm}^{k+1}\boldsymbol{\alpha}_{i},^{k+1}\boldsymbol{\beta}_{i}, ^{k+1}\textbf{d}_{i}) \label{obtain_e}\\
    \boldsymbol{\tau} -^{k+1}\mathbf{s}_{i}\end{bmatrix}\\
  ^{k+1}\overline{\boldsymbol{\xi}}_{i} &= \text{arg} \min_{\overline{\boldsymbol{\xi}}_{i}}\mathcal{L}({\boldsymbol{\xi}}_{i}, {^{k+1}}\mathbf{e}_i,  ^{k+1}\boldsymbol{\lambda}_{i},
) \label{comp_zeta}
\end{align}
\end{subequations}
\normalsize
\vspace{-0.2cm}

\noindent where that stacking of right-hand sides of \eqref{obtain_e} and \eqref{lag} provide the function $\mathbf{h}$ presented~in \eqref{fixed_point_1} and  \eqref{comp_zeta} is a representation of \eqref{proj_cost}-\eqref{reform_bound_p}. Also, the steps \eqref{comp_alpha}-\eqref{comp_d} have closed form solutions in terms of ${^k}\overline{\boldsymbol{\xi}}_i$ (see analysis of Algorithm \ref{alg_1} and Algorithm \ref{alg_2} in Sections \ref{main_result_1} and \ref{main_results_2}). Note that for each \gls{am} step, I only optimize one group of variables while others are held fixed. 

\vspace{0.45cm}
\noindent \textbf{GPU Accelerated Batch Operation:} The proposed method involves projecting multiple sampled trajectories onto the feasible set (see Figure\ref{fig_overview}). However, as mentioned before, performing these projections sequentially can be computationally burdensome. Fortunately, my projection optimizer exhibits certain structures conducive to batch and parallelized operations. 
To delve into this concept, consider the matrix $\mathbf{M}$ in \eqref{affine_trans}, which remains constant regardless of the input trajectory sample $\boldsymbol{\xi}_{i}$. In essence, $\mathbf{M}$ remains unchanged irrespective of the specific trajectory sample requiring projection onto the feasible set. This characteristic enables us to express the solution \eqref{affine_trans} for all trajectory samples $\boldsymbol{\xi}_{i}, i = 1,2,\dots,N_b$ as a single large matrix-vector product. This computation can be effortlessly parallelized across GPUs.
Similarly, acceleration is attainable for the function $\mathbf{h}$, which primarily involves element-wise products and sums. This allows for efficient parallelization and leveraging GPU capabilities to enhance overall performance.

\begin{figure}[!h]
    \centering
        \includegraphics[scale=0.55]{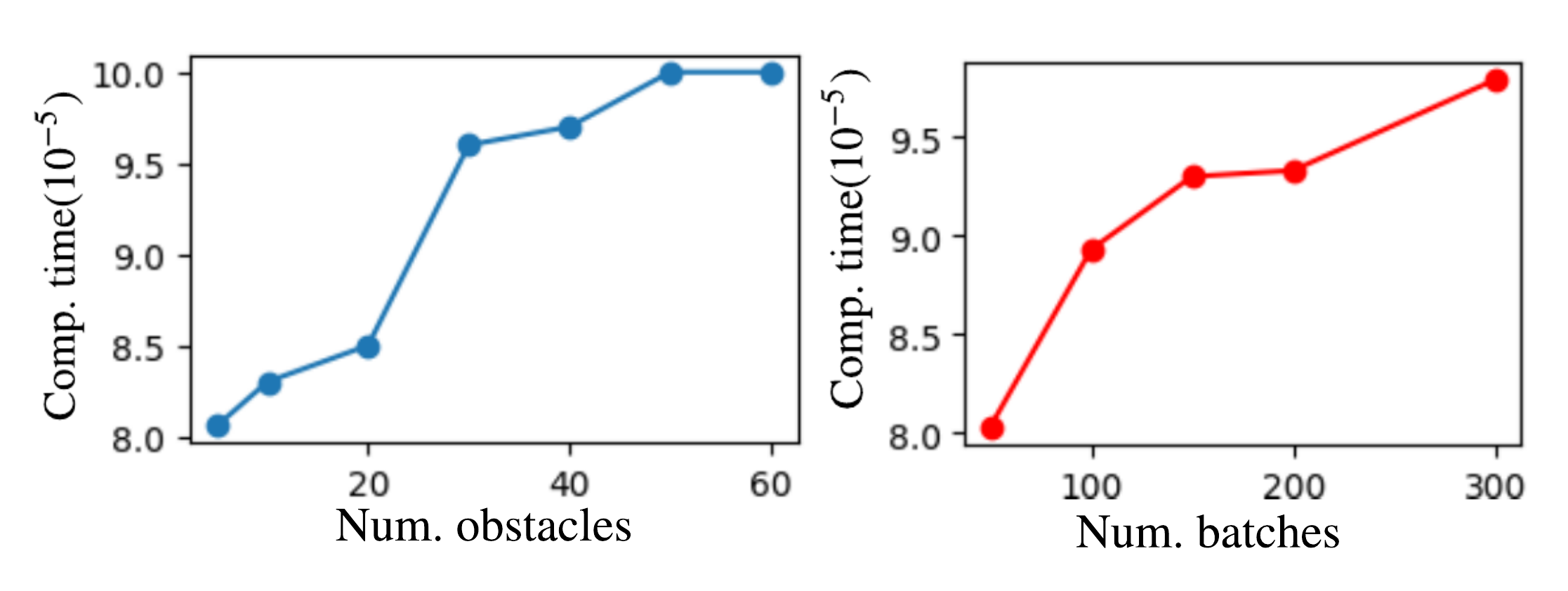}
    \caption[PRIEST scalability]{Scalability of per-iteration computation time of my projection optimizer to number of obstacles and batch size}
    \label{comp_time}
\end{figure}

\vspace{0.45cm}
\noindent \textbf{Scalability:} The matrix $\mathbf{M}$ in \eqref{affine_trans} requires a one-time computation, as $\mathbf{A}$ and $\mathbf{F}$ remain constant throughout the projection iteration. Conversely, the matrix $\boldsymbol{\eta}$ is recalculated at each iteration, with its computational cost primarily dominated by $\hspace{-0.05cm}\mathbf{F}^{T}\hspace{0.1cm}{^{k+1}}\mathbf{e}_{i}$. The number of rows in both $\mathbf{F}$ and $\mathbf{e}$ increases linearly with the planning horizon, the number of obstacles, or the batch size. This characteristic, in conjunction with GPU acceleration, shows my approach with remarkable scalability for long-horizon planning in highly cluttered environments.
Figure \ref{comp_time} presents the average per-iteration time of the projection optimizer. It also demonstrates how it scales with the number of obstacles and batch size. 

\subsection{Projection Guided Sampling-Based Optimizer}
\noindent\textbf{Algorithm Description:} Algorithm \ref{alg_3} presents another core contribution of this paper. It starts by generating $N_{b}$ samples of polynomial coefficients $\boldsymbol{\xi}$ from a Gaussian distribution with $\mathcal{N}({^l}\boldsymbol{\mu},\hspace{-0.08cm}{^l}\boldsymbol{\Sigma})$ at iteration $l\hspace{-0.1cm}=\hspace{-0.1cm}0$ (line 3). The~sampled $\hspace{-0.05cm}\boldsymbol{\xi}_i\hspace{-0.09cm}$ is subsequently projected onto the feasible set (lines~4-5). Due to real-time constraints, it may not be impractical to execute the projection optimization long enough to push all the sampled trajectories to feasibility. Thus, in line 6, I compute the constraint residuals $r(\overline{\boldsymbol{\xi}}_{i})$ associated with each sample. In line 7, I select the top $N_{proj}$ samples with the least constraint residuals and append them to the list $ConstraintEliteSet$. Lines 8-9 involve the creation of an augmented cost, $c_{aug}$, achieved by appending the residuals to the primary cost function. Subsequently, $c_{aug}$ is evaluated on the samples from $ConstraintEliteSet$.
In line 11, I once again select the top $N_{elite}$ samples with the lowest $c_{aug}$ and append them to the list $EliteSet$. Finally, moving to line 12, I update the distribution based on the samples of the $EliteSet$ and the associated $c_{aug}$ values. The final output of the optimizer is the sample from the $EliteSet$ with the lowest $c_{aug}$.

\vspace{-0.47cm}
 \small
    \begin{subequations}
        \begin{align}
         ^{l+1}\boldsymbol{\mu} &= (1-\sigma)\hspace{0.1cm} ^{l}\boldsymbol{\mu} +\hspace{-0.05cm} \sigma (\frac{1}{\sum\limits_{m \in C} c_{m}})
         \sum\limits_{m\in C}\overline{\boldsymbol{\xi}}_{m} c_{m}, \label{update_mean} \\
        \hspace{-0.27cm}^{l+1} \boldsymbol{\Sigma}\hspace{-0.1cm} &= \hspace{-0.1cm}(1\hspace{-0.1cm}-\hspace{-0.1cm}\sigma)  \hspace{0.05cm}^{l}\boldsymbol{\Sigma} \hspace{-0.1cm}+ \hspace{-0.1cm} \sigma \frac{\sum\limits_{m \in C} c_{m}(\overline{\boldsymbol{\xi}}_{m}-^{~l+1}\boldsymbol{\mu})(\overline{\boldsymbol{\xi}}_{m}-^{~l+1}\boldsymbol{\mu})^{T}}{\sum\limits_{m \in C} c_{m}}, 
         \label{update_covariance} \\
         c_{m} &= \exp\big({\gamma^{-1}(c_{aug}(\overline{\boldsymbol{\xi}}_{m}) - \delta )}\big),
        \end{align}
    \end{subequations}
    \normalsize

\vspace{-0.15cm}
\noindent where the scalar constant $\sigma$ is the so-called learning rate. The set $C$ consists of the top $N_{elite}$ selected trajectories (line 11). The constant $\gamma$ specifies the sensitivity of the exponentiated cost function $c_{aug}(\overline{\boldsymbol{\xi}}_{m})$ for top selected trajectories. $\delta = \min c_{aug}(^{l}\overline{\boldsymbol{\xi}}_{m})$ is defined to prevent numerical instability.

\begin{algorithm}[t]
\scriptsize
\DontPrintSemicolon
\SetAlgoLined
\SetNoFillComment
\caption{\scriptsize Projection Guided Sampling-Based Optimization (PRIEST) }\label{alg_3}
\KwInput{Initial states}
\KwInitialization{Initiate $^{l}\boldsymbol{\mu}$ and $^{l}\boldsymbol{\Sigma} $ at $i=0$}
\For{$l \leq N$}{
\hspace{-0.15cm}Initialize $CostList = []$
    
\hspace{-0.15cm}Draw $N_{b}$ samples $\boldsymbol{\xi}_{1},...,\boldsymbol{\xi}_{N_{b}}$ from $\mathcal{N}(^{l}\boldsymbol{\mu},^{l}\boldsymbol{\Sigma})$
   
\hspace{-0.15cm}Solve the inner convex optimizer to obtain $\overline{\boldsymbol{\xi}}_{i}$
    \begin{subequations}
        \begin{align}
            &\min_{\overline{\boldsymbol{{\xi}}}_{i}} \frac{1}{2} \left\Vert \overline{\boldsymbol{\xi}}_{i} - \boldsymbol{\xi}_{i}  \right\Vert ^{2}_{2}  
            \label{proj_cost} \\
        &~\text{s.t.: }\bold{A}\overline{\boldsymbol{\xi}}_{i}= \bold{b}_{eq}\label{eq_modify_p} \\
           &~~~~~~~~ \mathbf{g}(\overline{\boldsymbol{\xi}}_{i}) \leq \mathbf{0} \label{reform_bound_p},
        \end{align}
    \end{subequations}

    \vspace{-0.15cm}
   \hspace{-0.15cm} Compute the residuals set $r(\overline{\boldsymbol{\xi}}_{i})$ 
   
\hspace{-0.15cm} $ConstraintEliteSet\leftarrow$   Select $N_{proj}$ samples from $r(\overline{\boldsymbol{\xi}}_{i})$ with the lowest values. 
    
\hspace{-0.15cm}Evaluate the cost
    
    $c_{aug} \leftarrow  c_1(\overline{\boldsymbol{\xi}}_{i})+ r(\overline{\boldsymbol{\xi}}_{i})$
    
    
   $EliteSet \leftarrow$  Select $N_{elite}$ top samples with the lowest cost obtained from the $CostList$.
   
    
    Update the new mean and covariance, $^{l+1}\boldsymbol{\mu}$~and~$^{l+1}\boldsymbol{\Sigma}$, using \eqref{update_mean}-\eqref{update_covariance}
    
}
\hspace{-0.07cm}\textbf{Return} \hspace{-0.06cm}$\overline{\boldsymbol{\xi}}_{i}$\hspace{-0.09cm} corresponding to the lowest cost in \hspace{-0.06cm}$\small EliteSet \normalsize$
\end{algorithm}
\vspace{0.45cm}
\section{Validation and Benchmarking}

\noindent \textbf{Implementation Details:} I implemented Algorithm \ref{alg_3}, \gls{priest}, using Python with the JAX \cite{bradbury2018jax} library as the GPU-accelerated algebra backend. Our simulation framework for experiments was developed on the ROS \cite{koubaa2017robot} platform, utilizing the Gazebo physics simulator. All benchmarks were conducted on a Legion7 Lenovo laptop featuring an Intel Core i7 processor and an Nvidia RTX 2070 GPU. I employed the open3d library for downsampling PointCloud data \cite{zhou2018open3d}.
I selected $t_{f}\hspace{-0.1cm}=\hspace{-0.1cm}10,N \hspace{-0.1cm}=\hspace{-0.1cm}13,N_{b}\hspace{-0.1cm}=\hspace{-0.1cm}110$, $N_{proj}\hspace{-0.1cm}=\hspace{-0.1cm}80$ and $N_{elite}\hspace{-0.1cm}=\hspace{-0.1cm}20$. All the compared baselines operated with the same planning horizon. Furthermore, to provide the best possible opportunity for \gls{mppi} and log-MPPI, they were executed with their default sample size of 2496, which is over 20 times higher than that utilized by \gls{priest}. The benchmarking against other baselines is based on the following metrics:

\begin{itemize}
    \item \textbf{Success Rate:} A run is considered successful when the robot approaches the final point within a 0.5m radius without any collision. The success rate is calculated as the ratio of the total number of successful runs to the overall number of runs.
     \item \textbf{Travel Time:} This metric represents the duration it takes for the robot to reach the vicinity of the goal point.
     \item \textbf{Computation Time:} This metric quantifies the time required to calculate a solution trajectory.
\end{itemize}

Furthermore, I compared my approach with different baselines in four sets of benchmarks, including:

\begin{itemize}
\vspace{0.15cm}
    \item \textbf{Comparison on \gls{barn} Dataset \cite{perille2020benchmarking}:}
    In this benchmark, I used a holonomic mobile robot modeled as a double-integrator system. The robot is tasked with iteratively planning its trajectory through an obstacle field in a receding horizon manner. Consequently, the differential flatness function for extracting control inputs was defined as $\boldsymbol{\Phi} = (\ddot{x}(t),\ddot{y}(t))$. The cost function ($c_1$) took the following form:

\small
\begin{align}
    \sum_t{\ddot{x}(t)^2+\ddot{y}(t)^2+c_{\kappa}+c_{p}}
    \label{barn_cost}
\end{align}
\normalsize

\noindent where $c_{\kappa}=(\frac{\ddot{y}(t)\dot{x}(t)-\ddot{x}(t)\dot{y}(t) }{(\dot{x}(t)^2+\dot{x}(t)^2)^{1.5}})^2$ penalizing curvature, $c_{p}$ minimizes the orthogonal distance of the computed trajectory from a desired straight-line path to the goal. It is important to note that $c_{p}$ does not have an analytical form as it necessitates the computation of the projection of a sampled trajectory waypoint onto the desired path.
I utilized the \gls{barn} dataset, consisting of 300 environments with varying complexity, designed to create local-minima traps for the robot. The evaluation involves comparing our approach against \gls{dwa} \cite{fox1997dynamic}, \gls{teb} \cite{rosmann2017kinodynamic} implemented in the \gls{ros} navigation stack, \gls{mppi} \cite{williams2017model}, and log-MPPI \cite{mohamed2022autonomous}. All baselines, including \gls{priest}, had access only to the local cost map or point cloud. \gls{teb} and \gls{dwa} employed a combination of graph search and optimization, while \gls{priest} and log-MPPI were purely optimization-based approaches. 

\vspace{0.15cm}
\item \textbf{Point to Point Navigation with Differentiable Cost:} 
In this benchmark, our aim is to generate a single trajectory between a start and a goal location. The cost function  $c_1$ incorporates the first term of \eqref{barn_cost}. For comparison, I considered \gls{sota} gradient-based optimizers ROCKIT \cite{metz2003rockit} and \gls{fatrop} \cite{vanroye2023fatrop} and sampling-based optimizers \gls{cem}, and \gls{vpsto} \cite{jankowski2023vp}.

\vspace{0.15cm}
\item \textbf{Comparison in a Dynamic Environment:} In this evaluation, I benchmark against \gls{cem}, log-MPPI, \gls{mppi}, \gls{teb} and \gls{dwa}, employing the same cost function as used for \gls{barn} Dataset \eqref{barn_cost}. In this dynamic scenario, I introduced ten obstacles, each with a velocity of $0.1m/s$, moving in the opposite direction of the robot. Simulations were conducted across 30 distinct obstacle configurations and velocities. The robot model used for this benchmark is a nonholonomic mobile robot known as Jackal.

\end{itemize}

\subsection{Qualitative Results}

\noindent \textbf{A Simple Benchmark:}
Figure \ref{fig_teaser} presents a comparison of the behaviors of \gls{cem} and \gls{priest} in a scenario wherein all the initial sampled trajectories lie within a high-cost/infeasible region. The results clearly demonstrate that while \gls{cem} samples persistently remain in the infeasible region, \gls{priest} samples move toward the out-of-infeasible region. Additionally, Figure \ref{fig_teaser} (g-h) empirically validates Algorithm \ref{alg_3} by illustrating the gradual reduction and saturation of both constraint residuals and cost values as the iterations progress. The projection optimizer is essentially a set of analytical transformations over the sampled trajectories. Thus, Algorithm \ref{alg_3} retains the convergence properties of the base sampling-based optimizer upon which the projection part is embedded. For example, it will~inherit the properties of CEM when integrated with \gls{cem}-like a sampler. Moreover, the projection optimizer by construction satisfies the boundary constraints on the trajectories and ensures a reduction in the cost by pushing the samples toward the feasible region.

\begin{figure}[!h]
    \includegraphics[scale=0.5]{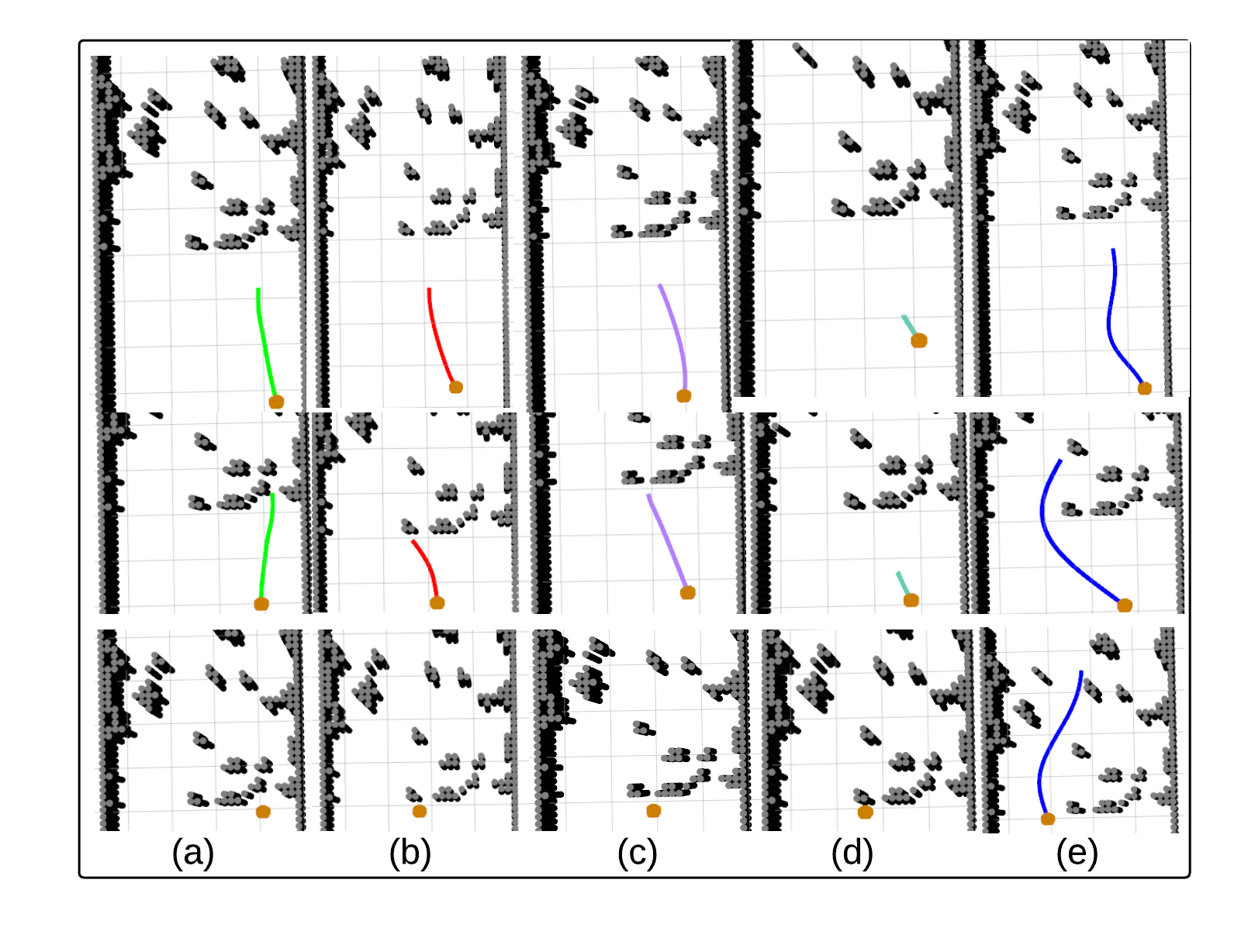}
    \vspace{-0.7cm}
    \caption[Comparative visualization of qualitative results from MPC built on MPPI, log-MPPI, TEB, DWA, and PRIEST]{\small{Comparative visualization of qualitative results from \gls{mpc} built on \gls{mppi} (a), log-MPPI(b), \gls{teb}(c), \gls{dwa}(d), and \gls{priest}(e). I showed the best trajectory obtained from each optimizer at three distinct snapshots.}}
    \label{methods_traj}
    \end{figure}

\begin{figure}[!h]
    \centering
    \includegraphics[scale=0.90]{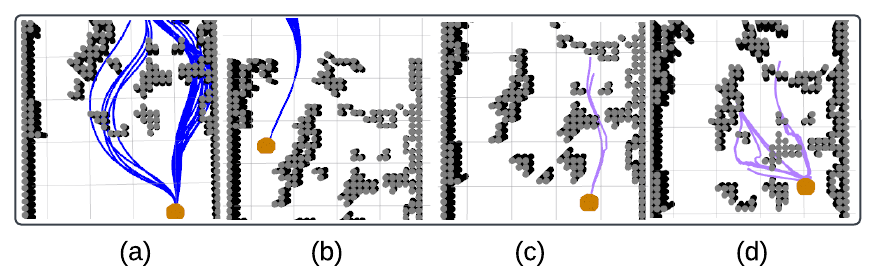}
    \vspace{-0.2cm}
    \caption[Qualitative result of MPC built on PRIEST and TEB]{\small{Qualitative result of MPC built on \gls{priest} (a-b) and \gls{teb} (c-d). Blue and purple trajectories show top samples in the PRIEST and TEB planner. }}
    \label{qual_homo}
\end{figure}

\vspace{0.45cm}
\noindent \textbf{Receding Horizon Planning on \gls{barn} Dataset:} 
In Figure \ref{methods_traj}, I show trajectories generated by \gls{priest} at three different snapshots within one of the \gls{barn} environments. These trajectories are compared with those produced by \gls{mppi}, \gls{teb}, \gls{dwa}, and log-MPPI in the same environment. As can be seen in Figure \ref{methods_traj}(e), \gls{priest} successfully generated collision-free trajectories while other methods faced challenges and got stuck (see the bottom row of Figure \ref{methods_traj}(a)-(d). Furthermore, to illustrate the exploration of various homotopies at each iteration, I compare the behavior of \gls{priest} with \gls{teb} in one of the \gls{barn} environments in Figure \ref{qual_homo}. While \gls{teb} employs graph search, \gls{priest} leverages the stochasticity inherent in the sampling process, guided by the projection optimizer. Consequently, \gls{priest} can explore over a more extended horizon and a broader state space. It is noteworthy that increasing the planning horizon of \gls{teb} significantly raises computation time, potentially degrading overall navigation performance rather than enhancing it. I provide a quantitative comparison with \gls{teb} and other baselines in the next section.

\begin{figure}[!h]
    \centering
        \includegraphics[scale=0.71]{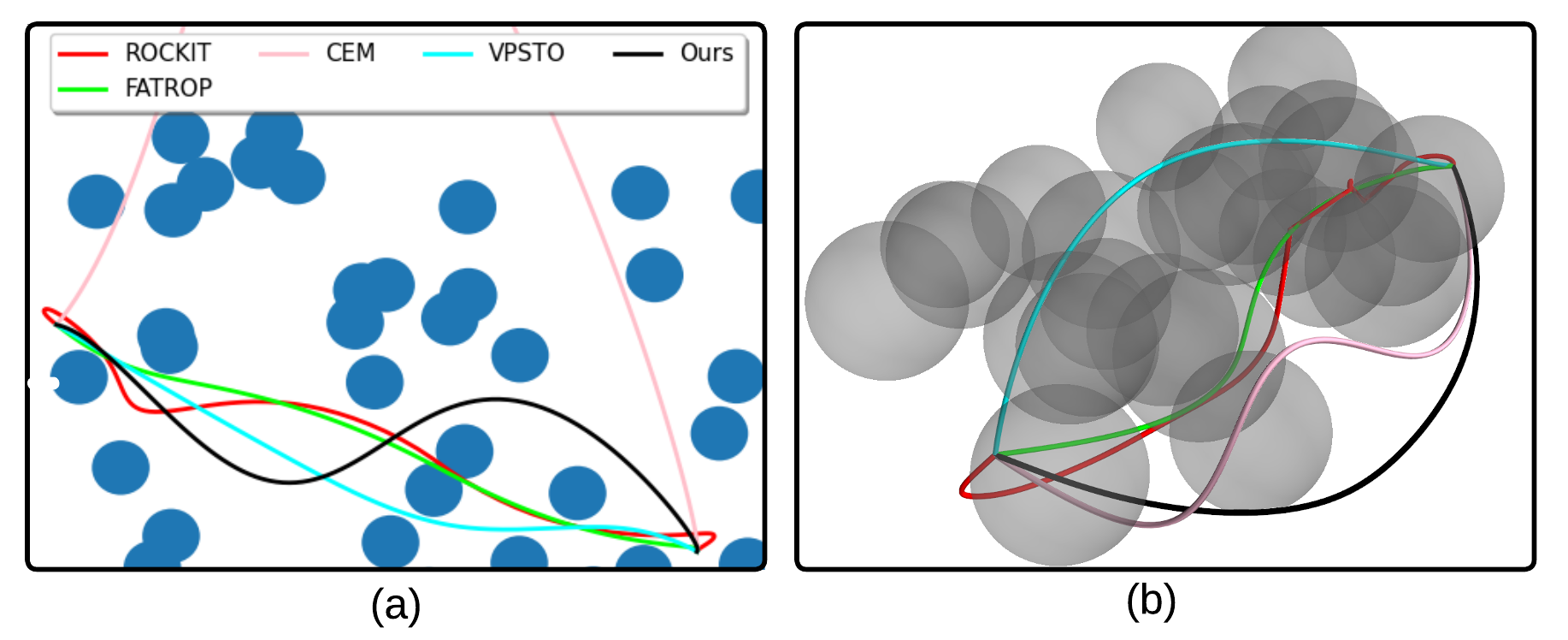}
    \caption[Qualitative result of point-to-point navigation.]{\small{Qualitative result of point-to-point navigation. Trajectories for different approaches have the same color in both 2D and 3D configurations.} }
    \label{traj_plot}
\end{figure}

\noindent\textbf{Point-to-Point Navigation Benchmark:} 
Figure \ref{traj_plot} illustrates trajectories generated by \gls{priest}, as well as those produced by gradient-based optimizers ROCKIT and \gls{fatrop}, and sampling-based optimizers \gls{cem} and \gls{vpsto}. In the specific 2D example presented, both \gls{priest} and \gls{vpsto} successfully generated collision-free trajectories, surpassing the performance of other baselines. In the showcased 3D environment, only \gls{priest} and \gls{cem} achieved collision-free trajectories. In the next section, I present the quantitative statistical trends for all the baselines across different randomly generated environments.

\begin{figure}[!h]
\centering
    \includegraphics[scale=0.55]{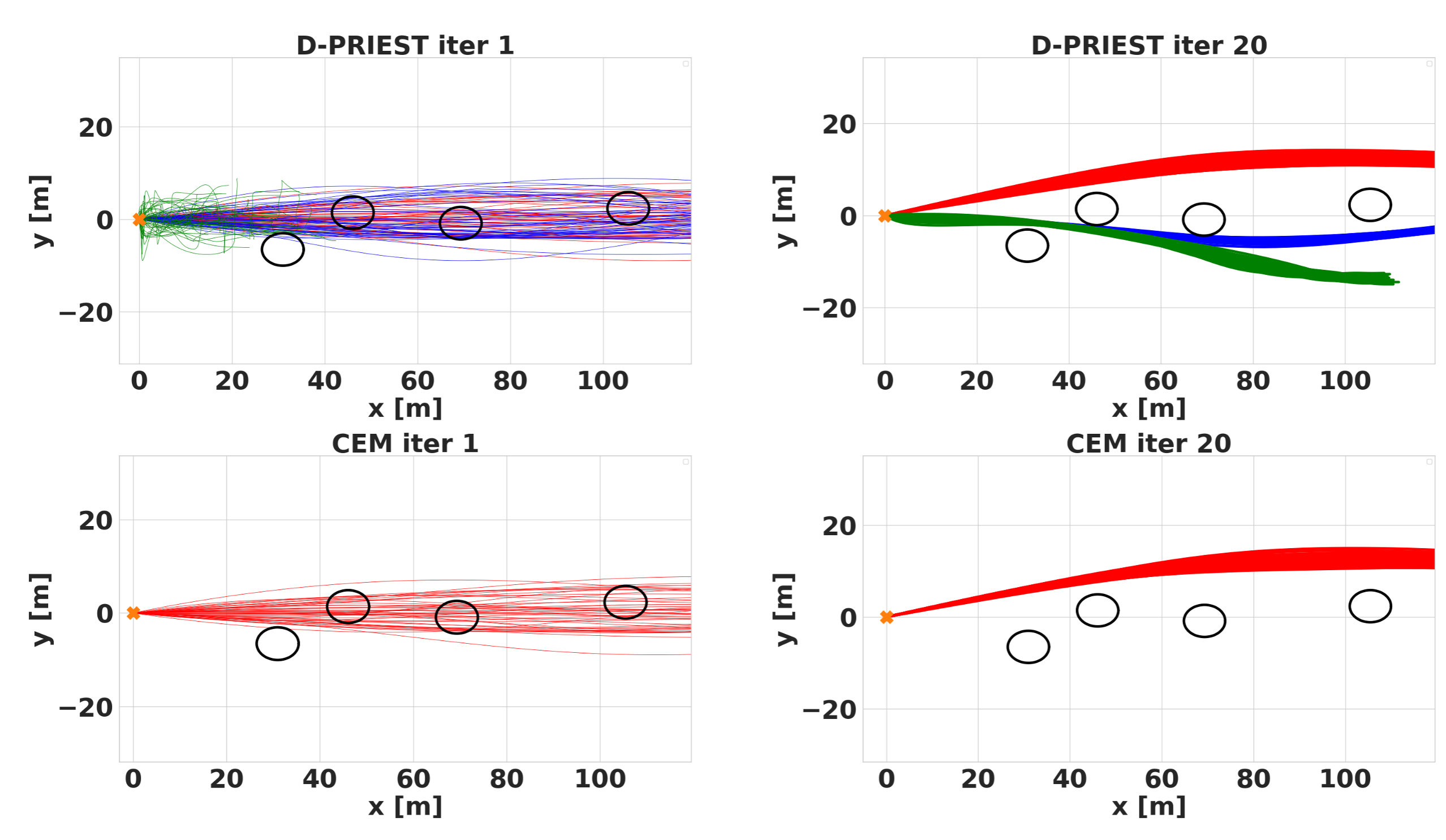}
    \caption[Comparison of D-PRIEST with baseline cem]{\small{Comparison of D-PRIEST with baseline \gls{cem}. As is shown, the former updates multiple parallel distributions, resulting in multi-modal optimal trajectory distribution upon convergence. D-PRIEST maintained three different distributions (shown in green, red, and blue) and thus could obtain multi-modal behavior. In contrast, \gls{cem}, which only has a single distribution, provides a limited set of options for collision-free trajectories.}}
    \label{com_plot}
\end{figure}

\noindent\textbf{Decenteralized Variant:}
Figure \ref{com_plot} shows the application of D-PRIEST for car-like vehicle trajectory planning. The cost function $c_1$ penalizes the magnitude of axis-wise accelerations and steering angle. Leveraging the differential flatness property, I express the steering angle as a function of axis-wise velocity and acceleration terms \cite{han2023efficient}. In Figure \ref{com_plot}, D-PRIEST demonstrates three different distributions (depicted in green, red, and blue) in parallel, resulting in multi-modal behaviors. These maneuvers intuitively correspond to overtaking static obstacles (depicted in black) from left to right or slowing down and shifting to another lane. In contrast, traditional \gls{cem} could only obtain a single maneuver for the vehicle.

\subsection{Quantiative Results}

\noindent\textbf{Comparison with \gls{mppi}, Log-MPPI, \gls{teb}, \gls{dwa}:} Table 1 summarizes the quantitative results. \gls{priest} achieves a 90\% success rate, outperforming the best baseline (\gls{teb}) with an 83\% success rate. \gls{teb}, along with \gls{dwa}, employs graph search and complex recovery maneuvers to improve success rates, albeit with increased travel time. Conversely, purely optimization-based approaches \gls{mppi} and log-MPPI exhibit a 32\% and 35\% lower success rate than PRIEST, accompanied by slightly higher travel times. Although \gls{priest} shows a slightly higher mean computation time, it remains fast enough for real-time applications.
  \setlength{\arrayrulewidth}{1pt}
\begin{table}[h] 
\small
\caption{Comparisons on the \gls{barn} 
Dataset}\label{Barn}
\centering
\begin{tabular}{|l|c|c|c|c|}
\hline
                             Method    & \cellcolor[HTML]{4EE6ED}\begin{tabular}[c]{@{}c@{}}Success\\  rate\end{tabular} & \cellcolor[HTML]{4EE6ED}\begin{tabular}[c]{@{}c@{}}Travel time (s)\\    Mean/ Min/Max\end{tabular}  & \cellcolor[HTML]{4EE6ED}\begin{tabular}[c]{@{}c@{}}Computation time (s)\\ Mean/ Min/Max\end{tabular} \\ \hline
\cellcolor[HTML]{9AFF99}DWA      & 76\%                                                                          & \begin{tabular}[c]{@{}c@{}}52.07/ 33.08/145.79\end{tabular}                                                                   & \begin{tabular}[c]{@{}c@{}}0.037/ 0.035/0.04\end{tabular}                                            \\ \hline
\cellcolor[HTML]{9AFF99}TEB      & 83\%                                                                          & \begin{tabular}[c]{@{}c@{}}52.34/ 42.25/106.32\end{tabular}                                                                        & \begin{tabular}[c]{@{}c@{}}0.039/0.035/0.04\end{tabular}                                            \\ \hline
\cellcolor[HTML]{9AFF99}MPPI     & 58\%                                                                            & \begin{tabular}[c]{@{}c@{}}36.66/ 31.15/99.62\end{tabular}                                                                         & \begin{tabular}[c]{@{}c@{}}0.019/0.018/0.02\end{tabular}                                            \\ \hline
\cellcolor[HTML]{9AFF99}log-MPPI & 55\%                                                                            & \begin{tabular}[c]{@{}c@{}}36.27/30.36/58.84\end{tabular}                                                                          & \begin{tabular}[c]{@{}c@{}}0.019/0.018/0.02\end{tabular}                                            \\ \hline
\cellcolor[HTML]{9AFF99}{\textbf{PRIEST}}   & \textbf{90\%}                                                                 & \begin{tabular}[c]{@{}c@{}}\textbf{33.59}/ 30.03/70.98\end{tabular}                                                                           & \begin{tabular}[c]{@{}c@{}}0.071/0.06/0.076\end{tabular}                                            \\ \hline
\end{tabular}
\normalsize
\end{table}

\vspace{0.45cm}
 \noindent\textbf{Comparison with Additional Gradient-Based and Sampling-based Optimizers:} Table \ref{grad_based} compares the performance of \gls{priest} with all the baselines in 2D and 3D cluttered environments (see Figure \ref{traj_plot}). ROCKIT and \gls{fatrop} were initialized with simple straight-line trajectories between the start and the goal, typically not collision-free. Due to conflicting gradients from neighboring obstacles, both methods often failed to obtain a collision-free trajectory. Interestingly, the sampling-based approaches did not fare much better, as both \gls{cem} and \gls{vpsto} reported a large number of failures. We attribute the failures of \gls{vpsto} and \gls{cem} to two reasons. First, most of the sampled trajectories for both \gls{cem} and \gls{vpsto} fell into the high-cost/infeasible area, creating a pathologically difficult case for sampling-based optimizers. Second, both \gls{cem} and \gls{vpsto} roll constraints into the cost as penalties and can be sensitive to tuning the individual cost terms. In summary, Table \ref{grad_based} highlights the importance of PRIEST, which uses convex optimization to guide trajectory samples toward constraint satisfaction. 
  
  \gls{priest} also shows superior computation time than ROCKIT and \gls{fatrop}. The \gls{cem} run times are comparable to PRIEST. Although \gls{vpsto} numbers are high, we note that the original author implementation that we use may not have been optimized for computation speed.

\begin{table}[t]
\vspace{-0.01cm}
\caption{Comparing \gls{priest} with Gradient/Sampling-Based Optimizers }
\small
\label{grad_based}
\centering
\begin{tabular}{|c|c|c|c|}
\hline
Method & Success rate  & \begin{tabular}[c]{@{}c@{}}Computation time (s) \color{black}(Mean/Min/Max)\end{tabular} \\ \hline
\rowcolor[HTML]{CBCEFB} 
ROCKIT-2D & 46\%          & 2.57/0.6/6.2                       \\ \hline
\rowcolor[HTML]{CBCEFB} 
FATROP-2D & 64\%          & 0.63/0.07/2.87                                                           \\ \hline
\rowcolor[HTML]{CBCEFB} 
\textbf{PRIEST-2D}   & \textbf{95\%} & \textbf{0.043/0.038/0.064}                            
\\ \hline
\rowcolor[HTML]{CBCEFB} 
CEM-2D   & 78\%         & 0.017/0.01/0.03                                                              \\ \hline
\rowcolor[HTML]{CBCEFB} 
VPSTO-2D & 66\%         & 1.63/0.78/4.5                                                                                      \\ \hline
\rowcolor[HTML]{C0C0C0} 
ROCKIT-3D &      65\%     &      1.65/0.68/5                                                                            \\ \hline
\rowcolor[HTML]{C0C0C0} 
FATROP-3D &    81\%      &    0.088/0.034/0.23                                                                    \\ \hline
\rowcolor[HTML]{C0C0C0} 
\textbf{PRIEST-3D}  & \textbf{90\%} & \textbf{0.053/0.044/0.063}                                                       \\ \hline
\rowcolor[HTML]{C0C0C0} 
CEM-3D   & 74\%         & 0.028/0.026/0.033                                      \\ \hline
\rowcolor[HTML]{C0C0C0} 
VPSTO-3D & 37\%          & 3.5/0.93/3.5                                  \\ \hline
\end{tabular}
\normalsize
\end{table}
\vspace{0.45cm}
 \noindent \textbf{ Combination of Gradient-Based and Sampling-based Optimizers:} \noindent A simpler alternative to PRIEST can be just to use a sampling-based optimizer to compute a good guess for the gradient-based solvers \cite{kim2022mppi}. However, such an approach will only be suitable for problems with differentiable costs. Nevertheless, we evaluate this alternative for the point-to-point benchmark of Figure \ref{traj_plot}. We used CEM to compute an initial guess for ROCKIT and FATROP. The results are summarized in Table \ref{sto_based}. As can be seen, while the performance of both ROCKIT and FATROP improved in 2D environments, the success rate of the latter decreased substantially in the 3D variant. The main reason for this conflicting trend is that the CEM (or any initial guess generation) is unaware of the exact capabilities of the downstream gradient-based optimizer.  This unreliability forms the core motivation behind PRIEST, which outperforms all ablations in Table \color{red}\ref{sto_based}\color{black}. By embedding the projection optimizer within the sampling process itself (refer Algorithm\ref{alg_3}) and augmenting the projection residual to the cost function, we ensure that the sampling and projection complement each other.
\begin{table}[!h]
\caption{Comparing \gls{priest} with Hybrid Gradient-Sampling Baselines. }
\small
\centering
\label{sto_based}
\begin{tabular}{|c|c|c|c|}
\hline
Method   & Success rate & \begin{tabular}[c]{@{}c@{}}Computation time (s) \color{black} (Mean/Min/Max)\end{tabular}  \\ \hline
\rowcolor[HTML]{CBCEFB} 
ROCKIT-CEM & 94\%         & 2.07/0.69/6.0                                                         \\ \hline
\rowcolor[HTML]{CBCEFB} 
FATROP-CEM & 84\%          & 0.34/0.06/0.96   
\\ \hline
\rowcolor[HTML]{CBCEFB} 
\textbf{PRIEST-2D}\color{black}   & \textbf{95\%} & \textbf{0.043/0.038/0.064}                     
\\ \hline
\rowcolor[HTML]{C0C0C0} 
ROCKIT-CEM & \textbf{100\%}          & 1.19/0.7/2.56                                             \\ \hline
\rowcolor[HTML]{C0C0C0} 
FATROP-CEM & 25\%          & 0.056/0.039/0.079                                  \\ \hline
\rowcolor[HTML]{C0C0C0} 
\textbf{PRIEST-3D}\color{black}   & 90\% & \textbf{0.053/0.044/0.063}                              \\ \hline                       
\end{tabular}
\normalsize
\end{table}
\noindent\textbf{}{Benchmarking in Dynamic Environments:}
Table \ref{dynamic} presents the results obtained from the experiments in dynamic environments. Herein, the projection optimizer of PRIEST ensures collision constraint satisfaction with respect to the linear prediction of the obstacles' motions. By having a success rate of 83\%, our method outperforms other approaches. Furthermore, our method shows competitive efficiency with a mean travel time of 11.95 seconds. Overall, the results show the superiority of our approach in dealing with the complexities of cluttered dynamic environments, making it a promising solution for real-world applications in human-habitable environments.

\begin{table}[!h]
\caption{Comparisons in cluttered and dynamic environments}
\small
\centering
\label{dynamic}
\begin{tabular}{|
>{\columncolor[HTML]{4EE6ED}}p{2cm}|p{1cm}|p{4.1cm}|p{0.5cm}|}
\hline
\cellcolor[HTML]{FFFFFF}Method & \cellcolor[HTML]{9AFF99}Success rate  & \cellcolor[HTML]{9AFF99}Travel time(s)(mean,min/max) \\ \hline
log-MPPI                       &60\%                                 & 18.80/15.68/24.3                                                                            \\ \hline
MPPI                           & 53\%                                 &19.38,16.28/27.08                                                  \\ \hline
CEM                            & 46\%                                 & 11.95/9.57/14.21 
\\ \hline
DWA                            & 66\%                                 &      33.4,31.4/37.17  
\\ \hline\textbf{PRIEST}                          & \textbf{83\%}                        & \textbf{11.95}/\textbf{11.43/13.39}                                                              \\ \hline
\end{tabular}
\end{table}
\normalsize

\subsection{Real-world Demonstration}

\begin{figure}[!h]
\centering
    \includegraphics[scale=0.67]{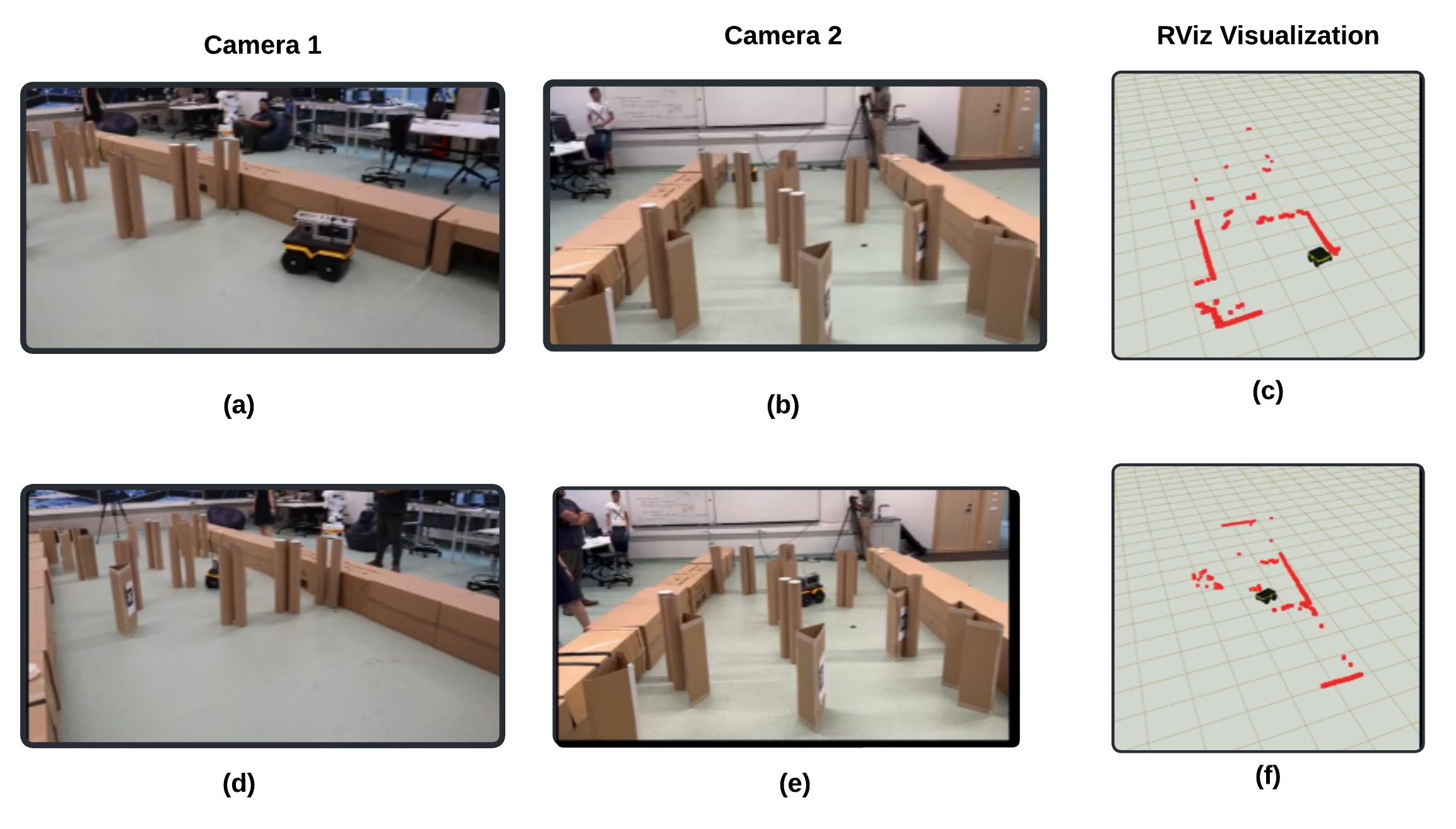}
    \caption[Real-world experiment for PRIEST]{\small{Two snapshots of PRIEST in BARN-like environment. Two cameras are used to show the environment. Also, for more clarity, the RViz visualization is added as well.}}
    \label{real_pre}
\end{figure}

To conduct real-world experiments and compare the performance of \gls{priest} against \gls{teb}, \gls{dwa}, \gls{mppi}, and log-MPPI, I created a series of random indoor cluttered environments simulating scenarios similar to those available in the \gls{barn} dataset, with dimensions of $4m \times 8m$. The experimental platform involved a Clearpath Jackal equipped with Velodyne \gls{lidar}, and the robot's state was tracked using the OptiTrack Motion Capture System. I plotted two snapshot of PRIEST implementation in a \gls{barn}-like environment in Figure \ref{real_pre}. More results of these experiments can be accessed on our website \footnote{\url{https://sites.google.com/view/priest-optimization}}. The website includes videos demonstrating that \gls{priest} consistently outperforms the considered baselines in terms of the success rate metric.

\section{Connection to the Rest of Thesis}

I extended the idea of proposing several initial guesses for trajectory optimization from our previous papers and utilized their techniques to leverage the optimization problem. 
    \chapter{Paper IV: Multi-agent Trajectory Optimization}\label{paper_4}

\section{Context}

Joint (or centralized) trajectory optimization for multiple agents is traditionally considered to be intractable. With existing approaches, generating joint trajectories for as few as 10 agents can take several seconds. This chapter/paper of the thesis challenges some of the established notions of joint trajectory optimization. In particular, I derive a novel optimizer that can compute trajectories of tens of agents in cluttered environments in a few tens of milliseconds.

\section{Problem Formulation}

The overall trajectory optimization is just a multi-agent version of the problem \eqref{g_1}-\eqref{g_3} introduced in Chapter \ref{literature}.

\begin{subequations}
\begin{align}
    &\min_{x_{i}(t), y_{i}(t), z_{i}(t)} \sum_{i=1}^{N_{a}}\sum_{t=0}^{n_{p}} \Big(\ddot{x}^{2}_{i}(t)+\ddot{y}^{2}_{i}(t)+\ddot{z}^{2}_{i}(t) \Big)\label{cost_multiagent}  \\
    &~\text{s.t.:} \nonumber \\
    &~
    (x_{i}(t),\dot{x}_{i}(t),\ddot{x}_{i}(t) y_{i}(t), \dot{y}_{i}(t),\ddot{y}_{i}(t) z_{i}(t),\dot{z}_{i}(t),\ddot{z}_{i}(t))|_{t=t_{0}} = \mathbf{b}_{0} \label{eq_multiagent_start}\\
   &~ (x_{i}(t),\dot{x}_{i}(t),\ddot{x}_{i}(t) y_{i}(t), \dot{y}_{i}(t),\ddot{y}_{i}(t) z_{i}(t),\dot{z}_{i}(t),\ddot{z}_{i}(t))|_{t=t_{f}} = \mathbf{b}_{f} \label{eq_multiagent_final}\\
  &~ -\frac{(x_{i}(t)-x_{o}(t))^{2}}{a^2}-\frac{(y_{i}(t)-y_{o}(t))^{2}}{a^2}-\frac{(z_{i}(t)-z_{o}(t))^{2}}{b^2}+1\leq 0, \label{quadratic_multiagent}
\end{align}
\end{subequations}

\noindent where in the cost function \eqref{cost_multiagent}, I aim to minimize the acceleration along the $x$, $y$, and $z$ axes. The vectors $\mathbf{b}_{0}$ and $\mathbf{b}_{f}$ denote the boundary values for each motion axis and its derivatives. The inequality constraint \eqref{quadratic_multiagent} captures collision avoidance constraints, modeling obstacles as spheroids with dimensions $(a, a, b)$. Solving this problem poses two main challenges: firstly, the complexity scales linearly with the number of agents, and secondly, the non-convex collision avoidance constraints exhibit exponential growth, ${n_o\choose{2}}$. In addition, $N_{a}$ and $n_{p}$ stand for the number of agents and number of planning steps, respectively. In the next sections, I elaborate on how I handle these challenges.


\section{High-Level Overview of the Main Algorithmic Results}

Just like in previous chapters, I show in this chapter that, at each iteration,  the core computations of the joint trajectory optimization can be reduced to solving a \gls{qp} of the following form:

\begin{align}
    &\min_{\boldsymbol{\xi}} \frac{1}{2}\boldsymbol{\xi}^T\mathbf{Q}\boldsymbol{\xi}+{^k}\mathbf{q}^T\boldsymbol{\xi} \\
   &~ \text{s.t.: }\mathbf{A}_{eq} \boldsymbol{\xi} = \mathbf{b}_{eq}.\label{over_multi}
\end{align}

 The most important thing to note is that only the vector $\mathbf{q}$ changes across iterations. This allows us to cache the most expensive matrix factorization and reduces the entire numerical steps to just computing matrix-matrix products that can be trivially accelerated over \gls{gpu}s.

In addition, I later show that the exact formulation derived even has more simplified structure where the trajectories along each motion axis can be decoupled and solved in parallel.

\section{Contribution}

The primary algorithmic challenge in multiagent trajectory optimization arises from the non-convex quadratic nature of inter-agent collision avoidance constraints, whose complexity escalates with an increasing number of agents. In this study, I address this challenge by first modeling non-convex collision avoidance as non-linear equality constraints through polar representation. Subsequently, I propose an \gls{am} procedure that organizes the optimization variables into specific blocks and optimizes them sequentially. An essential insight is that each block within the proposed optimization problem exhibits a \gls{qp} structure with a fixed inverse component applicable to all agents. This allows us to precompute and cache the inverse part offline, effectively decomposing the large optimization problem into several parallel single-variable optimization subproblems.

\noindent The proposed optimizer offers several distinct advantages over the \gls{sota} at the time of developing this paper:

\begin{enumerate}
    \item \textbf{Ease of Implementation and \gls{gpu} Acceleration:} The proposed optimizer's numerical computation primarily involves element-wise operations on matrix-matrix products. These operations can be efficiently accelerated on \gls{gpu}s using libraries such as CUPY \cite{nishino2017cupy} and JAX \cite{bradbury2018jax}. I also provide an open-source implementation, which can compute trajectories for 32 agents in just 0.7 seconds on a desktop computer equipped with an RTX-2080 \gls{gpu}.
    \item \textbf{\gls{sota} Performance:} The proposed optimizer significantly outperforms the computation time of joint trajectory optimization methods \cite{augugliaro2012generation} while achieving trajectories of comparable quality. It also surpasses the current \gls{sota} sequential approaches \cite{park2020efficient} by producing shorter trajectories in benchmark tests. Moreover, it exhibits improved computation times on several benchmarks despite conducting a more rigorous joint search over the agents' trajectory space.
    \item \textbf{Suitability for Edge Devices:} The proposed optimizer can efficiently compute trajectories for 16 agents in approximately 2 seconds on an Nvidia Jetson-TX2. This performance is nearly two orders of magnitude faster than the computation time of \cite{augugliaro2012generation} on standard desktop computers. As a result, my work enhances the onboard decision-making capabilities of agents like quadrotors, which may have limited computational resources. To my knowledge, there are no existing works that achieve similar performance on edge devices at the time of working on this paper.

\end{enumerate}

\section{Main Results}
Considering the same motivation as Section \ref{main_result_1}, I reformulate collision avoidance constraint, $\mathbf{f}_{o} =0$ as

\vspace{-0.2cm}
\small
\begin{align}
     \mathbf{f}_{o} = \left \{ \begin{array}{lcr}
\colorbox{my_green}{$x_{i}(t) -x_{j}(t)-a d_{ij}(t)\sin\beta_{ij}(t)\cos\alpha_{ij}(t)$}\\
\colorbox{my_pink}{$y_{i}(t) -y_{j}(t)-a d_{ij}(t)\sin\beta_{ij}(t)\sin\alpha_{ij}(t)$}\\ 
\colorbox{my_blue}{$z_{i}(t) -z_{j}(t)-b d_{ij}\cos\beta_{ij}(t)$} \\
\end{array} \right \},d_{ij}\geq 1
\label{sphere_proposed}
\end{align}
\normalsize

\noindent where $d_{ij}(t), \alpha_{ij}(t)$ and $\beta_{ij}(t)$ are optimization variables which are needed to be computed. Similar to previous works, to exploit the hidden convex structure within \eqref{cost_multiagent}-\eqref{eq_multiagent_final} and \eqref{sphere_proposed}, I utilize the augmented Lagrangian method for the proposed optimization problem.

\small
\begin{align}
&\min_{\scalebox{0.7}{$\begin{matrix}x_{i}(t), y_{i}(t), z_{i}(t), \\
\alpha_{ij}(t), \beta_{ij}(t), d_{ij}(t)\end{matrix}$}}  \hspace{-0.1cm}\sum_{i=1}^{N_{a}}\sum_{t=t_1}^{n_{p}}\hspace{-0.1cm}\Big( \ddot{x}^{2}_{i}(t)\hspace{-0.1cm}+\hspace{-0.05cm}\ddot{y}^{2}_{i}(t)\hspace{-0.1cm}+\hspace{-0.1cm}\ddot{z}^{2}_{i}(t)
\hspace{-0.1cm}+\hspace{-0.3cm}\sum_{j=1, j \neq i}^{j=N_{a}}\hspace{-0.1cm}\big (\frac{\rho_{o}}{2}(\hspace{-0.05cm}\frac{\lambda_{x_{ij}}(\hspace{-0.02cm}t\hspace{-0.02cm})}{\rho_{o}}\hspace{-0.01cm}\colorbox{my_green}{\hspace{-0.1cm}$+x_{i}(t)\hspace{-0.1cm}-x_{j}(t)$}\nonumber \\&~~~~~~~~~\colorbox{my_green}{\hspace{-0.2cm}$-ad_{ij}(\hspace{-0.02cm}t\hspace{-0.02cm})\sin\beta_{ij}(\hspace{-0.02cm}t\hspace{-0.02cm})\cos\alpha_{ij}(\hspace{-0.02cm}t\hspace{-0.02cm})$})^2  \hspace{-0.1cm}+\hspace{-0.1cm}\frac{\rho_{o}}{2}(\colorbox{my_blue}{$z_{i}(\hspace{-0.02cm}t\hspace{-0.02cm})\hspace{-0.1cm}-\hspace{-0.1cm}z_{j}(\hspace{-0.02cm}t\hspace{-0.02cm})\hspace{-0.1cm}-\hspace{-0.1cm}bd_{ij}(\hspace{-0.02cm}t\hspace{-0.02cm})\cos\beta_{ij}(t)\hspace{-0.1cm}+\hspace{-0.1cm}\frac{\lambda_{z_{ij}}(\hspace{-0.02cm}t\hspace{-0.02cm})}{\rho_{o}})$}^2\big)\nonumber \\&~~~~~~~~
    +\frac{\rho_{o}}{2}(\colorbox{my_pink}{$y_{i}(t) -y_{j}(t)-ad_{ij}(t)\sin\beta_{ij}(t)\sin\alpha_{ij}(t)$}+\frac{\lambda_{y_{ij}}(t)}{\rho_{o}})^2 \Big)
    \label{aug_lagrange_multi}
\end{align}
\normalsize

\noindent where $\lambda_{x_{ij}}(t), \lambda_{y_{ij}}(t), \lambda_{z_{ij}}(t)$ are time-dependent Lagrange multipliers and  $\rho_{o}$ is a scalar constant. I can utilize the \gls{am} method \cite{jain2017non} to solve \eqref{aug_lagrange_multi} with respect to optimization variables. Algorithm \ref{alg_5} provides a summary of the steps involved in solving \eqref{aug_lagrange_multi} subject to \eqref{eq_multiagent_start} and \eqref{eq_multiagent_final}.

\vspace{0.3cm}
\noindent \textbf{Analysis and Description of Algorithm \ref{alg_5}:} Now, I analyze each step of the proposed optimizer.

\noindent \textbf{Line 2}: 
The optimization problem \eqref{aug_lagrange_multi} can be converted into \gls{qp} forms where matrices remain constant across iterations. To achieve this, I examine \eqref{aug_lagrange_multi} and restructure it by including only terms involving $x_{i}(t)$ as \eqref{com_xi}.  Subsequently, using \eqref{parameter}, I parameterize the optimization variable, $x_{i}(t)$ and $x_{j}(t)$ and rewrite each terms of \eqref{com_xi} as

\small
\begin{align}
\hspace{-0.2cm}\min_{\boldsymbol{\xi}_{x}} \frac{1}{2}\boldsymbol{\xi}_{x}^T\mathbf{Q}_{x} \boldsymbol{\xi}_x\hspace{-0.1cm}+\hspace{-0.1cm}\hspace{-0.1cm}\Big(\frac{1}{2}\rho_{o}\boldsymbol{\xi}_{x}^T\mathbf{A}_{f_o}^T\mathbf{A}_{f_o}\boldsymbol{\xi}_{x}\hspace{-0.1cm}-\hspace{-0.1cm}(\rho_{o}\mathbf{A}_{f_o}^T{^k}\mathbf{b}_{f_o}^x)^T\boldsymbol{\xi}_x\Big),~
\text{s.t.:}~~\mathbf{A}_{eq}\boldsymbol{\xi}_x\hspace{-0.1cm}=\hspace{-0.1cm}\mathbf{b}_{eq}^x \label{eq_step_x}
\end{align}
\normalsize

\noindent where $\boldsymbol{\xi}_{x}$ is the stack of $\boldsymbol{\xi}_{x_{i}}$ for all the agents and $\boldsymbol{\xi}_{x_{i}}$ is the coefficient associated with the basis functions. The initial two terms in \eqref{eq_step_x} are the parameterized and simplified representations of the corresponding terms in \eqref{com_xi}. In the following, I show how this simplification is done and define each matrix and vector used in \eqref{eq_step_x}

\vspace{-0.2cm}
\small
\begin{subequations}
\begin{align}
   \text{First term: }&\sum_{t}\sum_{i}\ddot{x}_i(t)^2 \Rightarrow \frac{1}{2}\boldsymbol{\xi}_x^T\mathbf{Q}_x\boldsymbol{\xi}_x,  \mathbf{Q}_x = \begin{bmatrix}
    \ddot{\mathbf{P}}^T\ddot{\mathbf{P}} & & \\
    & \ddots  & \\
    & & \ddot{\mathbf{P}}^T\ddot{\mathbf{P}}
    \end{bmatrix}\\\text{Second term:}&
    \sum_{t}\sum_{i}\sum_{j} \frac{\rho_{o}}{2}\Big(x_i(t)\hspace{-0.1cm}-\hspace{-0.1cm}x_j(t)\hspace{-0.1cm}-\hspace{-0.1cm}a{^k}d_{ij}(t)\sin{^k}\beta_{ij}(t)\cos{^k}\alpha_{ij}(t)\hspace{-0.1cm}+\hspace{-0.1cm}\frac{{^k}\lambda_{x_{ij}}(t)}{\rho}\Big)^2  \nonumber \\&
        \Rightarrow \frac{\rho}{2}\Vert \mathbf{A}_{f_o} \boldsymbol{\xi}_x-{^k}\mathbf{b}_{f_o}^x\Vert_2^2,  
          {^k}\mathbf{b}_{f_o}^x = a{^k}\mathbf{d} \sin{^k}\boldsymbol{\beta}\cos{^k}\boldsymbol{\alpha}-\frac{\boldsymbol{{^k}\lambda_{x_{ij}}}}{\rho}
\nonumber \\
&
\mathbf{A}_{f_c}\hspace{-0.1cm}=\hspace{-0.1cm}\begin{bmatrix}
    \mathbf{A}_{1}\hspace{-0.1cm}& & \\
    & \hspace{-0.1cm}\ddots \hspace{-0.1cm}& \\
    & &\hspace{-0.1cm}\mathbf{A}_{n_{a}}
  \end{bmatrix},
\mathbf{A}_{n_{a}}\hspace{-0.1cm}=\hspace{-0.1cm}\begin{bmatrix}
     \begin{pmatrix}
     \mathbf{P}\\
     \mathbf{P}\\
      \vdots\\
      \mathbf{P}
     \end{pmatrix}_{\hspace{-0.1cm}\times N_{a}-n_{a}}\hspace{-0.2cm}&\hspace{-1.2cm}\begin{bmatrix}
    -\mathbf{P} & & \\
    & \ddots & \\
    & & -\mathbf{P}
  \end{bmatrix}_{\hspace{-0.1cm}\times N_{a}-n_{a}}
    \end{bmatrix}
    \label{A_f} \\&
  \mathbf{A}_{eq} = \begin{bmatrix}
    \mathbf{A} & & \\
    & \ddots & \\
    & & \mathbf{A}
    \end{bmatrix}, \mathbf{A} = \begin{bmatrix}
    \mathbf{P}_{0}&
    \dot{\mathbf{P}}_{0}&
    \ddot{\mathbf{P}}_{0}&
    \mathbf{P}_{-1}&
    \dot{\mathbf{P}}_{-1}&
    \ddot{\mathbf{P}}_{-1}&
    \end{bmatrix}^{T}
\end{align}
\end{subequations}
\normalsize

\noindent Now, the problem \eqref{eq_step_x} can be reduced to a set of linear equations as 

\small
\begin{align}
    \overbrace{\begin{bmatrix}
(\mathbf{Q}_x+\rho\mathbf{A}_{f_o}^T\mathbf{A}_{f_o}) & \mathbf{A}_{eq}^T\\
    \mathbf{A}_{eq} & \mathbf{0}
    \end{bmatrix} }^{\widetilde{\mathbf{Q}}_x} \begin{bmatrix}
    \boldsymbol{\xi}_x\\
    \boldsymbol{\nu}
    \end{bmatrix} = \overbrace{\begin{bmatrix}
    \rho\mathbf{A}_{f_o}^T{^k}\mathbf{b}_{f_o}^x\\
    \mathbf{b}_{eq}^x
    \end{bmatrix}}^{\widetilde{\mathbf{q}}_x}
\label{kkt_step_x}
\end{align}
\normalsize

\noindent where $\boldsymbol{\nu}$ is the dual optimization variable. 

\noindent \textbf{Line 3:} I adopt the geometrical intuition from the main results of the first provided paper, Section \ref{main_result_1} to solve \eqref{com_alpha_i}. 

\noindent \textbf{Line 4:} Similar to previous step, I obtain \eqref{com_beta_i}.

\noindent \textbf{Line 5-6:} As the optimization variables $d_{ij}(t)$ are independent across different time instances and agents, the optimization problem \eqref{com_d_i} can be decomposed into $n_{p}\times{N_{a}\choose{2}}$ \gls{qp} problems, each with a single variable. Consequently, I can readily express the problem symbolically. It is important to note that the bounds on $d_{ij}(t)$ are maintained by clipping the values to the interval $[0,1]$ at each iteration.

\noindent \textbf{Line 7:} I update the Lagrange multipliers as

\vspace{-0.3cm}
\small
\begin{subequations}
    \begin{align}
        \hspace{-0.15cm}{^{k+1}\hspace{-0.05cm}}\lambda_{x_{ij}}\hspace{-0.05cm}(\hspace{-0.032cm}t\hspace{-0.032cm})\hspace{-0.1cm}&=\hspace{-0.15cm}^{k\hspace{-0.05cm}}\lambda_{x_{ij}}\hspace{-0.05cm}\hspace{-0.05cm}(\hspace{-0.033cm}t\hspace{-0.033cm})\hspace{-0.1cm}+\hspace{-0.1cm}\rho_{o}(\hspace{-0.033cm}{^{k+1}\hspace{-0.05cm}}x_i(\hspace{-0.033cm}t\hspace{-0.033cm})\hspace{-0.1cm}-\hspace{-0.17cm}{^{k+1}\hspace{-0.05cm}}x_j(\hspace{-0.033cm}t\hspace{-0.033cm})\hspace{-0.1cm}-\hspace{-0.1cm}a{^{k+1}\hspace{-0.05cm}}d_{ij}(\hspace{-0.033cm}t\hspace{-0.033cm})\sin{^{k+1}\hspace{-0.05cm}}\beta_{ij}(\hspace{-0.033cm}t\hspace{-0.033cm})\cos{^{k+1}\hspace{-0.05cm}}\alpha_{ij}(\hspace{-0.033cm}t\hspace{-0.033cm})\hspace{-0.033cm}) \label{update_lamda_xi} \\
 \hspace{-0.15cm}{^{k+1}\hspace{-0.05cm}}\lambda_{y_{ij}}\hspace{-0.05cm}(\hspace{-0.032cm}t\hspace{-0.033cm})\hspace{-0.1cm}&=\hspace{-0.15cm} {^{k}\hspace{-0.05cm}}\lambda_{y_{ij}}\hspace{-0.05cm}(\hspace{-0.033cm}t\hspace{-0.033cm})\hspace{-0.1cm}+\hspace{-0.1cm}\rho_{o}({^{k+1}\hspace{-0.05cm}}y_i(\hspace{-0.033cm}t\hspace{-0.033cm})\hspace{-0.1cm}-\hspace{-0.17cm}{^{k+1}\hspace{-0.05cm}}y_j(\hspace{-0.033cm}t\hspace{-0.033cm})\hspace{-0.1cm}-\hspace{-0.1cm}a{^{k+1}\hspace{-0.05cm}}d_{ij}(\hspace{-0.033cm}t\hspace{-0.033cm})\sin{^{k+1}\hspace{-0.05cm}}\beta_{ij}(\hspace{-0.033cm}t\hspace{-0.033cm})\sin{^{k+1}\hspace{-0.05cm}}\alpha_{ij}(\hspace{-0.033cm}t\hspace{-0.033cm})\hspace{-0.033cm}) \label{update_lamda_yi} \\
 \hspace{-0.1cm}{^{k+1}}\lambda_{z_{ij}}(\hspace{-0.02cm}t\hspace{-0.02cm}) &=  {^{k}}\lambda_{z_{ij}}(\hspace{-0.02cm}t\hspace{-0.02cm})+\rho_{o}({^{k+1}}z_i(\hspace{-0.02cm}t\hspace{-0.02cm}) -{^{k+1}}z_j(\hspace{-0.02cm}t\hspace{-0.02cm})-b{^{k+1}}d_{ij}(\hspace{-0.02cm}t\hspace{-0.02cm})\cos{^{k+1}\hspace{-0.05cm}}\beta_{ij}(\hspace{-0.02cm}t\hspace{-0.02cm})).\label{update_lamda_zi}
    \end{align}
\end{subequations}
\normalsize

\begin{algorithm}[!h]
\DontPrintSemicolon
\SetAlgoLined
\SetNoFillComment
\scriptsize
\caption{\scriptsize Alternating Minimization for Solving Multi-agent Trajectory Optimization \eqref{cost_multiagent}-\eqref{quadratic_multiagent}}\label{alg_5}
\KwInitialization{Initiate $^{k}d_{ij}(t)$, $^{k}\alpha_{ij}(t)$, $^{k}\beta_{ij}(t)$  }
\While{$k \leq maxiter$}{

\hspace{-0.15cm}Compute $^{k+1}x_{i}(t)$,$^{k+1}y_{i}(t)$ and $^{k+1}z_{i}(t)$ 
\vspace{-0.4cm}
\begin{subequations}
    \begin{align}
^{k+1}x_{i}(t)&=\arg \min_{x_{i}(t)}\sum_{t=0}^{t=n_p}\sum_{i=1}^{i=N_{a}}\ddot{x}_{i}^{2}(t) \nonumber \\
&~~+\hspace{-0.1cm}\sum_{t=0}^{t=n_p}\sum_{i=1}^{i=N_{a}}\sum_{j=1,j\neq i}^{j=N_{a}}\hspace{-0.1cm}\frac{\rho_{o}}{2}(\colorbox{my_green}{$\hspace{-0.1cm}x_{i}(\hspace{-0.03cm}t\hspace{-0.03cm})\hspace{-0.1cm}-\hspace{-0.1cm}x_{j}(\hspace{-0.03cm}t\hspace{-0.03cm})\hspace{-0.1cm}-\hspace{-0.1cm}a\hspace{0.03cm}^{k}d_{ij}(\hspace{-0.03cm}t\hspace{-0.03cm})\hspace{0.05cm}\cos{^{k}\alpha_{ij}}(\hspace{-0.02cm}t\hspace{-0.03cm}) \hspace{0.03cm}\sin{^{k}\beta_{ij}}(\hspace{-0.03cm}t\hspace{-0.03cm})$\hspace{-0.1cm}}\hspace{-0.1cm}+\hspace{-0.1cm}\frac{^{k}\lambda_{x_{ij}}}{\rho_{o}})^{2} \label{com_xi}\\
   ^{k+1}y_{i}(t)
  &=\arg \min_{y_{i}(t)}\sum_{t=0}^{t=n_p}\sum_{i=1}^{i=N_{a}}\ddot{y}_{i}^{2}(t)\nonumber \\
&~~+\hspace{-0.1cm}\sum_{t=0}^{t=n_p}\sum_{i=1}^{i=N_{a}}\sum_{j=1,j\neq i}^{j=N_{a}}\hspace{-0.1cm}\frac{\rho_{o}}{2}(\colorbox{my_pink}{$\hspace{-0.1cm}y_{i}(\hspace{-0.03cm}t\hspace{-0.03cm})\hspace{-0.07cm}-\hspace{-0.07cm}y_{j}(\hspace{-0.03cm}t\hspace{-0.03cm})
\hspace{-0.07cm}-\hspace{-0.07cm}a\hspace{0.03cm}^{k}d_{ij}(\hspace{-0.03cm}t\hspace{-0.03cm})\hspace{0.03cm}\sin{^{k}\alpha_{ij}}(\hspace{-0.03cm}t\hspace{-0.03cm})\hspace{0.03cm}\sin{^{k}\beta_{ij}}(\hspace{-0.03cm}t\hspace{-0.03cm})\hspace{-0.1cm}$}\hspace{-0.1cm}+\hspace{-0.1cm}\frac{^{k}\lambda_{y_{ij}}}{\rho_{o}})^2 \label{com_yi}\\
^{k+1}z_{i}(t)&=\arg \min_{z_{i}(t)}\sum_{t=0}^{t=n_p}\sum_{i=1}^{i=N_{a}}\ddot{z}_{i}^{2}(t)\nonumber \\
&~~+\hspace{-0.2cm}\sum_{t=0}^{t=n_p}\sum_{i=1}^{i=N_{a}}\sum_{j=1,j\neq i}^{j=N_{a}}\frac{\rho_{o}}{2}(\colorbox{my_blue}{$z_{i}(t)-z_{j}(t) -b\hspace{0.05cm}^{k}d_{ij}\hspace{0.05cm}\cos{^{k}\beta_{ij}}(t)$}+\frac{^{k}\lambda_{z_{ij}}}{\rho_{o}} )^{2} \label{com_zi}
\end{align}
\end{subequations}
\vspace{-0.45cm}

\hspace{-0.15cm}Compute $\hspace{0.05cm}^{k+1}\alpha_{ij}(t)$
\vspace{-0.3cm}
\begin{align}
    \hspace{0.05cm}^{k+1}\alpha_{ij}(t) =  \arctan 2\Big((^{k+1}y_{i}(t)-^{k+1}y_{j}(t)),(^{k+1}x_{i}(t)-^{k+1}x_{j}(t))\Big) \label{com_alpha_i}
\end{align}
\vspace{-0.4cm}

\hspace{-0.15cm}Compute $\hspace{0.05cm}^{k+1}\beta_{ij}(t)$
\vspace{-0.3cm}
\begin{align}
    \hspace{0.05cm}^{k+1}\beta_{ij}(t) = \arctan 2\Big((\frac{^{k+1}x_{i}(t)-^{k+1}x_{j}(t)}{a \cos{^{k+1}\alpha_{ij}(t)}}),(\frac{^{k+1}z_{i}(t)-^{k+1}z_{j}(t)}{b })\Big)
    \label{com_beta_i}
\end{align}
\vspace{-0.5cm}

\hspace{-0.15cm}Compute $\hspace{0.05cm}^{k+1}d_{ij}(t)$ through 
\begin{align}
    \hspace{-0.5cm}^{k+1}d_{ij}(t)\hspace{-0.05cm}&=\hspace{-0.05cm}\arg \min_{d_{ij}(t)}\sum_{t}\sum_{i}\sum_{j}\hspace{-0.07cm}\frac{\rho_{o}}{2}\hspace{-0.04cm}\Big(\hspace{-0.07cm}(\colorbox{my_blue}{\hspace{-0.1cm}$^{k+1}z_{i}(t)-^{k+1}z_{j}(t) -b\hspace{0.05cm}^{k+1}d_{ij}\hspace{0.05cm}\cos{^{k+1}\beta_{ij}}(t)$\hspace{-0.2cm}}\hspace{-0.05cm}+\frac{^{k}\lambda_{z_{ij}}}{\rho_{o}} )^{2}
    \nonumber \\&~~+(\colorbox{my_green}{$\hspace{-0.1cm}^{k+1}x_{i}(\hspace{-0.03cm}t\hspace{-0.03cm})\hspace{-0.1cm}-\hspace{-0.1cm}^{k+1}x_{j}(\hspace{-0.03cm}t\hspace{-0.03cm})\hspace{-0.1cm}-\hspace{-0.1cm}a\hspace{0.03cm}^{k+1}d_{ij}(\hspace{-0.03cm}t\hspace{-0.03cm})\cos{^{k+1}\alpha_{ij}}(\hspace{-0.02cm}t\hspace{-0.03cm}) \sin{^{k+1}\beta_{ij}}(\hspace{-0.03cm}t\hspace{-0.03cm})$\hspace{-0.1cm}}\hspace{-0.1cm}+\hspace{-0.1cm}\frac{^{k}\lambda_{x_{ij}}}{\rho_{o}})^{2} \nonumber \\&~~+(\colorbox{my_pink}{$\hspace{-0.1cm}^{k+1}y_{i}(\hspace{-0.03cm}t\hspace{-0.03cm})\hspace{-0.07cm}-\hspace{-0.07cm}^{k+1}y_{j}(\hspace{-0.03cm}t\hspace{-0.03cm})
\hspace{-0.07cm}-\hspace{-0.07cm}a\hspace{0.03cm}^{k+1}d_{ij}(\hspace{-0.03cm}t\hspace{-0.03cm})\hspace{0.03cm}\sin{^{k+1}\alpha_{ij}}(\hspace{-0.03cm}t\hspace{-0.03cm})\hspace{0.03cm}\sin{^{k+1}\beta_{ij}}(\hspace{-0.03cm}t\hspace{-0.03cm})\hspace{-0.1cm}$}\hspace{-0.1cm}+\frac{^{k}\lambda_{y_{ij}}}{\rho_{o}})^2
    \label{com_d_i}
\end{align}

\hspace{-0.15cm}Update $\lambda_{x_{ij}}(t), \lambda_{y_{ij}}(t), \lambda_{z_{ij}}$ at $k+1$
}
\normalsize

\end{algorithm}




\section{Validation and Benchmarking}

\noindent \textbf{Implementation Details:}   Algorithm 4 is implemented using
Python with the JAX \cite{bradbury2018jax} and CUPY \cite{nishino2017cupy} library as the GPU-accelerated algebra backend. More specifically, the CUPY is used for trajectory optimization of up to 32 agents. All the benchmarks were conducted on a desktop computer with RTX 2080 featuring an Intel Core i7 processor and 32 GB RAM. 

Also, in the implementations, agents were modeled as spheres for simplicity, and a circumscribing sphere was constructed for static obstacles to integrate them into the proposed optimizer. In addition, to enhance the convergence of Algorithm \ref{alg_5}, I precomputed the inverse of $\widetilde{\mathbf{Q}}_x$ in \eqref{kkt_step_x} for ten incrementally higher values of $\rho_{o}$. These values were used in the latter iterations of Algorithm \ref{alg_5} and are inspired from \cite{xu2017admm}.
For a fair comparison with \cite{augugliaro2012generation} and \cite{park2020efficient}, I utilized their open-source implementations and datasets. In comparison with \cite{augugliaro2012generation}, I omitted hard bounds on position, velocities, and accelerations, resulting in a reduction in the number of inequality constraints. Similarly, I adjusted the "downwash" parameter in \cite{park2020efficient} to 1 to align with the proposed optimizer's implementation. As trajectories obtained from the proposed optimizer, \cite{augugliaro2012generation}, and \cite{park2020efficient} operate at different time scales, I employed the second-order finite difference of the position as a proxy for comparing accelerations across these three methods. In addition, I compared the proposed approach with different baselines in two sets of benchmarks, including:

\begin{itemize}
\item \textbf{Square Benchmark:} The agents are positioned along the perimeter of a square and tasked with reaching their antipodal positions.
\item \textbf{Random  Benchmark with static obstacles:} In this scenario, the starting and goal positions of the agents are randomly sampled. Additionally, I introduce a modified version of this benchmark where static obstacles are randomly distributed within the workspace.
\end{itemize}

\noindent It should be mentioned that three metrics, including smoothness cost, Arc-length cost, and computation time, are used for all the comparisons. 

\subsection{Qualitative Results}
\begin{figure}[!h]
    \centering
    \includegraphics[scale=0.33]{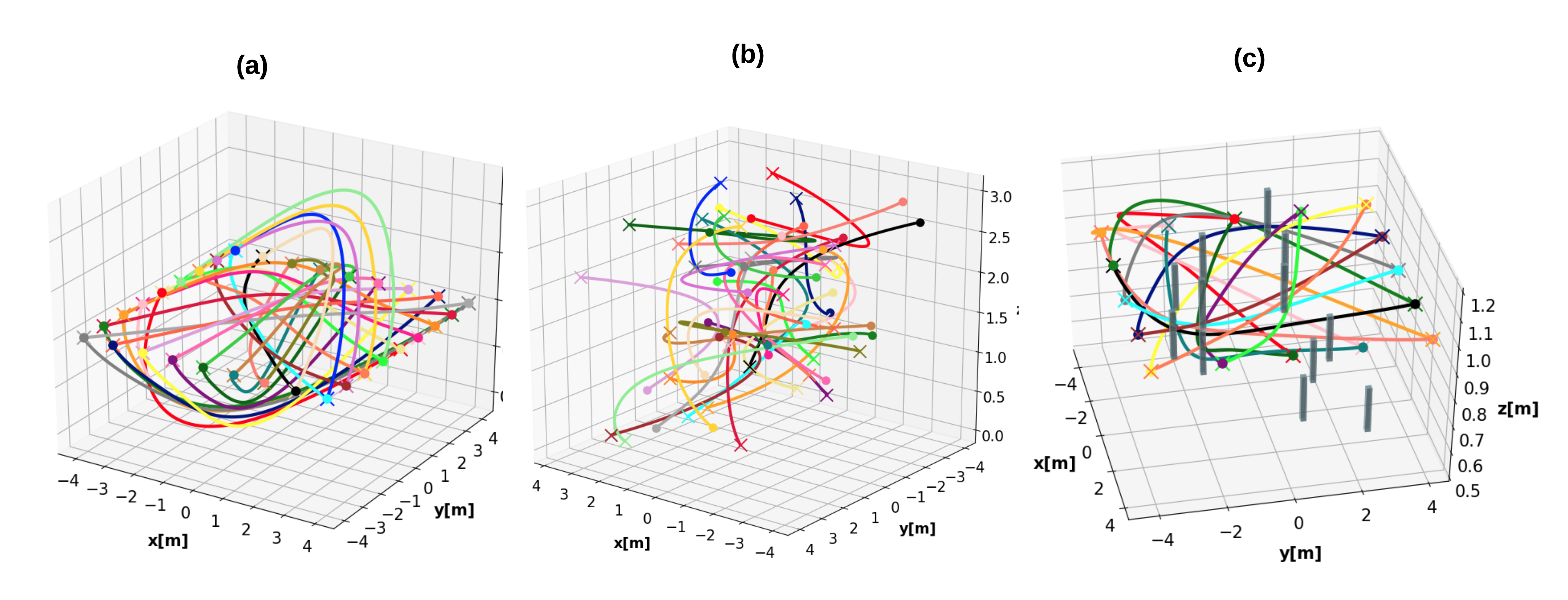}
    \caption[Qualitative trajectories for different benchmarks]{\small{Qualitative trajectories for square and random benchmarks without static obstacles and random benchmark with static obstacles.}}
    \label{multi-bench}
    \end{figure}

\begin{figure}[!h]
    \centering
    \includegraphics[scale=0.33]{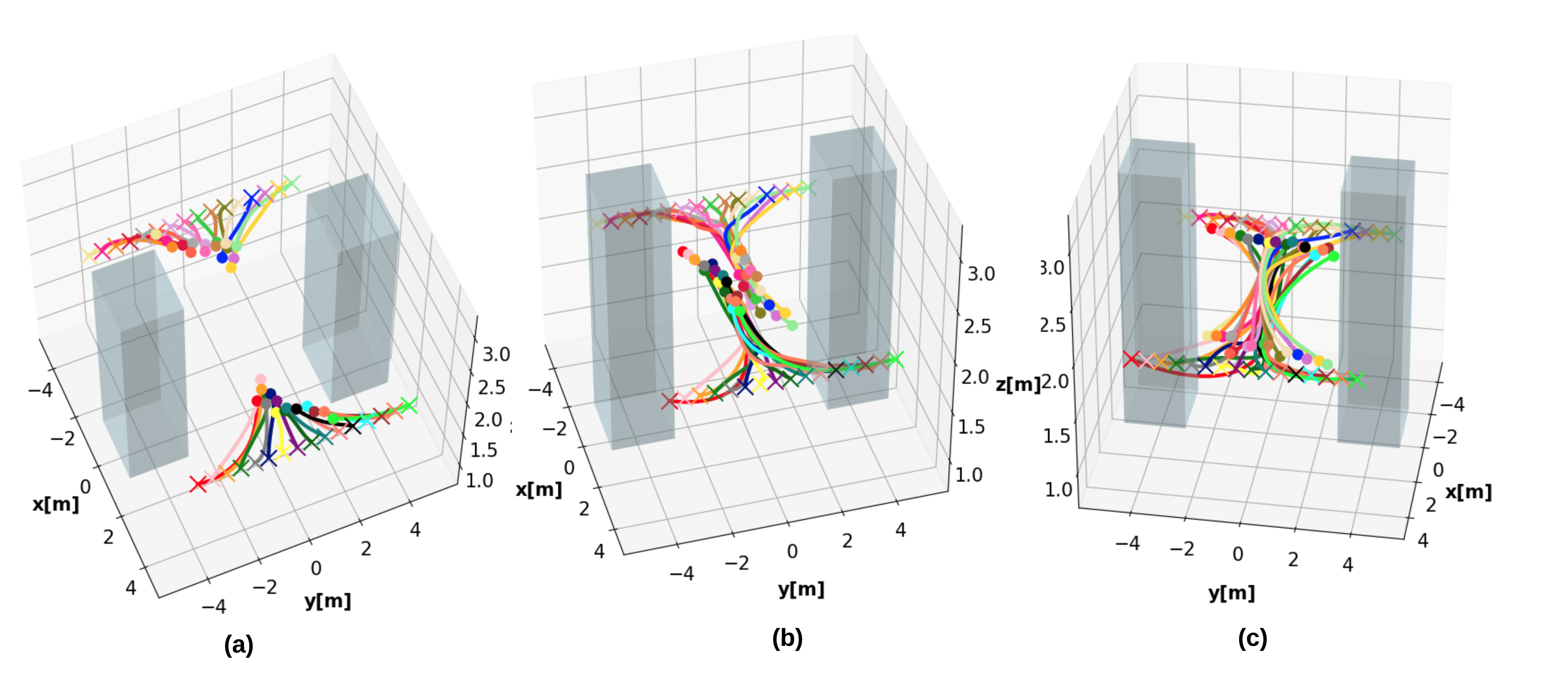}
    \caption[Qualitative trajectories for 32 agents in a narrow hallway]{\small{Collision avoidance snapshots of 32 agents exchanging positions in a
 narrow hallway is shown. The start and goal positions are marked with a "X" and a "o" respectively.}}
    \label{multi-snap}
    \end{figure}

\begin{figure}[!h]
    \centering
    \includegraphics[scale=0.37]{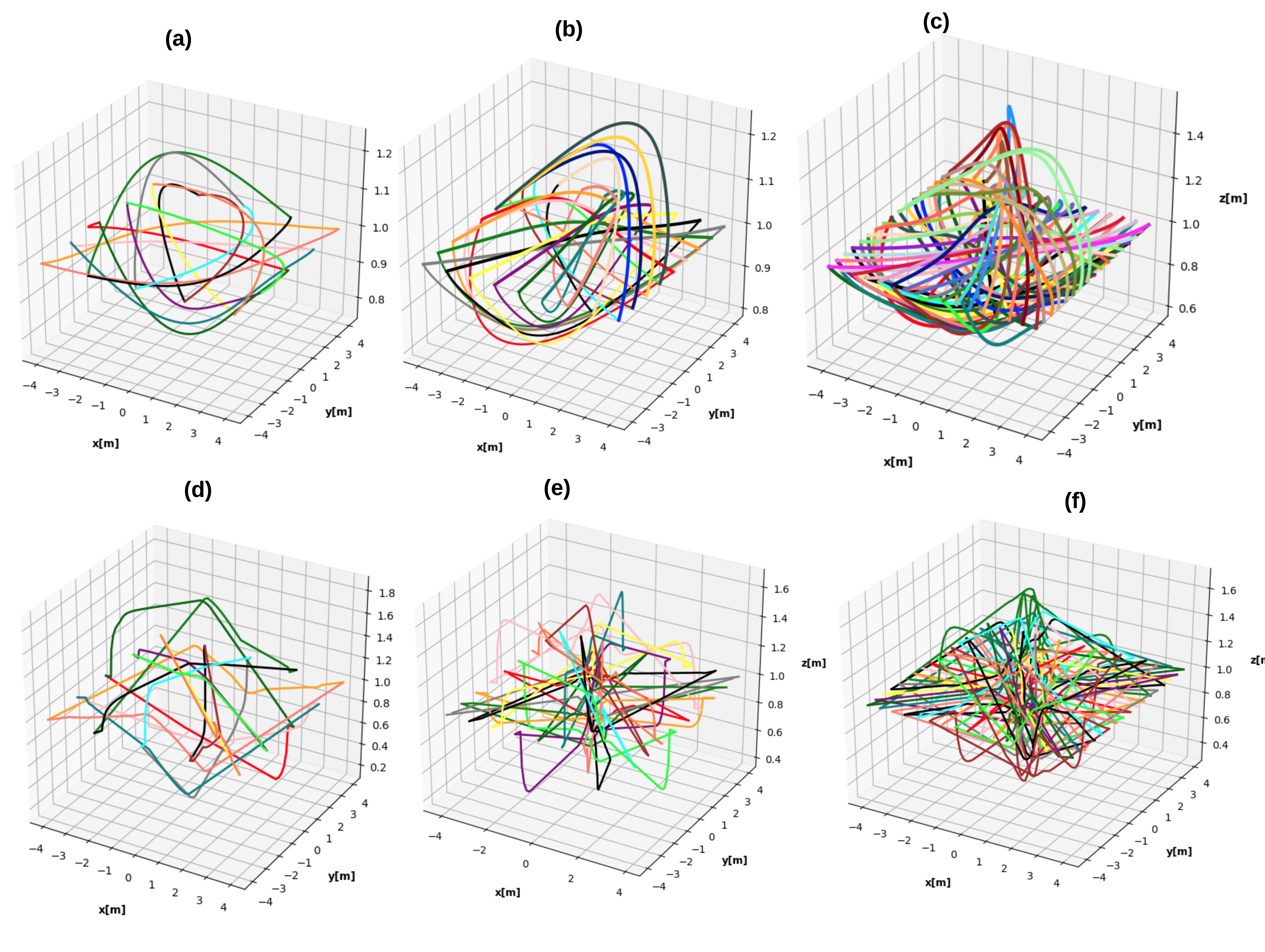}
    \caption[RVO vs. Algorithm \ref{alg_5} trajectories for different number of agents]{\small{Trajectories obtained for the proposed optimizer for the 16, 32, and 64 agents are shown in the first column, while trajectories obtained using RVO are shown in the second column}}
    \label{rvo_multi}
    \end{figure}

I showed the qualitative results for different benchmarks that are used in Figure \ref{multi-bench}. Additionally, I provided snapshots of 32 agents exchanging positions in a narrow hallway in Figure \ref{multi-snap} as an extra result. Also, I demonstrated trajectories obtained from the proposed algorithm and \gls{rvo} in Figure \ref{rvo_multi}. 

\subsection{Quantitative Results}

\begin{figure}[!h]
  \centering  
    \includegraphics[scale=0.3] {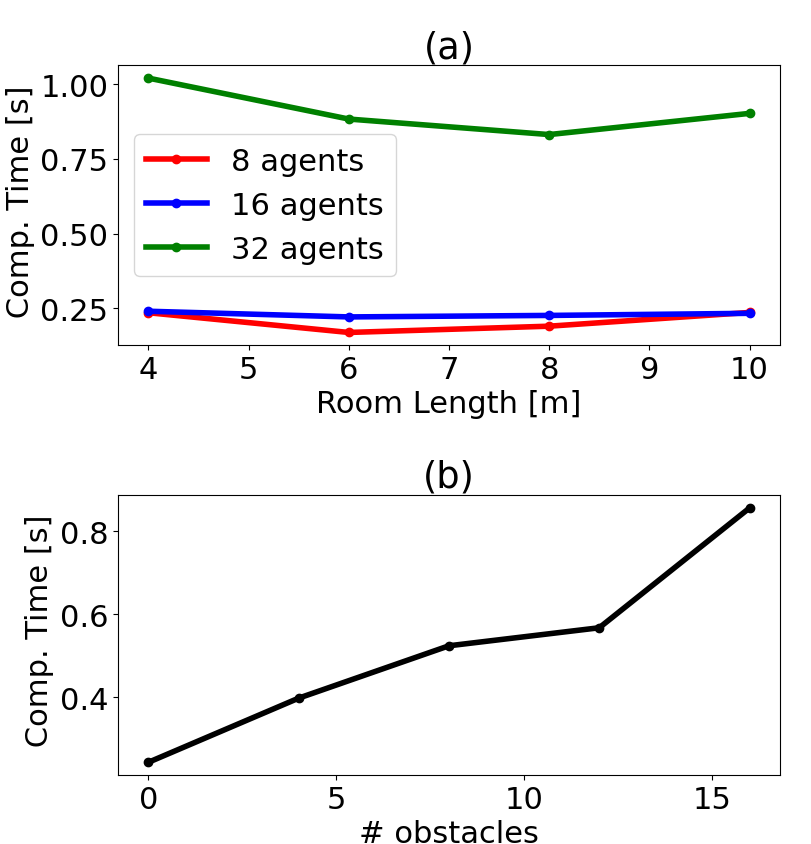}
\caption[Computation time]{ Figure (a) shows computation time for a varying number of agents
for benchmarks where I sample start and goal positions from a square with varying lengths. Figure (b) shows the linear scaling of computation time with obstacles for a given number of agents.  }
    \label{comp_time_room}
\end{figure}

\noindent\textbf{Computation time for varying numbers of agents:} Figure \ref{comp_time_room} (a) shows the average computation time of the proposed optimizer for varying numbers of agents (with radii of 0.4 meters), relative to the proximity of initial and final positions. Specifically, I randomly sampled initial and final positions within square rooms of varying dimensions to generate problem instances of differing complexity levels. As can be seen, even in the most challenging scenarios, the proposed optimizer successfully computed trajectories for 32 agents in approximately one second.

   \noindent\textbf{Computation time for varying numbers of static obstacles:} Figure \ref{comp_time_room} (b) depicts computation time for 16 agents under varying numbers of static obstacles. As can be seen, the proposed optimizer exhibits nearly linear scalability in computation time. This behavior arises because the inclusion of static obstacles primarily impacts the computational cost of obtaining $\mathbf{A}_{f_o}^T{^k}\mathbf{b}_{f_o}^x$ in equation (\ref{kkt_step_x}). Furthermore, both the matrix and vector dimensions increase linearly with the number of obstacles, and the resultant product is distributed across \gls{gpu}s.

\vspace{0.45cm}
\noindent \textbf{Comparison with \gls{rvo} \cite{van2008reciprocal}:}
The \gls{rvo} \cite{van2008reciprocal} is a widely used local, reactive planning approach in multi-agent navigation. While \gls{rvo} focuses on immediate collision avoidance in a single step, my optimizer tackles the more intricate task of global multi-agent trajectory optimization across multiple steps (100 in my implementation). Despite this fundamental difference, I benchmark the proposed optimizer against \gls{rvo} to establish a baseline for trajectory quality comparison.

It is worth noting that the computation time of \gls{rvo} is notably faster (three times faster in the benchmark with 16 agents) compared to the proposed optimizer. However, the trajectory comparison results, summarized in Table \ref{RVO-table}, reveal that \gls{rvo} produces slightly shorter trajectories than the proposed optimizer. This discrepancy arises because \gls{rvo} operates with single-integrator agents, allowing for abrupt velocity changes. In contrast, the proposed optimizer generates polynomial trajectories that prioritize higher-order differentiability.

Furthermore, while \gls{rvo}'s use of abrupt velocity changes enables shorter trajectories, it also incurs a substantially higher smoothness cost compared to the proposed optimizer. This trade-off highlights the differing optimization priorities between the two methods.

\setlength{\arrayrulewidth}{1pt}

\begin{table}[b]
\centering
\caption{\small{Comparison with \gls{rvo} \cite{van2008reciprocal}}}
\label{RVO-table}
\begin{tabular}{|c|c|c|c|}
\hline
\rowcolor[HTML]{BEFED4} 
Number of agents                                    & \multicolumn{1}{l|}{\cellcolor[HTML]{BEFED4}Benchmark} & \multicolumn{1}{l|}{\cellcolor[HTML]{BEFED4}Arc-length(m)} & \multicolumn{1}{l|}{\cellcolor[HTML]{BEFED4}Smoothness cost}     \\ \hline
\rowcolor[HTML]{9ADC9A} 
\cellcolor[HTML]{BEFED4}                            & \multicolumn{1}{l|}{\cellcolor[HTML]{9ADC9A}RVO}       & \multicolumn{1}{l|}{\cellcolor[HTML]{9ADC9A}9.491/1.12}    & \multicolumn{1}{l|}{\cellcolor[HTML]{9ADC9A}0.217/0.1}           \\ \cline{2-4} 
\rowcolor[HTML]{66BC8D} 
\multirow{-2}{*}{\cellcolor[HTML]{BEFED4}16 agents} & \multicolumn{1}{l|}{\cellcolor[HTML]{66BC8D}Our}       & \multicolumn{1}{l|}{\cellcolor[HTML]{66BC8D}9.877/1.33}    & \multicolumn{1}{l|}{\cellcolor[HTML]{66BC8D}\textbf{0.062/0.01}} \\ \hline
\rowcolor[HTML]{9ADC9A} 
\cellcolor[HTML]{BEFED4}                            & \multicolumn{1}{l|}{\cellcolor[HTML]{9ADC9A}RVO}       & \multicolumn{1}{l|}{\cellcolor[HTML]{9ADC9A}9.348/0.94}    & \multicolumn{1}{l|}{\cellcolor[HTML]{9ADC9A}0.26/0.11}           \\ \cline{2-4} 
\rowcolor[HTML]{66BC8D} 
\multirow{-2}{*}{\cellcolor[HTML]{BEFED4}32 agents} & \multicolumn{1}{l|}{\cellcolor[HTML]{66BC8D}Our}       & \multicolumn{1}{l|}{\cellcolor[HTML]{66BC8D}9.613/1.12}    & \multicolumn{1}{l|}{\cellcolor[HTML]{66BC8D}\textbf{0.06/0.01}}  \\ \hline
\rowcolor[HTML]{9ADC9A} 
\cellcolor[HTML]{BEFED4}                            & \multicolumn{1}{l|}{\cellcolor[HTML]{9ADC9A}RVO}       & \multicolumn{1}{l|}{\cellcolor[HTML]{9ADC9A}9.360.99}      & \multicolumn{1}{l|}{\cellcolor[HTML]{9ADC9A}0.228/0.09}          \\ \cline{2-4} 
\rowcolor[HTML]{66BC8D} 
\multirow{-2}{*}{\cellcolor[HTML]{BEFED4}64 agents} & Our                                                    & 9.439/1.07                                                 & \textbf{0.064/0.01}                                              \\ \cline{1-1} \hline
\end{tabular}
\end{table}
\vspace{0.3cm}

\begin{figure}[!h]
\centering
\includegraphics[scale=0.55] {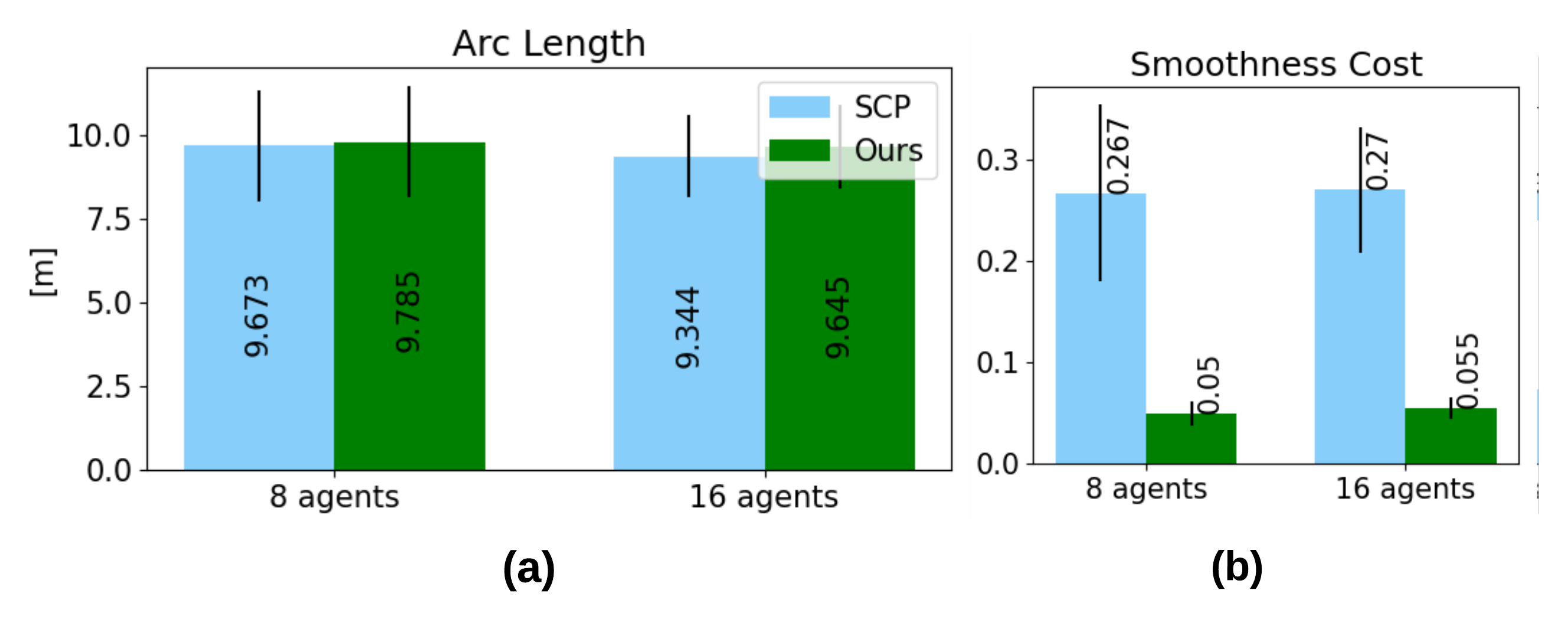}
\caption[Comparisons with state-of-the-art]{Comparisons with \gls{sota} \cite{augugliaro2012generation} in terms of arc-length and smoothness cost}
\label{scp_compare}
\end{figure}

\vspace{0.45cm}
\noindent\textbf{Comparison with \cite{augugliaro2012generation}:}
Figure \ref{scp_compare} provides a comparison between the proposed optimizer and \gls{scp} \cite{augugliaro2012generation} in terms of arc length and smoothness cost. Despite both optimizers converging to different trajectories, the arc-length statistics observed across all agents are remarkably similar. Moreover, the proposed optimizer demonstrates superior performance in terms of trajectory smoothness cost compared to \cite{augugliaro2012generation}.

Table \ref{scp_compare_time} also shows the computation time of the proposed method and \gls{scp} for different numbers of agents. As can be seen across various configurations with eight agents, the proposed optimizer 
computation time is 28 times faster than the \gls{scp} method. This gap became even bigger by increasing the number of agents. For 16 agents, the proposed optimizer was 613 times faster than the \gls{scp} method.

\setlength{\arrayrulewidth}{1pt}
\begin{table}[b]
\caption{\small{Computation time(s) comparison with 
 SCP}:}
\label{scp_compare_time}
\centering
\begin{tabular}{|l|l|l|}
\hline
\rowcolor[HTML]{C0C0C0} 
    & 8 agents       & 16 agents      \\ \hline
\rowcolor[HTML]{009901} 
\textbf{\begin{tabular}[c]{@{}l@{}}Our \\ Computation time\end{tabular}} & \textbf{0.242} & \textbf{0.262} \\ \hline
\rowcolor[HTML]{68CBD0} 
\begin{tabular}[c]{@{}l@{}}SCP\\ Computation time\end{tabular}           & 6.79           & 160.76         \\ \hline
\end{tabular}
\end{table}



\begin{figure}[h]
\centering
\includegraphics[scale=0.55] {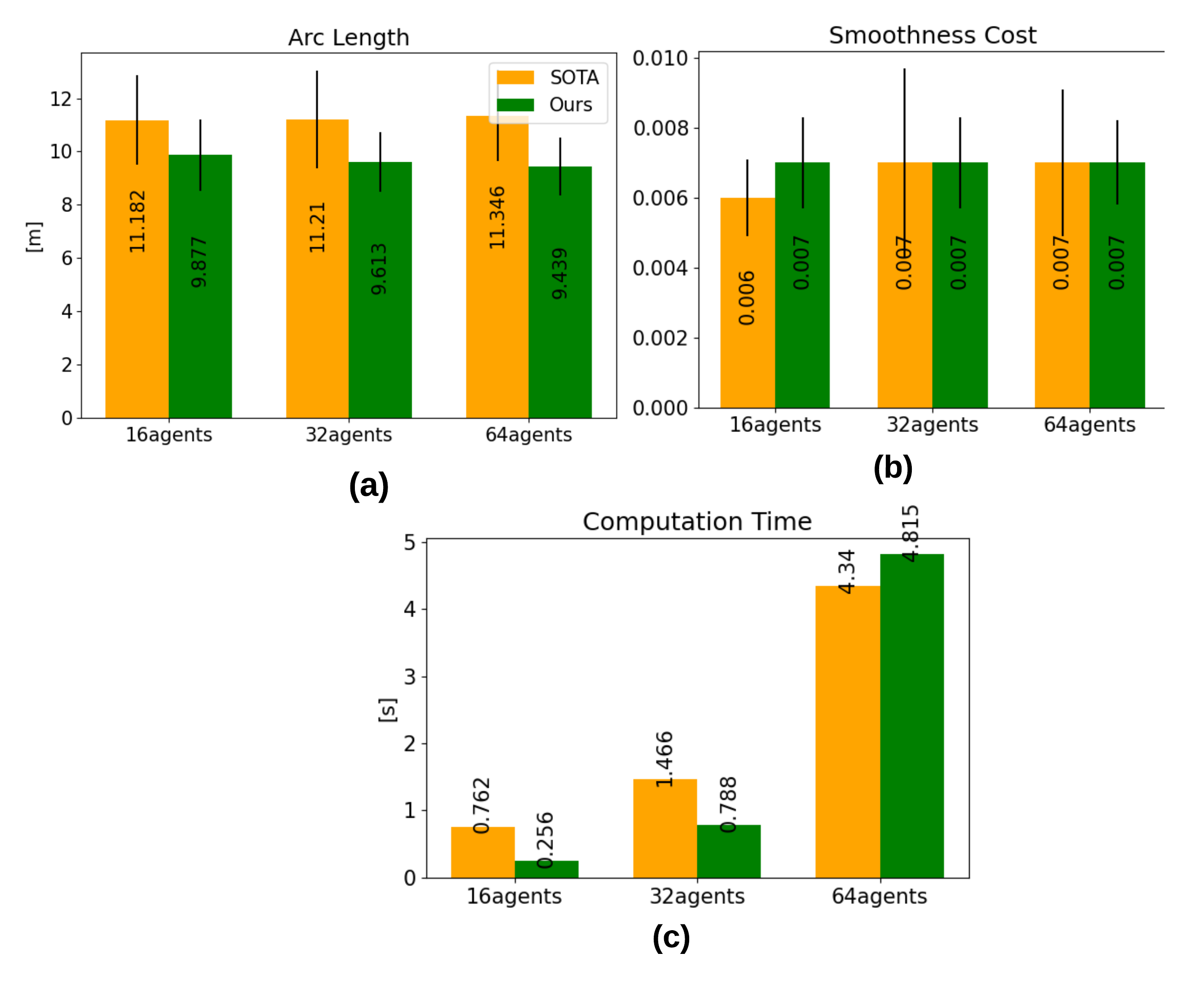}
\caption[Comparisons with state-of-the-art]{Comparisons with \gls{sota} \cite{park2020efficient} in terms of arc-length and smoothness cost and computation time}
\label{rbp_compare}
\end{figure}

\vspace{0.45cm}
\noindent \textbf{Comparison with \cite{park2020efficient}:}
Figure \ref{rbp_compare} shows the most important results of this study, where I compare the proposed optimizer with the current \gls{sota} method, \cite{park2020efficient}. The cited work adopts a sequential approach but with a batch
of agents. It also leverages the parallel \gls{qp}-solving ability of
CPLEX \cite{cplex} on multi-core \gls{cpu}s. My optimizer provides trajectories with comparable smoothness to \cite{park2020efficient} but with significantly shorter arc lengths. This trend can be attributed to the reduced feasible space accessible to a sequential approach. Additionally, the proposed optimizer surpasses \cite{park2020efficient} in terms of computation time for 16 and 32 agent benchmarks while achieving comparable performance on the 64 agent benchmark. It is crucial to contextualize these timings by acknowledging that the proposed optimizer conducts a much more exhaustive search than \cite{park2020efficient} across the joint trajectory space of the agents. The trends in computation time can be explained as follows: for a smaller number of agents, the computation time of \cite{park2020efficient} is primarily influenced by the trajectory initialization derived from sampling-based planners. Additionally, the overhead of \gls{cpu} parallelization is substantial for fewer agents, but this overhead is offset by the computational speed-up attained for a larger number of agents.

\vspace{0.45cm}
\noindent\textbf{Performance on Jetson TX2:}
Table \ref{table_jetson} displays the computation time for varying numbers of agents in the square benchmark conducted on the Nvidia Jetson TX2 platform. The starting and ending positions are randomly selected within an 8-meter square area. The results demonstrate that the proposed optimizer facilitates rapid on-board decision-making for up to 16 agents. Furthermore, even with 32 agents, the computation time remains sufficiently low to be applicable in practical scenarios.

\setlength{\arrayrulewidth}{1pt}

\begin{table}[h]
\centering
\caption{\small{Computation time on Nvidia-Jetson TX2}:}
\label{table_jetson}
\begin{tabular}{|l|l|} 
\hline
\rowcolor[HTML]{BEFED4} 
{Square Benchmark} & { Computation time} \\ \hline
\rowcolor[HTML]{99FEF7} 
8 agents, radius $= 0.1/0.6/1.2$        & 1.01/1.32/1.27                          \\ \hline
\rowcolor[HTML]{99FEF7} 
16 agents, radius $= 0.3/0.6$           & 2.10/2.34                               \\ \hline
\rowcolor[HTML]{99FEF7} 
32 agents, radius $= 0.25$              & 7.70                                    \\ \hline
\end{tabular}
\end{table}

\subsection{Algorithm Validation}
As mentioned in previous works, a key to validating the proposed trajectory optimization algorithm is to show that residuals are going to zero over iterations across various benchmarks. The residuals plot is generated by averaging residuals from 20 distinct problem instances. Typically, around 150 iterations were adequate to achieve residuals of approximately $0.01$. It is worth noting that these residuals represent the norm of a vector comprising tens of thousands (${n\choose 2} m$) of elements. For instance, with 64 robots, each of the collision avoidance vectors along the x, y and z axis, $\mathbf{f}_c^x, \mathbf{f}_c^y, \mathbf{f}_c^z$ contains more than $2\times10^6$ elements. Hence, it's crucial to monitor both the norm and the maximum magnitude across these vector elements to ascertain the convergence of Algorithm \ref{alg_5}. I observed that individual elements of the residual vector often have magnitude around $10^{-3}$ or lower even when the residual is around $0.01$. I can allow the optimizer to run for more iterations to obtain even lower residuals. My implementation uses an additional practical trick.  I inflate the radius of the agents by four times the typical residual I observe after 150 iterations of the proposed optimizer. In practice, this increased the agent's dimensions by around $4 cm$.

\begin{figure}[h]
\centering
\includegraphics[scale=0.3] {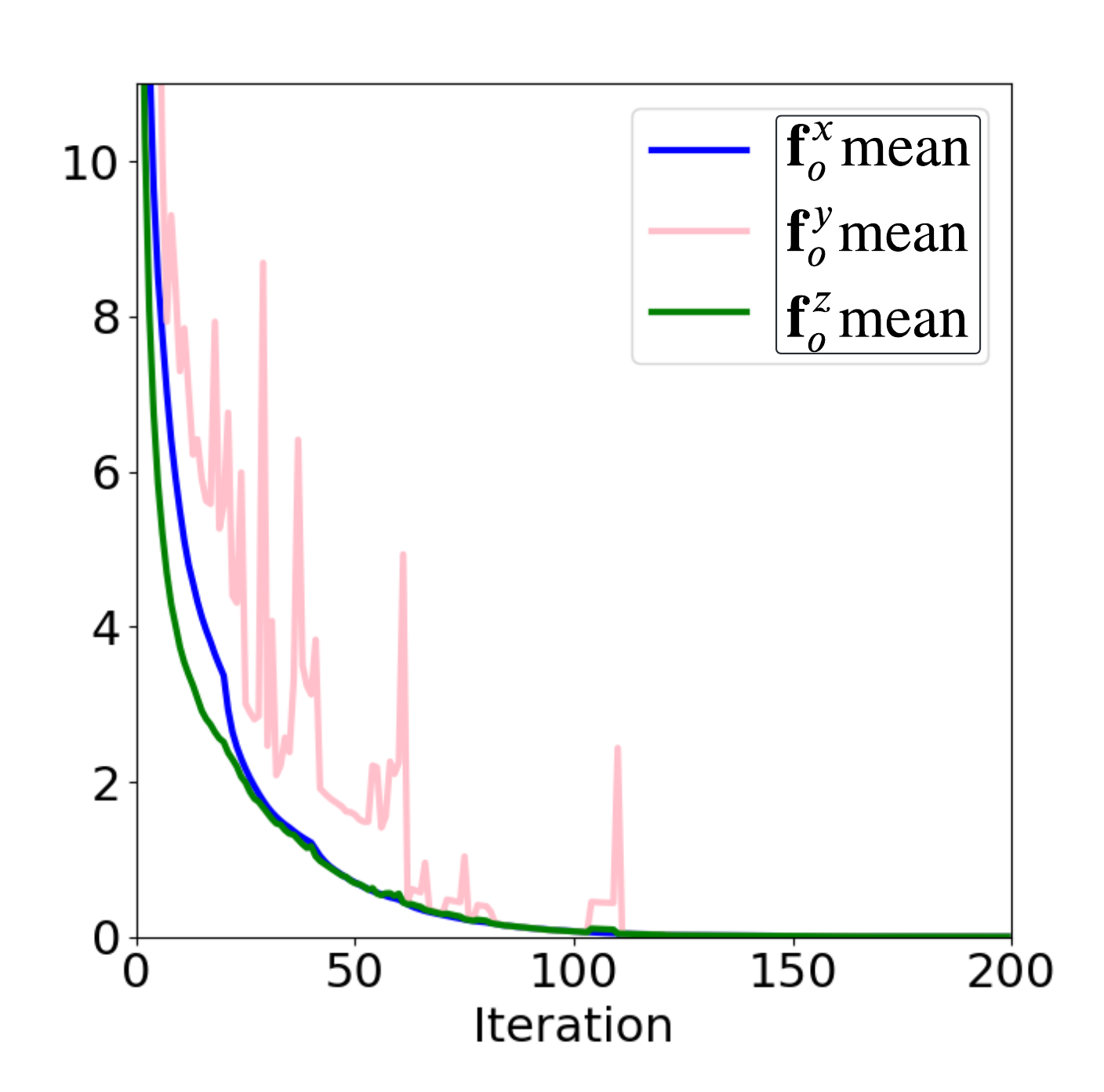}
\caption[Residuals]{The general trend in residual observed across several instances.  }
\label{ress}
\end{figure}

\subsection{Real-world Demonstration}
\begin{figure}[h]
\centering
\includegraphics[scale=0.6] {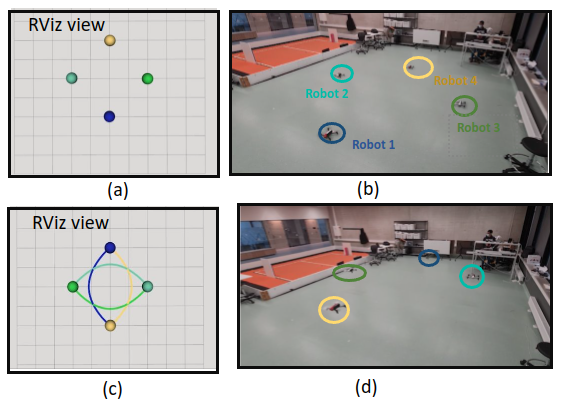}
\caption[Multi-agent real-world demonstration]{Multi-agent real-world demonstration using four robots. (a) and (c) shows the trajectory of the robots on RViz. For clarity, each robot position is marked with a specific color.  }
\label{real_multi}
\end{figure}

I have presented two instances from an actual experiment involving four Parrot Bebop 2 robots moving simultaneously. These instances are depicted in Figure \ref{real_multi}.  A video of this experiment can be found at Youtube  \footnote{https://youtu.be/hUuq9yiNoxQ}. For better understanding, the robots' trajectories are plotted over RViz.

\section{Connection to Other Chapters}

The proposed optimizer extends the single-agent trajectory optimization method introduced in paper I to a multi-agent setting. More specifically, I develop the multi-agent version of the collision avoidance model and reconfigure the underlying matrix algebra to facilitate distributed computations on \gls{gpu}. This approach explores a feature that has not been previously leveraged in previous work.


    \chapter{Conclusion}\label{conclusion}
In conclusion, this thesis has addressed the challenge of achieving reliable and efficient trajectory optimization in cluttered environments.

The first significant contribution of this research involves the development of a novel trajectory optimizer algorithm that is scalable with a number of constraints. To reach this scalable structure, Algorithm \ref{alg_1} leverages non-convex constraints, and explores hidden convex structures through multiple layers of reformulations in the optimization problem. The proposed optimizer shows superior performance in computation time compared to \gls{sota} method, \gls{ccp}. Also, our proposed algorithm is scalable with increasing the number of constraints.

Building upon the insights gained from the preceding work, the second contribution introduces an innovative GPU-accelerated batchable trajectory optimizer tailored for autonomous navigation in cluttered environments. This novel approach overcomes the challenge of initialization in trajectory optimization problems by incorporating hundreds of initial guesses. Our optimizer significantly enhances computational efficiency across various challenging scenarios by leveraging \gls{gpu} accelerations. We also benchmark our work in terms of success rate, acceleration, and tracking error with the \gls{sota} method, \gls{cem}.

Furthermore, as the third contribution, the thesis introduces a real-time projection-based trajectory optimization technique that guides initial guesses toward feasible regions at each iteration. Our proposed algorithm can handle any arbitrary cost function and removes the condition of having a convex cost. The proposed optimizer surpasses \gls{sota} methods such as \gls{mppi}, \gls{cem}, and \gls{dwa} in terms of computation time and success rate across various static and dynamic benchmarks set in highly cluttered environments.

Lastly, the thesis presents a trajectory optimization method designed specifically for multi-agent robots, constituting the final contribution. This method addresses the complexities inherent in coordinating multiple agents within cluttered environments, further extending the applicability and robustness of trajectory optimization techniques in real-world scenarios.

Overall, the culmination of these contributions represents a significant advancement in the field of trajectory optimization, offering promising avenues for enhancing the reliability, scalability,  and efficiency of robotic navigation in cluttered environments. 
 
\vspace{2cm}
\section{Limitations}
In this section, I introduce the limitations that each optimizer has

\noindent \textbf{Chapter \ref{papper_1}:} A key limitation of the optimizer proposed in this chapter is its dependency on the convexity of the cost function. This shows that it is best suited for holonomic robots. Additionally, this optimizer is susceptible to getting stuck in local minima.

\noindent \textbf{Chapter \ref{paper_2}:} In this chapter, similar to the previous one, the requirement for the cost function to be convex exists. This is a significant constraint that can limit the applicability of the optimizer. Additionally, the parameters for the distribution of samples are fixed. This means that even with hundreds of samples, there is a possibility that our optimizer may not be able to refine the samples effectively. As a result, it might fail to generate feasible trajectories. This highlights the need for careful parameter selection and potential improvements in the sampling process.

\noindent \textbf{Chapter \ref{paper_3}:} One of the limitations of our optimizer in this chapter is that although PRIEST can plan over a reasonably long horizon, it is still a local planner. Thus, it is expected to struggle in maze-like environments without some guidance from a global plan or some learning-based methods. 

Moreover, an additional limitation stems from our assumption of having accurate knowledge of obstacle positions. In scenarios where point cloud data is uncertain or imprecise, there is a risk of PRIEST becoming trapped in local minima and colliding with obstacles. 

\noindent \textbf{Chapter \ref{paper_4}:}
Finally, for multi-agent trajectory optimizer, we use only one initial guess for each agent. Thus, our optimizer is prone to get stuck in local minima. Also, similar to chapter \ref{papper_1}, the cost function in our optimizer has to be convex.

\section{Future Works  }
As part of my future work, I am interested in focusing on the design of trajectory optimization algorithms that can address the limitations identified in the algorithms presented in this thesis. Specifically, I aim to explore the following concepts:

\begin{itemize}
    \item \textbf{Integration of learning-based methods:} Investigate the integration of learning-based methods, such as reinforcement learning or imitation learning, into trajectory optimization algorithms. By leveraging the power of machine learning, we can enhance the adaptability and robustness of trajectory planning in dynamic and maze-like environments.
    \item \textbf{Combination of perception methods with trajectory optimization:} Explore the integration of perception methods, such as LiDAR and camera sensors, with trajectory optimization algorithms. By incorporating real-time environmental awareness into the trajectory planning process, we can improve the adaptability and robustness of motion planning algorithms. This integration may involve techniques such as feature extraction, object detection, and scene understanding to provide a richer context for trajectory optimization.
    \item \textbf{Batch multi-agent trajectory optimization:} Design batch trajectory optimization for multi-agent optimization problems. By leveraging parallelization and optimization techniques tailored for multi-agent scenarios, we can mitigate the risk of getting stuck in local minima and improve overall optimization performance.
\end{itemize}

In addition to the mentioned subjects, another avenue for improvement involves incorporating arbitrary dynamics and control constraints within the formulation. By considering a broader range of dynamics and control constraints, we can develop more versatile and adaptable trajectory optimization algorithms that are capable of addressing a wider array of real-world scenarios and challenges. This expansion of the optimization framework will further enhance the applicability and robustness of our algorithms in practical robotic systems.

    \printbibliography[heading=bibintoc]

    %
    \appendix
    \include{sections/appendixA}
    \include{sections/appendixB}

    \setcounter{secnumdepth}{-1}
    
\chapter{Acknowledgements}

I would like to extend my gratitude to my supervisor, Arun Kumar Singh, for his support and mentorship during my academic journey. His guidance and patience have been crucial in my development as a researcher in robotics. I am grateful for the programming skills and research insights he has imparted to me. His dedication to my growth as a professional has been truly commendable.

\vspace{0.3cm}
\noindent I would like to express my gratitude to my co-supervisor, Alvo Aabloo, for his support and encouragement throughout my academic journey. 

\vspace{0.5cm}
\noindent I would like to express my heartfelt gratitude to my host supervisor, Jan Swevers, for his invaluable support and guidance during my time at KU Leuven University. Learning from him and his wonderful group, MECO, has been a truly rewarding experience. I thoroughly enjoyed every moment I spent there and felt privileged to be part of his research group. His mentorship has contributed significantly to my personal and professional growth, and I am sincerely thankful for his support throughout my stay.

\vspace{0.5cm}
\noindent I would like to express my heartfelt appreciation to my husband and also my best friend, Iman. His unwavering support, love, and encouragement have been instrumental in my success throughout this journey. He stood by my side during both challenging and joyful moments. When I felt tired or disheartened, he believed in me and encouraged me to stay strong and continue moving forward. 

\vspace{0.5cm}
\noindent I am profoundly grateful to Mozhgan Pourmoradnasseri and Andreas Müller for serving as the reviewer and the opponent of this thesis. 

\vspace{0.5cm}
\noindent I would like to extend my appreciation to my parents, Masoumeh and Torabali, and also my brothers, Alireza and Mohammadreza, whose unconditional love and emotional support served as a constant motivation throughout this endeavor.  

\vspace{0.5cm}
\noindent I would like to acknowledge dear friends Houman, Zahra, Rafieh, Shahla, Mahtab, Mehrnoosh, Atiyeh, Maryam, Paria, Atefeh, Sepideh, Yasaman, Faezeh, Ava, Javad, Karim, Kaveh, Elyad, Ebi, Bahman, Nima, Yashar, Siim, and Ali. Their substantial assistance played an important role in completing this work. Having these friends beside me has been truly valuable.
    \begin{otherlanguage}{estonian}
\chapter*{Sisukokkuv\~ote}
\addcontentsline{toc}{chapter}{Sisukokkuv\~ote (Summary in Estonian)}

\section*{Usaldusväärse reaalajas trajektoori optimeerimise suunas} 
Liikumise planeerimine on robootika põhiaspekt, mis võimaldab robotitel liikuda läbi keeruliste ja muutuvate keskkondade. Levinud lähenemisviis liikumise planeerimise probleemide lahendamiseks on trajektoori optimeerimine. Trajektoori optimeerimine võib matemaatiliste aparatuuride kaudu esindada robotite kõrgtasemelist käitumist. Siiski praegustel trajektoori optimeerimise lähenemisviisidel on kaks peamist väljakutset. Esiteks, sõltub nende lahendus suuresti esialgsest oletusest ja nad kipuvad takerduma kohalike miinimumidesse. Teiseks seisavad nad silmitsi mastaapsuse piirangutega, kuna kitsenduste arv suureneb.

Antud doktoritöö püüab nende väljakutsetega toime tulla, tutvustades nelja uuenduslikku trajektoori optimeerimise algoritmi, et parandada usaldusväärsust, mastaapsust ja arvutusliku efektiivsust.

Pakutud algoritmidel on kaks uudset aspekti. Esimene oluline uuendus on kinemaatiliste kitsenduste ja kokkupõrke vältimise kitsenduste ümberkujundamine. Teine oluline uuendus seisneb algoritmide väljatöötamises, mis kasutavad tõhusalt graafikaprotsessori kiirendite paralleelset arvutust. Kasutades ümbersõnastatud kitsendusi ja võimendades graafikaprotsessoritede arvutusvõimsust, näitavad selle lõputöö pakutud algoritmid oluliselt tõhususe ja mastaapsuse paranemist võrreldes olemasolevate meetoditega. Paralleelarvutus võimaldab kiiremat arvutusaega, võimaldades dünaamilistes keskkondades reaalajas otsuseid langetada. Lisaks on algoritmid loodud kohanema keskkonnamuutustega, tagades tugeva jõudluse isegi tundmatutes ja segastes tingimustes.

Iga pakutud optimeerija põhjalik võrdlusanalüüs kinnitab nende tõhusust. Tänu põhjalikule hindamisele ületavad pakutud algoritmid pidevalt tipptasemel meetodeid erinevate mõõdustike kaudu, näiteks sujuvuse kulude ja arvutusaja osas. Need tulemused rõhutavad pakutud trajektoori optimeerimise algoritmide potentsiaali robootikarakenduste liikumise planeerimise tipptasemel märkimisväärselt edendada.

Kokkuvõttes annab antud doktoritöö olulise panuse trajektoori optimeerimise algoritmide valdkonnale. See tutvustab uuenduslikke lahendusi, mis käsitlevad konkreetselt olemasolevate meetodite ees seisvaid väljakutseid. Kavandatud algoritmid sillutavad teed tõhusamatele ja jõulisematele liikumisplaneerimise lahendustele robootikas, võimendades paralleelset arvutust ja spetsiifilisi matemaatilisi struktuure.
\end{otherlanguage} 
    \IfThesisTypeIsCollection{\vfill\newpage\null\newpage
\chapter{Publications}
\thispagestyle{empty}

\vfill\newpage\null\newpage
\marginpar{
\begin{tikzpicture}
      \draw[fill,color=black] (0,0) rectangle (4cm,2cm);
      \draw[color=white] (1cm,1cm) node {\normalfont\huge\bfseries I};
\end{tikzpicture}
}

\phantomsection\addcontentsline{toc}{section}{A Novel Trajectory Optimization for Affine Systems: Beyond
Convex-Concave Procedure}
\thispagestyle{empty}\null\vfill
\begin{flushright}
F. Rastgar, A. K. Singh, H. Masnavi, K. Kruusamae, A. Aabloo\\
A Novel Trajectory Optimization for Affine Systems: Beyond
Convex-Concave Procedure\\
2020 IEEE/RSJ International Conference on Intelligent Robots and Systems (IROS), Las Vegas, NV, USA, 2020, pp. 1308-1315\\
\medskip
The article is reprinted with permission of the copyright owner.
\end{flushright}
\includepdf[pages=-]{publications/publication_1.pdf}

\vfill\newpage\null\newpage
\marginpar{
\begin{tikzpicture}
      \draw[fill,color=black] (0,0) rectangle (4cm,2cm);
      \draw[color=white] (1cm,1cm) node {\normalfont\huge\bfseries II};
\end{tikzpicture}
} 

\phantomsection\addcontentsline{toc}{section}{GPU Accelerated Batch Trajectory Optimization for Autonomous Navigation}
\thispagestyle{empty}\null\vfill
\begin{flushright}
F. Rastgar, H. Masnavi, K. Kruusamäe, A. Aabloo and A. K. Singh\\
GPU Accelerated Batch Trajectory Optimization for Autonomous Navigation\\
2023 American Control Conference (ACC), San Diego, CA, USA, 2023, pp. 718-725\\
\medskip
The article is reprinted with permission of the copyright owner.
\end{flushright}
\includepdf[pages=-]{publications/publication_3.pdf}

\vfill\newpage\null\newpage
\marginpar{
\begin{tikzpicture}
      \draw[fill,color=black] (0,0) rectangle (4cm,2cm);
  \draw[color=white] (1cm,1cm) node {\normalfont\huge\bfseries III};
\end{tikzpicture}
} 

\phantomsection\addcontentsline{toc}{section}{PRIEST: Projection Guided Sampling-Based Optimization For Autonomous Navigation}
\thispagestyle{empty}\null\vfill
\begin{flushright}
F. Rastgar, H. Masnavi, B. Sharma, A. Aabloo, J. Swevers, A. K. Singh\\
PRIEST: Projection Guided Sampling-Based Optimization For Autonomous Navigation\\
IEEE Robotics and Automation Letters\\
January 2024 \\
\medskip
The article is reprinted with permission of the copyright owner.
\end{flushright}
\includepdf[pages=-]{publications/publication_4.pdf}

\vfill\newpage\null\newpage
\marginpar{
\begin{tikzpicture}
      \draw[fill,color=black] (0,0) rectangle (4cm,2cm);
  \draw[color=white] (1cm,1cm) node {\normalfont\huge\bfseries IV};
\end{tikzpicture}
} 

\phantomsection\addcontentsline{toc}{section}{GPU Accelerated Convex Approximations for Fast
Multi-Agent Trajectory Optimization}
\thispagestyle{empty}\null\vfill
\begin{flushright}
F. Rastgar, H. Masnavi, J. Shrestha, K. Kruusamäe, A. Aabloo and A. K. Singh\\
GPU Accelerated Convex Approximations for Fast
Multi-Agent Trajectory Optimization\\
IEEE Robotics and Automation Letters, vol. 6, no. 2, pp. 3303-3310, April 2021, doi: 10.1109/LRA.2021.3061398\\
\medskip
The article is reprinted with permission of the copyright owner.
\end{flushright}
\includepdf[pages=-]{publications/publication_2.pdf}

}{}
    
    \vfill\newpage\null\newpage
    \chapter{Curriculum Vitae}

\section*{Personal data}

\begin{tabular}{@{}l@{\hskip7mm}l}
Name:       & Fatemeh Rastgar\\
Date of birth:       & 24.07.1991\\
Contact:       & fatemeh@ut.ee\\
Current Position:       & Junior Research Fellow in Robotics 
\end{tabular}

\section*{Education}

\begin{tabular}{@{}l@{\hskip7mm}p{90mm}}
2019--2024       & Ph.D. Candidate, University of Tartu, Tartu, Estonia\\
2015--2018        & MSc. Electrical and Control Engineering, Imam Khomeini International University \\
2010--2014        & BSc. Electrical and Electronic Engineering, Shahid Rajaee Teacher Training University
\end{tabular}

\section*{Employment}

\begin{tabular}{@{}l@{\hskip7mm}p{90mm}}
2020--2024         & Junior Research Fellow in Robotics,  Institute of Technology, University of Tartu\\
\end{tabular}

\section*{Scientific work}

Main fields of interest:
\begin{itemize}
  \item Motion Planning and Control
  \item Optimization
  \item Robotics
\end{itemize} 
    \begin{otherlanguage}{estonian}
\chapter*{Elulookirjeldus}
\addcontentsline{toc}{chapter}{Elulookirjeldus (Curriculum Vitae in Estonian)}

\section*{Isikuandmed}

\begin{tabular}{@{}l@{\hskip7mm}l}
Nimi:       & Fatemeh Rastgar\\
Sünniaeg:   & 24.07.1991\\
E-mail:     & fatemeh@ut.ee\\
Praegune positsioon:  & robootika nooremteadur
\end{tabular}

\section*{Haridus}

\begin{tabular}{@{}l@{\hskip7mm}p{90mm}}
2019--2024        & Tartu Ülikool, Loodus- ja täppisteaduste valdkond, tehnoloogiainstituut, doktoriõpe\\
2015--2018         & Imam Khomeini International Ülikool, Elektri- ja juhtimistehnika, magistriõpe (\textit{cum laude})\\
2010--2014         & Shahid Rajaee Teacher Training Ülikool, Elektri- ja elektroonikatehnika, bakalaureuseõpe, (\textit{cum laude})
\end{tabular}

\section*{Teenistusk\"aik}

\begin{tabular}{@{}l@{\hskip7mm}p{90mm}}
2020--2024         & Tartu Ülikool, Loodus- ja täppisteaduste valdkond, tehnoloogiainstituut, robootika nooremteadur\\
\end{tabular}

\section*{Teadustegevus}

Peamised uurimisvaldkonnad:
\begin{itemize}
\item Liikumise planeerimine ja juhtimine
\item Optimeerimine
\item Robootika
\end{itemize}
\end{otherlanguage}
   
\end{document}